\documentclass[11pt]{article}

\usepackage[preprint]{acl}

\usepackage{times}
\usepackage{latexsym}

\usepackage[T1]{fontenc}
\usepackage[utf8]{inputenc}

\usepackage{microtype}

\usepackage{inconsolata}

\usepackage{booktabs}
\usepackage[ruled,vlined]{algorithm2e}
\usepackage{amsmath,amssymb}
\usepackage{graphicx}
\usepackage{subcaption}

\newcommand{\pv}{\textit{p}-value}
\newcommand{\pvs}{\textit{p}-values}

\title{Improving Reproducibility in Evaluation through Multi-Level Annotator Modeling}

\author{
  Deepak Pandita \\
  Rochester Institute of Technology\\
  \texttt{deepak@mail.rit.edu} \\
  \And
  Flip Korn \\
  Google Research \\
  \texttt{flip@google.com} \\
  \AND
  Chris Welty \\
  Google Research \\
  \texttt{cawelty@gmail.com} \\
  \And
  Christopher M. Homan \\
  Rochester Institute of Technology\\
  \texttt{cmh@cs.rit.edu} \\
}

\begin{document}
\maketitle
\begin{abstract}

As generative AI models such as large language models (LLMs) become more pervasive, ensuring the safety, robustness, and overall trustworthiness of these systems is paramount. However, AI is currently facing a reproducibility crisis driven by unreliable evaluations and unrepeatable experimental results. While human raters are often used to assess models for utility and safety, they introduce divergent biases and subjective opinions into their annotations. Overcoming this variance is exceptionally challenging because very little data exists to study how experimental repeatability actually improves as the annotator pool grows. Standard evaluation practices typically rely on a small number of annotations per item (often 3 to 5) and lack the persistent rater identifiers necessary to model individual variance across items. In this work, we introduce a multi-level bootstrapping approach to model annotator behavior realistically. Leveraging datasets with a large number of ratings and persistent rater identifiers, we analyze the tradeoffs between the number of items ($N$) and the number of responses per item ($K$) required to achieve statistical significance.

\end{abstract}

\section{Introduction}
\label{sec:intro}

% \subsection{What is the problem?}
In a shockingly short amount of time, generative AI (GenAI) has become an essential tool of the white-collar workplace and life in general. While anecdotally, this technology seems to boost productivity on both menial and creative tasks, it is unreliable, usually requiring active human supervision to protect against harmful or incorrect content. When humans trust GenAI too much, tragedies can result~\citep{campbell2025examination}. 

Any scientific pathway to better GenAI trustworthiness requires standard, reliable, repeatable evaluation protocols. Yet reproducibility remains a problem of considerable concern~\citep{kapoor2022leakagereproducibilitycrisismlbased,semmelrock2024reproducibility}. The reasons why this remains the case are not entirely clear, but a major problem is an over-reliance on benchmarks where each input (or prompt) is paired with a single \emph{gold-standard output}. In an era when inference relies on randomness, and where the annotators who provide the gold standard outputs frequently disagree~\citep{Cabitza_Campagner_Basile_2023}, investigators are beginning to explore datasets with multiple,  \emph{disaggregated} outputs for each input. 

However, human annotations are an expensive resource, and a critical question becomes, \emph{How many annotations for each input item are needed for us to reliably distinguish the performance of one model from another?} This problem has received recent attention~\citep{Pandita_Korn_Welty_Homan_2026}, however, to our knowledge, prior approaches have treated the annotations for each item as independent from all other items, when in fact, in most datasets, most annotators rate multiple items. Failing to account for this dependency effectively increases the degrees of freedom assumed to exist, which can lead to overestimating the statistical power of the dataset.

In this paper, we explore the consequences of accounting for annotators across all items that they annotate and how doing so affects the number of items and responses per item needed to obtain reliable null hypothesis significance tests (NHSTs) for comparing machine learning models. We use simulations grounded in real-world data to generate results by under- and over-sampling with replacement from the few machine learning evaluation datasets that contain multiple, disaggregated annotations per item as well as identifiers that allow us to track annotators across multiple items. We ask the following research questions:

\begin{description}
\item[RQ1] Does accounting for rater behavior across items result in higher \pv\ estimates than when rater behavior is not accounted for?
\item[RQ2] Does accounting for rater behavior across items affect the tradeoff between the ideal number of raters and items needed to construct reliable hypothesis tests?
\item[RQ3] How does having multiple, distinct sets of raters working on batches of items impact \pv\ estimation?
\item[RQ4] How does downsampling the number of raters affect \pv\ estimation?
\end{description}

Our results show that failure to account for annotator behavior across items results in substantial underestimates of \pvs. Consequentially, we recommend that creators of evaluation datasets should seriously consider increasing the number of annotators who respond to each input.

\section{Related Work}
\label{sec:related_work}
Though there are exceptions, for the most part, reasoning on GenAI models is done stochastically. However, there are other sources of inferential variance, such as differences in the hardware on which the model is running. \citet{pham_problems_2020} found that 84\% of survey participants ($n= 901)$ were unsure or unaware of the sources of variance in their models. So, it is not a surprise that AI is facing a reproducibility crisis~\citep{baker_1500_2016,Gundersen_Kjensmo_2018,hutson_2018,mieskes-etal-2019-community,Gundersen_2020} in which researchers are not able to reproduce the results of previous studies~\citep{NEURIPS2019_c429429b}. Recognizing this, there have been initiatives where top venues in AI and NLP introduced reproducibility guidelines and checklists to mitigate the problem~\citep{review_acl_2024,neurips_2021,ijcai_2022,aaai_2023,icml_2023,dodge-etal-2019-show,rogers-etal-2021-just-think}. However, the effectiveness of such measures is yet to be determined.

In creating gold-standard data, it is common practice to gather responses from three to five annotators for each input item~\citep{snow-etal-2008-cheap} and then to aggregate these responses into a single response before the data is ever publicly released. Such an aggregation erases rater disagreement by treating it as noise rather than a signal of subjective complexity~\citep{barile2021toward,davani-etal-2022-dealing}. Recent work in NLP Perspectivism challenges this ``gold label'' paradigm, arguing that variance in human labels--especially in sensitive domains like toxicity or hate-speech--captures essential diverse viewpoints~\citep{Cabitza_Campagner_Basile_2023}. Consequently, researchers have begun to make the case that datasets should provide the disaggregated responses so that response variance can be accounted for~\citep{basile-etal-2021-need,prabhakaran-etal-2021-releasing,weerasooriya-etal-2023-disagreement,rizzi_soft_2024,pandita-etal-2024-rater}. 

\citet{wein-etal-2023-follow} introduced the Variance Estimation Toolkit (VET~\footnote{\url{https://github.com/google-research/vet}})
to evaluate system rankings by calculating \pvs\ for model comparisons. VET utilizes NHSTs to validate model advancements while deliberately accounting for sampling variance across both individual items and their corresponding responses. It investigates which specific measurement, aggregation, and sampling techniques most accurately estimate the true, ground-truth \pv\ from a single test set. The VET simulator generates synthetic responses representing a large human rater pool ($G$) alongside two distinct machine learning models ($A$ and $B$). To model $G$, it applies a \textit{multistage sampling} method to produce $K$ responses across $N$ items. First, a mean and standard deviation are drawn from specific uniform distributions for each item, and then $K$ responses are sampled from a normal distribution defined by those parameters. The simulated responses for models $A$ and $B$ rely on these identical parameters, with one key difference: model $B$'s means are adjusted by a minor perturbation $\epsilon$ (selected uniformly at random within a specific range). To construct data for the null hypothesis, the responses from models $A$ and $B$ are merged into a single pool and resampled. Finally, NHSTs are applied to this combined set to calculate \pv\ estimates across various sampling techniques and metrics.

\citet{homan-etal-2026-many} leveraged the VET simulator to determine the necessary $N$ and $K$ required to robustly compare two ML models, assuming a minimum performance difference of $\epsilon$ as measured by a specific metric $\Gamma$. \citet{homan-etal-2026-many} studied seven datasets with multiple human annotations and found that for regression models, increasing K is often more critical for significance than increasing N. \citet{Pandita_Korn_Welty_Homan_2026} extended the framework to categorical datasets and adopted a Bayesian approach for modeling to investigate the optimization problem of allocating a fixed human annotation budget ($N \times K$). \citet{Pandita_Korn_Welty_Homan_2026} experimented with five categorical datasets and found that increasing $K$ yields more reliable evaluations than increasing $N$. Furthermore, a modest budget ($N \times K \le 1000$, with $K > 10$) sufficiently captures the full human response distribution, and distribution-sensitive metrics particularly benefit from higher $K$.

While the prior work~\citep{wein-etal-2023-follow, Pandita_Korn_Welty_Homan_2026, homan-etal-2026-many} has looked into response variance, it assumes that responses are drawn from specific distributions and are conditionally independent across items. In practice, however, individual human raters typically annotate multiple items, introducing biases and subjective opinions across their annotations. To account for the hierarchical structure of rater-item interactions, we relax these assumptions by leveraging multi-level non-parametric bootstrap sampling to more accurately model rater dynamics.

\section{Methods}
\label{sec:methods}

\subsection{Bootstrap Sampling}
\label{sec:bootstrap}
We utilize multi-level bootstrap sampling to answer our research questions. Bootstrap sampling is used to estimate the distribution of a statistic while making no assumptions about the sampling distribution. It involves sampling examples with replacement from the original dataset to construct a new bootstrap dataset and then calculating a statistic. This process is repeated many times to obtain a bootstrap distribution of the statistic. We use the following multi-level sampling techniques for our study.

\paragraph{Sample Items and Raters (S1).} In this method, we assume that we have a dataset $D$ with $N$ items and $K$ raters, where all raters annotate each item. First, $N'$ items are sampled independently with replacement from $N$ items. Next, $K'$ raters are sampled independently with replacement from $K$ raters. Then, a dataset $D'$ is constructed using $N'$ items and $K'$ raters. This process is described in Algorithm \ref{alg:sample_items_and_raters}.

\begin{algorithm}[ht]
\SetKwInput{KwInput}{Inputs} 
\caption{Sample Items and Raters (S1)}
\label{alg:sample_items_and_raters}
\KwInput{Dataset, $D$, with $N$ items and $K$ raters, $N'$, $K'$}
% \SetKwFor{RepTimes}{Bootstrap: Repeat}{times}{end}
% \RepTimes{$num\_samples$}{
\textbf{Resample Items:} Sample $N'$ items independently with replacement from $N$\\
\textbf{Resample Raters:} Sample $K'$ raters independently with replacement from $K$\\
\textbf{Create new dataset}: Create $D'$ using resampled items $N'$ and raters $K'$\\
% }
\end{algorithm}

\paragraph{Sample Items and Raters Within an Item (S2).} In this method, we assume that we have a dataset $D$ with $N$ items and $K$ raters, where raters annotate a different number of items. First, $N'$ items are sampled independently with replacement from $N$ items. Next, for each item in $N'$, $K'$ raters are sampled independently with replacement from a pool of raters who annotated that item. Then, a dataset $D'$ is constructed using $N'$ items and corresponding ratings from $K'$ raters. This process is described in Algorithm \ref{alg:sample_items_and_raters_within_item}.

\begin{algorithm}[ht]
\SetKwInput{KwInput}{Inputs} 
\caption{Sample Items and Raters Within an Item (S2)}
\label{alg:sample_items_and_raters_within_item}
\KwInput{Dataset, $D$, with $N$ items and $K$ raters, $N'$, $K'$}
% \SetKwFor{RepTimes}{Bootstrap: Repeat}{times}{end}
% \RepTimes{$num\_samples$}{
\textbf{Resample Items:} Sample $N'$ items independently with replacement from $N$\\
\textbf{Resample Raters:} For each item in $N'$, sample $K'$ raters independently with replacement from a set of raters who annotated that item\\
\textbf{Create new dataset}: Create $D'$ using resampled items $N'$ and ratings from raters $K'$\\
% }
\end{algorithm}

\paragraph{Sample Stratified Batches (S3).} In this method, we assume that we have a dataset $D$ with $N$ items and $K$ raters, where items are divided into equal-sized ($N_b$) batches and are annotated by the same raters in a given batch. First, we sample batches based on $N'$ from the set of batches. Next, for each batch, $N_b'$ items are sampled independently with replacement from $N_b$ items. Then, for each batch, $K'$ raters are sampled independently with replacement from a pool of raters who annotated items $N_b$. Finally, a dataset $D'$ is constructed using $N'$ items and corresponding ratings from $K'$ raters. This process is described in Algorithm \ref{alg:sample_stratified_batches}.

\begin{algorithm}[ht]
\SetKwInput{KwInput}{Inputs} 
\caption{Sample Stratified Batches (S3)}
\label{alg:sample_stratified_batches}
\KwInput{Dataset, $D$, with $N$ items and $K$ raters, $N'$, $K'$}
% \SetKwFor{RepTimes}{Bootstrap: Repeat}{times}{end}
% \RepTimes{$num\_samples$}{
\textbf{Resample Batches:} Sample $B=ceil(N'/N_b)$ batches\\
\textbf{Resample Items:} For each batch, sample $N_b'$ items independently with replacement from $N_b$ such that $N_b' * B <= N'$\\
\textbf{Resample Raters:} For each batch, sample $K'$ raters independently with replacement from a pool of raters who annotated items in $N_b$\\
\textbf{Create new dataset}: Create $D'$ using resampled items $N'$ and ratings from raters $K'$\\
% }
\end{algorithm}

\subsection{Simulation Framework}
Our simulation framework draws inspiration and builds upon earlier work~\citep{wein-etal-2023-follow, Pandita_Korn_Welty_Homan_2026, homan-etal-2026-many}, and we use the VET toolkit for our simulations. A key difference is that our simulations are based on non-parametric bootstrapping that makes no assumptions about the underlying distribution, whereas earlier work uses parametric models and assumes specific distributions for the data. The VET simulator is designed to sample data for a pool of human raters, $G$, and two machine learning models, $A$ and $B$. The distribution for model $A$ is kept the same as $G$, making it a perfect representation of $G$, but the distribution for model $B$ is perturbed by a small amount $\epsilon$. NHSTs are then applied to calculate the \pvs\ and effect sizes under different perturbations and metrics.

We also use the same general idea to sample data for $G$, $A$, and $B$ and compare the models under different perturbations and metrics (Algorithm \ref{alg:simulation}). However, our use of bootstrap sampling enables rater modeling and substantially changes the sampling process. The data for $G$ is generated using the sampling methods described in Section \ref{sec:bootstrap}. The items for model $A$ are kept the same as in $G$, but the raters are resampled using the same sampling method as $G$. The data for model $B$ is generated similarly, but a small fraction ($\epsilon$) of the responses are perturbed randomly. This process ensures that model $A$ is an ideal representation of $G$, and model $B$ is slightly worse than model $A$. Any statistical test for model comparison should be able to determine that model $A$ wins over model $B$, and the \pvs\ should converge to zero given enough data and $\epsilon>0$.

\begin{algorithm}[ht]
\SetKwInput{KwInput}{Inputs} 
\caption{Simulation for $H_{alt}$}
\label{alg:simulation}
% \KwInput{$Dataset\,D\,with\,N\,items\,and\,K\,raters, N', K',\epsilon, num\_samples$}
\KwInput{Dataset, $D$, with $N$ items and $K$ raters, $N'$, $K'$, $\epsilon$, $num\_samples$}
\SetKwFor{RepTimes}{Bootstrap: Repeat}{times}{end}
\RepTimes{$num\_samples$}{
Resample $G$ from $D$ using Algorithm \ref{alg:sample_items_and_raters}, \ref{alg:sample_items_and_raters_within_item}, or \ref{alg:sample_stratified_batches}\\
Resample $A$ by keeping items the same as $G$ and sampling raters similar to $G$\\
Resample $B$ similar to $A$ and add noise $\epsilon$\\
Compute metrics
}
Compute \pvs
\end{algorithm}

\subsection{Hypothesis Testing}
The data for the alternative hypothesis ($H_{alt}$) is generated using the process described above. The data for the null hypothesis ($H_{null}$) is generated using a process similar to the alternative hypothesis, except that the data for the models $A$ and $B$ is randomly mixed. We use a metric $\Gamma(A, B, G)$ for each pair of samples $\{A, B\}$ and gold $G$ to obtain a score. $\Gamma(A, B, G) = \Gamma(A, G) - \Gamma(B, G)$, where larger is better and $\Gamma(A, B, G) = \Gamma(B, G) - \Gamma(A, G)$, where smaller is better. We obtain a distribution over the metric scores by repeating the sampling process a large number of times and calculate a \pv\ which is the proportion of samples in the null distribution $\Gamma^{null}$ that exceed the scores in the alternative distribution $\Gamma^{alt}$.

\section{Experiments}
\label{sec:experiments}

\subsection{Data}
We utilize three datasets for our experiments. Each of these datasets contains multiple human annotations per item.

\paragraph{DICES 350.} Diversity in Conversational AI Evaluation for Safety~\citep{NEURIPS2023_a74b697b} is a dataset of 350 conversations labeled for safety by a diverse set of 123 raters. It is one of the rare datasets where all raters annotate each item, forming a fully-crossed matrix.

\paragraph{Toxicity.} The Stanford toxicity dataset~\citep{kumar2021designing} comprises 107,620 social media comments annotated by 17,280 raters. The original dataset was annotated in batches of 20 comments by the same raters, representing a common crowdsourcing setting.

\paragraph{D3code.} D3code~\citep{davani_d3code_2024} is a cross-cultural dataset containing 4554 items annotated for offensiveness by 4309 raters of different ages and genders across 21 countries.

\begin{table*}[ht]
\centering
% \small
\begin{tabular}{c|c|cccccccc}
% \toprule
Dataset & Stat & Accuracy & MAE & Wins & Precision & Recall & F1-Score & KL-Div & JSD \\
\midrule
 & NK & 5000 & 1000 & 1000 & - & 2500 & 2500 & 2500 & 1000 \\
DICES & \pv\ & 0.045 & 0.032 & 0.035 & - & 0.050 & 0.029 & 0.042 & 0.037 \\
S1 & K & 40 & 100 & 100 & - & 40 & 40 & 100 & 100 \\
 & $\Delta$ & 0.154 & 0.095 & 0.799 & - & 0.241 & 0.250 & 0.138 & 0.117 \\
 \hline
 & NK & 5000 & 1000 & 1000 & 50000 & 2500 & 2500 & 2500 & 1000 \\
DICES & \pv\ & 0.019 & 0.020 & 0.026 & 0.036 & 0.039 & 0.018 & 0.031 & 0.026 \\
S2 & K & 40 & 40 & 100 & 1 & 40 & 40 & 100 & 40 \\
 & $\Delta$ & 0.160 & 0.083 & 0.799 & 0.013 & 0.238 & 0.247 & 0.140 & 0.098 \\
 \hline
 & NK & 10000 & 1000 & 1000 & 25000 & 5000 & 2500 & 2500 & 1000 \\
D3code & \pv\ & 0.018 & 0.034 & 0.044 & 0.014 & 0.025 & 0.019 & 0.021 & 0.033 \\
S2 & K & 100 & 100 & 100 & 1 & 1 & 9 & 100 & 60 \\
 & $\Delta$ & 0.198 & 0.106 & 0.691 & 0.020 & 0.023 & 0.111 & 0.067 & 0.074 \\
 \hline
 & NK & 500 & 500 & 500 & 1000 & 500 & 500 & 500 & 500 \\
Toxicity & \pv\ & 0.035 & 0.011 & 0.024 & 0.022 & 0.028 & 0.018 & 0.035 & 0.008 \\
S2 & K & 1 & 9 & 4 & 1 & 1 & 1 & 1 & 10 \\
 & $\Delta$ & 0.061 & 0.076 & 0.163 & 0.048 & 0.061 & 0.070 & 1.268 & 0.072 \\
 \hline
 & NK & 2500 & 1000 & 2500 & 2500 & 2500 & 1000 & 2500 & 1000 \\
Toxicity & \pv\ & 0.009 & 0.034 & 0.006 & 0.030 & 0.007 & 0.039 & 0.008 & 0.030 \\
S3 & K & 1 & 20 & 4 & 1 & 1 & 1 & 40 & 10 \\
 & $\Delta$ & 0.064 & 0.089 & 0.163 & 0.050 & 0.064 & 0.071 & 0.096 & 0.073 \\
% \bottomrule
\end{tabular}
\caption{Minimum \pv, $K$, and corresponding effect size ($\Delta$) for lowest $NK$ with $p<0.05$ ($ \epsilon=0.3$).}
\label{tab:low_k_for_p_lte_05_nk_min_p_e_0.3}
\end{table*}

\subsection{Metrics}
We use the following metrics for our experiments:

\paragraph{Accuracy.} Take the plurality vote for all items and then compute the accuracy of $A$ and $B$ against $G$.

\paragraph{MAE.} Compute the distribution of labels for all items and take the mean absolute difference across all items in $A$ and $B$ against $G$.

\paragraph{Wins.} Calculate the MAE for all items in $A$ and $B$ against $G$. Count the average number of wins of $A$ and $B$ by assigning a win to $A$ if $A$ has a lower score than $B$ and vice versa.

\paragraph{Precision.} Take the plurality vote for all items and then compute the macro-weighted precision of $A$ and $B$ against $G$.
by calculating the precision for each label class and taking their average, weighted by the number of true instances in each class.

\paragraph{Recall.} Take the plurality vote for all items and then compute the macro-weighted recall of $A$ and $B$ against $G$.
by calculating the recall for each label class and taking their average, weighted by the number of true instances in each class.

\paragraph{F1-Score.} Take the plurality vote for all items and then compute the macro-weighted F1-Score of $A$ and $B$ against $G$.
by calculating the F1-Score for each label class and taking their average, weighted by the number of true instances in each class.

\paragraph{KL-Divergence (KL-Div).} Compute the distribution of labels for all items and calculate the average Kullback-Leibler divergence to measure how the predicted probability distributions of $A$ and $B$ differ from the reference distribution of $G$.

\paragraph{Jensen-Shannon Distance (JSD).} Compute the distribution of labels for all items. Calculate the Jensen-Shannon distance for $A$ and $B$ against $G$ by taking the square root of the Jensen-Shannon divergence between their respective distributions and the reference distribution of $G$.

\begin{figure*}
  \centering
  \begin{subfigure}[b]{0.32\linewidth}
    \centering
    \includegraphics[width=\linewidth]{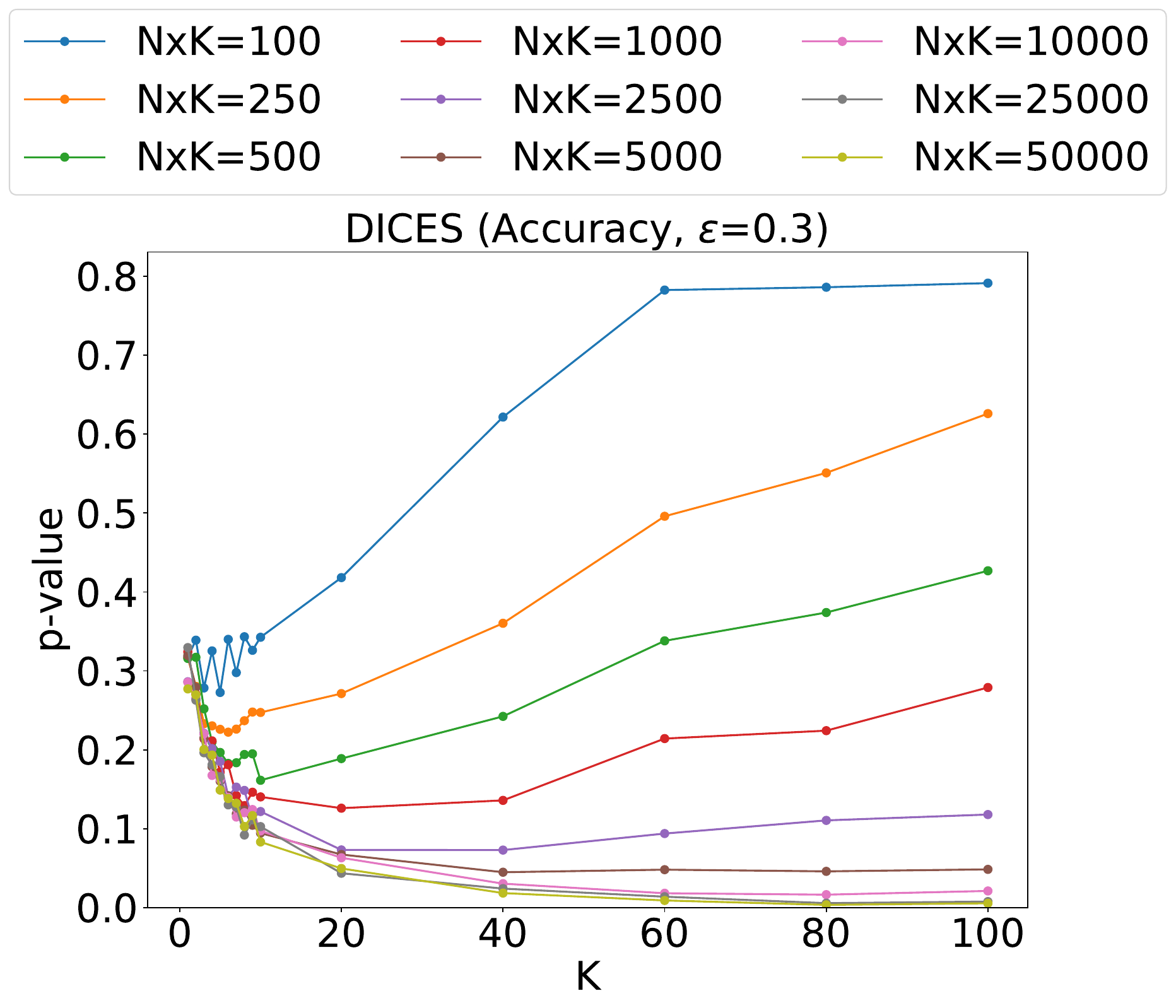}
    \caption{Accuracy, $\epsilon = 0.3$}
    \label{fig:main_dices_acc_e03}
  \end{subfigure} \hfill
  \begin{subfigure}[b]{0.32\linewidth}
    \centering
    \includegraphics[width=\linewidth]{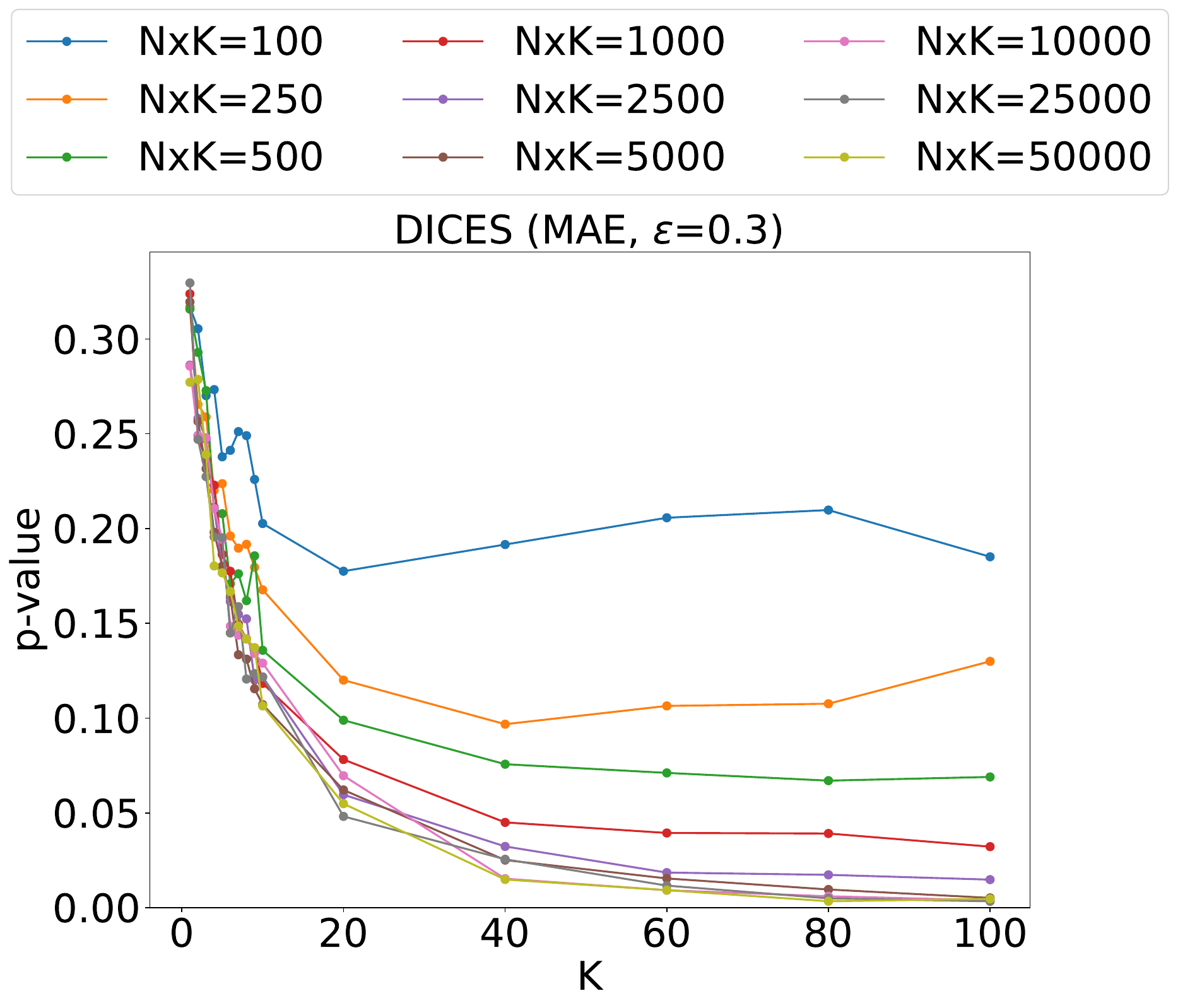}
    \caption{MAE, $\epsilon = 0.3$}
    \label{fig:main_dices_MAE_e03}
  \end{subfigure} \hfill
  \begin{subfigure}[b]{0.32\linewidth}
    \centering
    \includegraphics[width=\linewidth]{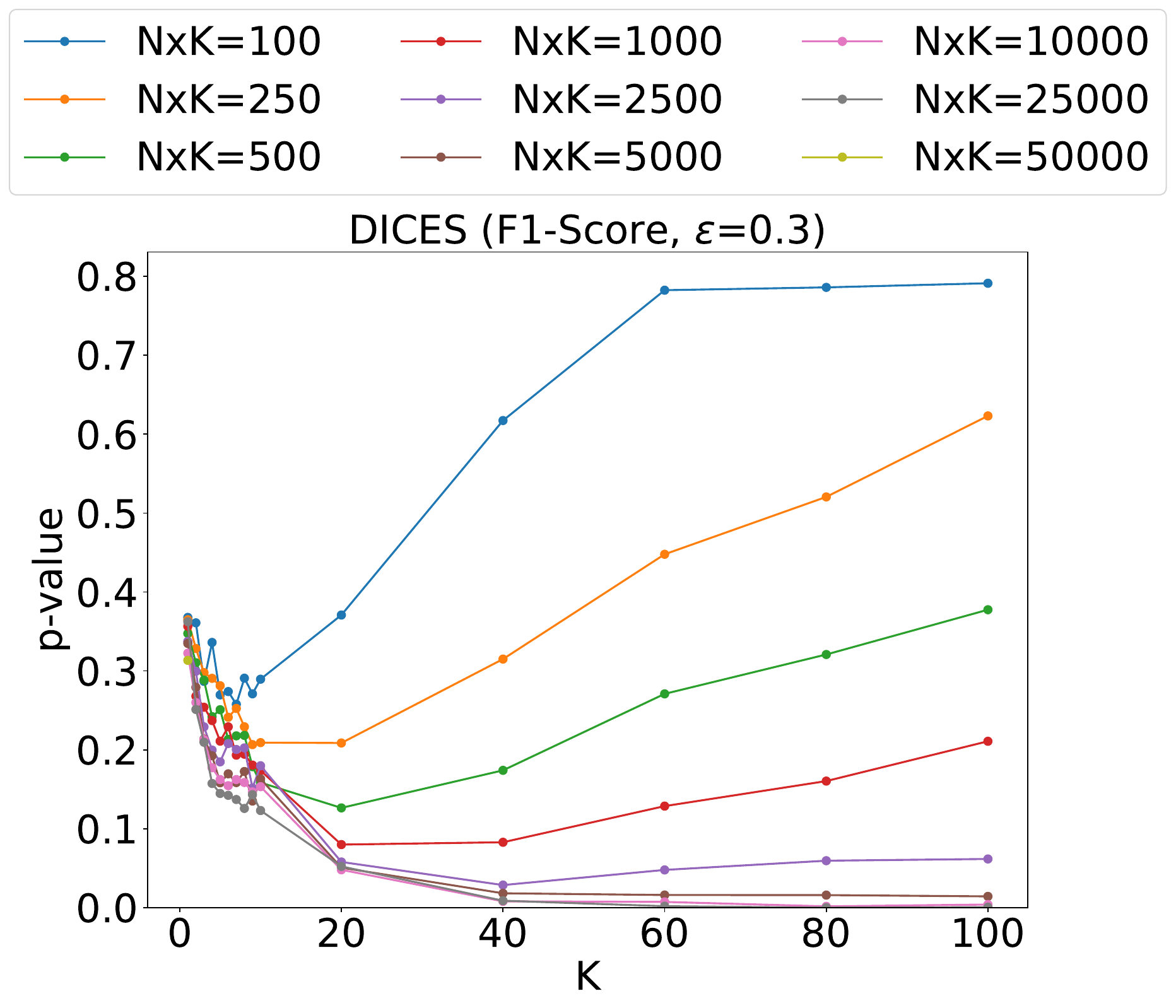}
    \caption{F1-Score, $\epsilon = 0.3$}
    \label{fig:main_dices_f1_e03}
  \end{subfigure} \hfill
  \begin{subfigure}[b]{0.32\linewidth}
    \centering
    \includegraphics[width=\linewidth]{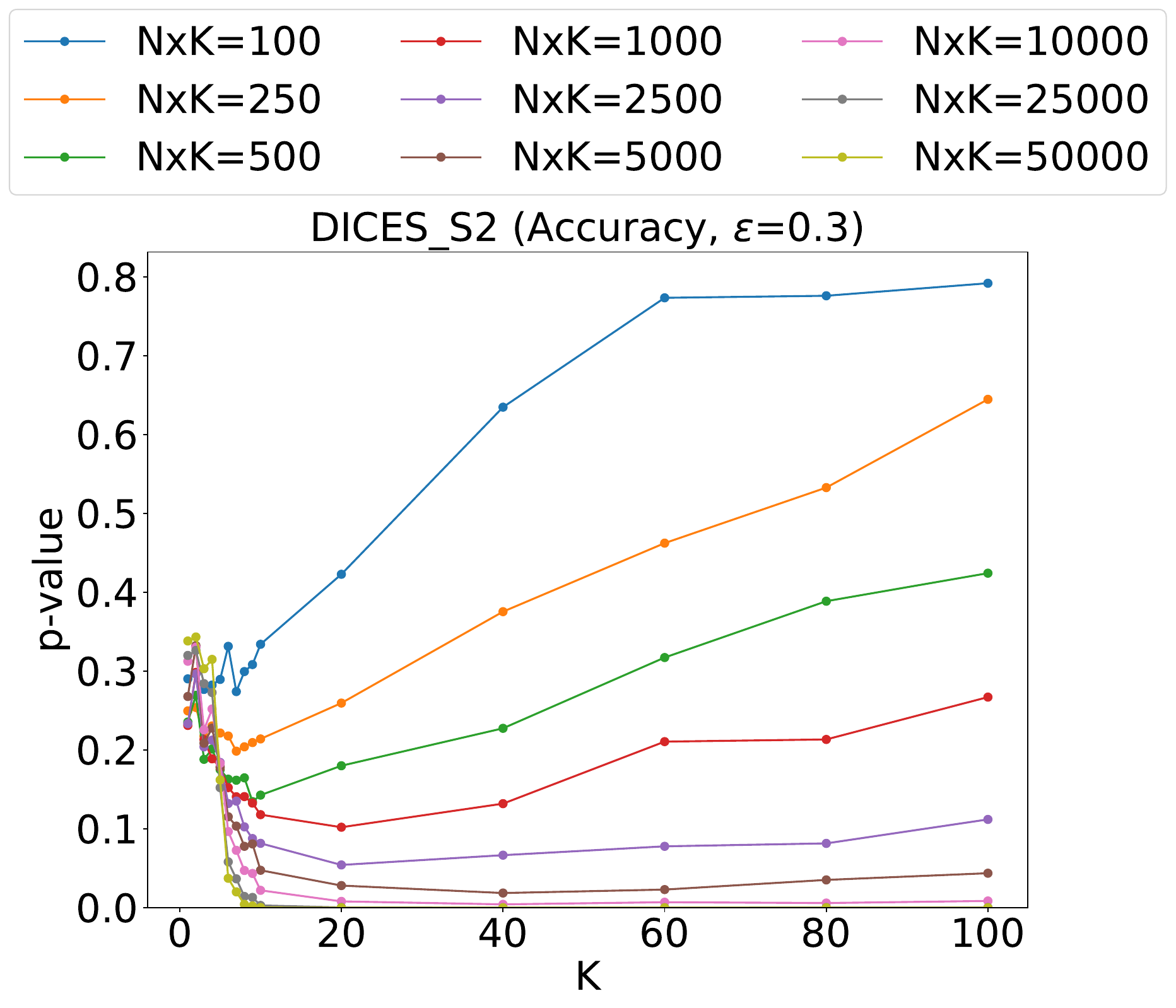}
    \caption{Accuracy, $\epsilon = 0.3$}
    \label{fig:main_dices_s2_acc_e03}
  \end{subfigure} \hfill
  \begin{subfigure}[b]{0.32\linewidth}
    \centering
    \includegraphics[width=\linewidth]{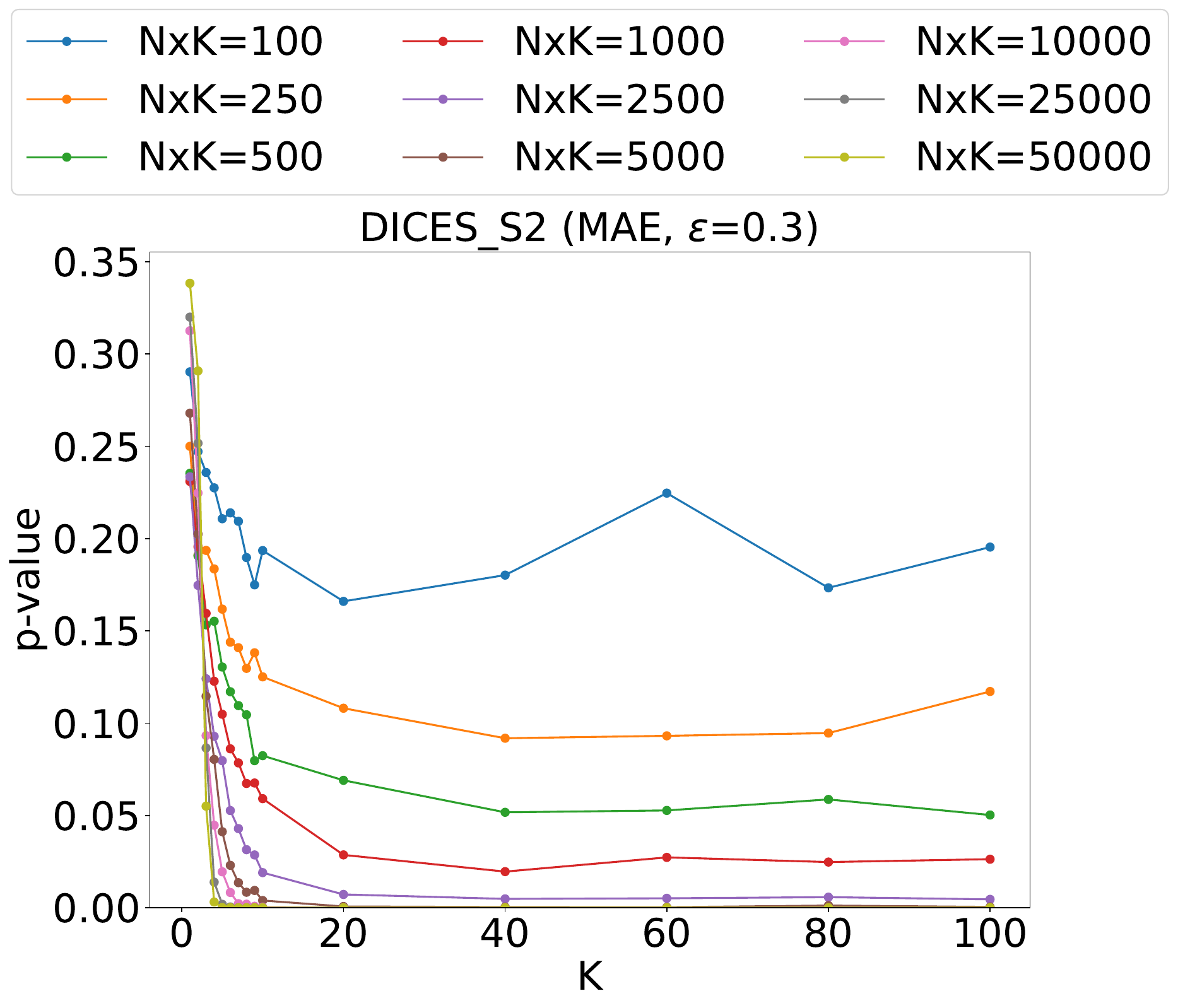}
    \caption{MAE, $\epsilon = 0.3$}
    \label{fig:main_dices_s2_MAE_e03}
  \end{subfigure} \hfill
  \begin{subfigure}[b]{0.32\linewidth}
    \centering
    \includegraphics[width=\linewidth]{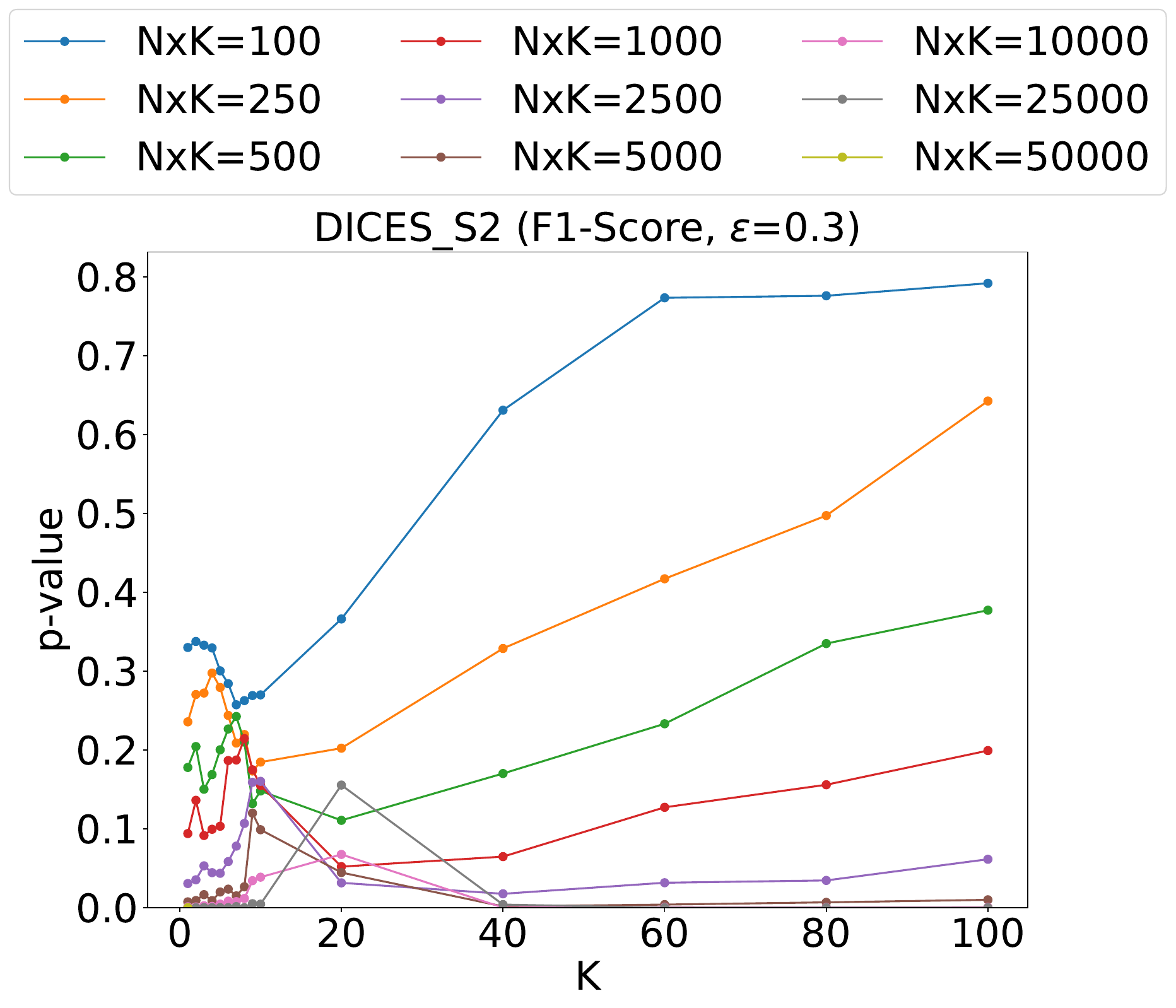}
    \caption{F1-Score, $\epsilon = 0.3$}
    \label{fig:main_dices_s2_f1_e03}
  \end{subfigure}
  \caption{Comparing S1 (top) vs S2 (bottom): P-value plots for DICES dataset with Accuracy and F1-Score}
  \label{fig:main_dices_accuracy_f1}
\end{figure*}

\subsection{Experimental Setup}
We conduct hypothesis testing experiments across varying total annotation budgets ($N \times K \in \{100$, $250$, $500$, $1000$, $2500$, $5000$, $10000$, $25000$, $50000\}$). We scale the number of responses per item ($K$) from 1 to 100, incrementing by 1 up to $K=10$, and by 20 thereafter. For each dataset, we evaluate eight distinct metrics across four different perturbations ($\epsilon \in \{0.1, 0.2, 0.3, 0.4\}$) and repeat the sampling process 1000 times.
To answer RQ4, we downsample the DICES dataset to include 5 raters, keeping items the same, and rerun our experiments as if it were the original dataset. Our experiments are conducted on either a local machine with a 16-core processor and 64GB RAM or a compute node with a 16-core processor and 40GB RAM. The experiments take anywhere between 12 to 18 hours to run for a single dataset.

\subsection{Results}

Table \ref{tab:low_k_for_p_lte_05_nk_min_p_e_0.3} shows the minimum required total budget ($NK$), the optimal number of responses per item ($K$), and the corresponding effect size ($\Delta$) required to achieve statistical significance ($p < 0.05$ at $\epsilon = 0.3$) across various metrics, datasets, and bootstrapping strategies (S1, S2, and S3).

Comparing the bootstrapping approach S1 to S2 on the DICES dataset, we notice that S2 results in overall lower \pvs. At a fixed budget $N\times K=5000$, the \pv\ for Accuracy decreases from 0.045 in S1 to 0.019 in S2. Similarly, MAE and JSD at $N\times K=1000$ yield a \pv\ of 0.032 and 0.037, respectively, under S1 but drop to 0.02 and 0.026, respectively, under S2, while the required number of responses per item drops from $K=100$ to $K=40$. Precision, which could not achieve significant results under S1, achieves significance under S2, though it requires a massive total budget of $NK=50000$. Wins, Recall, F1-Score, and KL-Div show lower \pvs\ under S2 than S1.

Comparing S2 with the stratified batch bootstrapping approach (S3) on the Toxicity dataset shows that the required budget to achieve significance increases. Under S2, Accuracy, Wins, Recall, and KL-Div are significant at $NK=500$; however, under S3, the required budget increases fivefold to $NK=2500$. Similarly, MAE, F1-Score, and JSD are significant at $NK=500$ but double to $NK=1000$ under S3. 

Under S2, DICES and D3code datasets show very interesting patterns where distribution-sensitive metrics such as MAE, Wins, and JSD achieve significance at a lower total budget of $NK=1000$ while requiring a higher $K$ from 40 to 100. However, metrics such as Accuracy, Precision, Recall, and F1-Score achieve significance at a higher total budget of $NK>=2500$. We also see similar results for other $\epsilon$.

\begin{figure*}[!ht]
  \centering
  \begin{subfigure}[b]{0.32\linewidth}
    \centering
    \includegraphics[width=\linewidth]{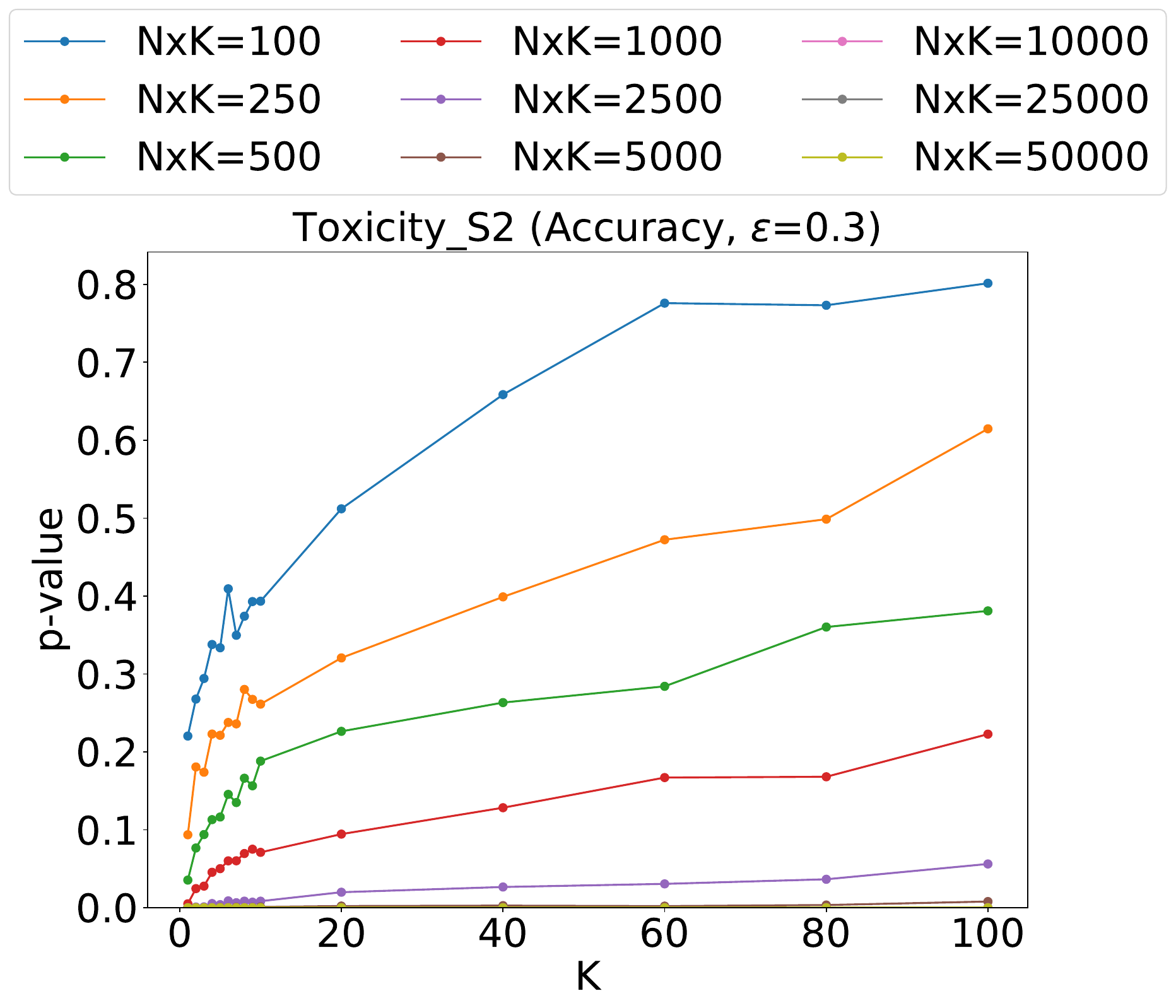}
    \caption{Accuracy, $\epsilon = 0.3$}
    \label{fig:main_toxicity_s2_acc_e03}
  \end{subfigure} \hfill
  \begin{subfigure}[b]{0.32\linewidth}
    \centering
    \includegraphics[width=\linewidth]{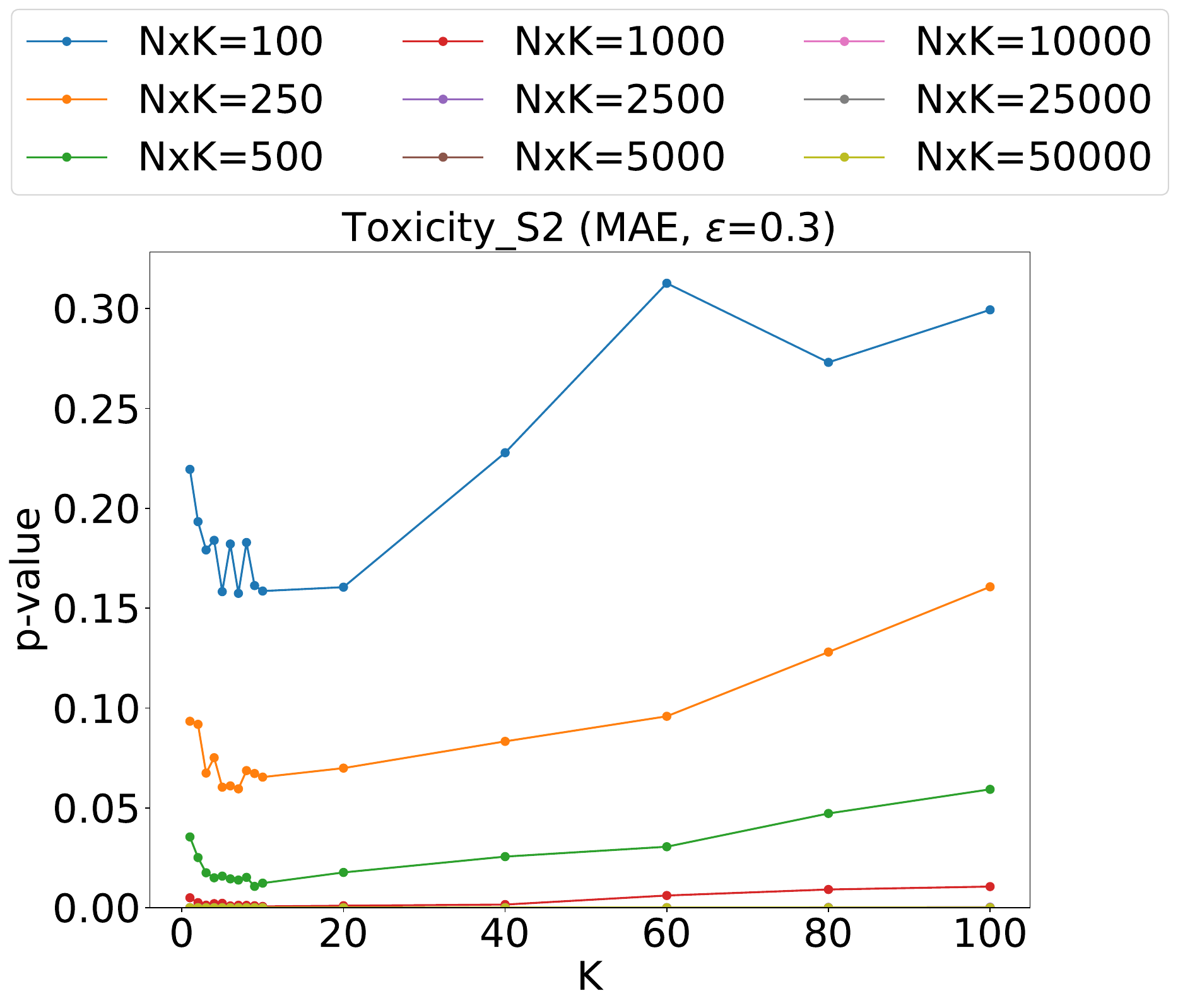}
    \caption{MAE, $\epsilon = 0.3$}
    \label{fig:main_toxicity_s2_mae_e03}
  \end{subfigure} \hfill
  \begin{subfigure}[b]{0.32\linewidth}
    \centering
    \includegraphics[width=\linewidth]{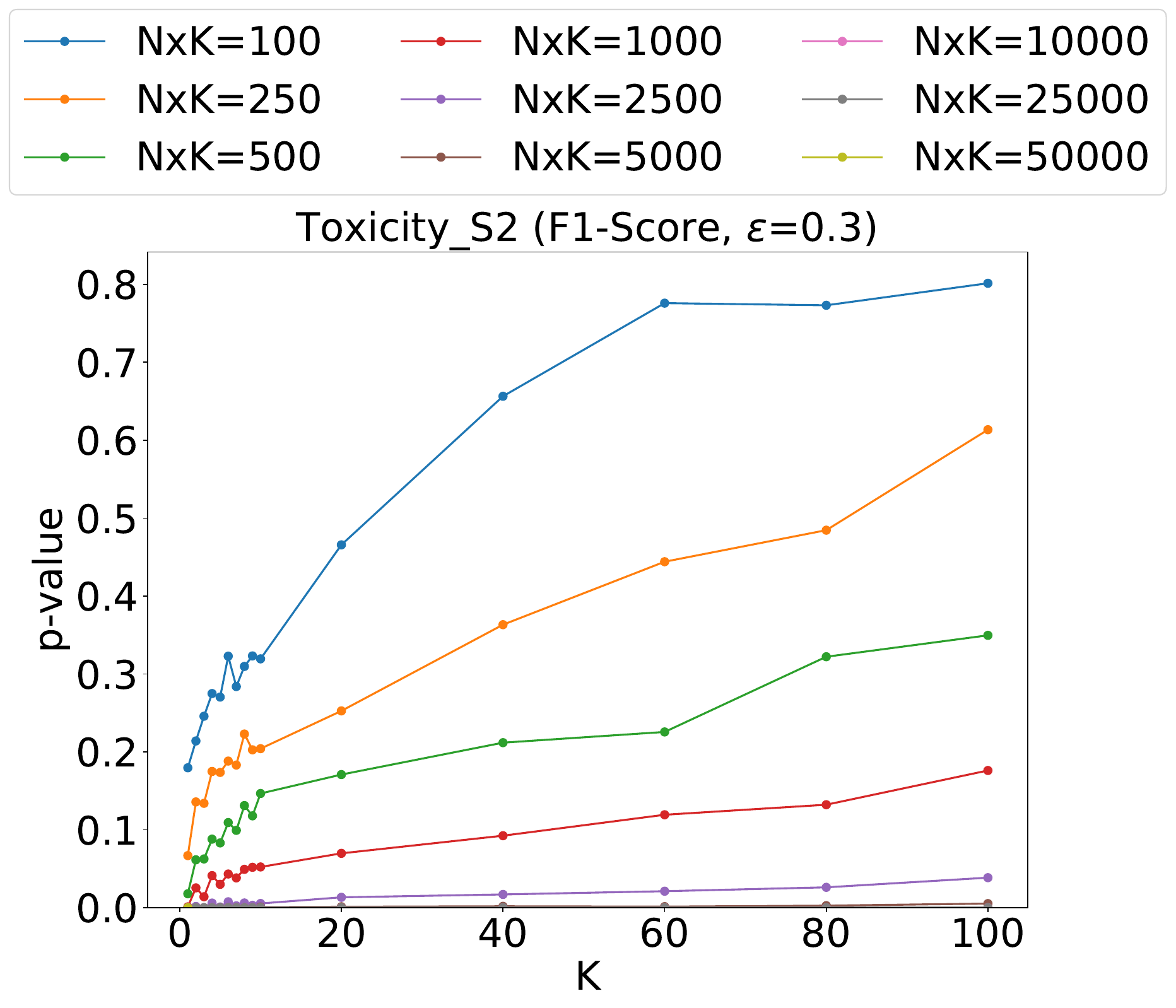}
    \caption{F1-Score, $\epsilon = 0.3$}
    \label{fig:main_toxicity_s2_f1_e03}
  \end{subfigure} \hfill
  \begin{subfigure}[b]{0.32\linewidth}
    \centering
    \includegraphics[width=\linewidth]{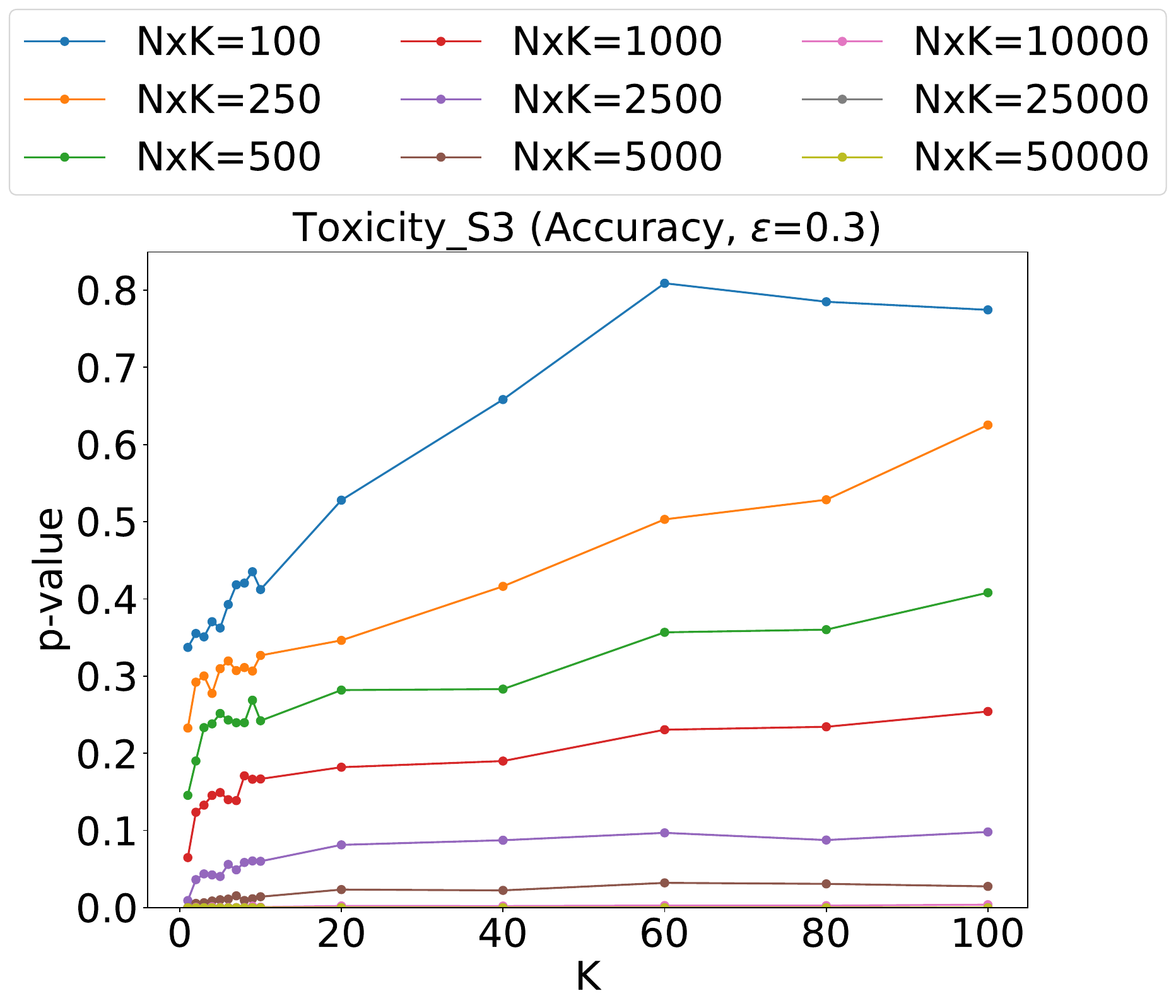}
    \caption{Accuracy, $\epsilon = 0.3$}
    \label{fig:main_toxicity_s3_acc_e03}
  \end{subfigure} \hfill
  \begin{subfigure}[b]{0.32\linewidth}
    \centering
    \includegraphics[width=\linewidth]{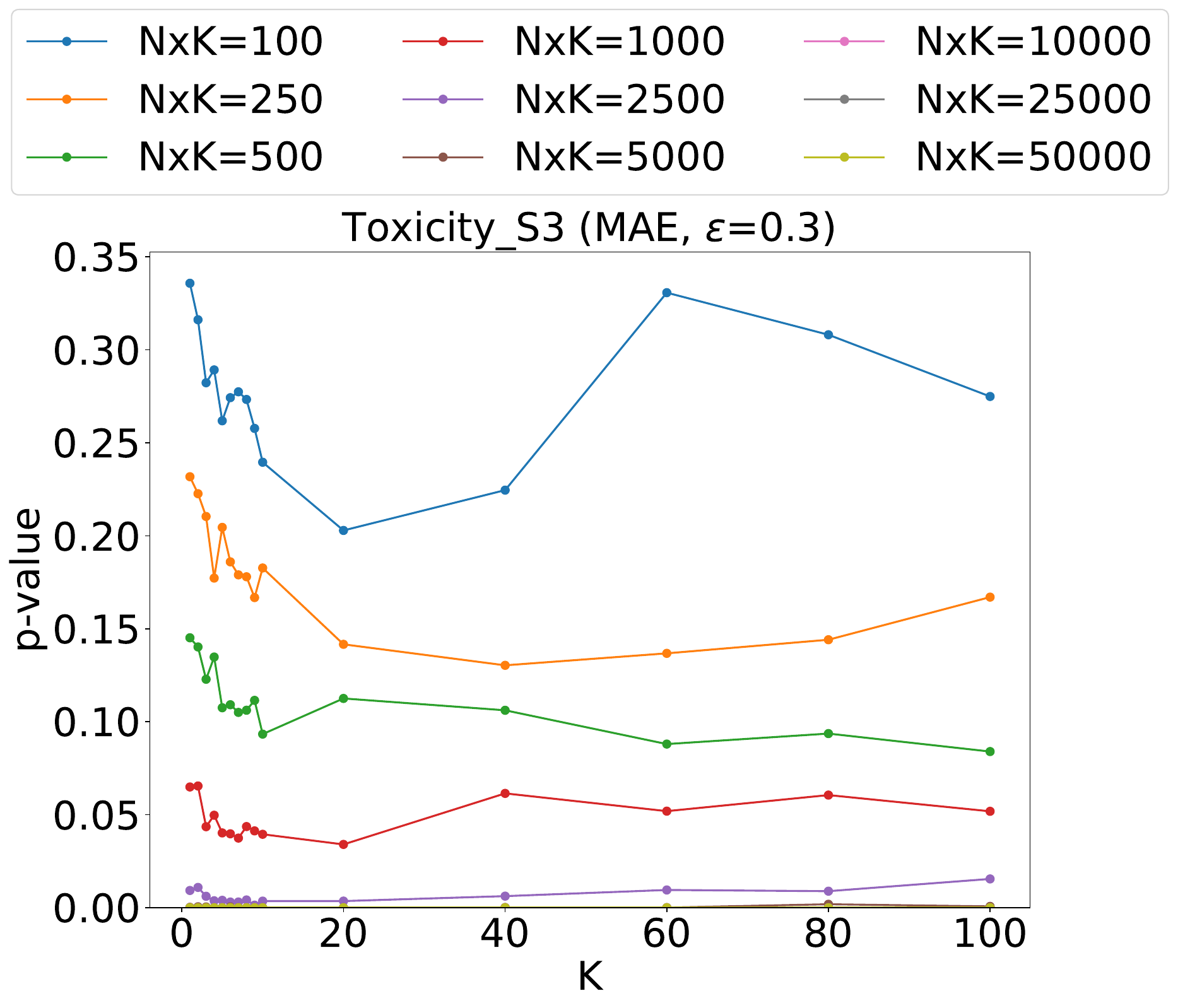}
    \caption{MAE, $\epsilon = 0.3$}
    \label{fig:main_toxicity_s3_mae_e03}
  \end{subfigure} \hfill
  \begin{subfigure}[b]{0.32\linewidth}
    \centering
    \includegraphics[width=\linewidth]{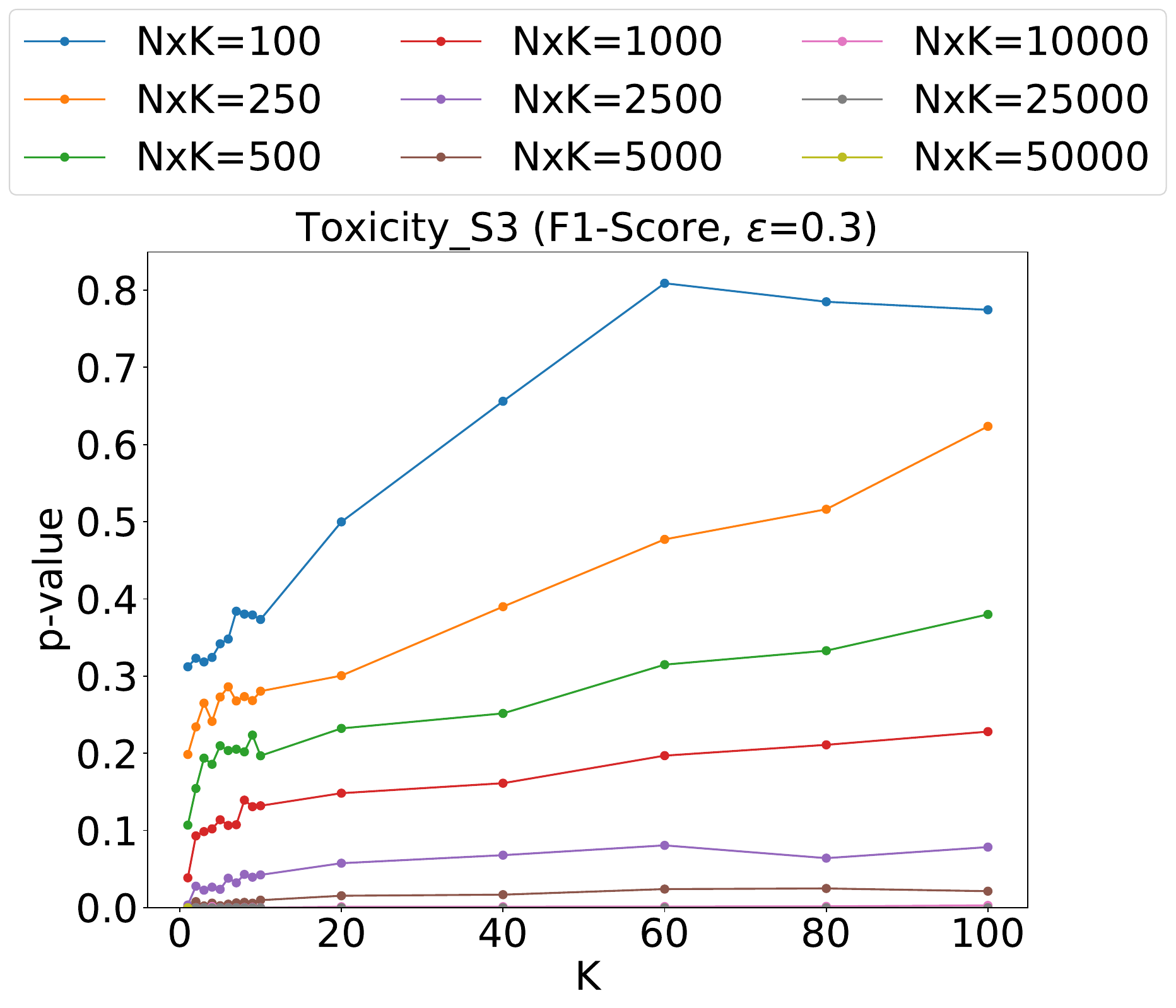}
    \caption{F1-Score, $\epsilon = 0.3$}
    \label{fig:main_toxicity_s2_f1_e04}
  \end{subfigure}
  \caption{Comparing S2 (top) vs S3 (bottom): P-value plots for Toxicity dataset with Accuracy and F1-Score}
  \label{fig:main_toxicity_accuracy_f1}
\end{figure*}

\section{Discussion}

\textbf{RQ1:} Does accounting for rater behavior across items result in higher \pv\ estimates than when rater behavior is not accounted for?

Yes, as shown in Figure \ref{fig:main_dices_accuracy_f1} and \ref{fig:main_toxicity_accuracy_f1}, our results indicate that when comparing approaches S1 (which models items and raters independently) vs S2 (which samples raters within an item without tracking their behavior across items) and S2 vs S3 (which models raters via stratified batches), S1 and S3 result in higher \pv\ estimates and need higher budgets. 

\textbf{RQ2:} Does accounting for rater behavior across items affect the tradeoff between the ideal number of raters and items needed to construct reliable hypothesis tests?

Accounting for rater behavior increases the budget required to obtain significant results. The number of raters required also increases when modeling raters, for example, for MAE, the optimal $K$ for the DICES dataset shifts from 40 under S2 to 100 under S1, and for the toxicity dataset, the optimal $K$ shifts from 9 under S2 to 20 under S3.

\textbf{RQ3:} How does having multiple, distinct sets of raters working on batches of items impact \pv\ estimation?

Having distinct sets of raters working on isolated batches (as modeled by S3) introduces batch-specific behavior, increasing the overall variance of the dataset. As a result, \pv\ estimation must counteract this effect and achieve statistical significance by scaling up the evaluation budget.

\textbf{RQ4:} How does downsampling the number of raters affect \pv\ estimation?

\begin{figure*}[!ht]
  \centering
  \begin{subfigure}[b]{0.32\linewidth}
    \centering
    \includegraphics[width=\linewidth]{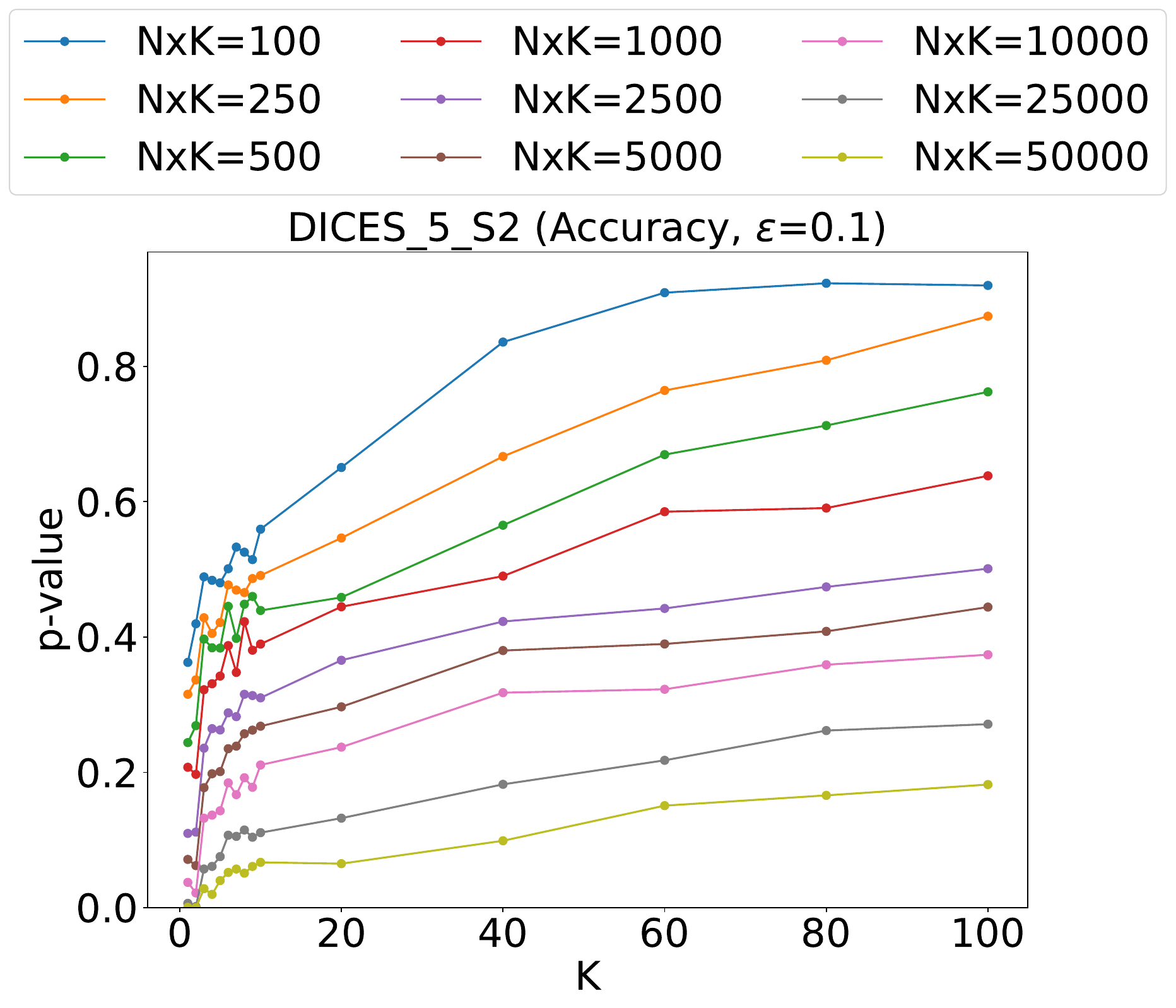}
    \caption{Accuracy, $\epsilon = 0.1$}
    \label{fig:main_dices_5_s2_acc_e01}
  \end{subfigure} \hfill
  \begin{subfigure}[b]{0.32\linewidth}
    \centering
    \includegraphics[width=\linewidth]{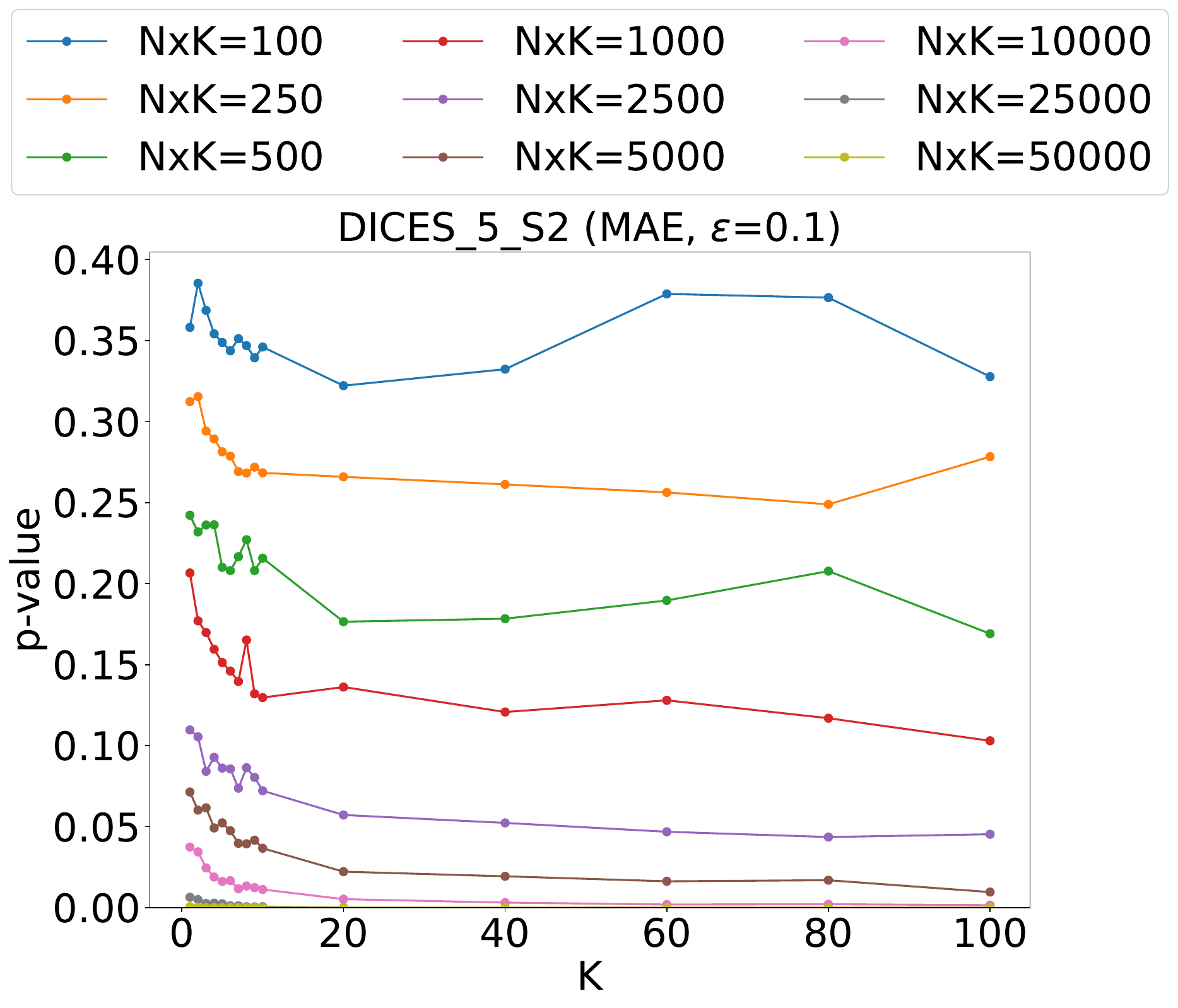}
    \caption{MAE, $\epsilon = 0.1$}
    \label{fig:main_dices_5_s2_mae_e01}
  \end{subfigure} \hfill
  \begin{subfigure}[b]{0.32\linewidth}
    \centering
    \includegraphics[width=\linewidth]{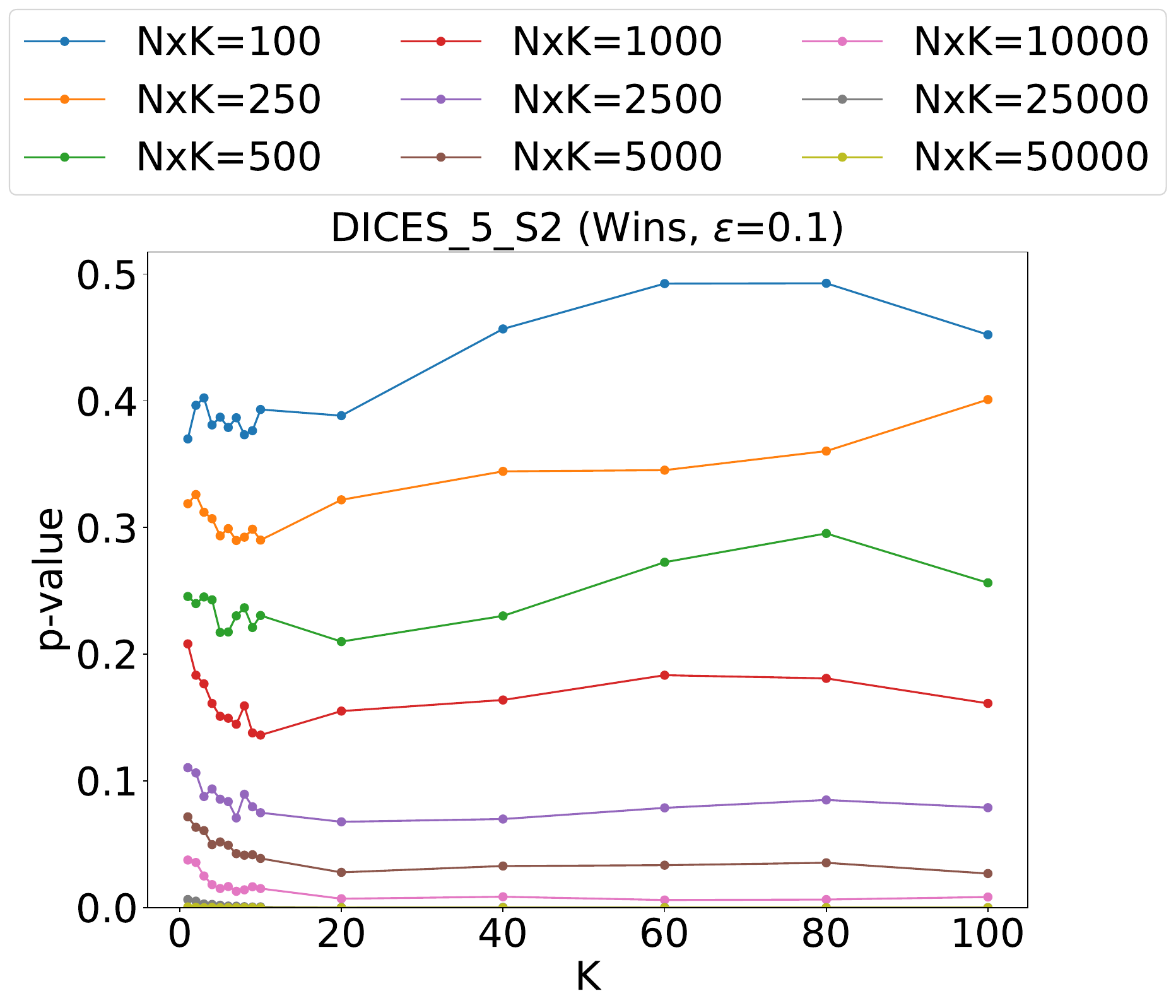}
    \caption{Wins, $\epsilon = 0.1$}
    \label{fig:main_dices_5_s2_wins_e01}
  \end{subfigure} \hfill
  \begin{subfigure}[b]{0.32\linewidth}
    \centering
    \includegraphics[width=\linewidth]{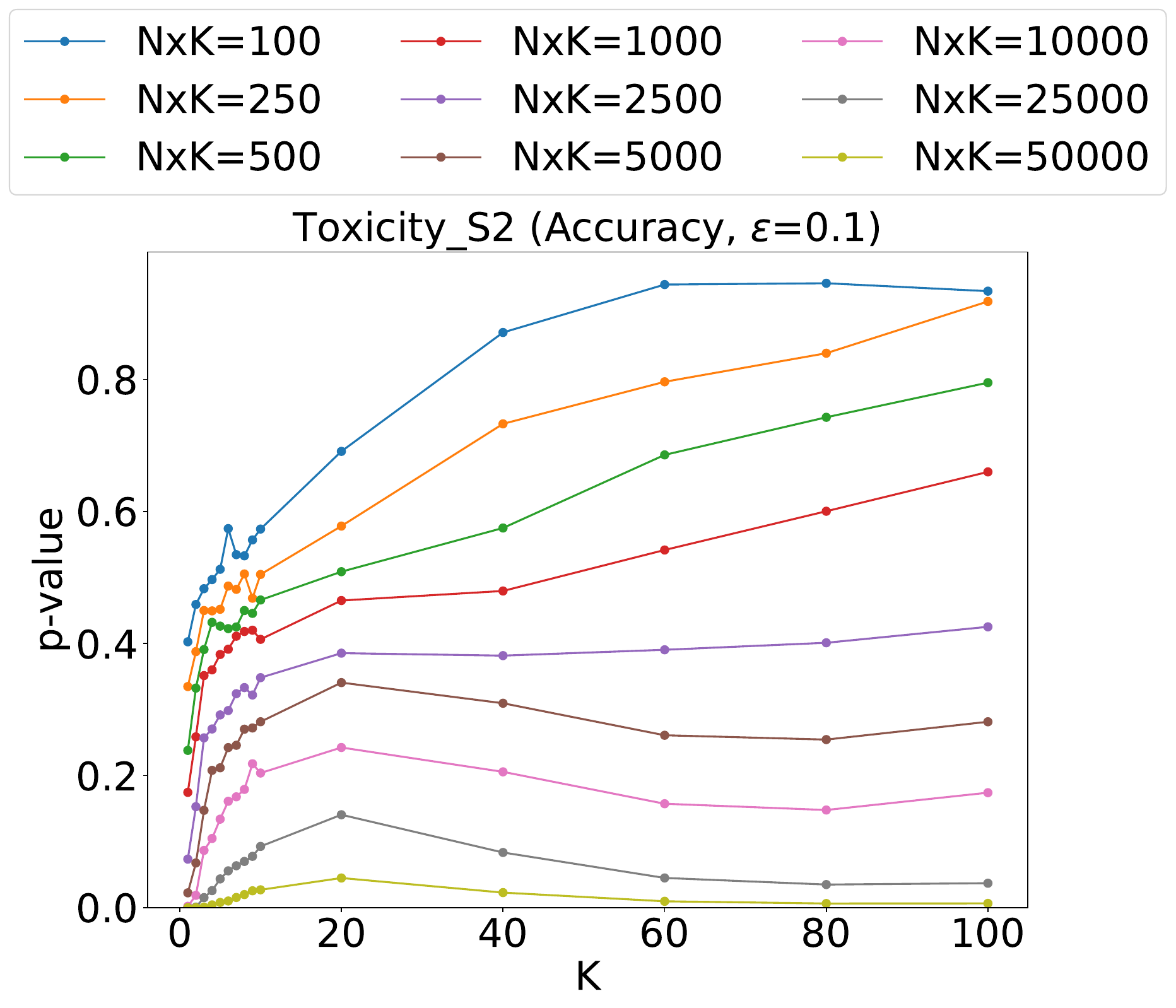}
    \caption{Accuracy, $\epsilon = 0.1$}
    \label{fig:main_toxicity_s2_acc_e01}
  \end{subfigure} \hfill
  \begin{subfigure}[b]{0.32\linewidth}
    \centering
    \includegraphics[width=\linewidth]{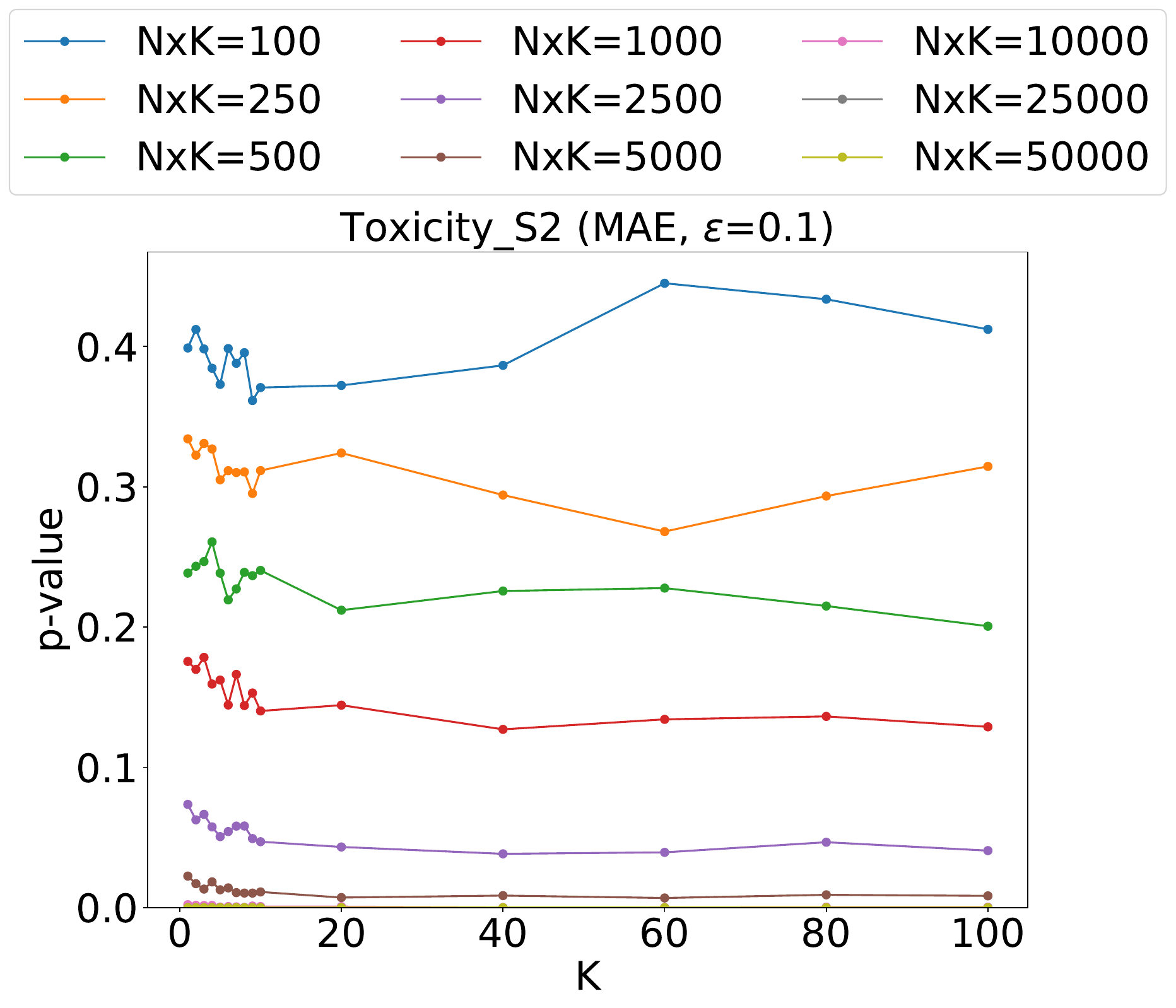}
    \caption{MAE, $\epsilon = 0.1$}
    \label{fig:main_toxicity_s2_mae_e01}
  \end{subfigure} \hfill
  \begin{subfigure}[b]{0.32\linewidth}
    \centering
    \includegraphics[width=\linewidth]{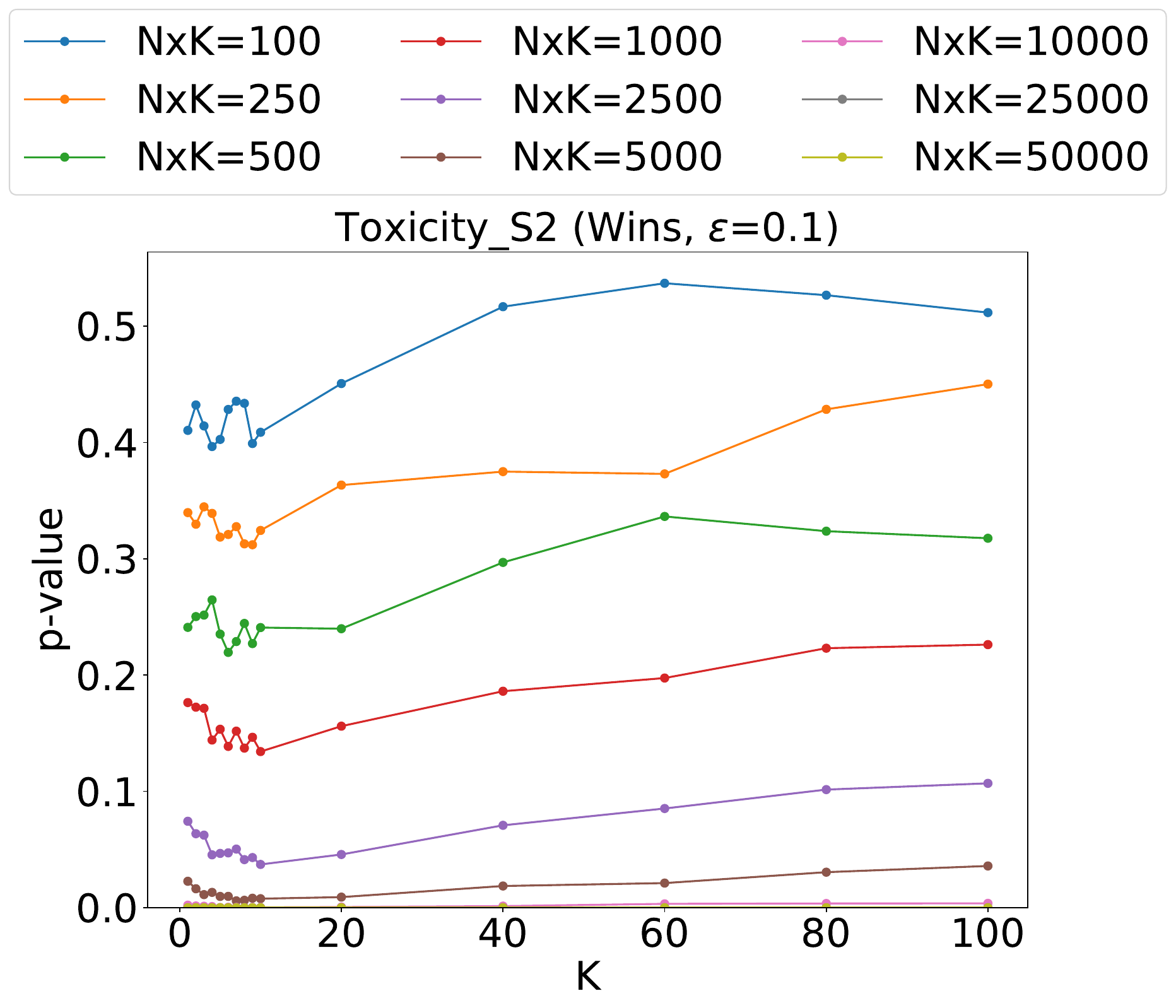}
    \caption{Wins, $\epsilon = 0.1$}
    \label{fig:main_toxicity_s2_wins_e01}
  \end{subfigure} \hfill
  \caption{S2: P-value plots for comparing the DICES 5 rater sample with the Toxicity dataset under different metrics}
  \label{fig:main_dices_5_s2_toxicity}
\end{figure*}

Figure \ref{fig:main_dices_5_s2_toxicity} shows that the \pvs\ across the metrics behave similarly to those of the toxicity dataset. We notice that the \pvs\ across budgets show a separation in both datasets, suggesting that it may be due to the small number of raters in both sets.

\section{Conclusion}

In this work, we introduced a multi-level bootstrapping framework to model rater variance and empirically analyze the tradeoffs between the number of raters ($K$) and items ($N$) required for statistical significance. Our results demonstrate that accounting for rater behavior results in higher \pv\ estimates and that overcoming the rater variance requires scaling up the budgets to obtain significant results. Furthermore, we show that using distribution-sensitive metrics and choosing a higher number of responses per item offers an efficient path to achieve statistical significance. Our method and framework offers a rigorous foundation and a strategic roadmap for budget optimization and repeatable evaluation of AI systems.

\section*{Limitations}
\label{sec:limitations}

A primary constraint of this study is the scarcity of suitable datasets. Our framework requires persistent rater identifiers and high rater-to-item ratios, metadata often omitted from public benchmarks. We hope these results encourage creators to move beyond sparse ``plurality-vote architectures''. Furthermore, while our non-parametric bootstrapping avoids restrictive distributional assumptions, the necessity of oversampling from finite rater pools remains a sub-optimal proxy for truly massive annotation efforts. Our three sampling strategies (S1–S3) were tailored to the structures of the DICES and Toxicity datasets; these may require further adaptation for varying annotation topologies, such as sparse or overlapping graphs. Future work will extend this framework to demographic sub-groups, investigating how $N$ vs $K$ tradeoffs shift when evaluating for fairness across specific cultural and social intersections.
.

% \section*{Acknowledgments}

% Bibliography entries for the entire Anthology, followed by custom entries
%\bibliography{custom,anthology-overleaf-1,anthology-overleaf-2}

% Custom bibliography entries only
\bibliography{custom, anthology-1, anthology-2, references}

\begin{thebibliography}{32}
\providecommand{\natexlab}[1]{#1}

\bibitem[{AAAI(2023)}]{aaai_2023}
AAAI. 2023.
\newblock \href {https://aaai.org/conference/aaai/aaai-23/reproducibility-checklist/} {Reproducibility {Checklist}}.
\newblock Accessed: 2027-08-03.

\bibitem[{ACL(2021)}]{review_acl_2024}
ACL. 2021.
\newblock \href {http://aclrollingreview.org/responsibleNLPresearch/} {{ACL} {Rolling} {Review}}.
\newblock Accessed: 2027-08-03.

\bibitem[{Aroyo et~al.(2023)Aroyo, Taylor, D\'{\i}az, Homan, Parrish, Serapio-Garc\'{\i}a, Prabhakaran, and Wang}]{NEURIPS2023_a74b697b}
Lora Aroyo, Alex Taylor, Mark D\'{\i}az, Christopher Homan, Alicia Parrish, Gregory Serapio-Garc\'{\i}a, Vinodkumar Prabhakaran, and Ding Wang. 2023.
\newblock \href {https://proceedings.neurips.cc/paper_files/paper/2023/file/a74b697bce4cac6c91896372abaa8863-Paper-Datasets_and_Benchmarks.pdf} {Dices dataset: Diversity in conversational ai evaluation for safety}.
\newblock In \emph{Advances in Neural Information Processing Systems}, volume~36, pages 53330--53342, New Orleans, Louisiana, USA. Curran Associates, Inc.

\bibitem[{Baker(2016)}]{baker_1500_2016}
Monya Baker. 2016.
\newblock \href {https://doi.org/10.1038/533452a} {1,500 scientists lift the lid on reproducibility}.
\newblock \emph{Nature}, 533(7604):452--454.

\bibitem[{Barile et~al.(2021)Barile, Najafian, Draws, Inel, Rieger, Hada, and Tintarev}]{barile2021toward}
Francesco Barile, Shabnam Najafian, Tim Draws, Oana Inel, Alisa Rieger, Rishav Hada, and Nava Tintarev. 2021.
\newblock \href {https://ceur-ws.org/Vol-2955/paper11.pdf} {Toward benchmarking group explanations: Evaluating the effect of aggregation strategies versus explanation}.
\newblock In \emph{Perspectives on the Evaluation of Recommender Systems Workshop 2021: co-located with the 15th ACM Conference on Recommender Systems (RecSys 2021)}, Amsterdam, The Netherlands. ACM New York, NY, USA.

\bibitem[{Basile et~al.(2021)Basile, Fell, Fornaciari, Hovy, Paun, Plank, Poesio, and Uma}]{basile-etal-2021-need}
Valerio Basile, Michael Fell, Tommaso Fornaciari, Dirk Hovy, Silviu Paun, Barbara Plank, Massimo Poesio, and Alexandra Uma. 2021.
\newblock \href {https://doi.org/10.18653/v1/2021.bppf-1.3} {We need to consider disagreement in evaluation}.
\newblock In \emph{Proceedings of the 1st Workshop on Benchmarking: Past, Present and Future}, pages 15--21, Online. Association for Computational Linguistics.

\bibitem[{Cabitza et~al.(2023)Cabitza, Campagner, and Basile}]{Cabitza_Campagner_Basile_2023}
Federico Cabitza, Andrea Campagner, and Valerio Basile. 2023.
\newblock \href {https://doi.org/10.1609/aaai.v37i6.25840} {Toward a perspectivist turn in ground truthing for predictive computing}.
\newblock \emph{Proceedings of the AAAI Conference on Artificial Intelligence}, 37(6):6860--6868.

\bibitem[{Campbell et~al.(2025)Campbell, Babb, Lambie, and Hayes}]{campbell2025examination}
Laurie~O Campbell, Kathryn Babb, Glenn~W Lambie, and B~Grant Hayes. 2025.
\newblock An examination of generative ai response to suicide inquires: content analysis.
\newblock \emph{JMIR Mental Health}, 12:e73623.

\bibitem[{Davani et~al.(2024)Davani, Díaz, Baker, and Prabhakaran}]{davani_d3code_2024}
Aida~Mostafazadeh Davani, Mark Díaz, Dylan Baker, and Vinodkumar Prabhakaran. 2024.
\newblock \href {https://doi.org/10.48550/arXiv.2404.10857} {{D3CODE}: {Disentangling} {Disagreements} in {Data} across {Cultures} on {Offensiveness} {Detection} and {Evaluation}}.
\newblock \emph{arXiv preprint}.
\newblock ArXiv:2404.10857 [cs].

\bibitem[{Dodge et~al.(2019)Dodge, Gururangan, Card, Schwartz, and Smith}]{dodge-etal-2019-show}
Jesse Dodge, Suchin Gururangan, Dallas Card, Roy Schwartz, and Noah~A. Smith. 2019.
\newblock \href {https://doi.org/10.18653/v1/D19-1224} {Show your work: Improved reporting of experimental results}.
\newblock In \emph{Proceedings of the 2019 Conference on Empirical Methods in Natural Language Processing and the 9th International Joint Conference on Natural Language Processing (EMNLP-IJCNLP)}, pages 2185--2194, Hong Kong, China. Association for Computational Linguistics.

\bibitem[{Gundersen(2020)}]{Gundersen_2020}
Odd~Erik Gundersen. 2020.
\newblock \href {https://doi.org/10.1609/aimag.v41i3.5318} {The reproducibility crisis is real}.
\newblock \emph{AI Magazine}, 41(3):103--106.

\bibitem[{Gundersen and Kjensmo(2018)}]{Gundersen_Kjensmo_2018}
Odd~Erik Gundersen and Sigbjørn Kjensmo. 2018.
\newblock \href {https://doi.org/10.1609/aaai.v32i1.11503} {State of the art: Reproducibility in artificial intelligence}.
\newblock \emph{Proceedings of the AAAI Conference on Artificial Intelligence}, 32(1):1644--1651.

\bibitem[{Homan et~al.(2026)Homan, Korn, Pandita, and Welty}]{homan-etal-2026-many}
Christopher~M. Homan, Flip Korn, Deepak Pandita, and Chris Welty. 2026.
\newblock \href {https://doi.org/10.18653/v1/2026.findings-eacl.223} {How many ratings per item are necessary for reliable significance testing?}
\newblock In \emph{Findings of the {A}ssociation for {C}omputational {L}inguistics: {EACL} 2026}, pages 4258--4273, Rabat, Morocco. Association for Computational Linguistics.

\bibitem[{Hutson(2018)}]{hutson_2018}
Matthew Hutson. 2018.
\newblock \href {https://doi.org/10.1126/science.359.6377.725} {Artificial intelligence faces reproducibility crisis}.
\newblock \emph{Science}, 359(6377):725--726.

\bibitem[{ICML(2023)}]{icml_2023}
ICML. 2023.
\newblock \href {https://icml.cc/Conferences/2023/PaperGuidelines} {{ICML} 2023 {Paper} {Guidelines}}.
\newblock Accessed: 2027-08-03.

\bibitem[{IJCAI(2022)}]{ijcai_2022}
IJCAI. 2022.
\newblock \href {https://ijcai-22.org/reproducibility/} {Reproducibility {Guidelines} – {IJCAI}-{ECAI} 2022}.
\newblock Accessed: 2027-08-03.

\bibitem[{Kapoor and Narayanan(2022)}]{kapoor2022leakagereproducibilitycrisismlbased}
Sayash Kapoor and Arvind Narayanan. 2022.
\newblock \href {https://arxiv.org/abs/2207.07048} {Leakage and the reproducibility crisis in ml-based science}.
\newblock \emph{Preprint}, arXiv:2207.07048.

\bibitem[{Kumar et~al.(2021)Kumar, Kelley, Consolvo, Mason, Bursztein, Durumeric, Thomas, and Bailey}]{kumar2021designing}
Deepak Kumar, Patrick~Gage Kelley, Sunny Consolvo, Joshua Mason, Elie Bursztein, Zakir Durumeric, Kurt Thomas, and Michael Bailey. 2021.
\newblock \href {https://www.usenix.org/conference/soups2021/presentation/kumar} {Designing toxic content classification for a diversity of perspectives}.
\newblock In \emph{Seventeenth Symposium on Usable Privacy and Security (SOUPS 2021)}, pages 299--318.

\bibitem[{Mieskes et~al.(2019)Mieskes, Fort, N{\'e}v{\'e}ol, Grouin, and Cohen}]{mieskes-etal-2019-community}
Margot Mieskes, Kar{\"e}n Fort, Aur{\'e}lie N{\'e}v{\'e}ol, Cyril Grouin, and Kevin Cohen. 2019.
\newblock \href {https://doi.org/10.26615/978-954-452-056-4_089} {Community perspective on replicability in natural language processing}.
\newblock In \emph{Proceedings of the International Conference on Recent Advances in Natural Language Processing (RANLP 2019)}, pages 768--775, Varna, Bulgaria. INCOMA Ltd.

\bibitem[{Mostafazadeh~Davani et~al.(2022)Mostafazadeh~Davani, D{\'i}az, and Prabhakaran}]{davani-etal-2022-dealing}
Aida Mostafazadeh~Davani, Mark D{\'i}az, and Vinodkumar Prabhakaran. 2022.
\newblock \href {https://doi.org/10.1162/tacl_a_00449} {Dealing with disagreements: Looking beyond the majority vote in subjective annotations}.
\newblock \emph{Transactions of the Association for Computational Linguistics}, 10:92--110.

\bibitem[{NeurIPS(2021)}]{neurips_2021}
NeurIPS. 2021.
\newblock \href {https://neurips.cc/Conferences/2021/PaperInformation/PaperChecklist} {{PaperChecklist}}.
\newblock Accessed: 2027-08-03.

\bibitem[{Pandita et~al.(2026)Pandita, Korn, Welty, and Homan}]{Pandita_Korn_Welty_Homan_2026}
Deepak Pandita, Flip Korn, Chris Welty, and Christopher~M Homan. 2026.
\newblock \href {https://doi.org/10.1609/aaai.v40i29.39659} {Forest vs tree: The (n, k) trade-off in reproducible ml evaluation}.
\newblock \emph{Proceedings of the AAAI Conference on Artificial Intelligence}, 40(29):24736--24744.

\bibitem[{Pandita et~al.(2024)Pandita, Weerasooriya, Dutta, Luger, Ranasinghe, KhudaBukhsh, Zampieri, and Homan}]{pandita-etal-2024-rater}
Deepak Pandita, Tharindu~Cyril Weerasooriya, Sujan Dutta, Sarah~K. Luger, Tharindu Ranasinghe, Ashiqur~R. KhudaBukhsh, Marcos Zampieri, and Christopher~M. Homan. 2024.
\newblock \href {https://doi.org/10.18653/v1/2024.findings-emnlp.296} {Rater cohesion and quality from a vicarious perspective}.
\newblock In \emph{Findings of the Association for Computational Linguistics: EMNLP 2024}, pages 5149--5162, Miami, Florida, USA. Association for Computational Linguistics.

\bibitem[{Pham et~al.(2020)Pham, Qian, Wang, Lutellier, Rosenthal, Tan, Yu, and Nagappan}]{pham_problems_2020}
Hung~Viet Pham, Shangshu Qian, Jiannan Wang, Thibaud Lutellier, Jonathan Rosenthal, Lin Tan, Yaoliang Yu, and Nachiappan Nagappan. 2020.
\newblock \href {https://ieeexplore.ieee.org/abstract/document/9286042} {Problems and {Opportunities} in {Training} {Deep} {Learning} {Software} {Systems}: {An} {Analysis} of {Variance}}.
\newblock In \emph{2020 35th {IEEE}/{ACM} {International} {Conference} on {Automated} {Software} {Engineering} ({ASE})}, pages 771--783.

\bibitem[{Prabhakaran et~al.(2021)Prabhakaran, Mostafazadeh~Davani, and Diaz}]{prabhakaran-etal-2021-releasing}
Vinodkumar Prabhakaran, Aida Mostafazadeh~Davani, and Mark Diaz. 2021.
\newblock \href {https://doi.org/10.18653/v1/2021.law-1.14} {On releasing annotator-level labels and information in datasets}.
\newblock In \emph{Proceedings of the Joint 15th Linguistic Annotation Workshop (LAW) and 3rd Designing Meaning Representations (DMR) Workshop}, pages 133--138, Punta Cana, Dominican Republic. Association for Computational Linguistics.

\bibitem[{Raff(2019)}]{NEURIPS2019_c429429b}
Edward Raff. 2019.
\newblock \href {https://proceedings.neurips.cc/paper_files/paper/2019/file/c429429bf1f2af051f2021dc92a8ebea-Paper.pdf} {A step toward quantifying independently reproducible machine learning research}.
\newblock In \emph{Advances in Neural Information Processing Systems}, volume~32, Vancouver, Canada. Curran Associates, Inc.

\bibitem[{Rizzi et~al.(2024)Rizzi, Leonardelli, Poesio, Uma, Pavlovic, Paun, Rosso, and Fersini}]{rizzi_soft_2024}
Giulia Rizzi, Elisa Leonardelli, Massimo Poesio, Alexandra Uma, Maja Pavlovic, Silviu Paun, Paolo Rosso, and Elisabetta Fersini. 2024.
\newblock \href {https://aclanthology.org/2024.nlperspectives-1.9/} {Soft metrics for evaluation with disagreements: an assessment}.
\newblock In \emph{Proceedings of the 3rd {Workshop} on {Perspectivist} {Approaches} to {NLP} ({NLPerspectives}) @ {LREC}-{COLING} 2024}, pages 84--94, Torino, Italia. ELRA and ICCL.

\bibitem[{Rogers et~al.(2021)Rogers, Baldwin, and Leins}]{rogers-etal-2021-just-think}
Anna Rogers, Timothy Baldwin, and Kobi Leins. 2021.
\newblock \href {https://doi.org/10.18653/v1/2021.findings-emnlp.414} {`just what do you think you{'}re doing, dave?' a checklist for responsible data use in {NLP}}.
\newblock In \emph{Findings of the Association for Computational Linguistics: EMNLP 2021}, pages 4821--4833, Punta Cana, Dominican Republic. Association for Computational Linguistics.

\bibitem[{Semmelrock et~al.(2025)Semmelrock, Ross-Hellauer, Kopeinik, Theiler, Haberl, Thalmann, and Kowald}]{semmelrock2024reproducibility}
Harald Semmelrock, Tony Ross-Hellauer, Simone Kopeinik, Dieter Theiler, Maximilian Haberl, Stefan Thalmann, and Dominik Kowald. 2025.
\newblock Reproducibility in machine-learning-based research: Overview, barriers, and drivers.
\newblock \emph{AI Magazine}, 46:e70002.
\newblock \url{https://doi.org/10.1002/aaai.70002}.

\bibitem[{Snow et~al.(2008)Snow, O{'}Connor, Jurafsky, and Ng}]{snow-etal-2008-cheap}
Rion Snow, Brendan O{'}Connor, Daniel Jurafsky, and Andrew Ng. 2008.
\newblock \href {https://aclanthology.org/D08-1027/} {Cheap and fast {--} but is it good? evaluating non-expert annotations for natural language tasks}.
\newblock In \emph{Proceedings of the 2008 Conference on Empirical Methods in Natural Language Processing}, pages 254--263, Honolulu, Hawaii. Association for Computational Linguistics.

\bibitem[{Weerasooriya et~al.(2023)Weerasooriya, Ororbia, Bhensadadia, KhudaBukhsh, and Homan}]{weerasooriya-etal-2023-disagreement}
Tharindu~Cyril Weerasooriya, Alexander Ororbia, Raj Bhensadadia, Ashiqur KhudaBukhsh, and Christopher~M. Homan. 2023.
\newblock \href {https://doi.org/10.18653/v1/2023.findings-acl.287} {Disagreement matters: Preserving label diversity by jointly modeling item and annotator label distributions with {D}is{C}o}.
\newblock In \emph{Findings of the Association for Computational Linguistics: ACL 2023}, pages 4679--4695, Toronto, Canada. Association for Computational Linguistics.

\bibitem[{Wein et~al.(2023)Wein, Homan, Aroyo, and Welty}]{wein-etal-2023-follow}
Shira Wein, Christopher~M. Homan, Lora Aroyo, and Chris Welty. 2023.
\newblock \href {https://doi.org/10.18653/v1/2023.findings-acl.196} {Follow the leader(board) with confidence: Estimating p-values from a single test set with item and response variance}.
\newblock In \emph{Findings of the Association for Computational Linguistics: ACL 2023}, pages 3138--3161, Toronto, Canada. Association for Computational Linguistics.

\end{thebibliography}

\appendix

\section{Example Appendix}
\label{sec:appendix}

\begin{table*}
\centering
% \small
\begin{tabular}{l|c|cccccccc}
% \toprule
Dataset & Stat & Accuracy & MAE & Wins & Precision & Recall & F1-Score & KL-Div & JSD \\
\midrule
 & NK & NaN & 2500 & 5000 & NaN & NaN & 50000 & 25000 & 2500 \\
DICES:S1 & \pv\ & NaN & 0.039 & 0.042 & NaN & NaN & 0.032 & 0.037 & 0.048 \\
 & K & NaN & 100 & 100 & NaN & NaN & 100 & 100 & 100 \\
 & $\Delta$ & NaN & 0.057 & 0.663 & NaN & NaN & 0.171 & 0.068 & 0.075 \\
 \hline
 & NK & 25000 & 1000 & 2500 & 50000 & 10000 & 5000 & 10000 & 2500 \\
DICES:S2 & \pv\ & 0.017 & 0.046 & 0.024 & 0.042 & 0.049 & 0.045 & 0.020 & 0.027 \\
 & K & 20 & 80 & 100 & 1 & 1 & 1 & 100 & 80 \\
 & $\Delta$ & 0.057 & 0.056 & 0.655 & 0.010 & 0.014 & 0.024 & 0.066 & 0.071 \\
 \hline
 & NK & 25000 & 2500 & 2500 & 25000 & 10000 & 5000 & 2500 & 2500 \\
D3code:S2 & \pv\ & 0.012 & 0.030 & 0.048 & 0.038 & 0.039 & 0.047 & 0.045 & 0.031 \\
 & K & 100 & 100 & 100 & 1 & 1 & 1 & 100 & 100 \\
 & $\Delta$ & 0.092 & 0.059 & 0.505 & 0.015 & 0.016 & 0.023 & 0.032 & 0.047 \\
 \hline
 & NK & 1000 & 1000 & 1000 & 2500 & 1000 & 1000 & 1000 & 500 \\
Toxicity:S2 & \pv\ & 0.038 & 0.017 & 0.026 & 0.009 & 0.035 & 0.026 & 0.023 & 0.040 \\
 & K & 1 & 8 & 3 & 1 & 1 & 1 & 40 & 20 \\
 & $\Delta$ & 0.042 & 0.043 & 0.092 & 0.035 & 0.041 & 0.045 & 0.052 & 0.054 \\
 \hline
 & NK & 5000 & 2500 & 2500 & 5000 & 5000 & 2500 & 2500 & 2500 \\
Toxicity:S3 & \pv\ & 0.019 & 0.029 & 0.036 & 0.041 & 0.015 & 0.043 & 0.033 & 0.014 \\
 & K & 1 & 10 & 10 & 1 & 1 & 1 & 40 & 10 \\
 & $\Delta$ & 0.041 & 0.045 & 0.194 & 0.034 & 0.041 & 0.045 & 0.051 & 0.047 \\
\bottomrule
\end{tabular}
\caption{Minimum \pv, $K$, and corresponding effect size ($\Delta$) for lowest $NK$ with $p<0.05$ ($ \epsilon=0.2$).}
\label{tab:low_k_for_p_lte_05_nk_min_p_e_0.2}
\end{table*}

\subsection{S1}

\paragraph{DICES}

\begin{figure*}
  \centering
  \begin{subfigure}[b]{0.24\linewidth}
    \centering
    \includegraphics[width=\linewidth]{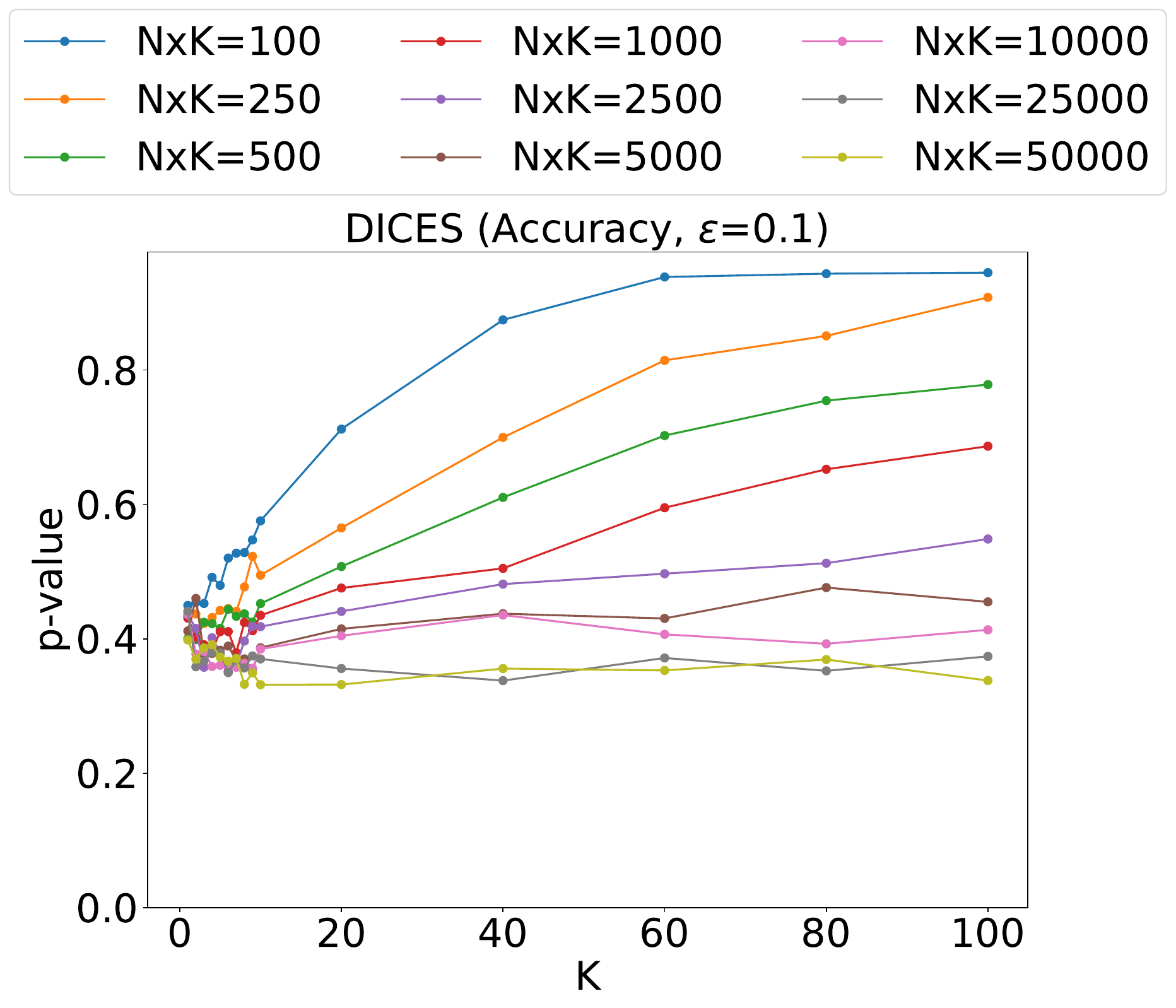}
    \caption{$\epsilon = 0.1$}
    \label{fig:dices_acc_e01}
  \end{subfigure} \hfill
  \begin{subfigure}[b]{0.24\linewidth}
    \centering
    \includegraphics[width=\linewidth]{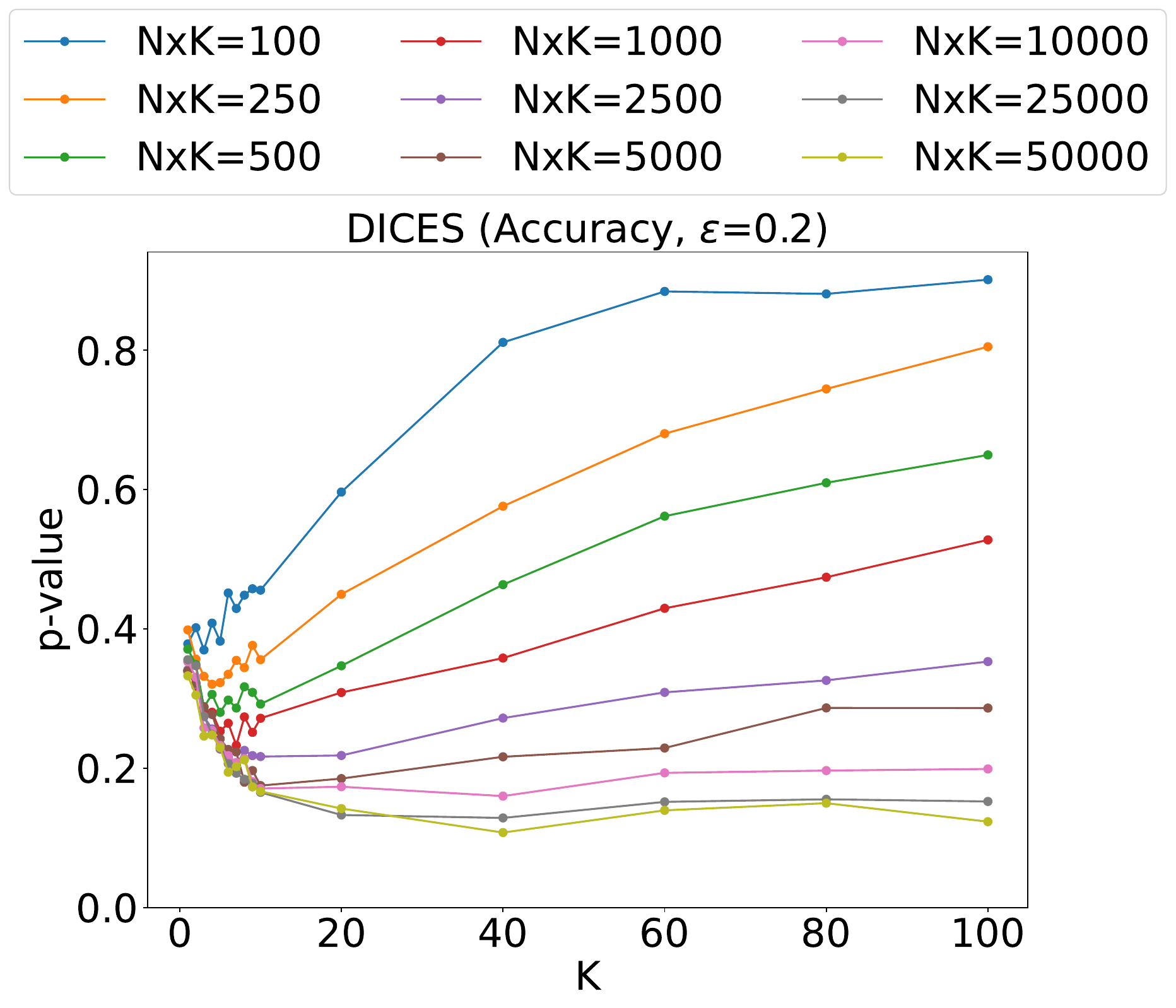}
    \caption{$\epsilon = 0.2$}
    \label{fig:dices_acc_e02}
  \end{subfigure} \hfill
  \begin{subfigure}[b]{0.24\linewidth}
    \centering
    \includegraphics[width=\linewidth]{figures/pvals_plots/DICES/DICES_p_vals_Accuracy_K_100_e_0.3.pdf}
    \caption{$\epsilon = 0.3$}
    \label{fig:dices_acc_e03}
  \end{subfigure} \hfill
  \begin{subfigure}[b]{0.24\linewidth}
    \centering
    \includegraphics[width=\linewidth]{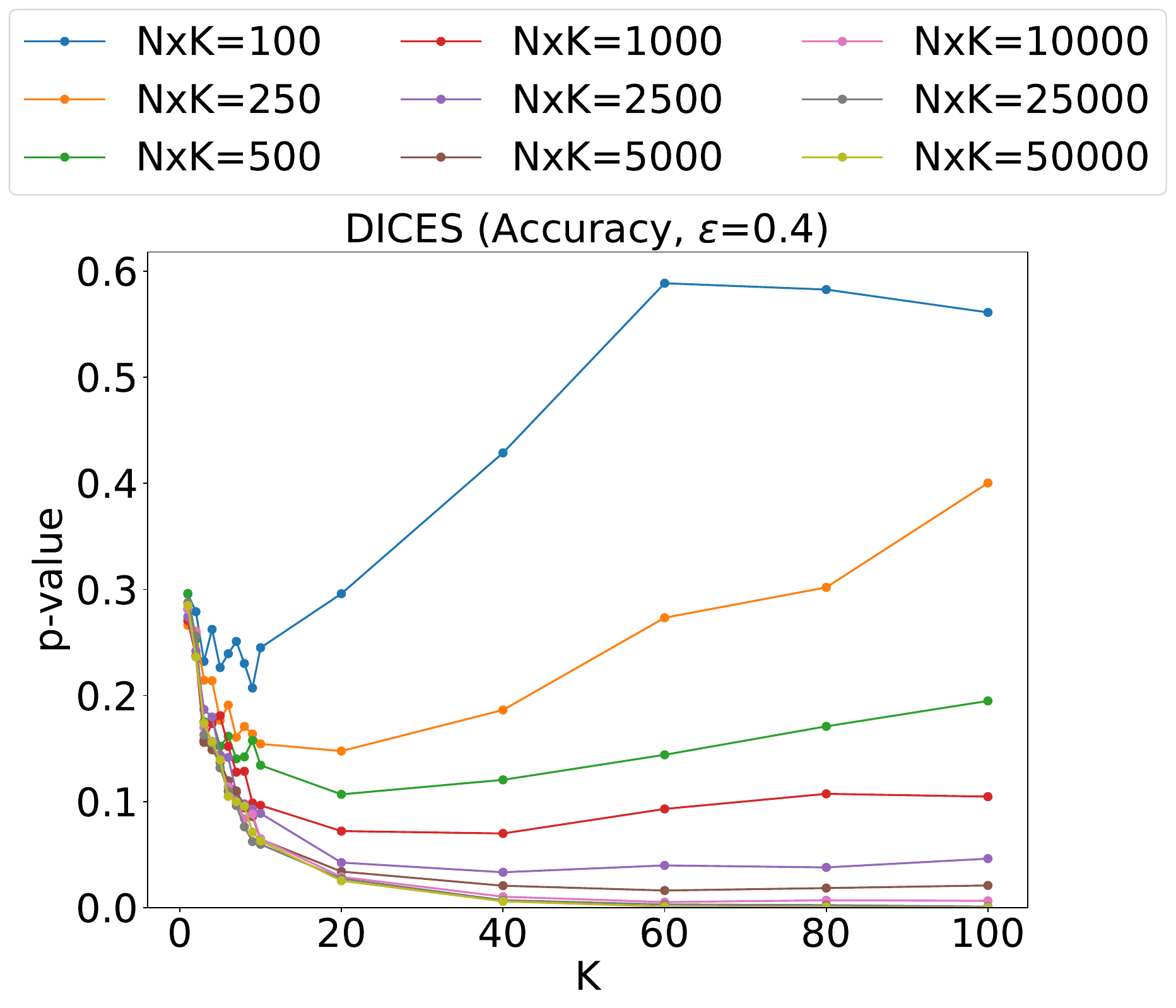}
    \caption{$\epsilon = 0.4$}
    \label{fig:dices_acc_e04}
  \end{subfigure}
  \caption{S1: P-value plots for DICES dataset with Accuracy as the metric}
  \label{fig:dices_accuracy}
\end{figure*}

\begin{figure*}
  \centering
  \begin{subfigure}[b]{0.24\linewidth}
    \centering
    \includegraphics[width=\linewidth]{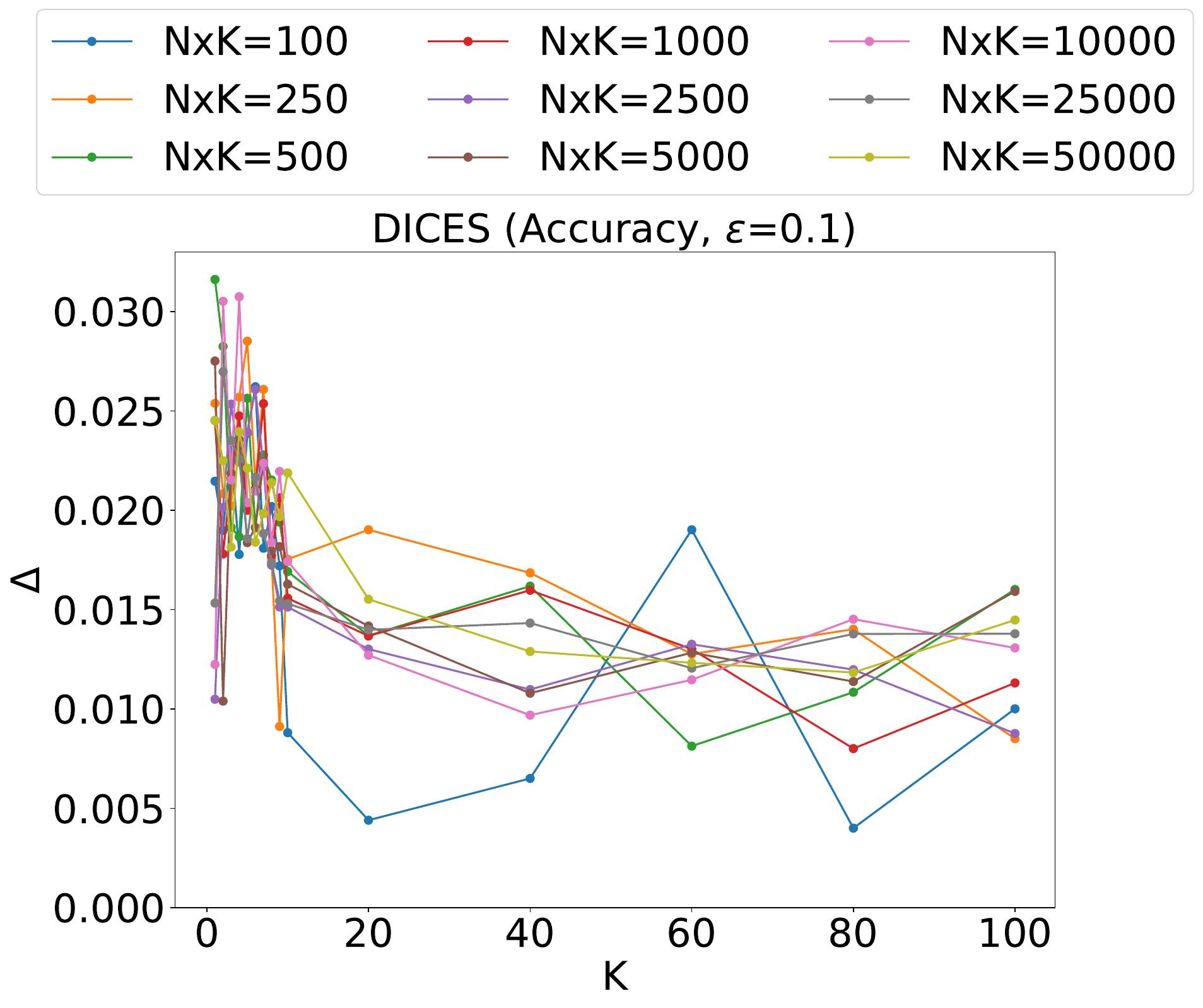}
    \caption{$\epsilon = 0.1$}
    \label{fig:dices_delta_acc_e01}
  \end{subfigure} \hfill
  \begin{subfigure}[b]{0.24\linewidth}
    \centering
    \includegraphics[width=\linewidth]{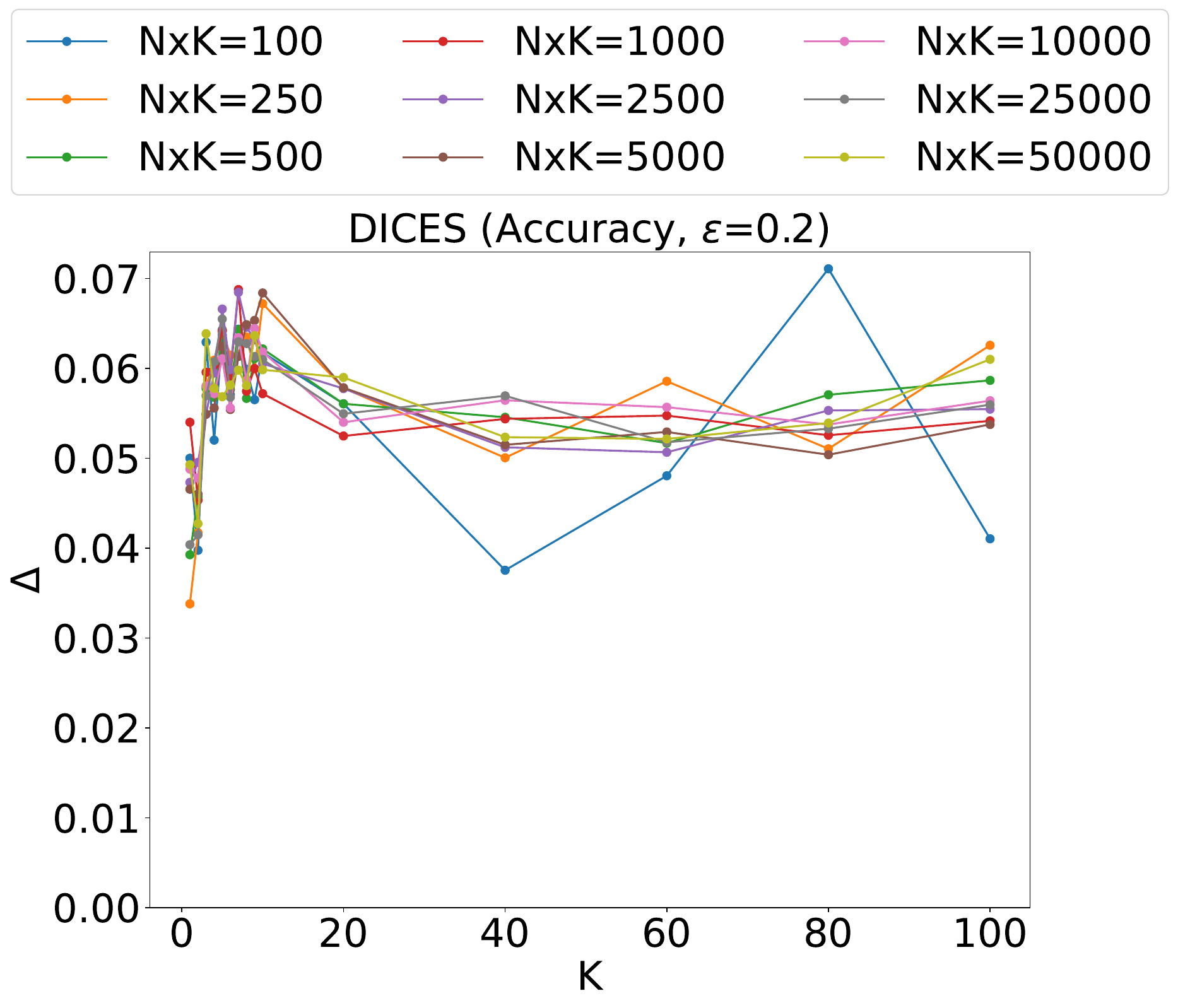}
    \caption{$\epsilon = 0.2$}
    \label{fig:dices_delta_acc_e02}
  \end{subfigure} \hfill
  \begin{subfigure}[b]{0.24\linewidth}
    \centering
    \includegraphics[width=\linewidth]{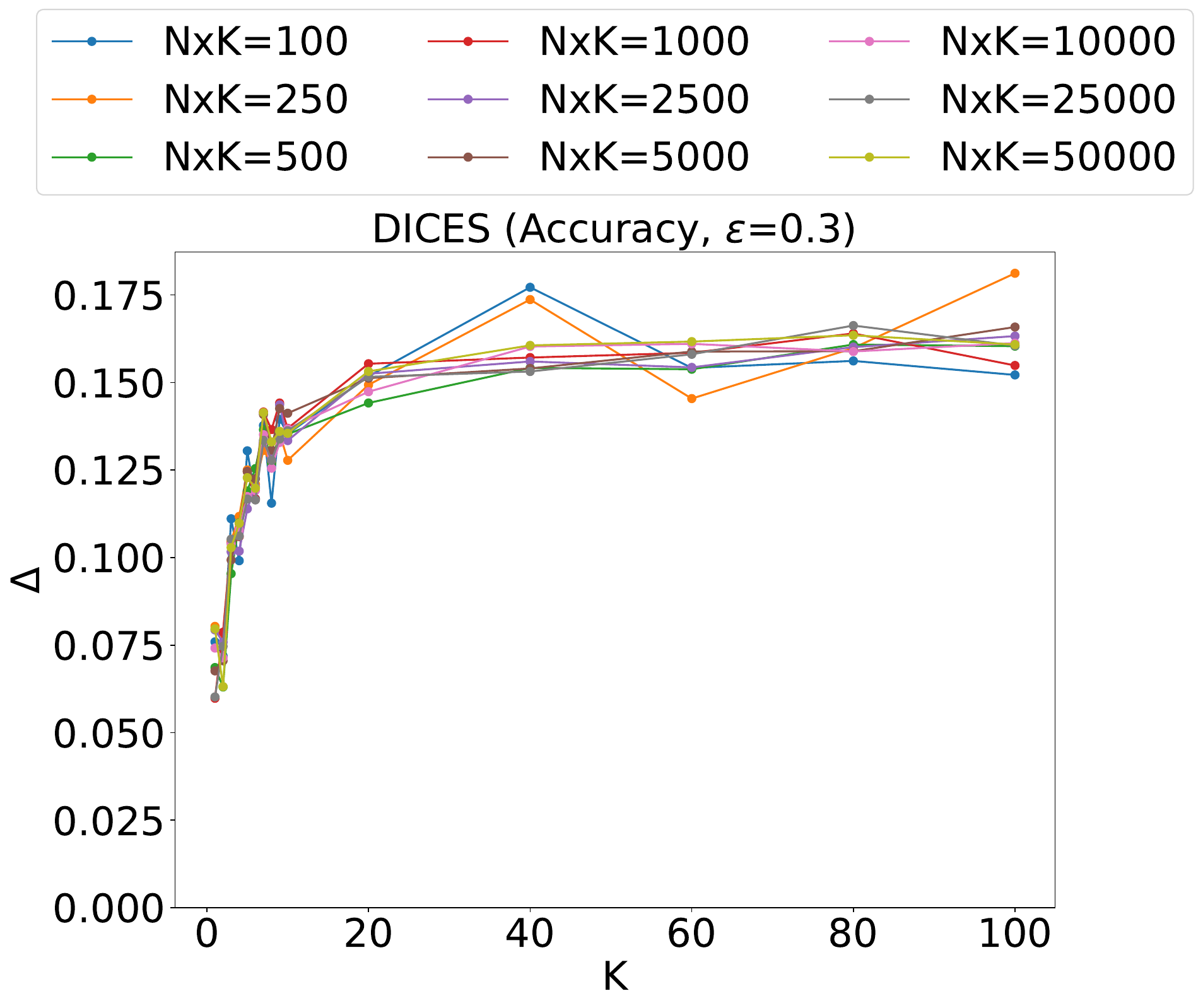}
    \caption{$\epsilon = 0.3$}
    \label{fig:dices_delta_acc_e03}
  \end{subfigure} \hfill
  \begin{subfigure}[b]{0.24\linewidth}
    \centering
    \includegraphics[width=\linewidth]{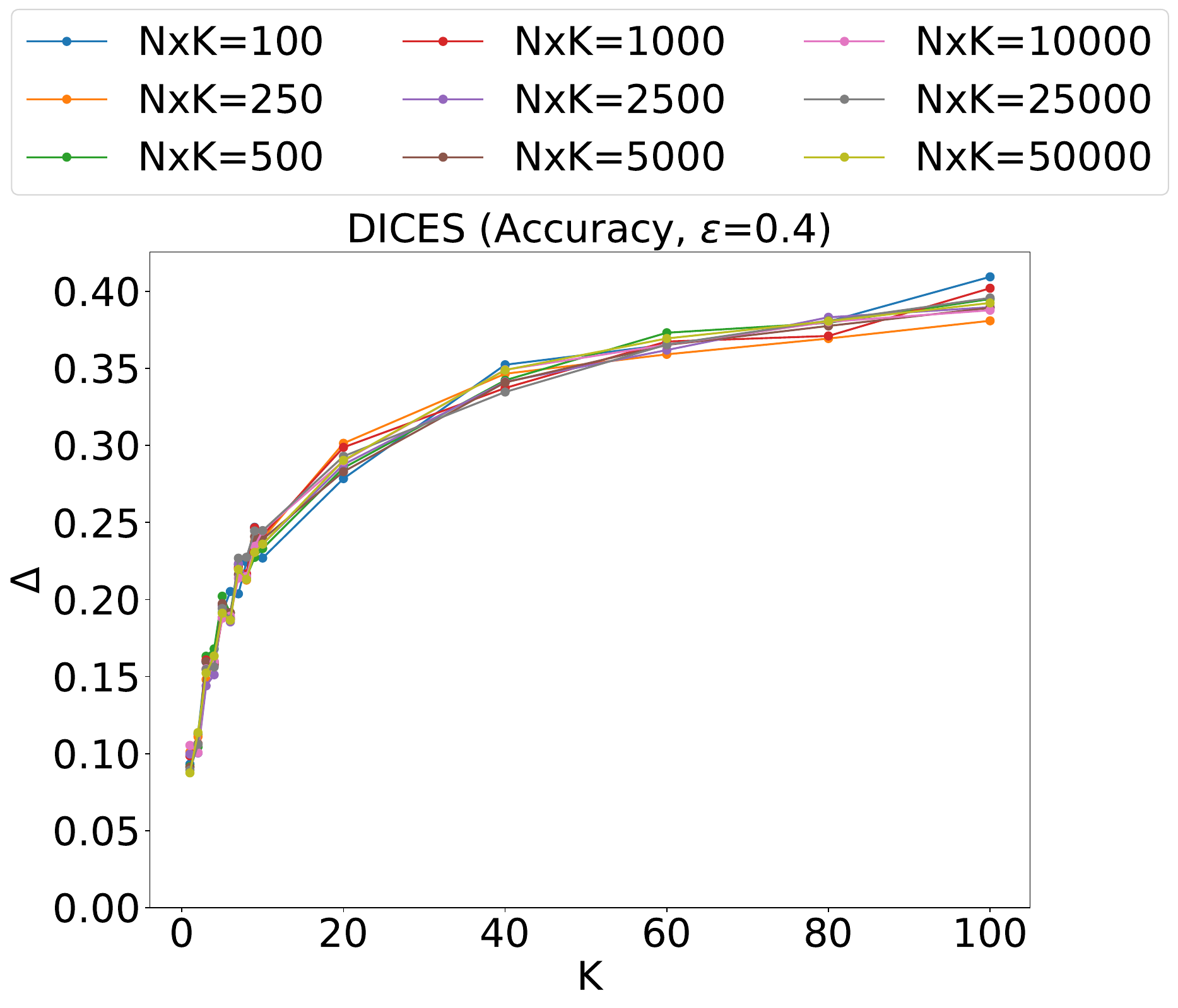}
    \caption{$\epsilon = 0.4$}
    \label{fig:dices_delta_acc_e04}
  \end{subfigure}
  \caption{S1: Effect sizes ($\Delta$) for DICES dataset with Accuracy as the metric}
  \label{fig:dices_delta_accuracy}
\end{figure*}

\begin{figure*}
  \centering
  \begin{subfigure}[b]{0.24\linewidth}
    \centering
    \includegraphics[width=\linewidth]{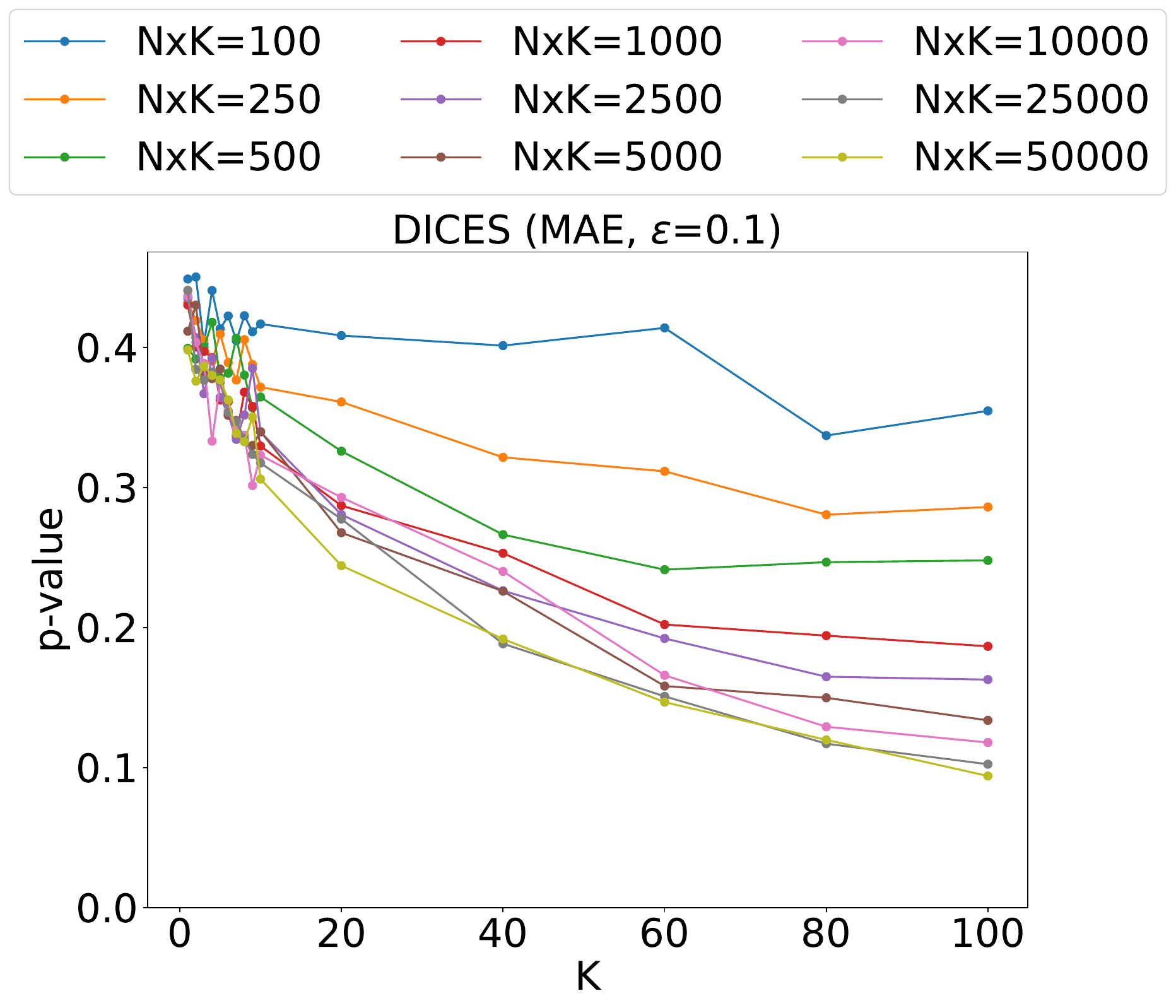}
    \caption{$\epsilon = 0.1$}
    \label{fig:dices_MAE_e01}
  \end{subfigure} \hfill
  \begin{subfigure}[b]{0.24\linewidth}
    \centering
    \includegraphics[width=\linewidth]{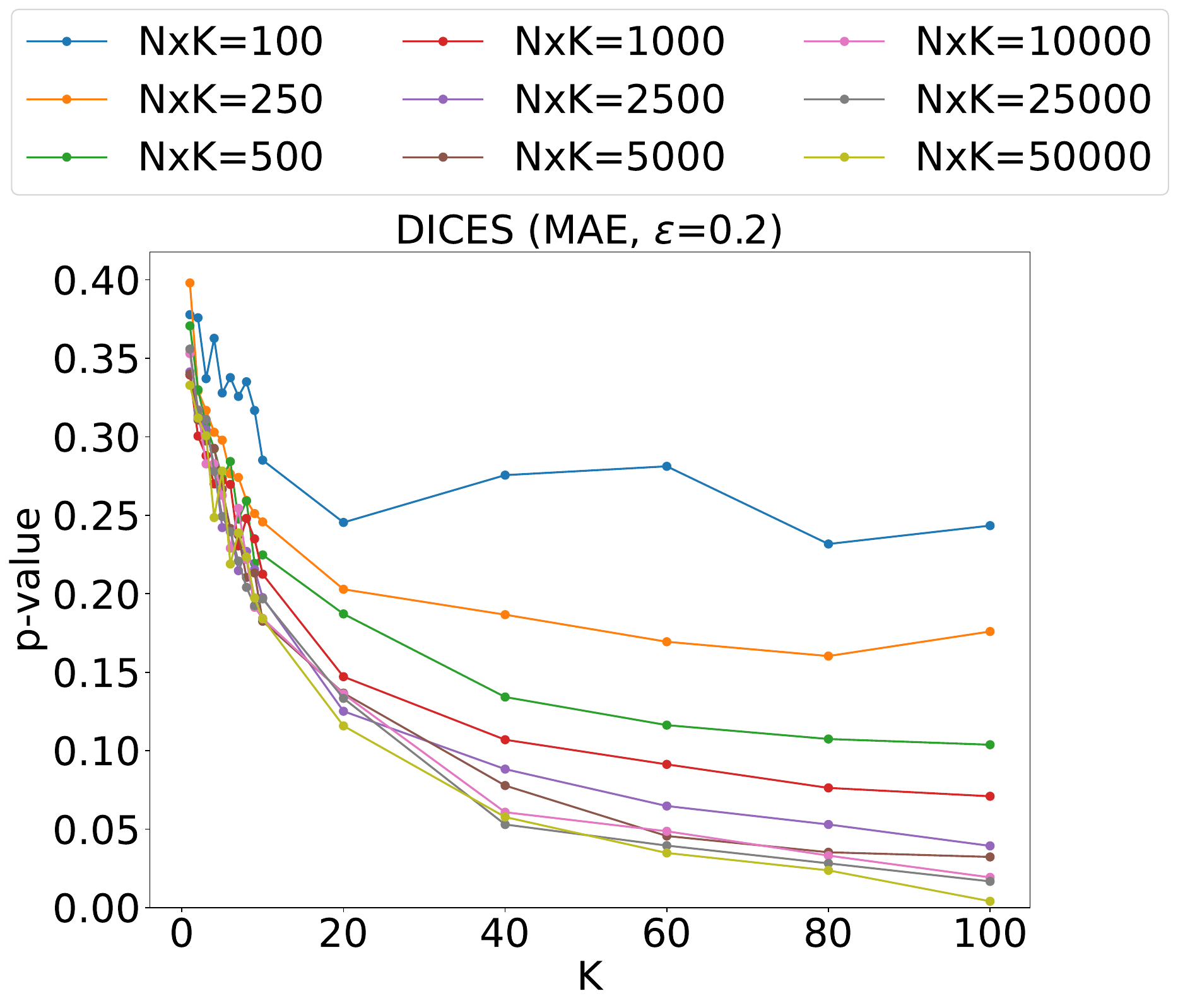}
    \caption{$\epsilon = 0.2$}
    \label{fig:dices_MAE_e02}
  \end{subfigure} \hfill
  \begin{subfigure}[b]{0.24\linewidth}
    \centering
    \includegraphics[width=\linewidth]{figures/pvals_plots/DICES/DICES_p_vals_MAE_K_100_e_0.3.pdf}
    \caption{$\epsilon = 0.3$}
    \label{fig:dices_MAE_e03}
  \end{subfigure} \hfill
  \begin{subfigure}[b]{0.24\linewidth}
    \centering
    \includegraphics[width=\linewidth]{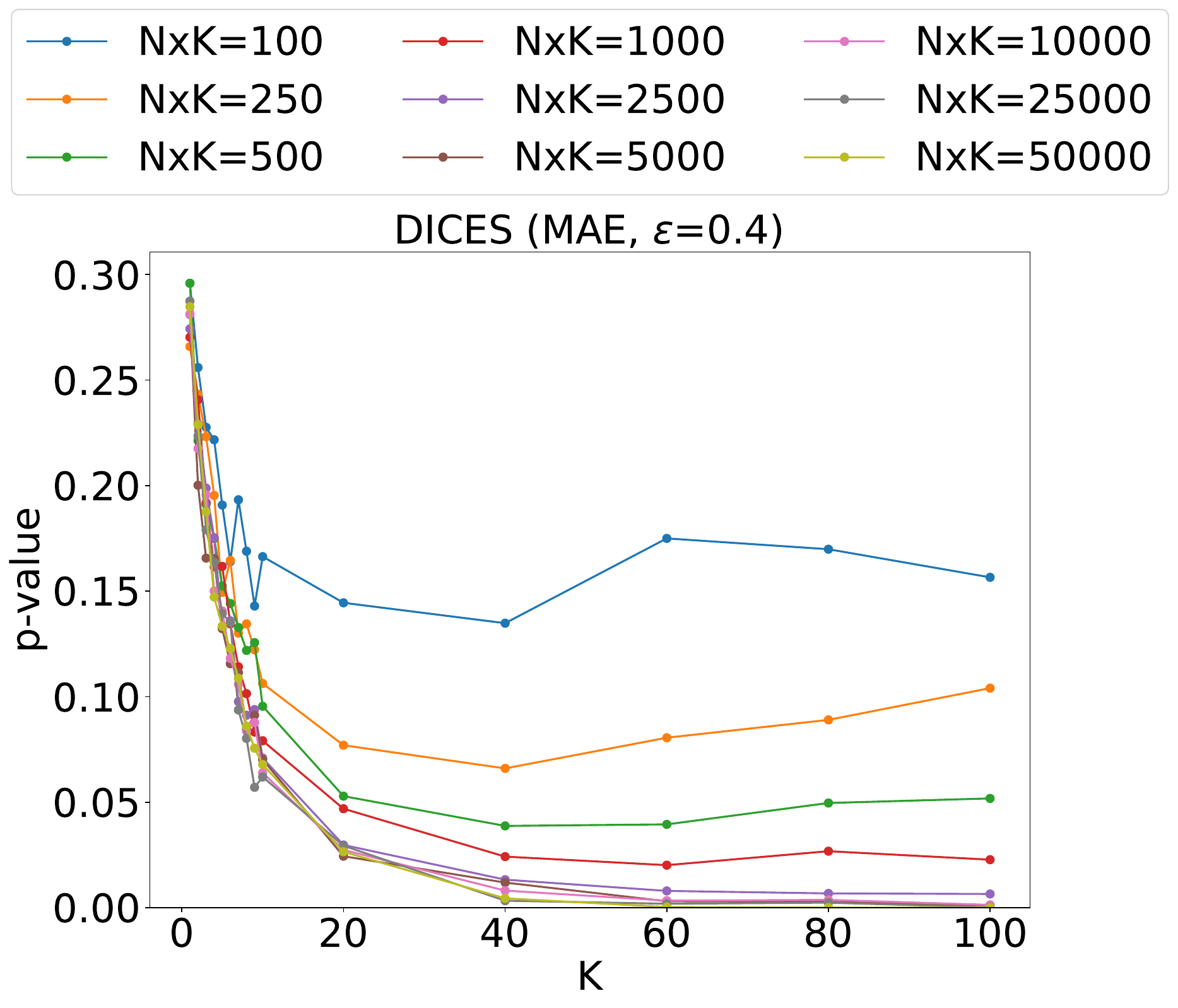}
    \caption{$\epsilon = 0.4$}
    \label{fig:dices_MAE_e04}
  \end{subfigure}
  \caption{S1: P-value plots for DICES dataset with MAE as the metric}
  \label{fig:dices_MAE}
\end{figure*}

\begin{figure*}
  \centering
  \begin{subfigure}[b]{0.24\linewidth}
    \centering
    \includegraphics[width=\linewidth]{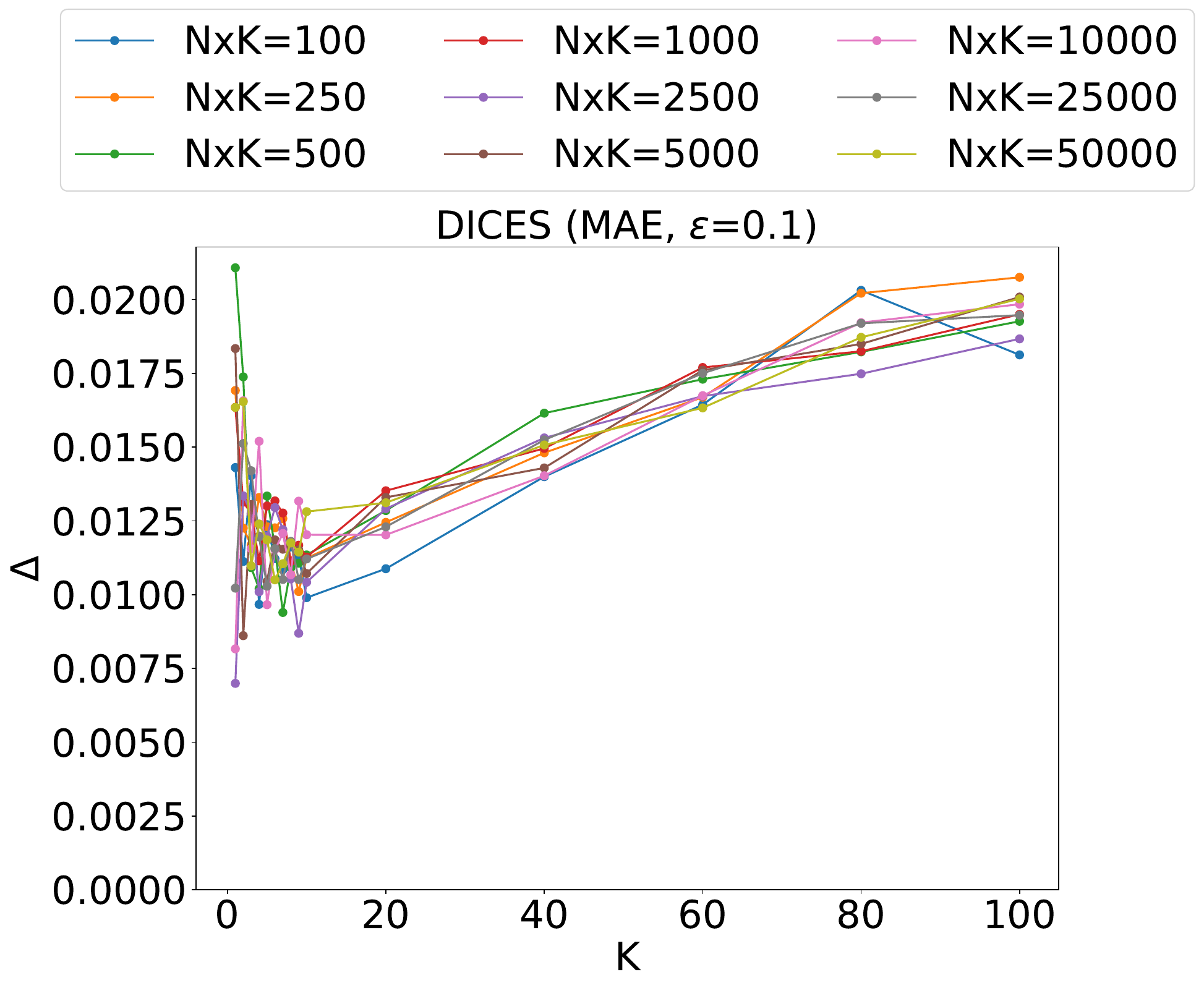}
    \caption{$\epsilon = 0.1$}
    \label{fig:dices_delta_MAE_e01}
  \end{subfigure} \hfill
  \begin{subfigure}[b]{0.24\linewidth}
    \centering
    \includegraphics[width=\linewidth]{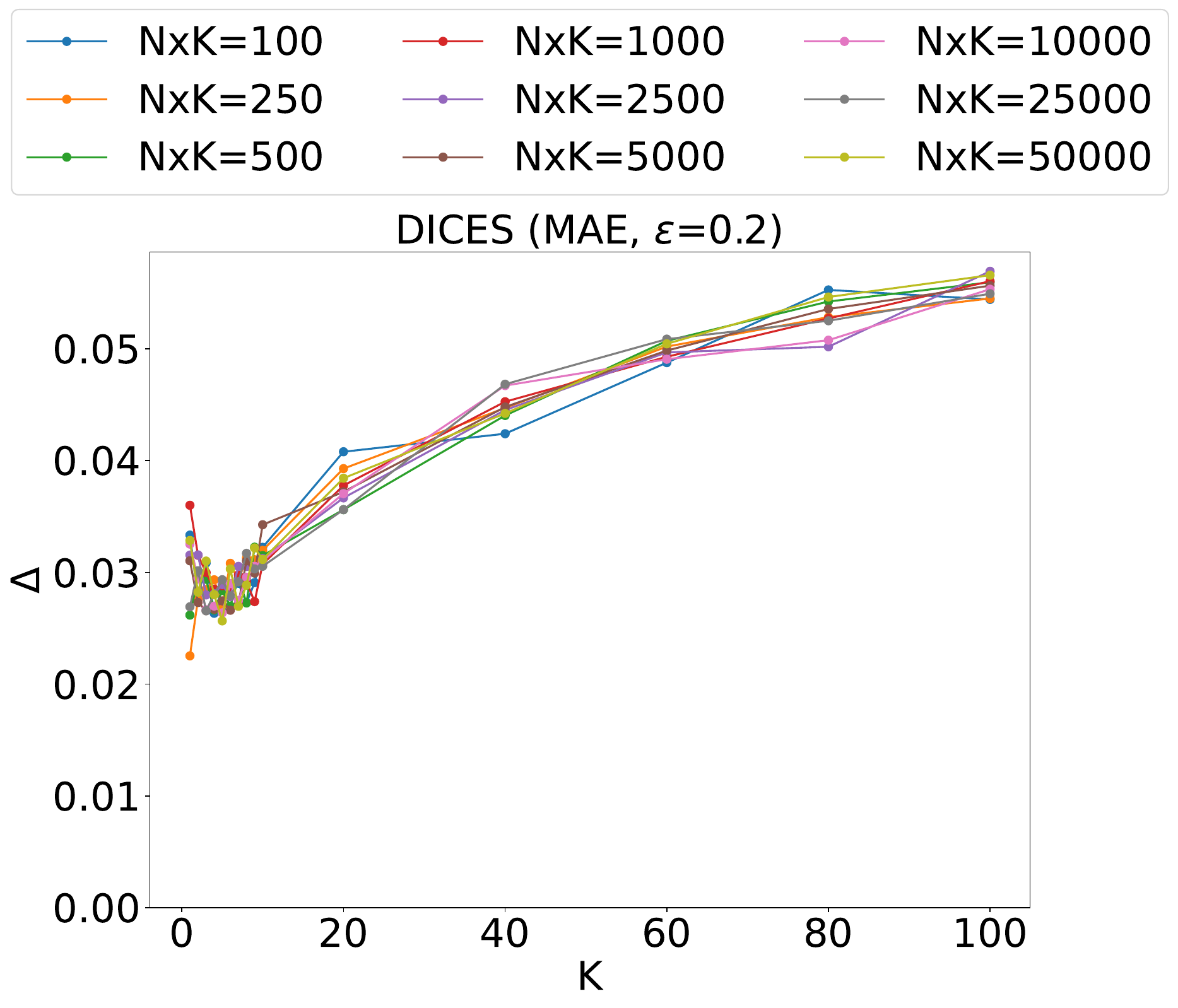}
    \caption{$\epsilon = 0.2$}
    \label{fig:dices_delta_MAE_e02}
  \end{subfigure} \hfill
  \begin{subfigure}[b]{0.24\linewidth}
    \centering
    \includegraphics[width=\linewidth]{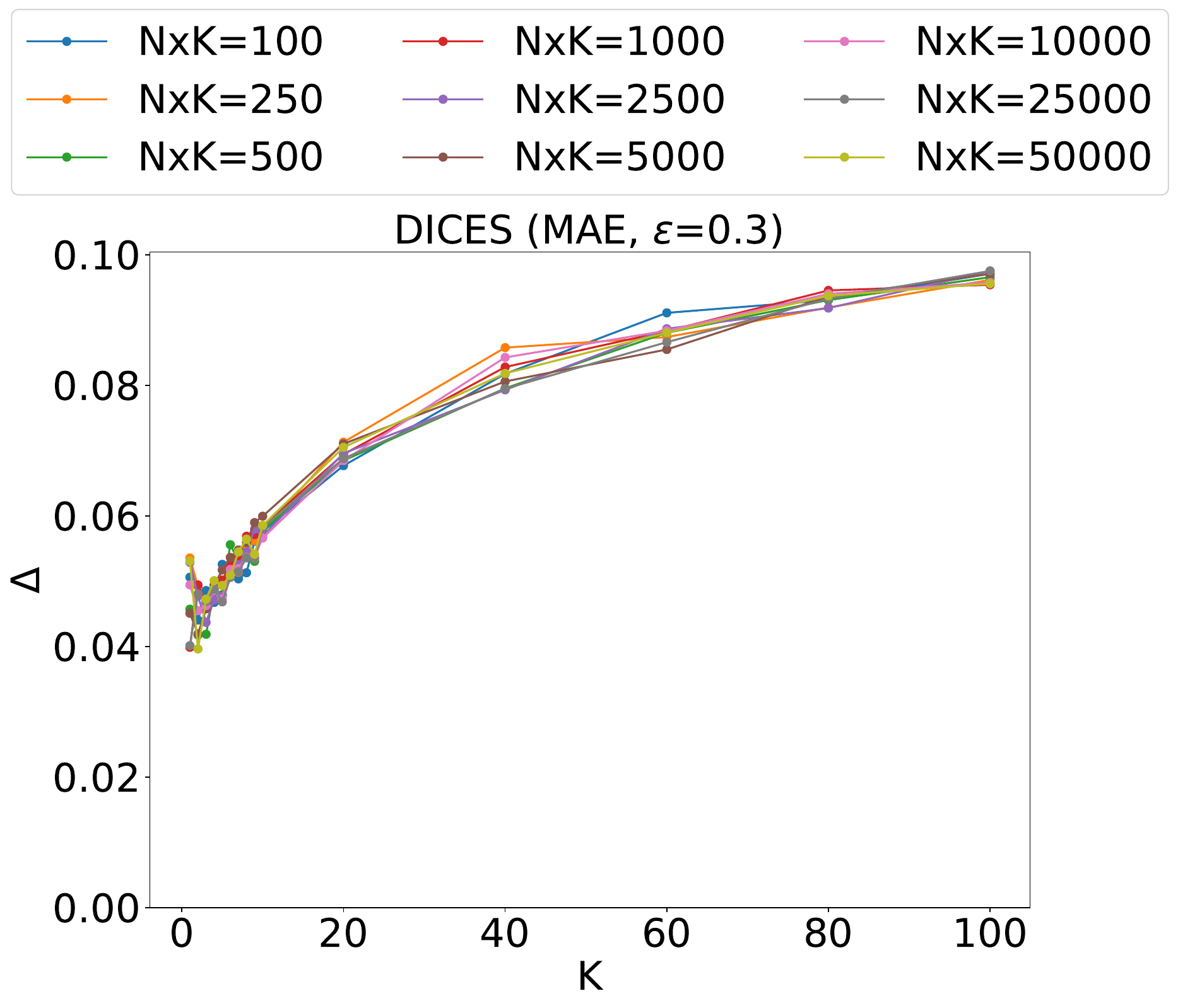}
    \caption{$\epsilon = 0.3$}
    \label{fig:dices_delta_MAE_e03}
  \end{subfigure} \hfill
  \begin{subfigure}[b]{0.24\linewidth}
    \centering
    \includegraphics[width=\linewidth]{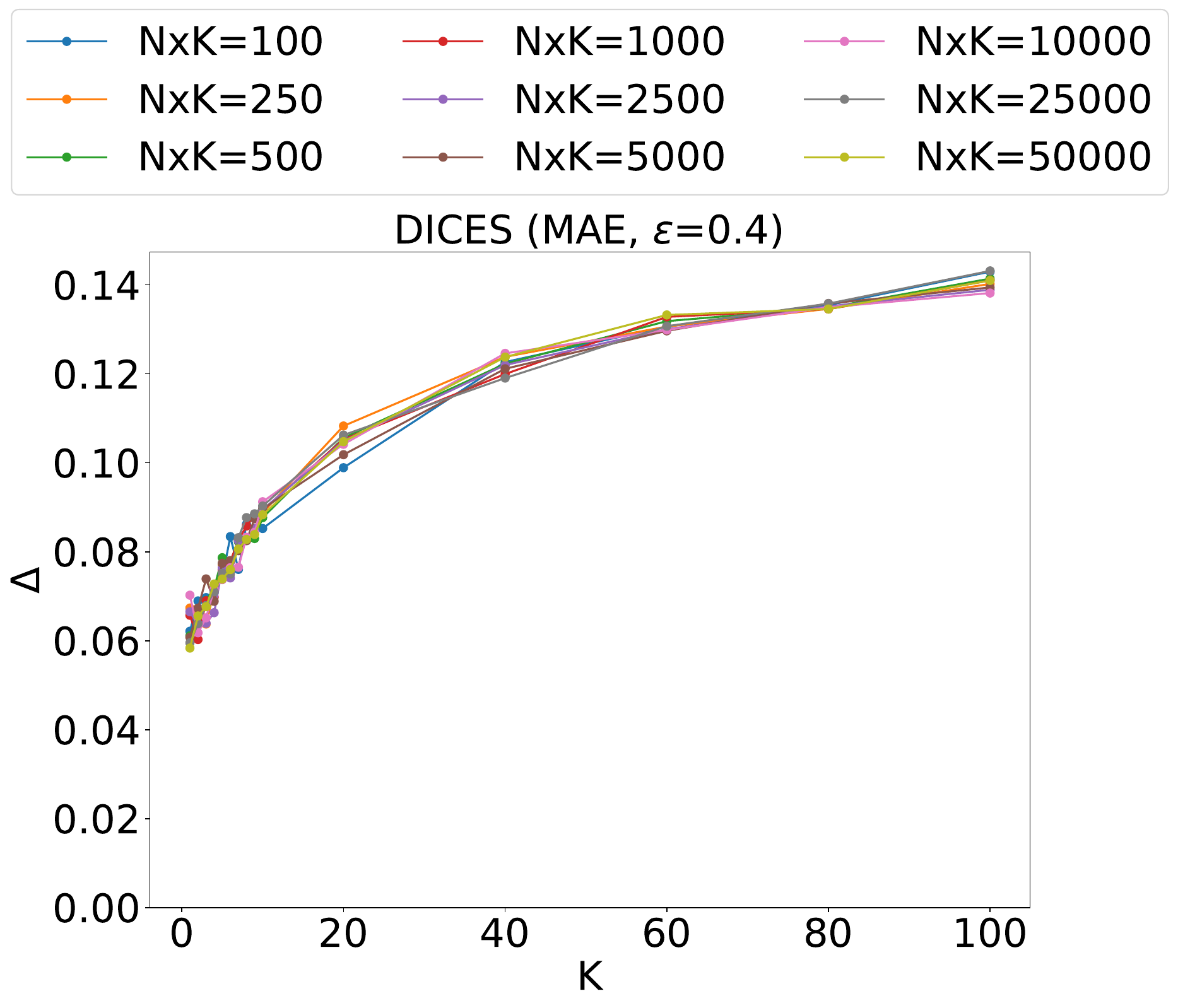}
    \caption{$\epsilon = 0.4$}
    \label{fig:dices_delta_MAE_e04}
  \end{subfigure}
  \caption{S1: Effect sizes ($\Delta$) for DICES dataset with MAE as the metric}
  \label{fig:dices_delta_MAE}
\end{figure*}

\begin{figure*}
  \centering
  \begin{subfigure}[b]{0.24\linewidth}
    \centering
    \includegraphics[width=\linewidth]{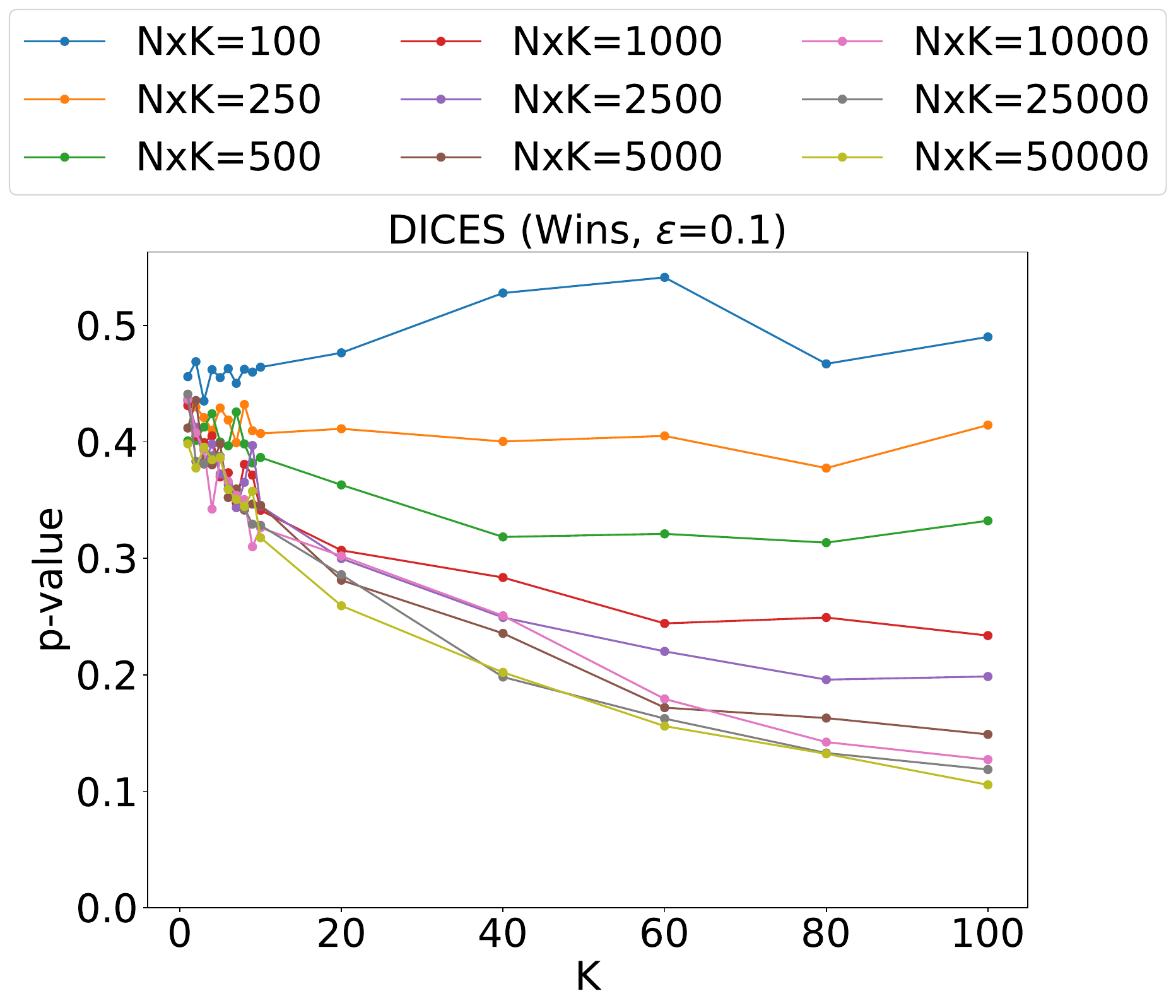}
    \caption{$\epsilon = 0.1$}
    \label{fig:dices_wins_e01}
  \end{subfigure} \hfill
  \begin{subfigure}[b]{0.24\linewidth}
    \centering
    \includegraphics[width=\linewidth]{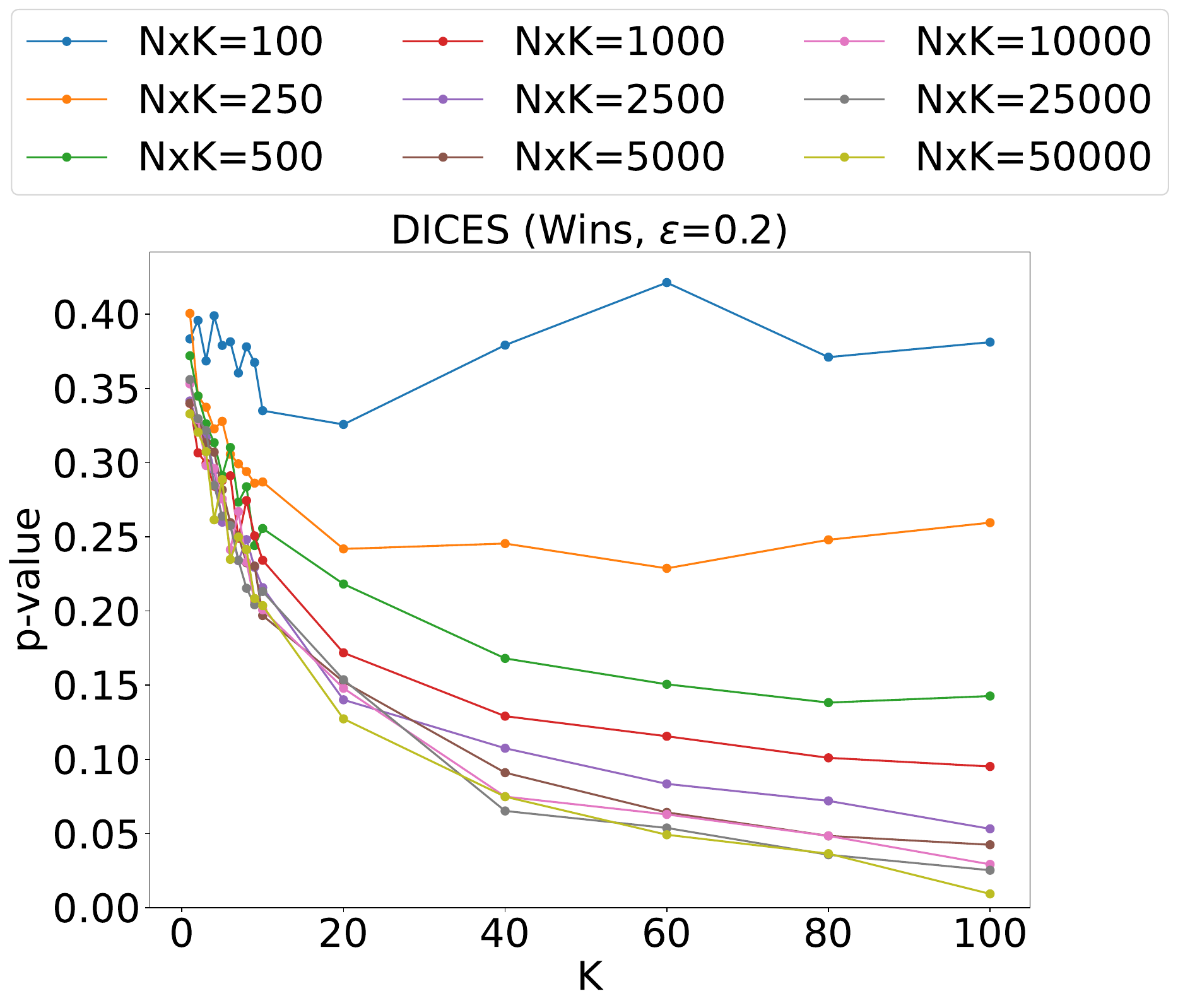}
    \caption{$\epsilon = 0.2$}
    \label{fig:dices_wins_e02}
  \end{subfigure} \hfill
  \begin{subfigure}[b]{0.24\linewidth}
    \centering
    \includegraphics[width=\linewidth]{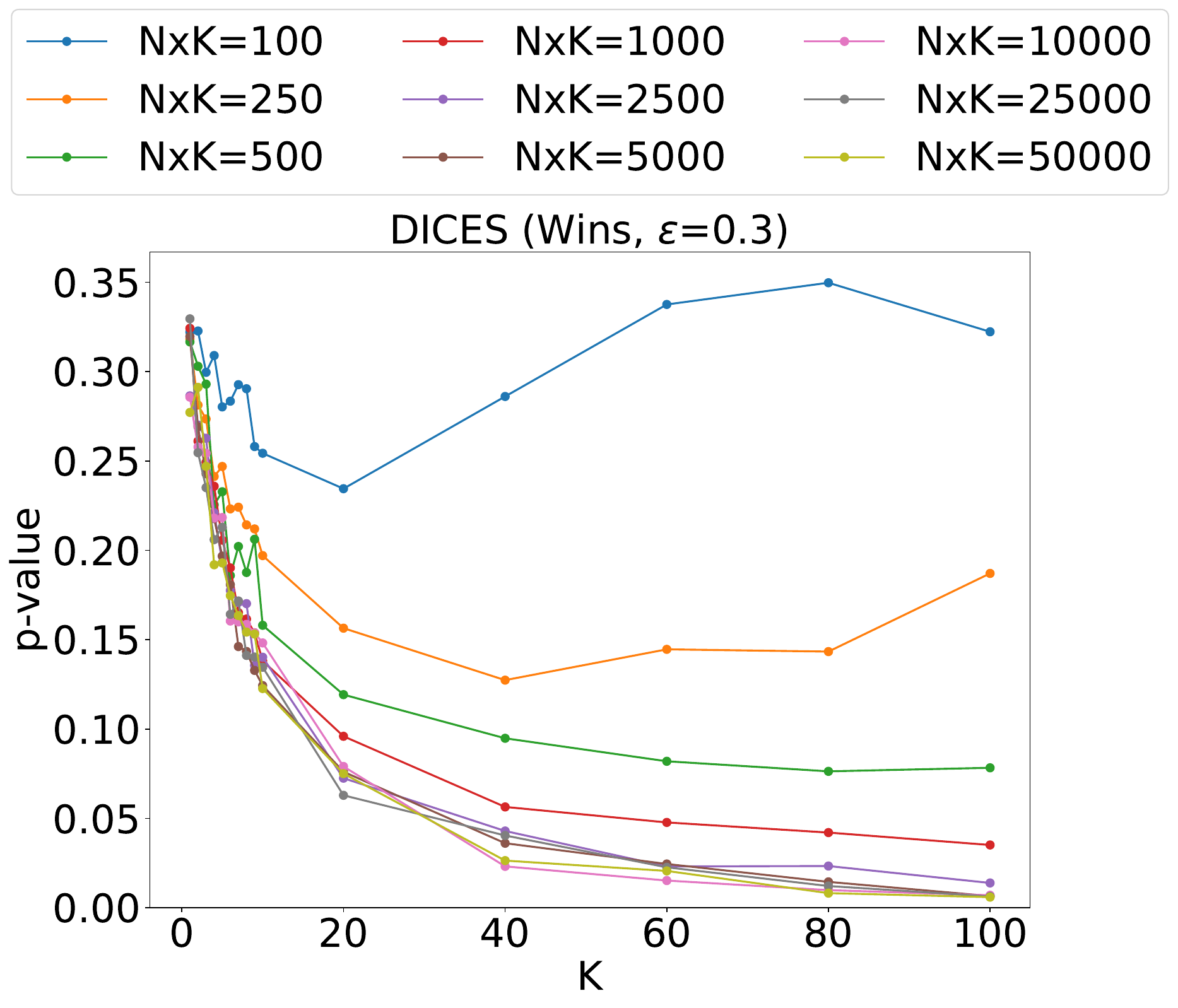}
    \caption{$\epsilon = 0.3$}
    \label{fig:dices_wins_e03}
  \end{subfigure} \hfill
  \begin{subfigure}[b]{0.24\linewidth}
    \centering
    \includegraphics[width=\linewidth]{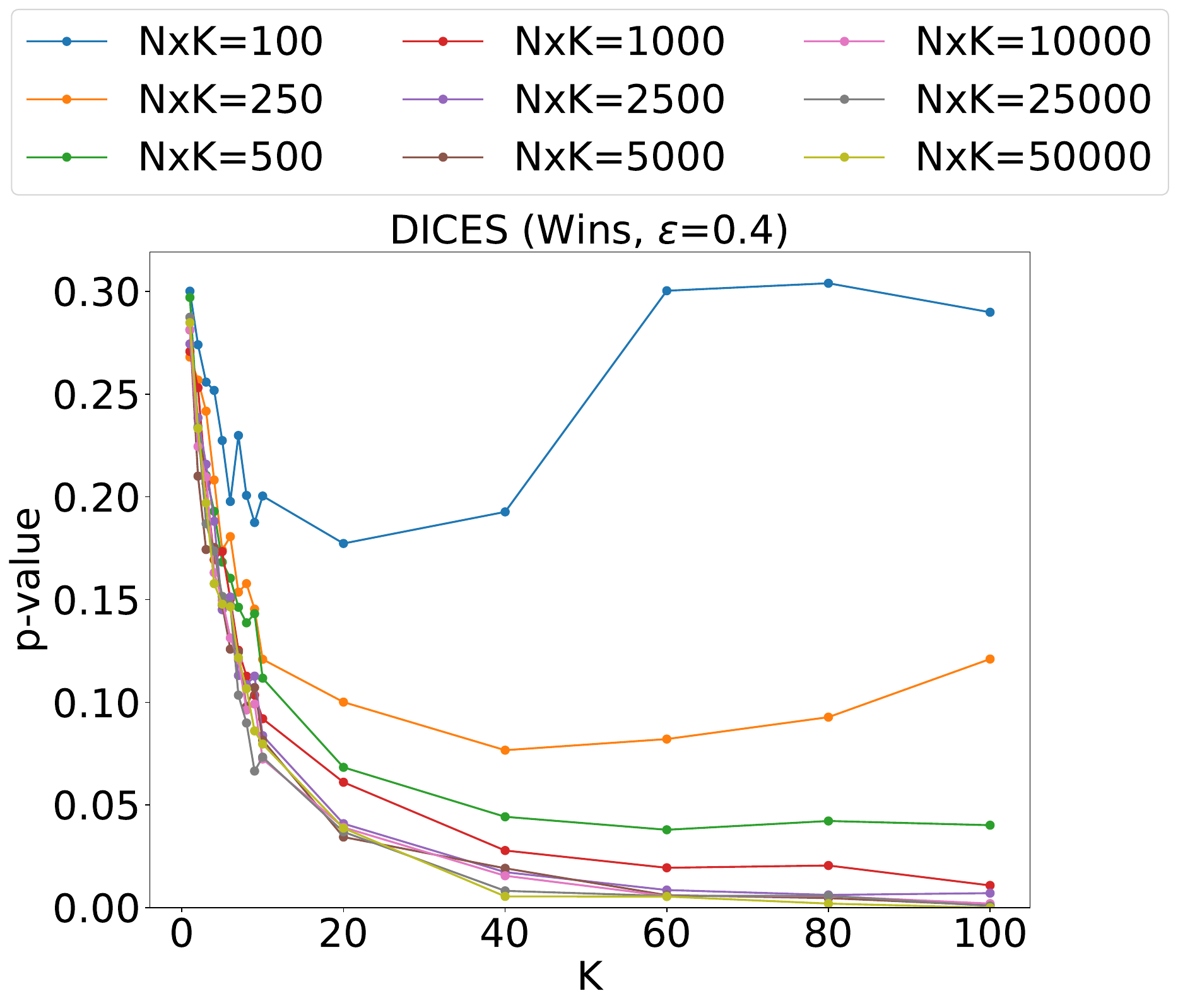}
    \caption{$\epsilon = 0.4$}
    \label{fig:dices_wins_e04}
  \end{subfigure}
  \caption{S1: P-value plots for DICES dataset with Wins as the metric}
  \label{fig:dices_wins}
\end{figure*}

\begin{figure*}
  \centering
  \begin{subfigure}[b]{0.24\linewidth}
    \centering
    \includegraphics[width=\linewidth]{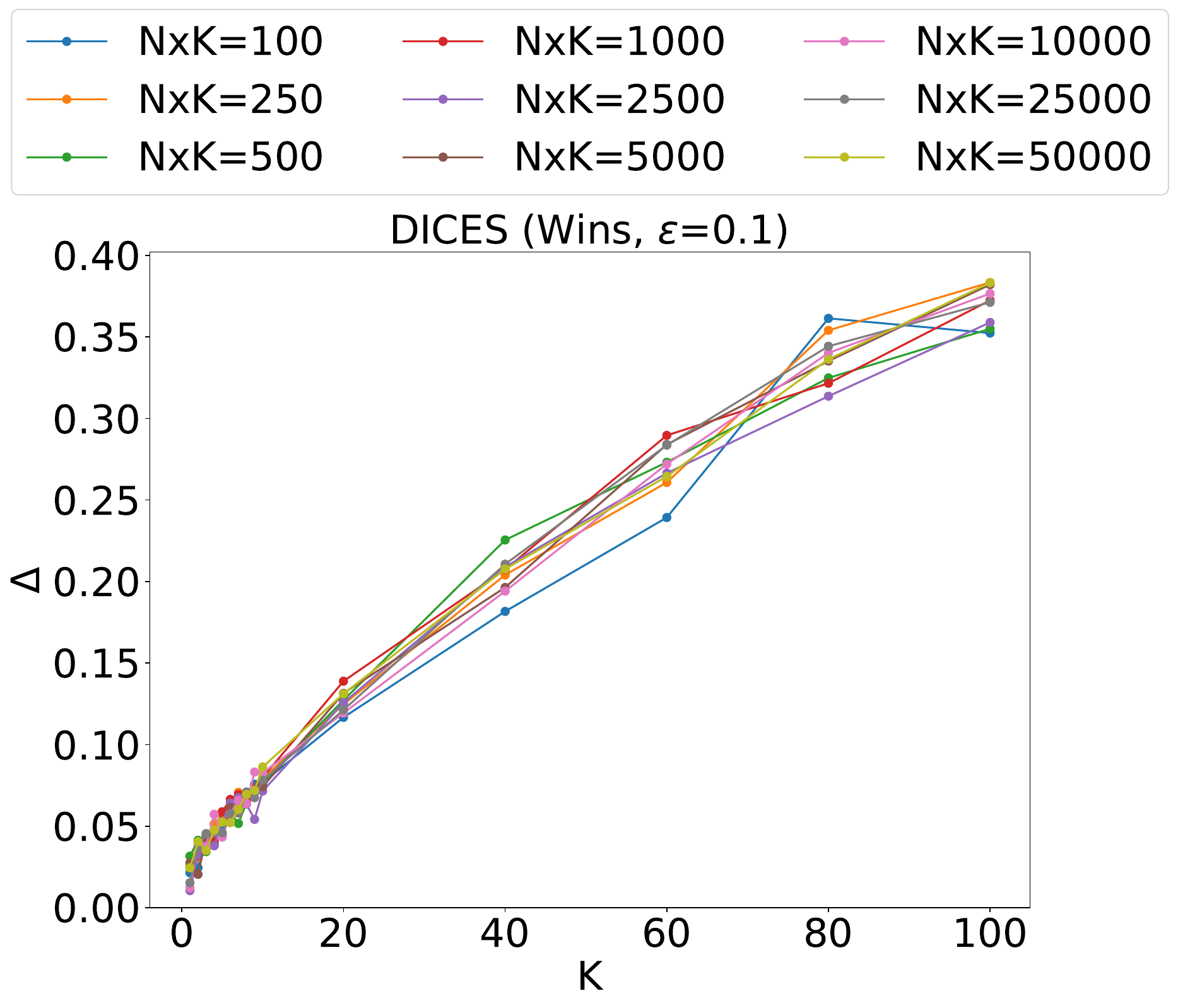}
    \caption{$\epsilon = 0.1$}
    \label{fig:dices_delta_wins_e01}
  \end{subfigure} \hfill
  \begin{subfigure}[b]{0.24\linewidth}
    \centering
    \includegraphics[width=\linewidth]{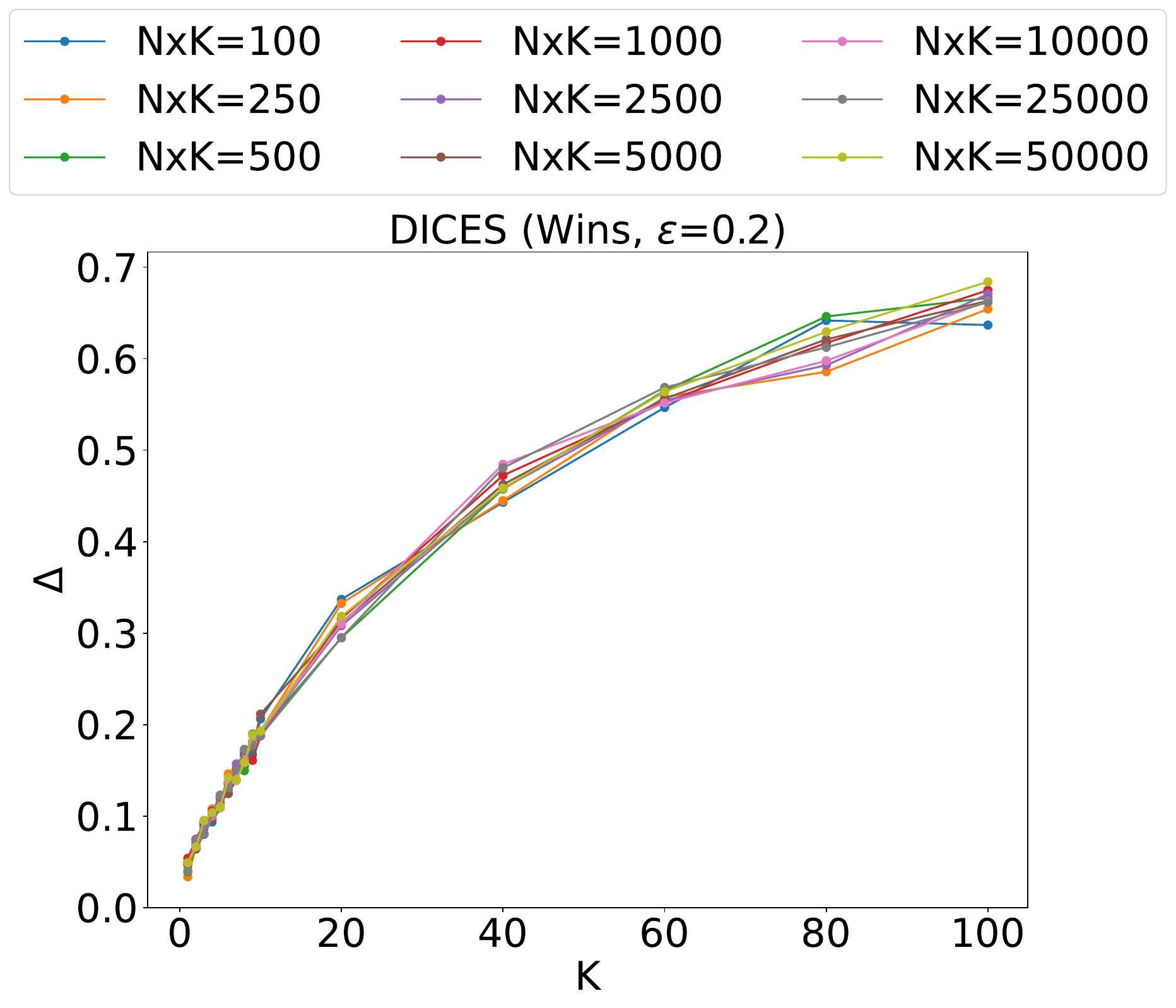}
    \caption{$\epsilon = 0.2$}
    \label{fig:dices_delta_wins_e02}
  \end{subfigure} \hfill
  \begin{subfigure}[b]{0.24\linewidth}
    \centering
    \includegraphics[width=\linewidth]{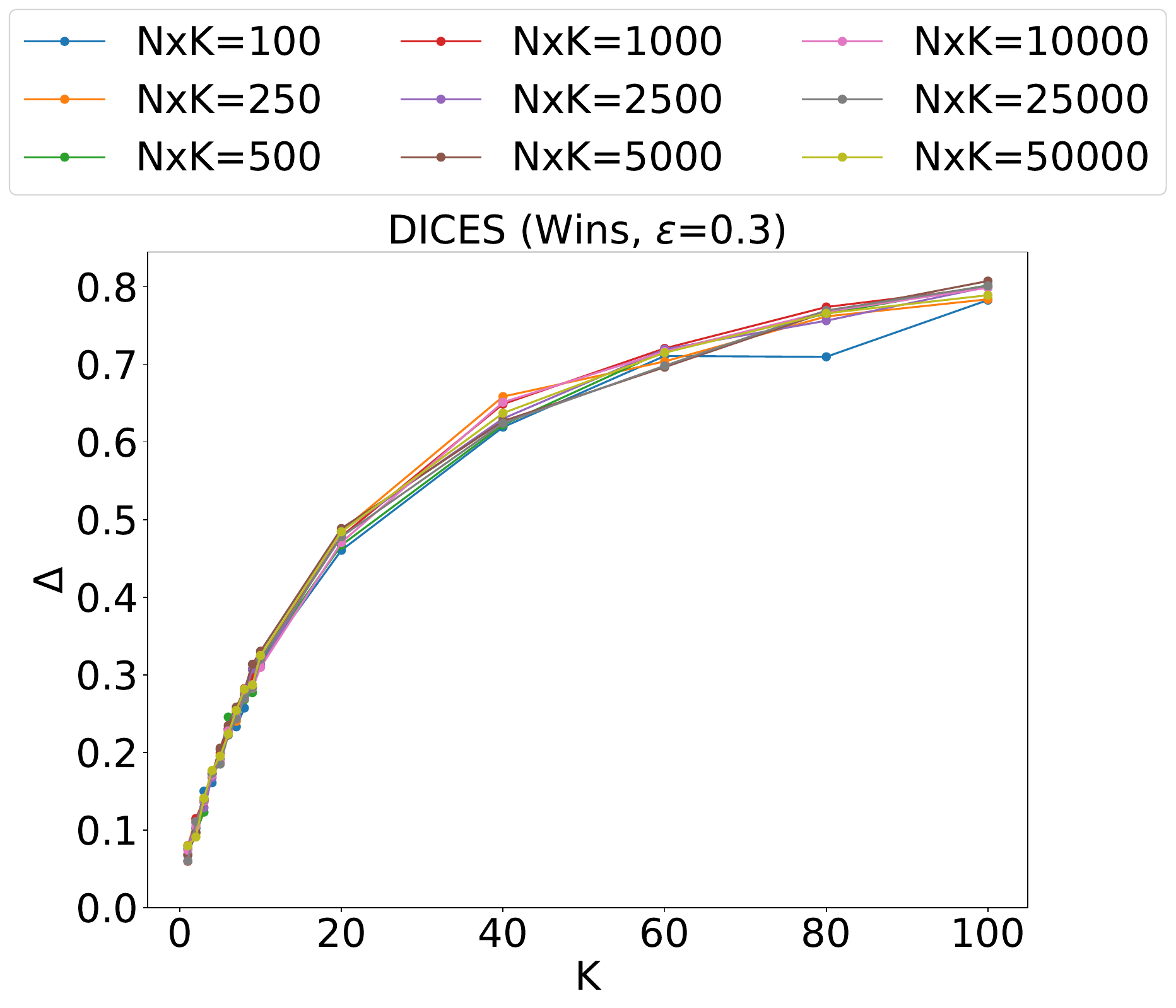}
    \caption{$\epsilon = 0.3$}
    \label{fig:dices_delta_wins_e03}
  \end{subfigure} \hfill
  \begin{subfigure}[b]{0.24\linewidth}
    \centering
    \includegraphics[width=\linewidth]{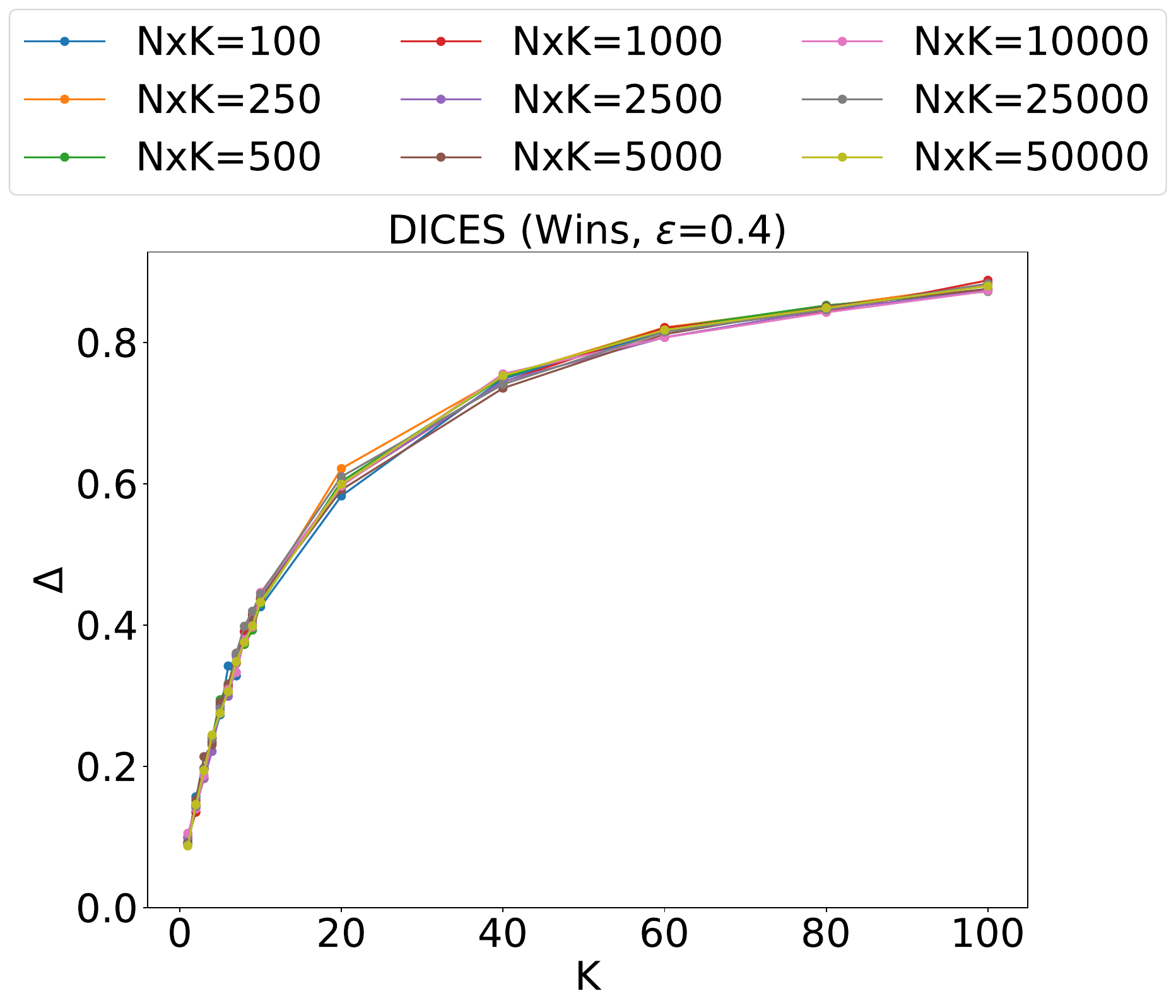}
    \caption{$\epsilon = 0.4$}
    \label{fig:dices_delta_wins_e04}
  \end{subfigure}
  \caption{S1: Effect sizes ($\Delta$) for DICES dataset with Wins as the metric}
  \label{fig:dices_delta_wins}
\end{figure*}

\paragraph{DICES - 5 Rater Sample}

\begin{figure*}
  \centering
  \begin{subfigure}[b]{0.24\linewidth}
    \centering
    \includegraphics[width=\linewidth]{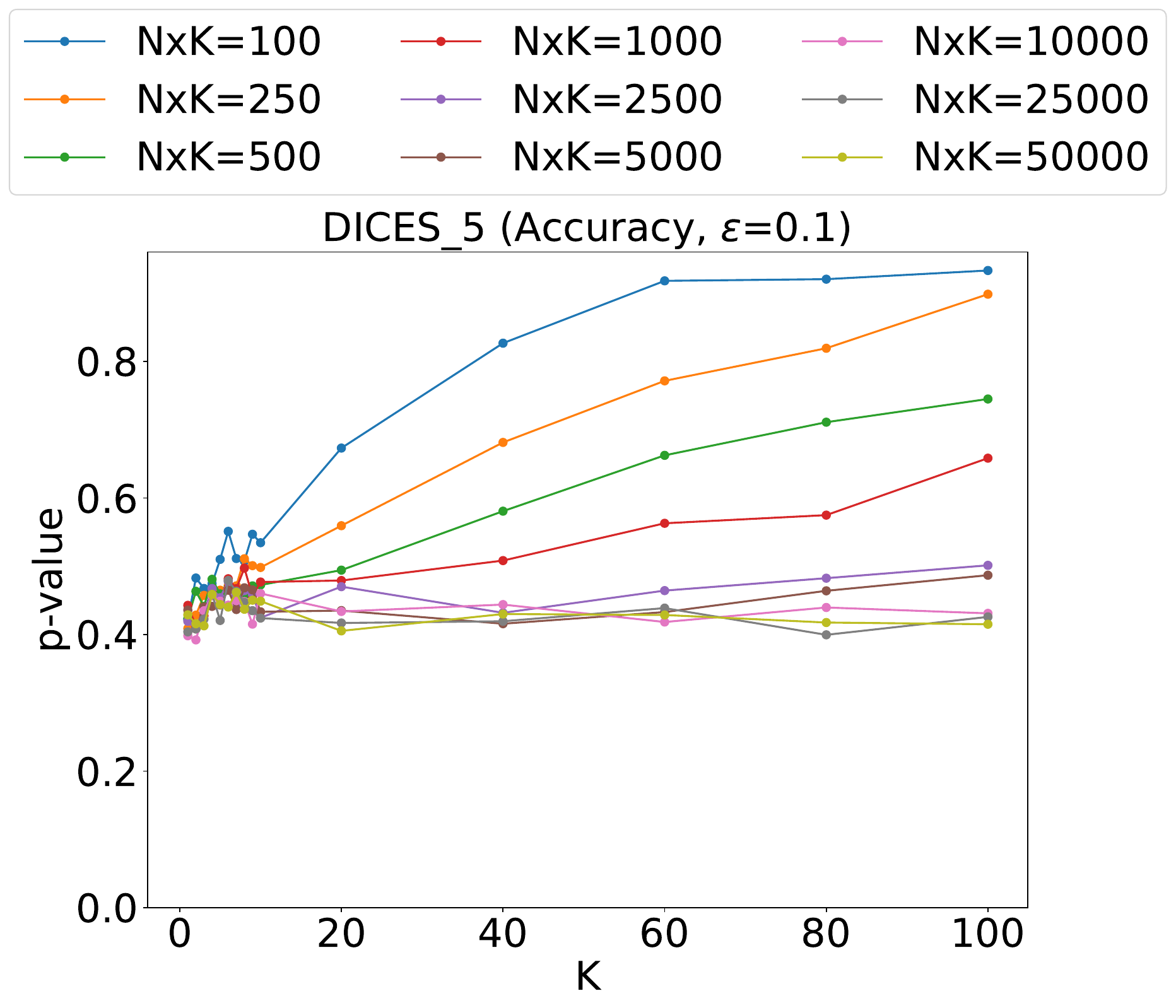}
    \caption{$\epsilon = 0.1$}
    \label{fig:dices_5_acc_e01}
  \end{subfigure} \hfill
  \begin{subfigure}[b]{0.24\linewidth}
    \centering
    \includegraphics[width=\linewidth]{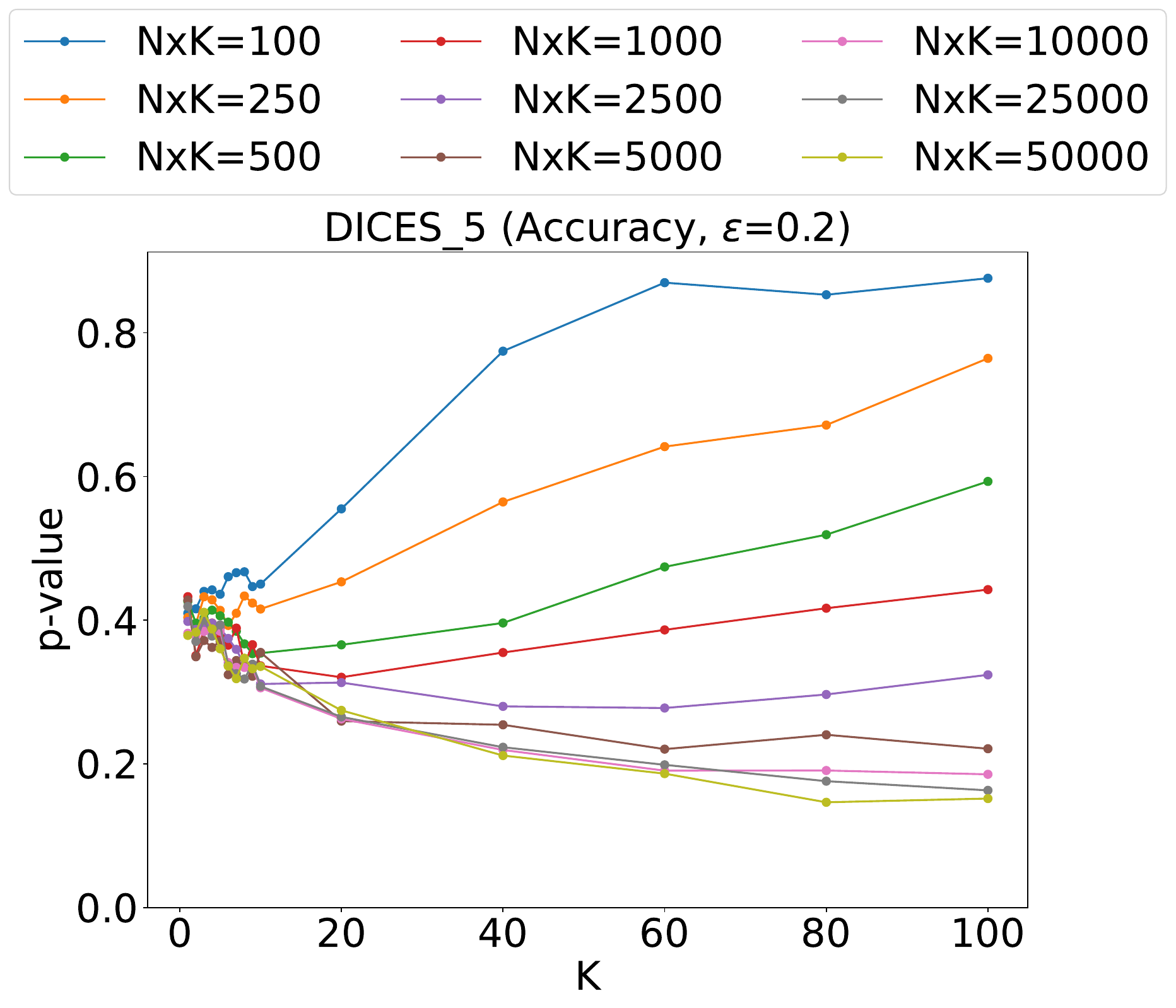}
    \caption{$\epsilon = 0.2$}
    \label{fig:dices_5_acc_e02}
  \end{subfigure} \hfill
  \begin{subfigure}[b]{0.24\linewidth}
    \centering
    \includegraphics[width=\linewidth]{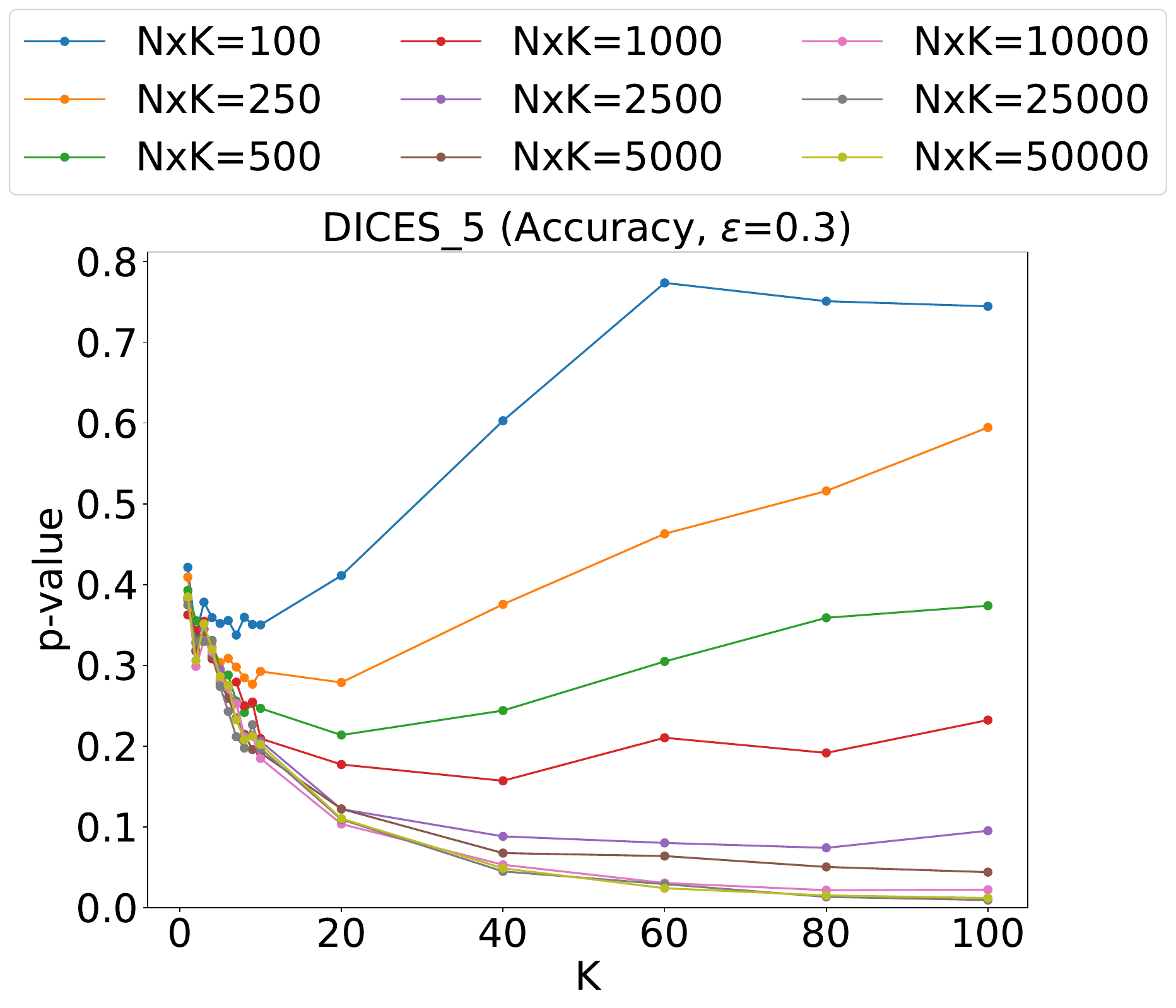}
    \caption{$\epsilon = 0.3$}
    \label{fig:dices_5_acc_e03}
  \end{subfigure} \hfill
  \begin{subfigure}[b]{0.24\linewidth}
    \centering
    \includegraphics[width=\linewidth]{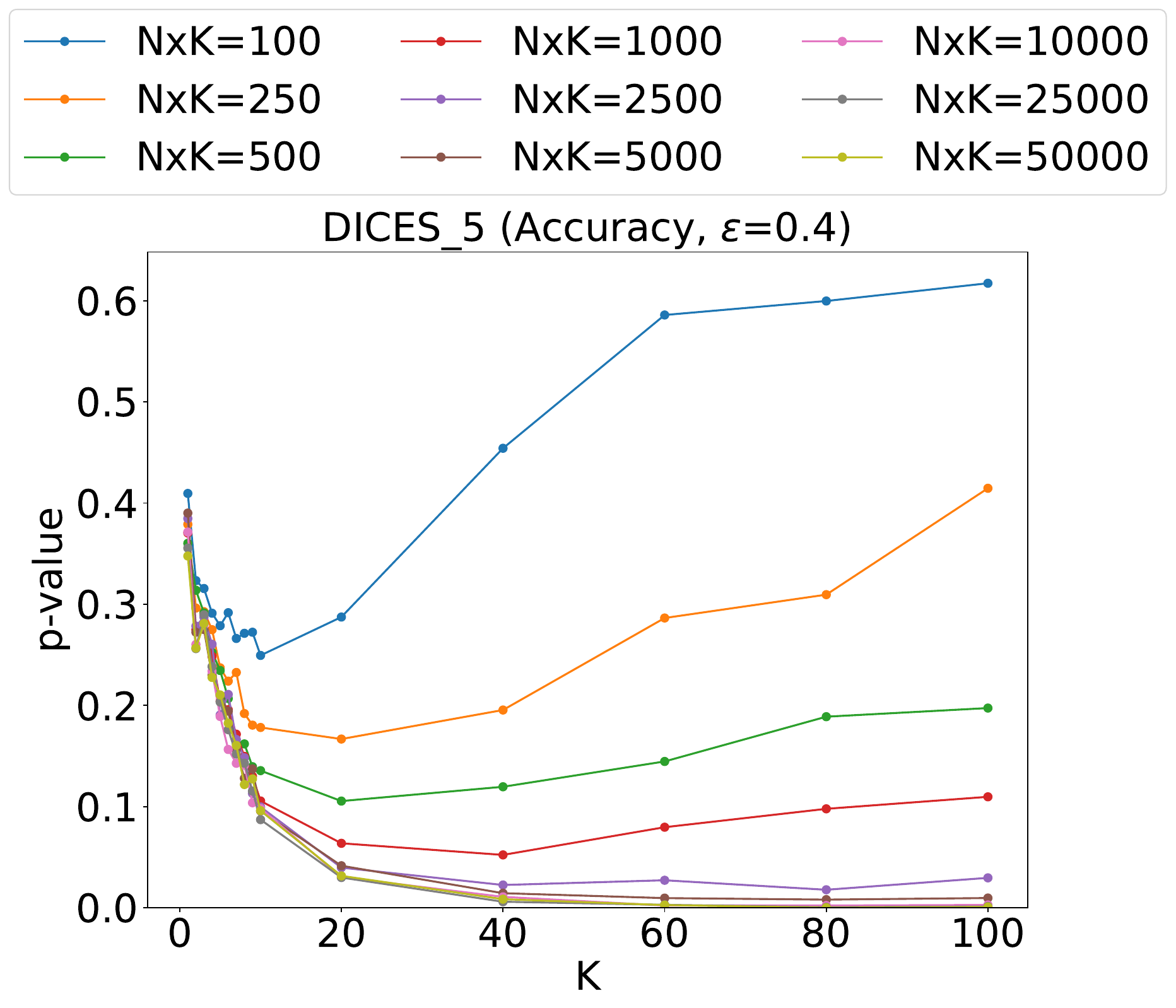}
    \caption{$\epsilon = 0.4$}
    \label{fig:dices_5_acc_e04}
  \end{subfigure}
  \caption{S1: P-value plots for DICES 5 rater sample with Accuracy as the metric}
  \label{fig:dices_5_accuracy}
\end{figure*}

\begin{figure*}
  \centering
  \begin{subfigure}[b]{0.24\linewidth}
    \centering
    \includegraphics[width=\linewidth]{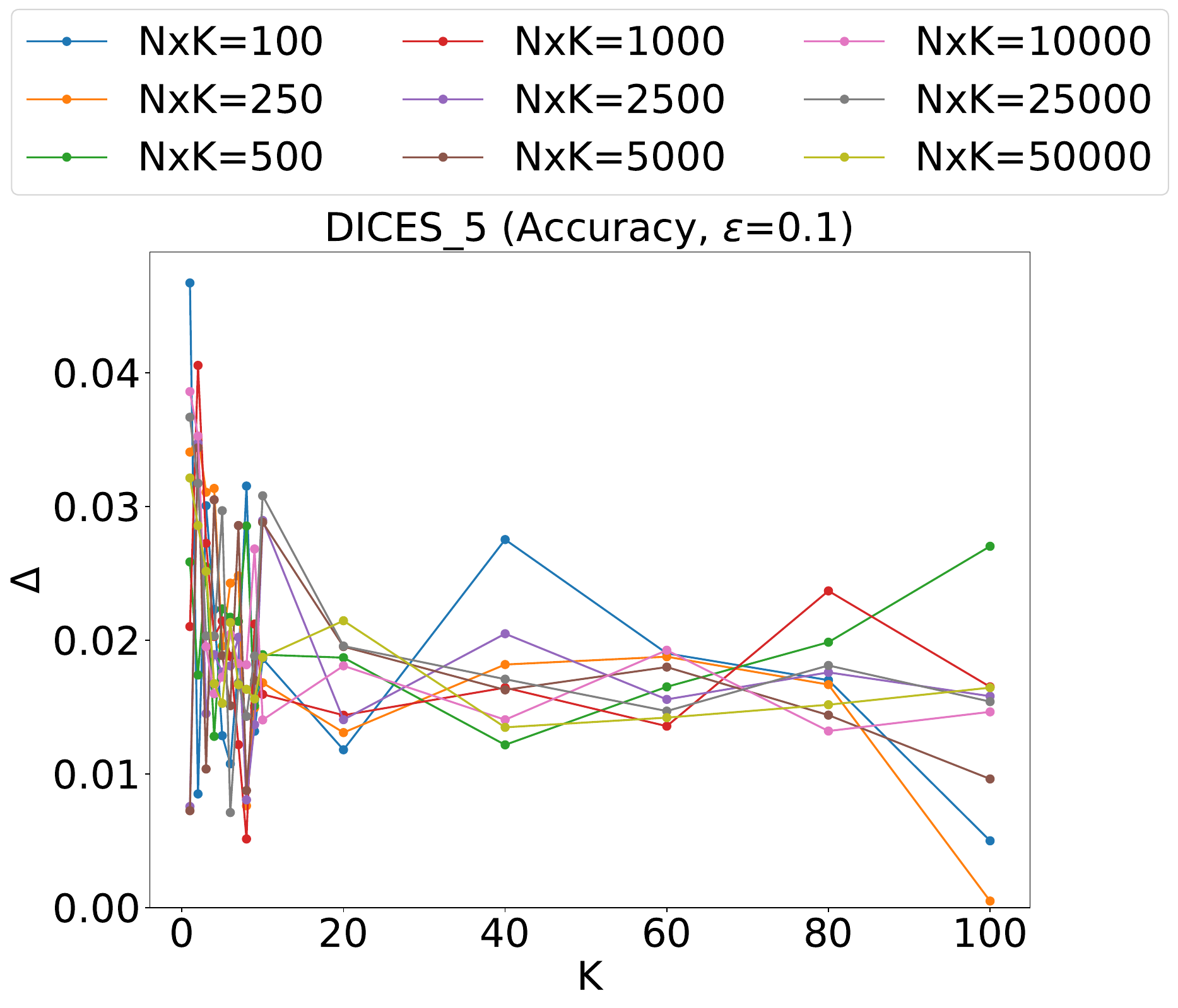}
    \caption{$\epsilon = 0.1$}
    \label{fig:dices_5_delta_acc_e01}
  \end{subfigure} \hfill
  \begin{subfigure}[b]{0.24\linewidth}
    \centering
    \includegraphics[width=\linewidth]{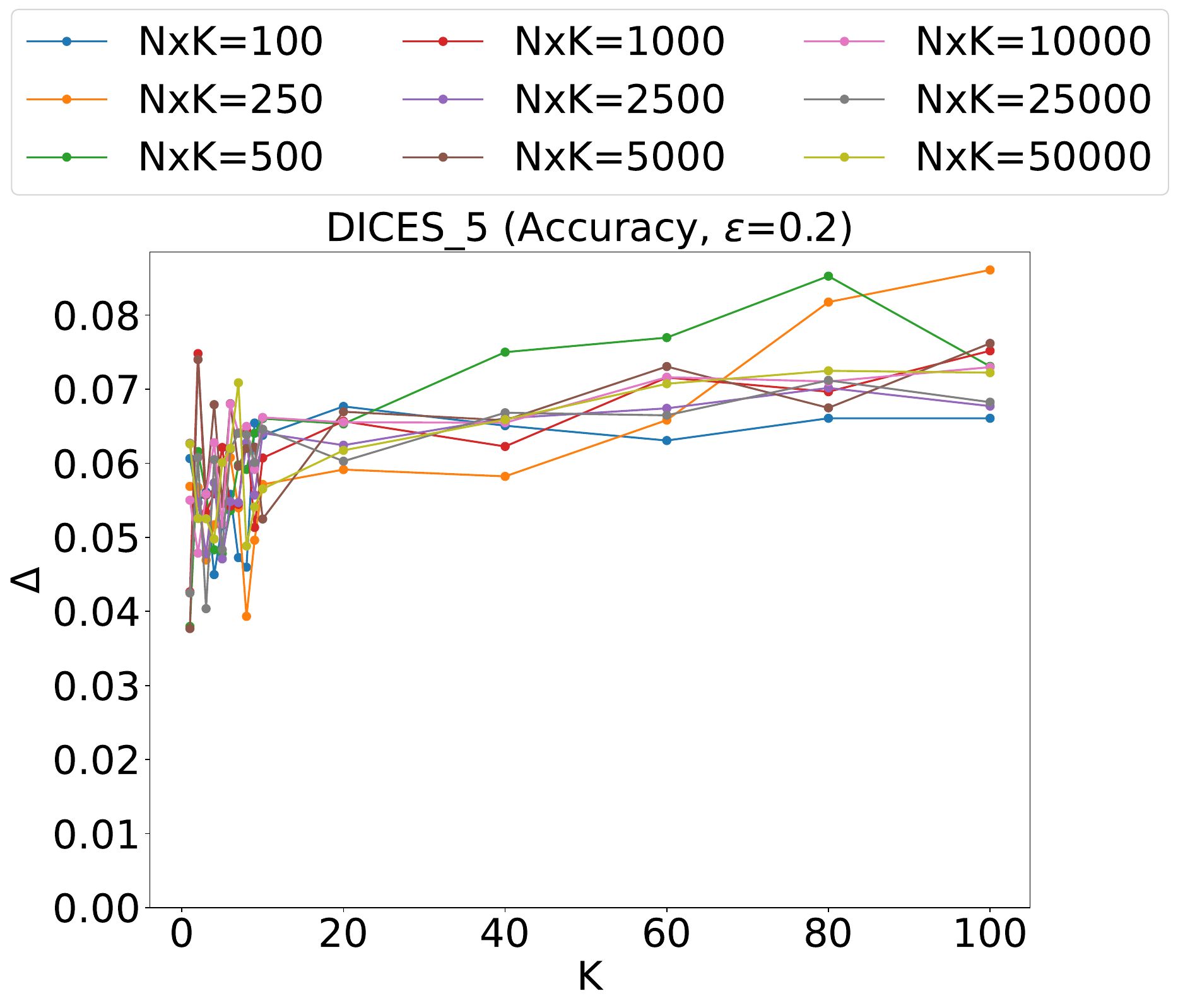}
    \caption{$\epsilon = 0.2$}
    \label{fig:dices_5_delta_acc_e02}
  \end{subfigure} \hfill
  \begin{subfigure}[b]{0.24\linewidth}
    \centering
    \includegraphics[width=\linewidth]{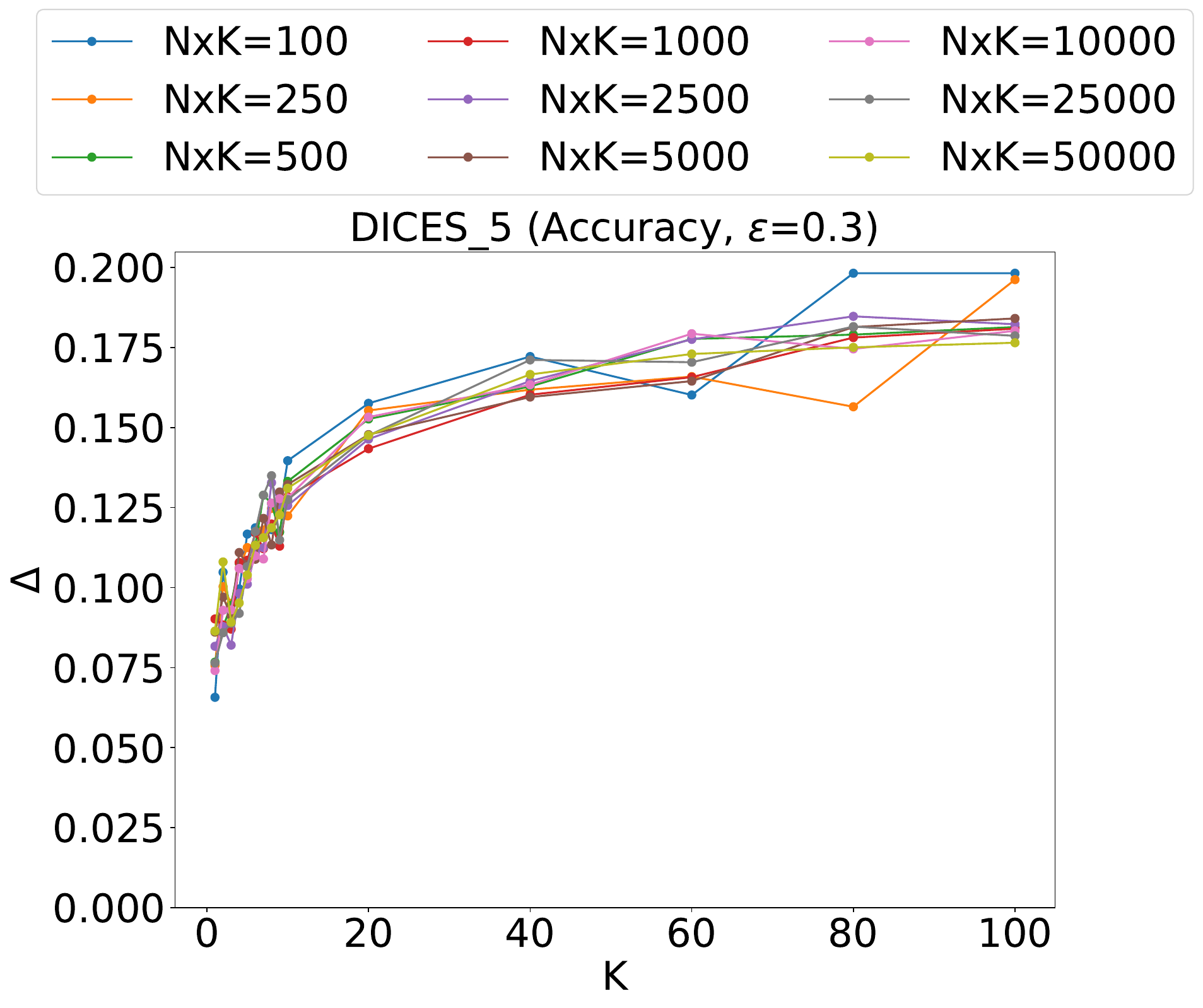}
    \caption{$\epsilon = 0.3$}
    \label{fig:dices_5_delta_acc_e03}
  \end{subfigure} \hfill
  \begin{subfigure}[b]{0.24\linewidth}
    \centering
    \includegraphics[width=\linewidth]{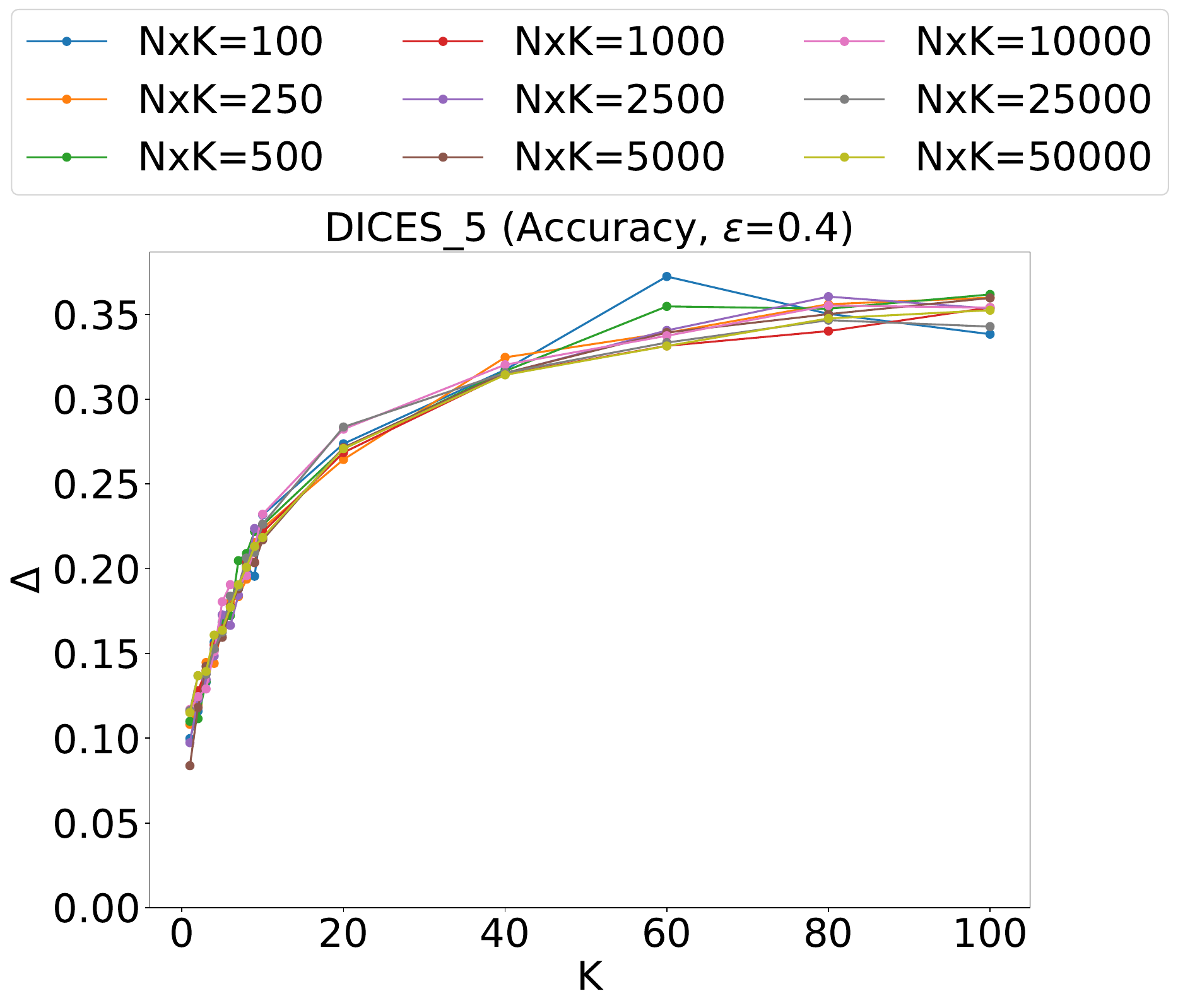}
    \caption{$\epsilon = 0.4$}
    \label{fig:dices_5_delta_acc_e04}
  \end{subfigure}
  \caption{S1: Effect sizes ($\Delta$) for DICES 5 rater sample with Accuracy as the metric}
  \label{fig:dices_5_delta_accuracy}
\end{figure*}

\begin{figure*}
  \centering
  \begin{subfigure}[b]{0.24\linewidth}
    \centering
    \includegraphics[width=\linewidth]{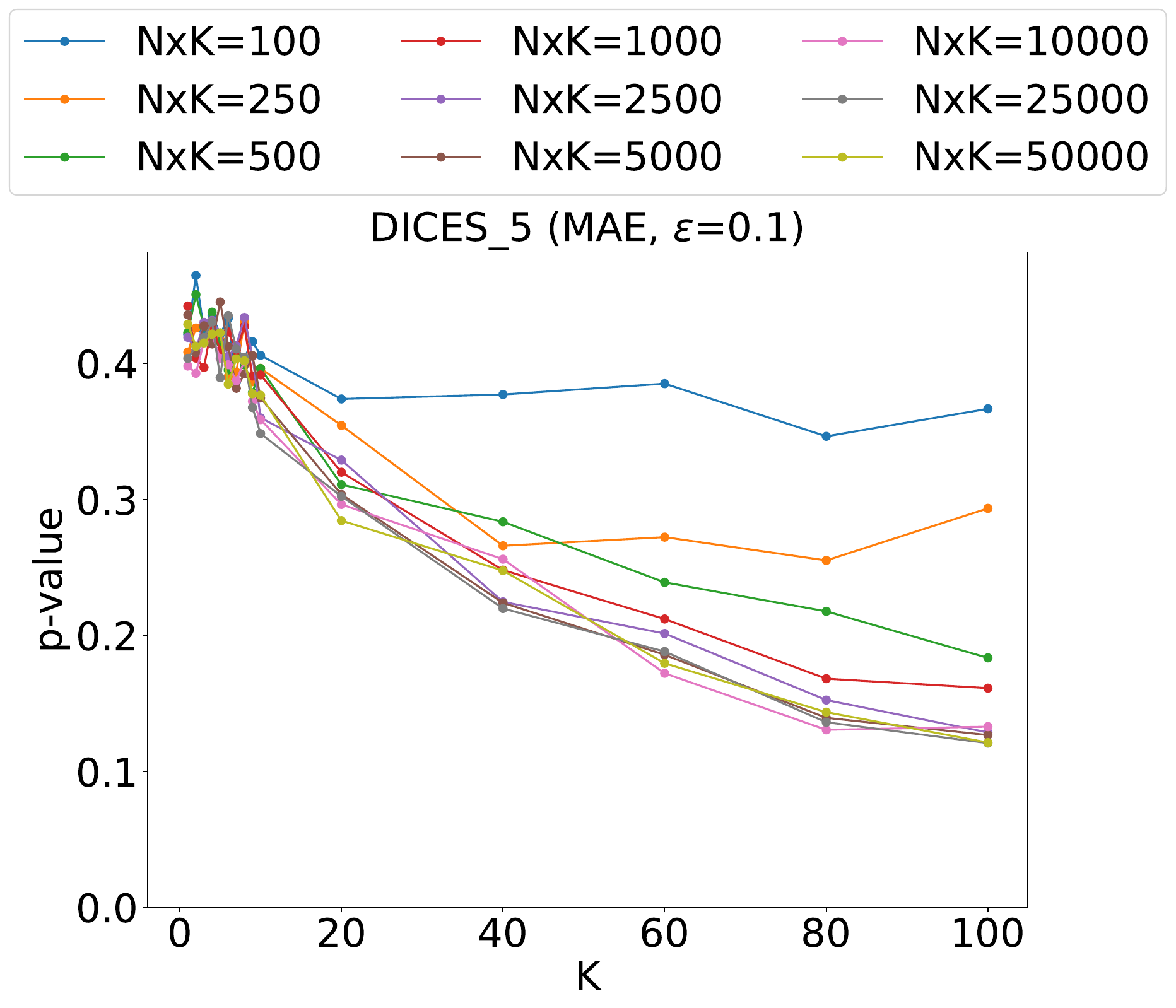}
    \caption{$\epsilon = 0.1$}
    \label{fig:dices_5_MAE_e01}
  \end{subfigure} \hfill
  \begin{subfigure}[b]{0.24\linewidth}
    \centering
    \includegraphics[width=\linewidth]{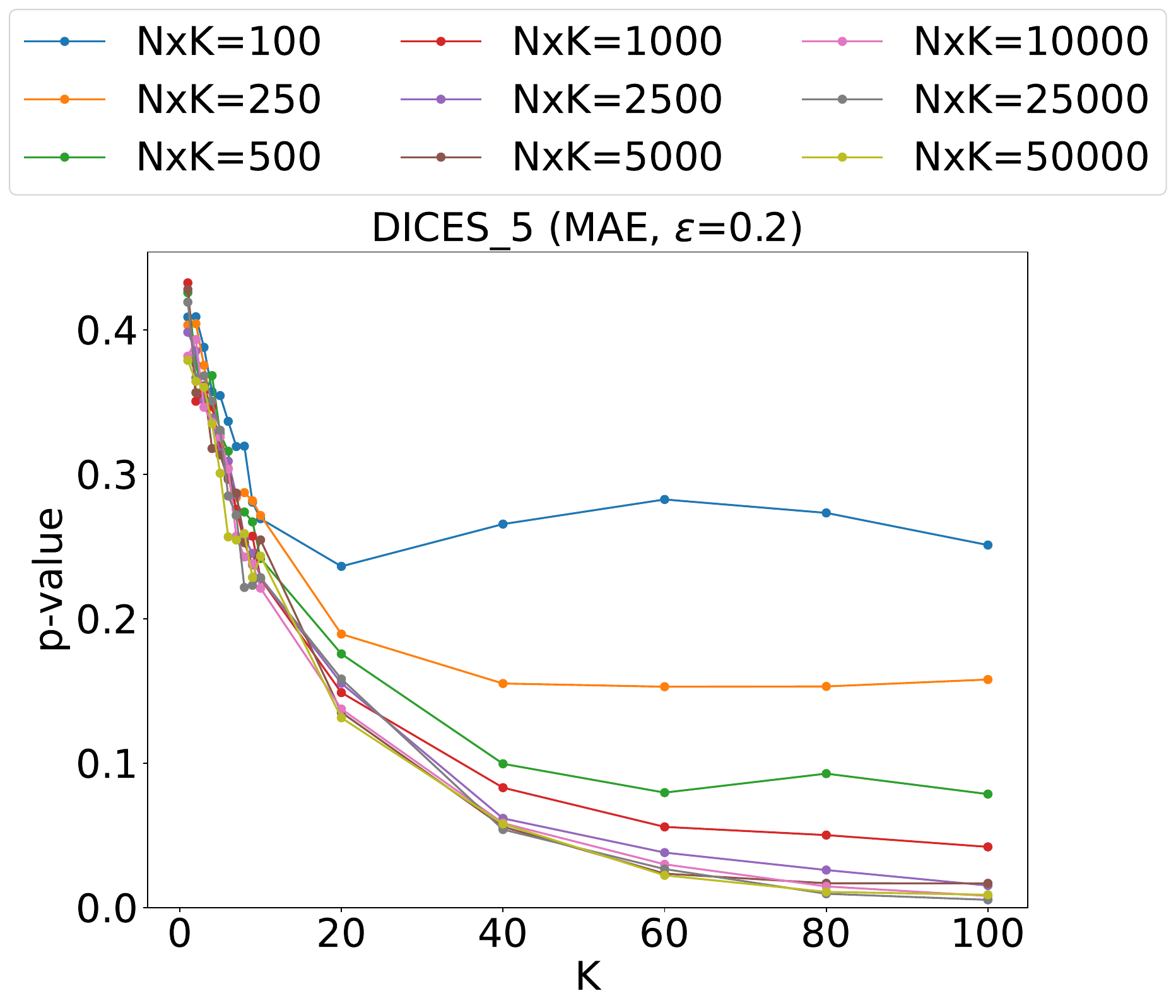}
    \caption{$\epsilon = 0.2$}
    \label{fig:dices_5_MAE_e02}
  \end{subfigure} \hfill
  \begin{subfigure}[b]{0.24\linewidth}
    \centering
    \includegraphics[width=\linewidth]{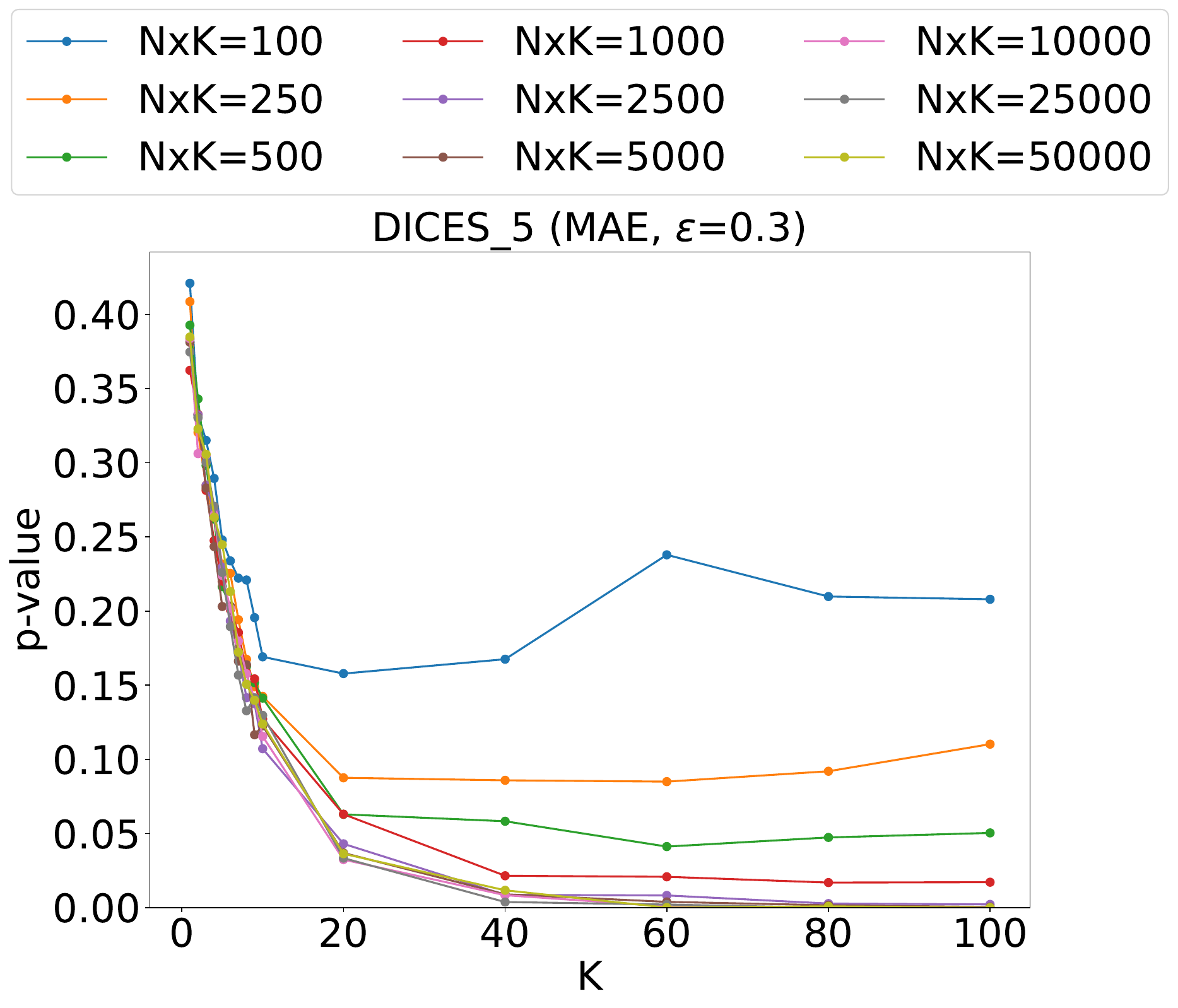}
    \caption{$\epsilon = 0.3$}
    \label{fig:dices_5_MAE_e03}
  \end{subfigure} \hfill
  \begin{subfigure}[b]{0.24\linewidth}
    \centering
    \includegraphics[width=\linewidth]{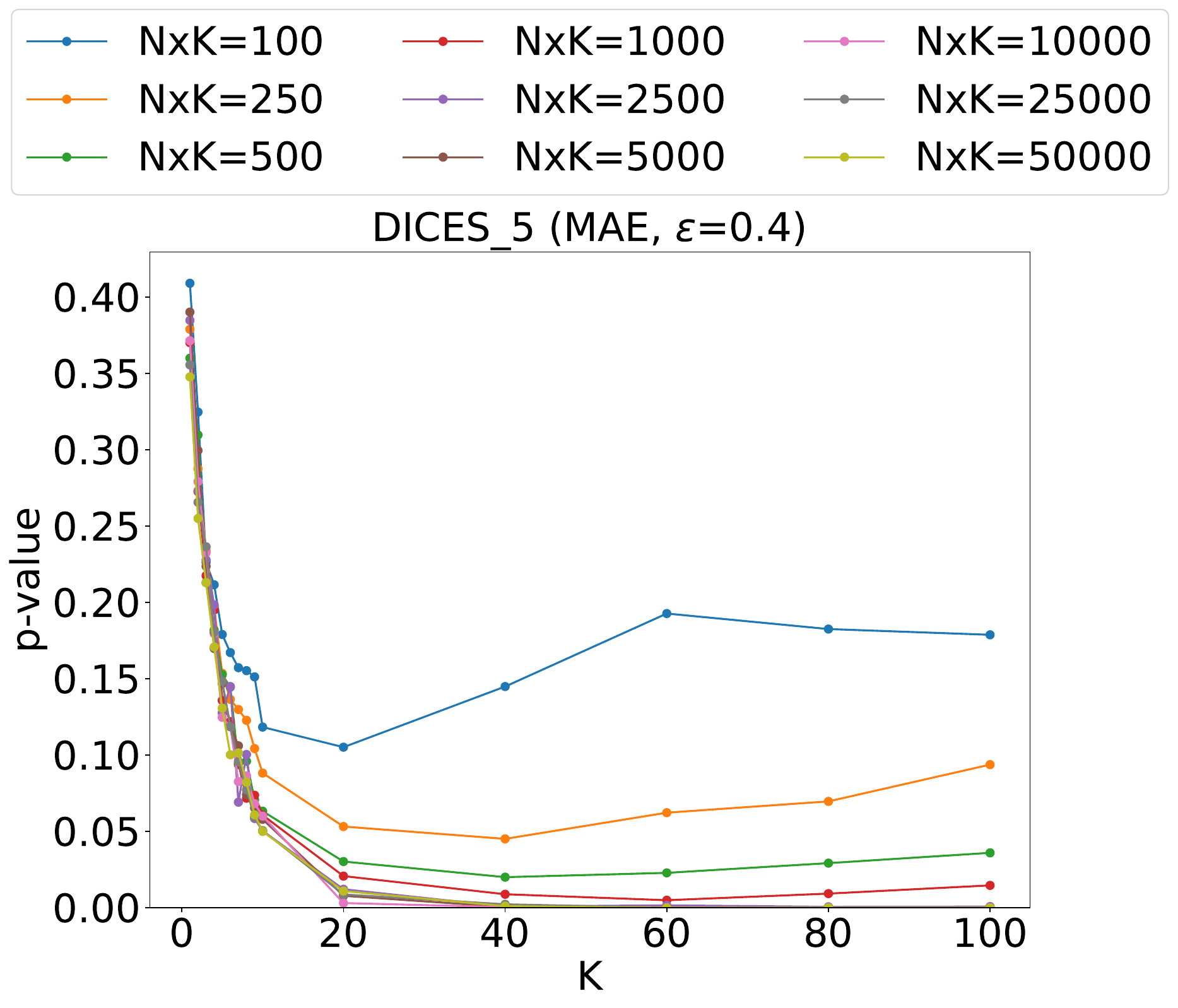}
    \caption{$\epsilon = 0.4$}
    \label{fig:dices_5_MAE_e04}
  \end{subfigure}
  \caption{S1: P-value plots for DICES 5 rater sample with MAE as the metric}
  \label{fig:dices_5_MAE}
\end{figure*}

\begin{figure*}
  \centering
  \begin{subfigure}[b]{0.24\linewidth}
    \centering
    \includegraphics[width=\linewidth]{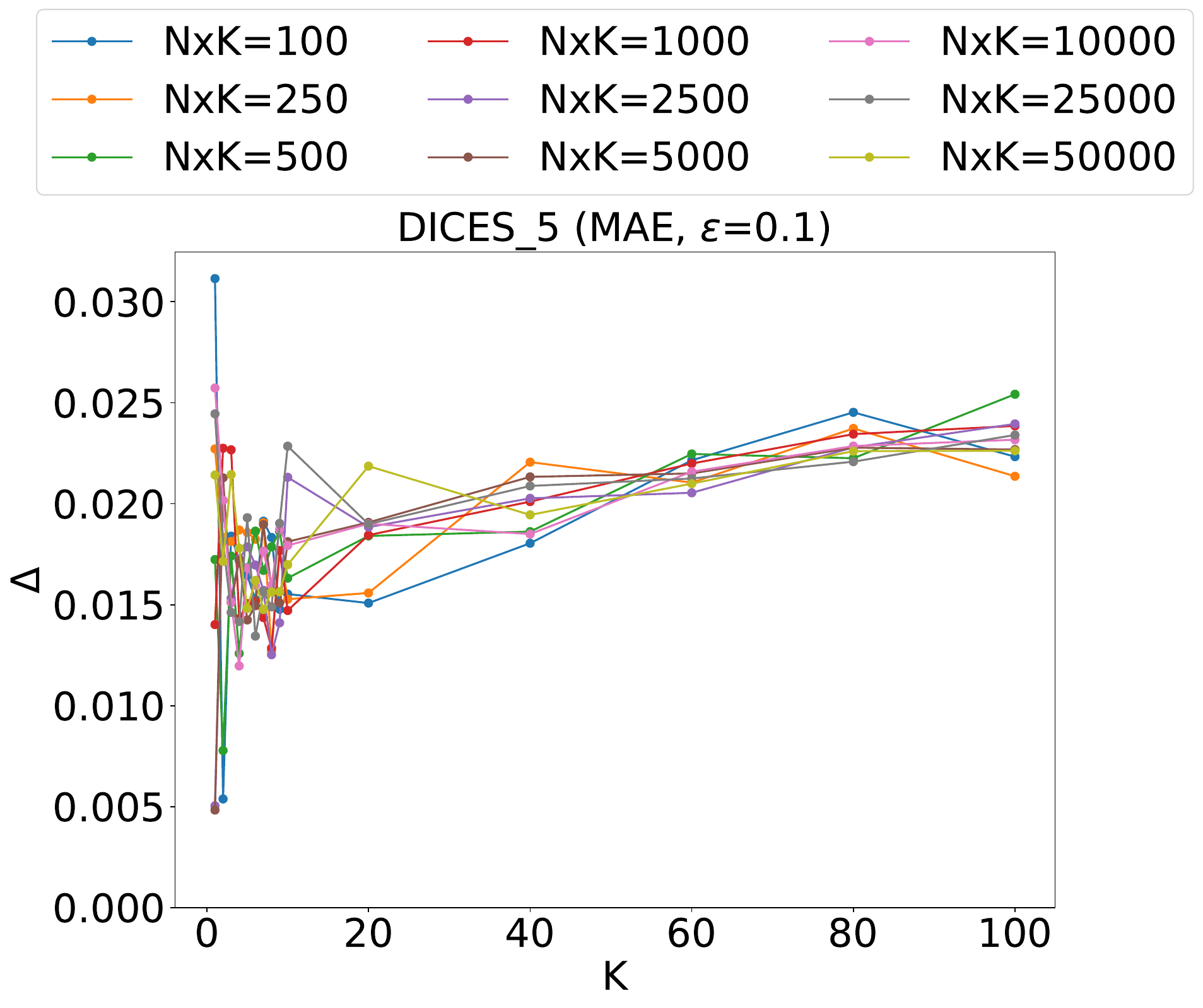}
    \caption{$\epsilon = 0.1$}
    \label{fig:dices_5_delta_MAE_e01}
  \end{subfigure} \hfill
  \begin{subfigure}[b]{0.24\linewidth}
    \centering
    \includegraphics[width=\linewidth]{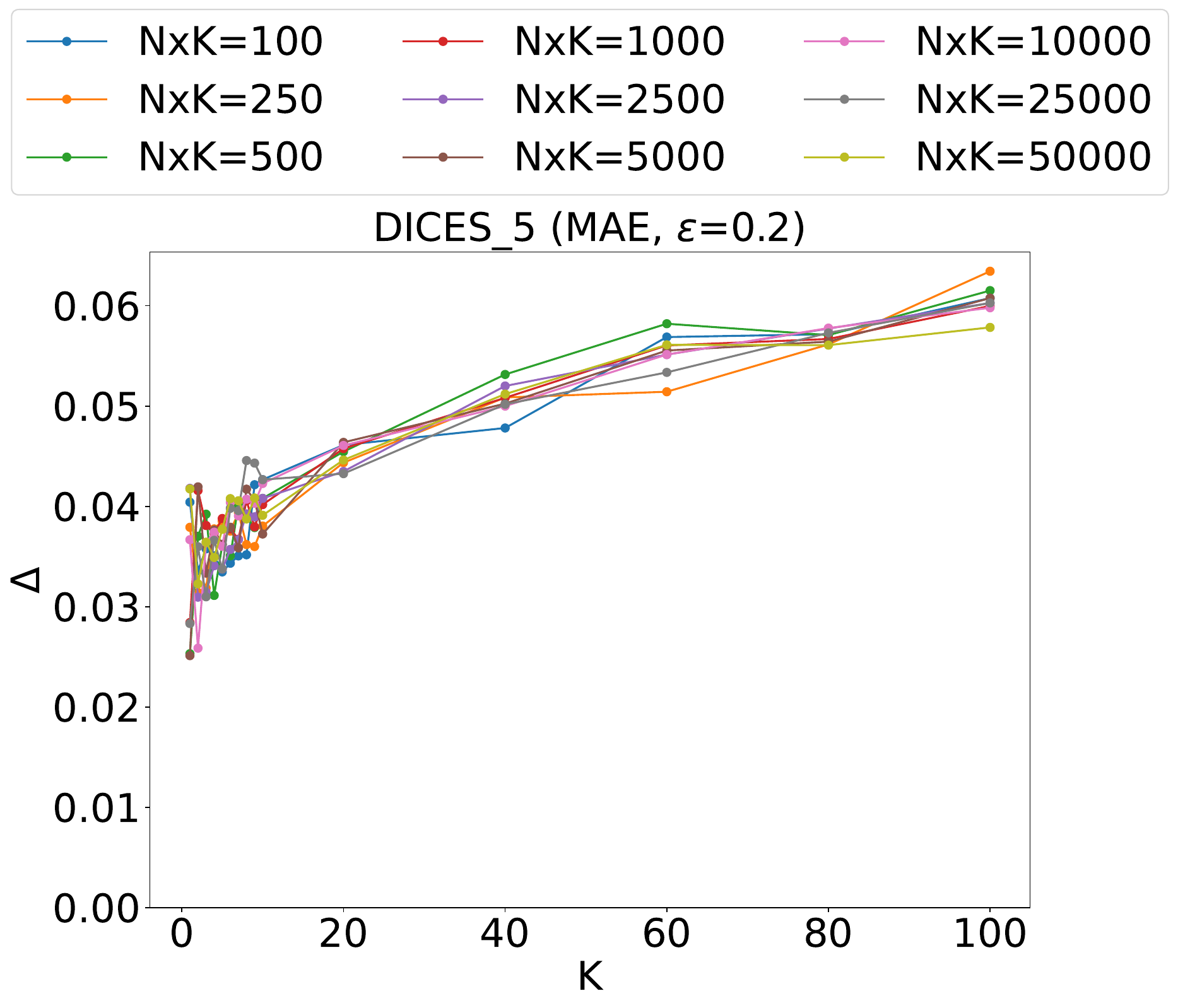}
    \caption{$\epsilon = 0.2$}
    \label{fig:dices_5_delta_MAE_e02}
  \end{subfigure} \hfill
  \begin{subfigure}[b]{0.24\linewidth}
    \centering
    \includegraphics[width=\linewidth]{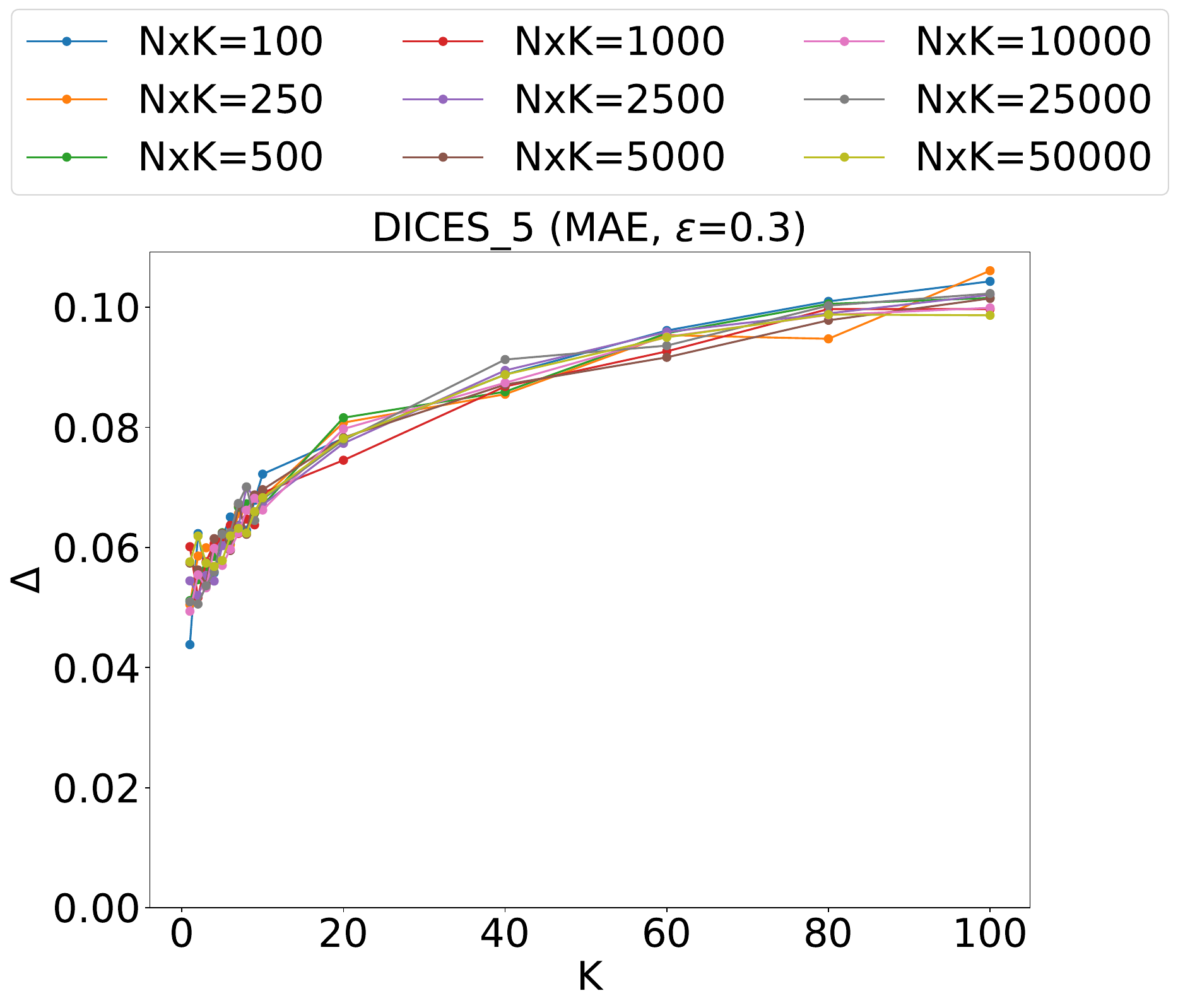}
    \caption{$\epsilon = 0.3$}
    \label{fig:dices_5_delta_MAE_e03}
  \end{subfigure} \hfill
  \begin{subfigure}[b]{0.24\linewidth}
    \centering
    \includegraphics[width=\linewidth]{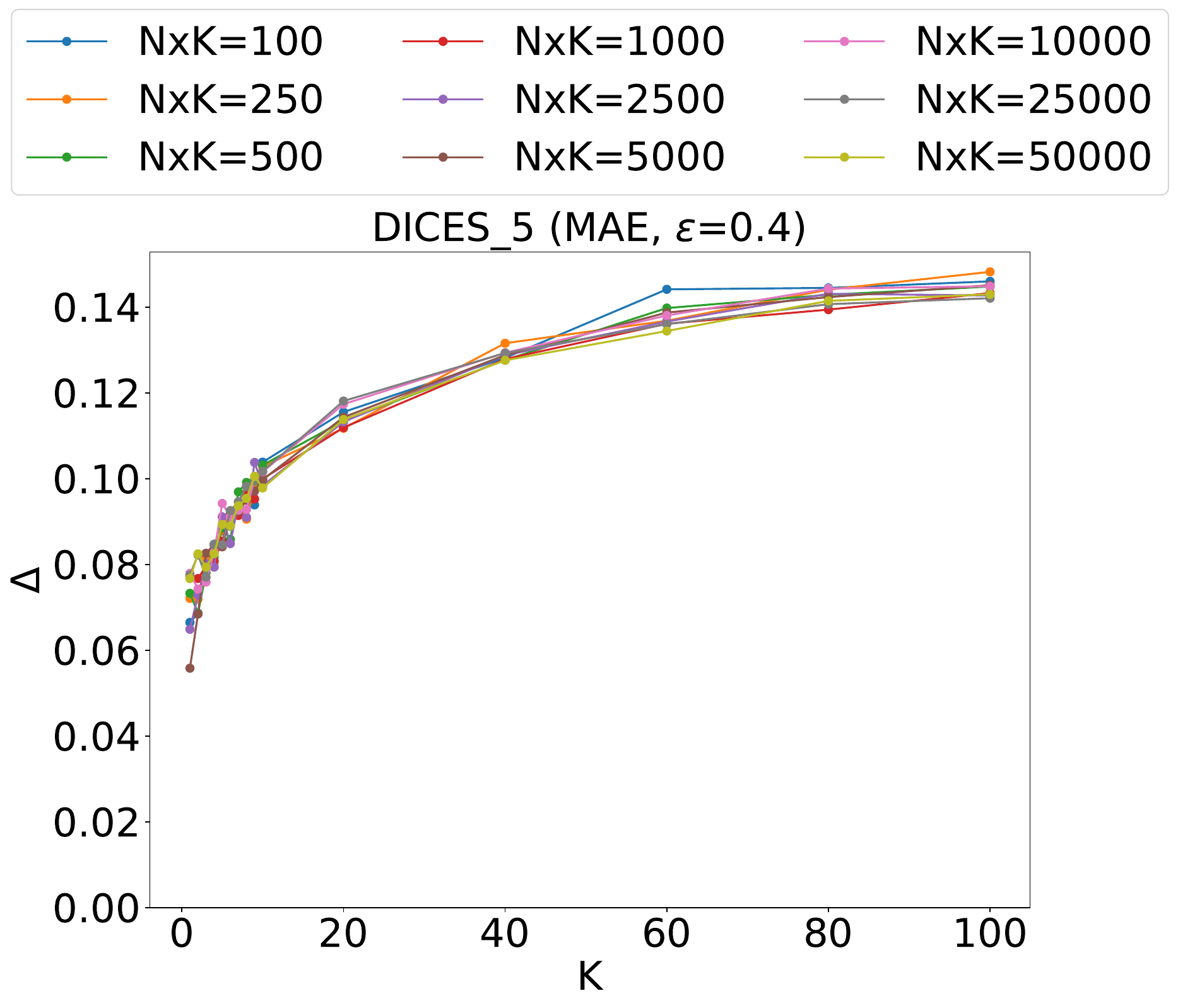}
    \caption{$\epsilon = 0.4$}
    \label{fig:dices_5_delta_MAE_e04}
  \end{subfigure}
  \caption{S1: Effect sizes ($\Delta$) for DICES 5 rater sample with MAE as the metric}
  \label{fig:dices_5_delta_MAE}
\end{figure*}

\begin{figure*}
  \centering
  \begin{subfigure}[b]{0.24\linewidth}
    \centering
    \includegraphics[width=\linewidth]{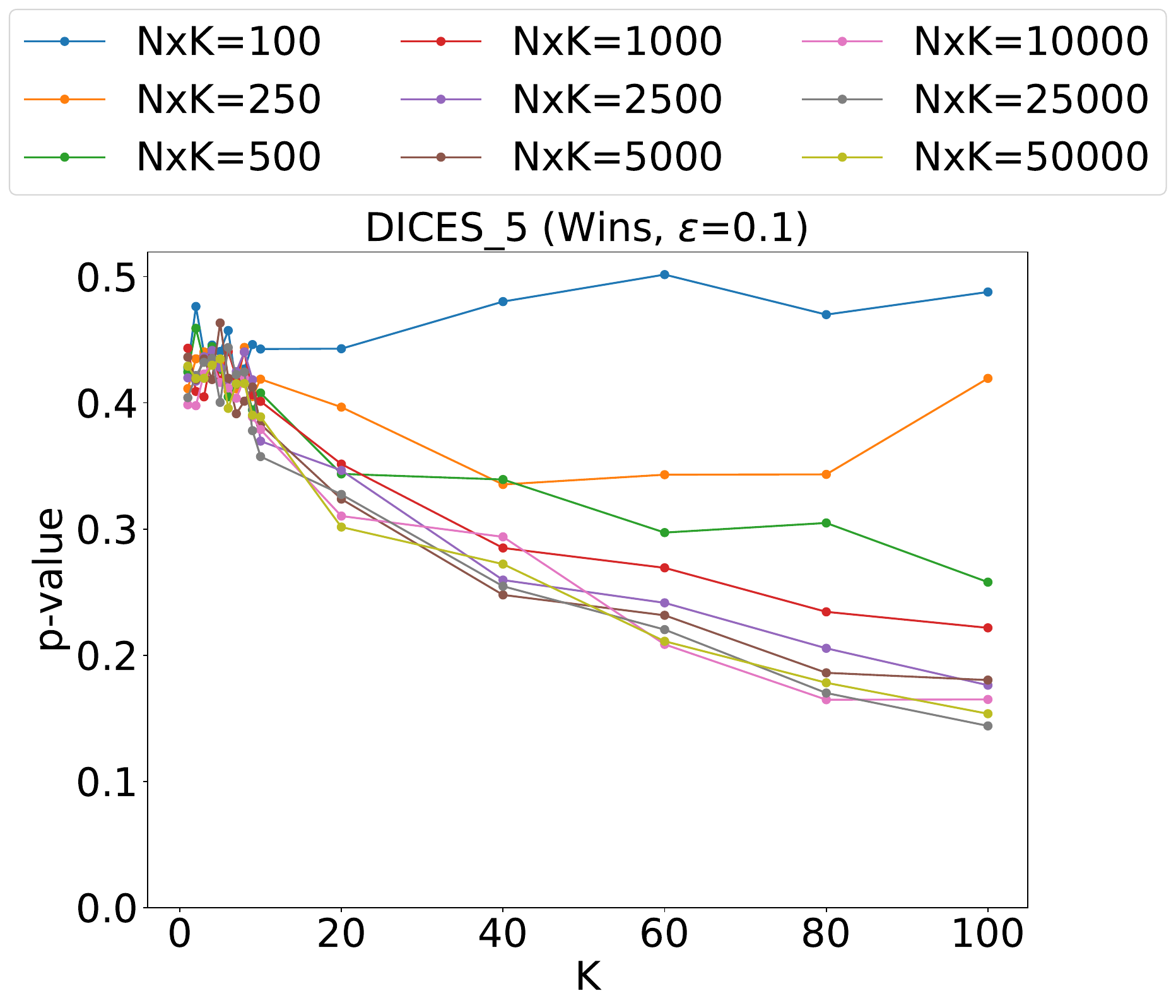}
    \caption{$\epsilon = 0.1$}
    \label{fig:dices_5_wins_e01}
  \end{subfigure} \hfill
  \begin{subfigure}[b]{0.24\linewidth}
    \centering
    \includegraphics[width=\linewidth]{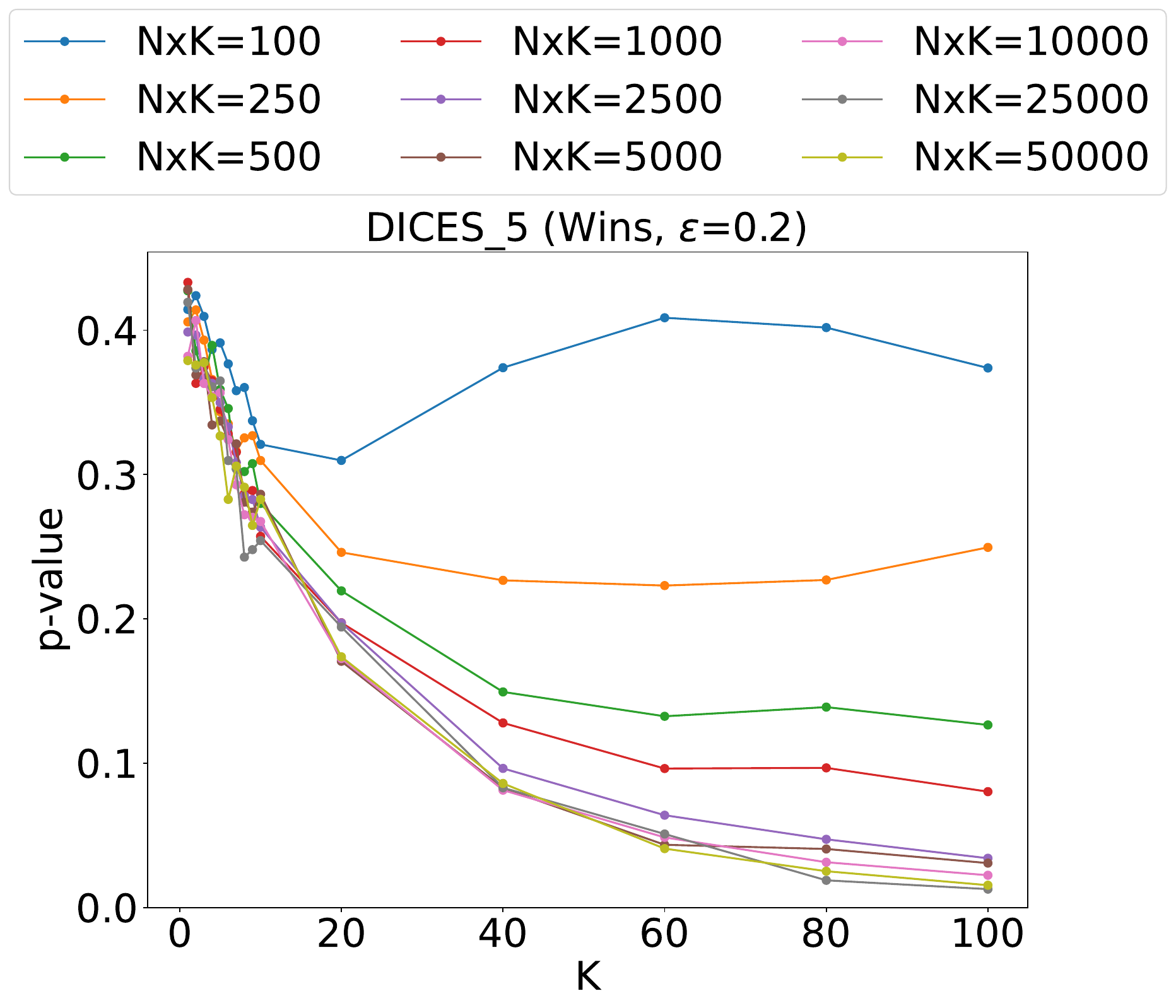}
    \caption{$\epsilon = 0.2$}
    \label{fig:dices_5_wins_e02}
  \end{subfigure} \hfill
  \begin{subfigure}[b]{0.24\linewidth}
    \centering
    \includegraphics[width=\linewidth]{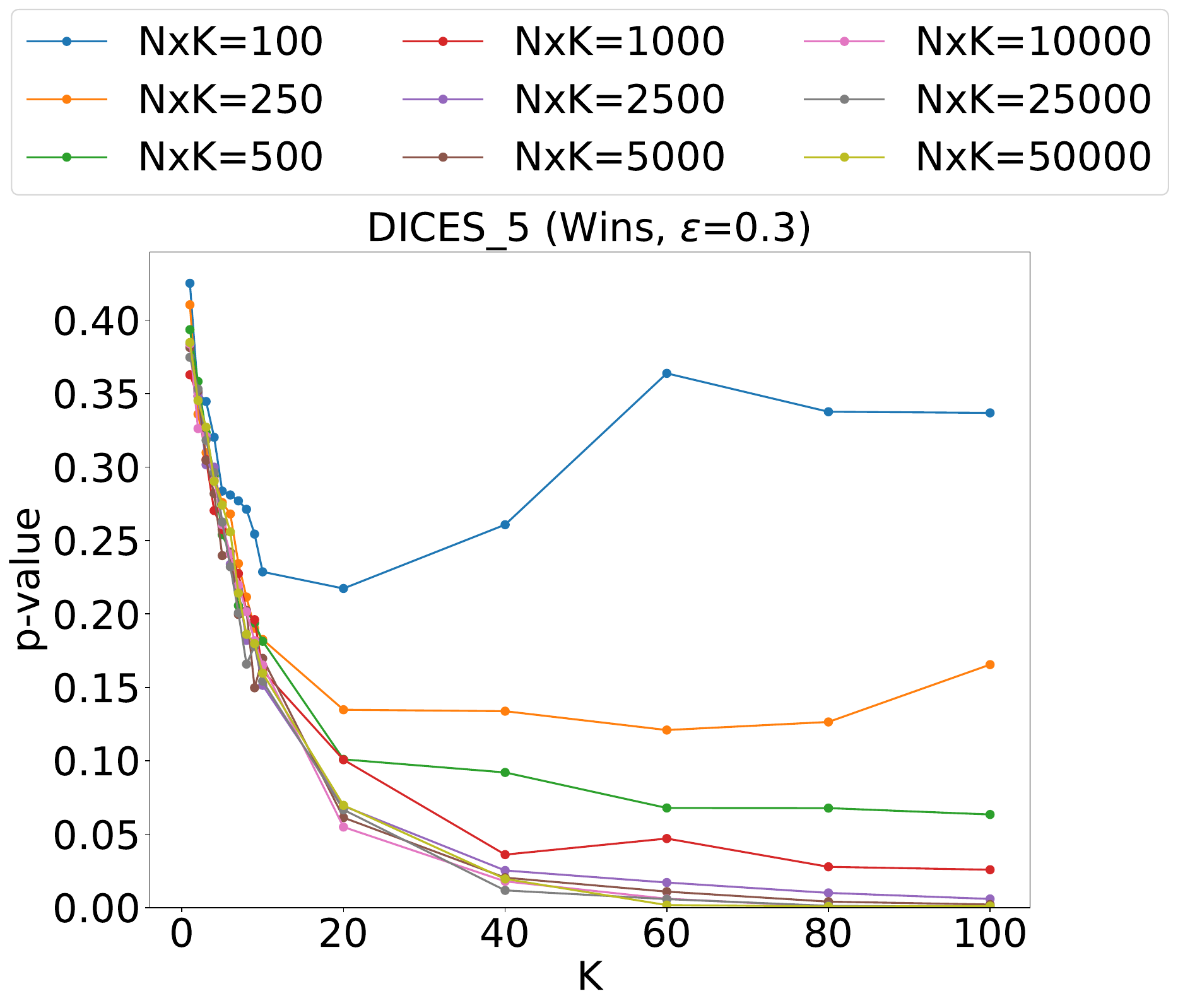}
    \caption{$\epsilon = 0.3$}
    \label{fig:dices_5_wins_e03}
  \end{subfigure} \hfill
  \begin{subfigure}[b]{0.24\linewidth}
    \centering
    \includegraphics[width=\linewidth]{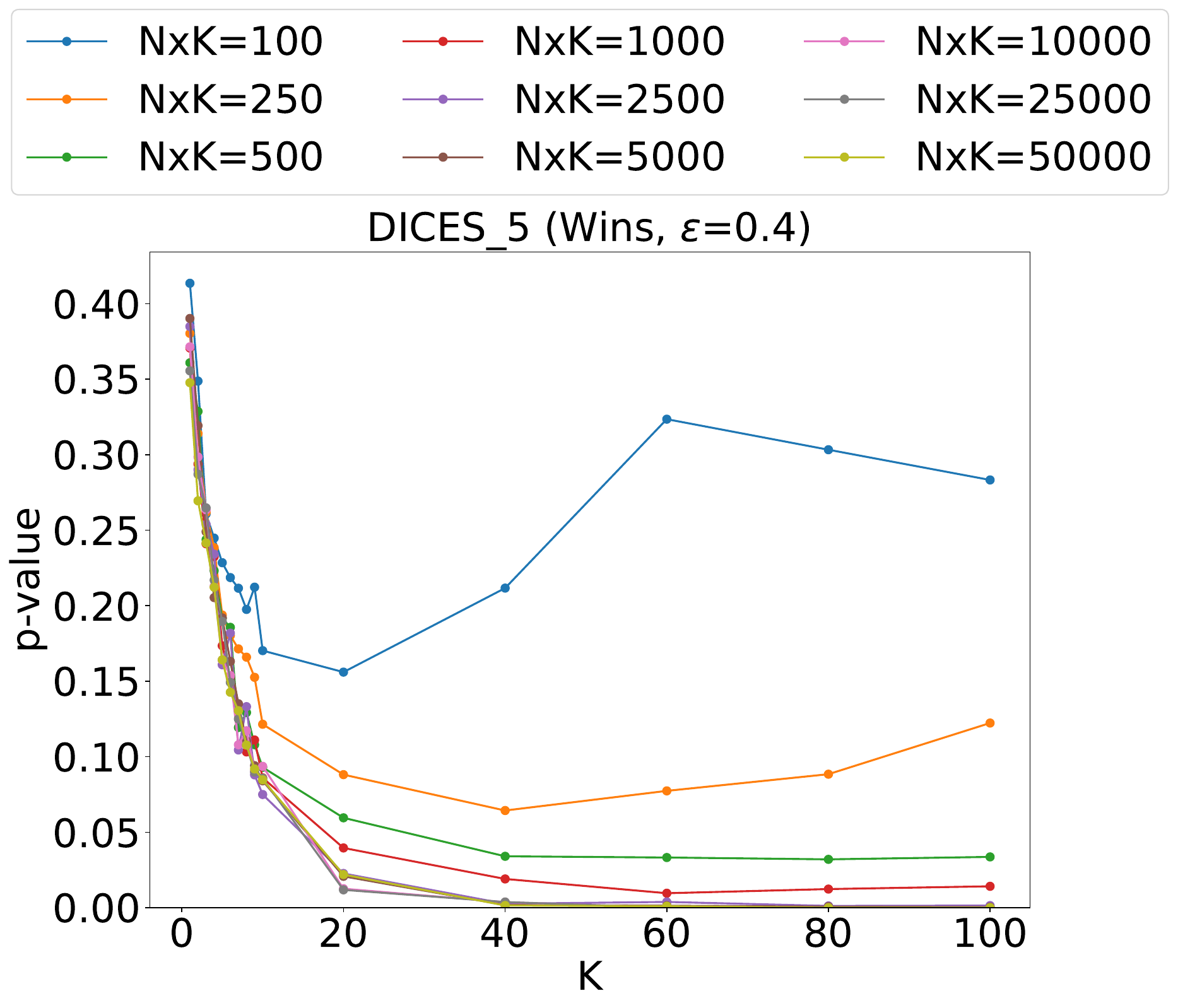}
    \caption{$\epsilon = 0.4$}
    \label{fig:dices_5_wins_e04}
  \end{subfigure}
  \caption{S1: P-value plots for DICES 5 rater sample with Wins as the metric}
  \label{fig:dices_5_wins}
\end{figure*}

\begin{figure*}
  \centering
  \begin{subfigure}[b]{0.24\linewidth}
    \centering
    \includegraphics[width=\linewidth]{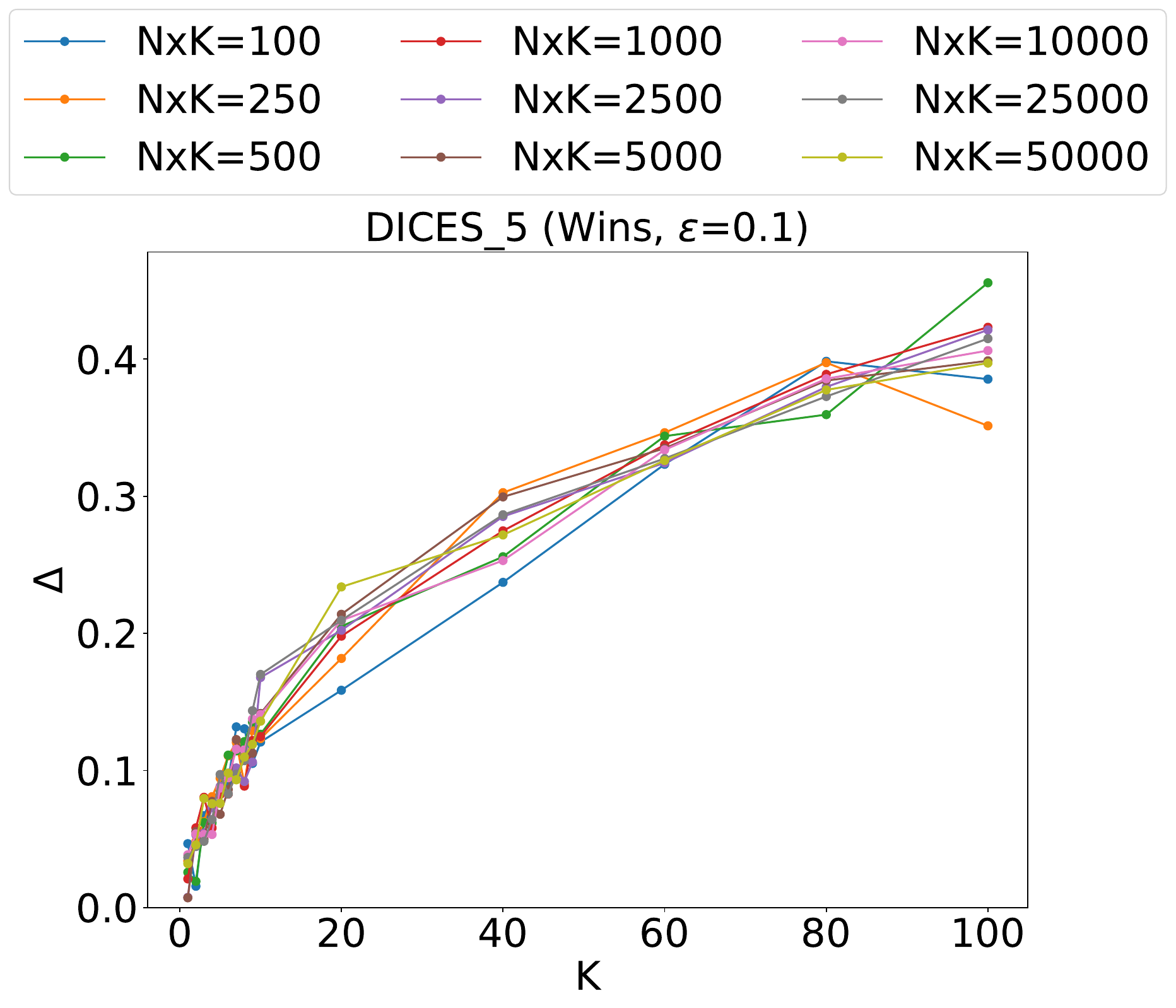}
    \caption{$\epsilon = 0.1$}
    \label{fig:dices_5_delta_wins_e01}
  \end{subfigure} \hfill
  \begin{subfigure}[b]{0.24\linewidth}
    \centering
    \includegraphics[width=\linewidth]{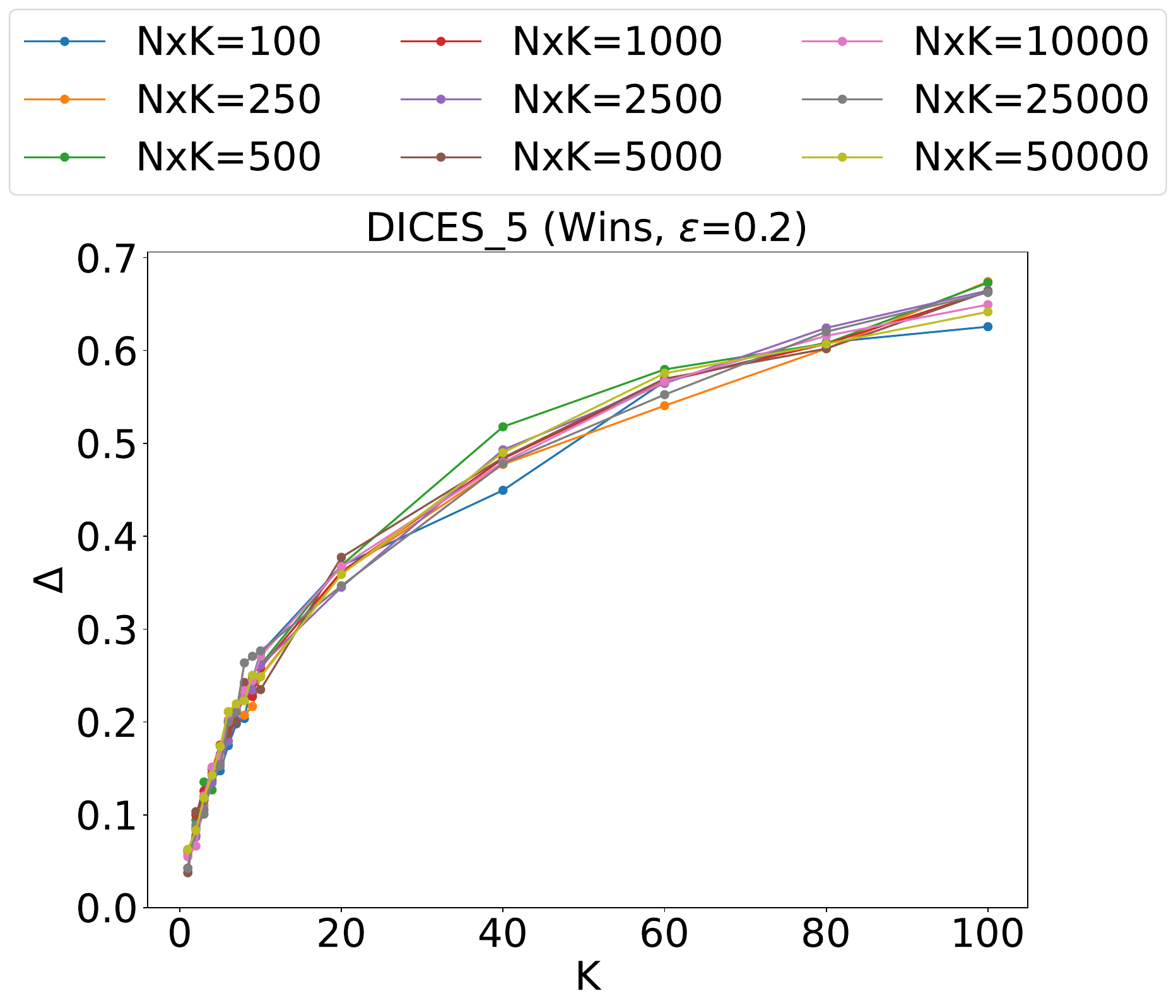}
    \caption{$\epsilon = 0.2$}
    \label{fig:dices_5_delta_wins_e02}
  \end{subfigure} \hfill
  \begin{subfigure}[b]{0.24\linewidth}
    \centering
    \includegraphics[width=\linewidth]{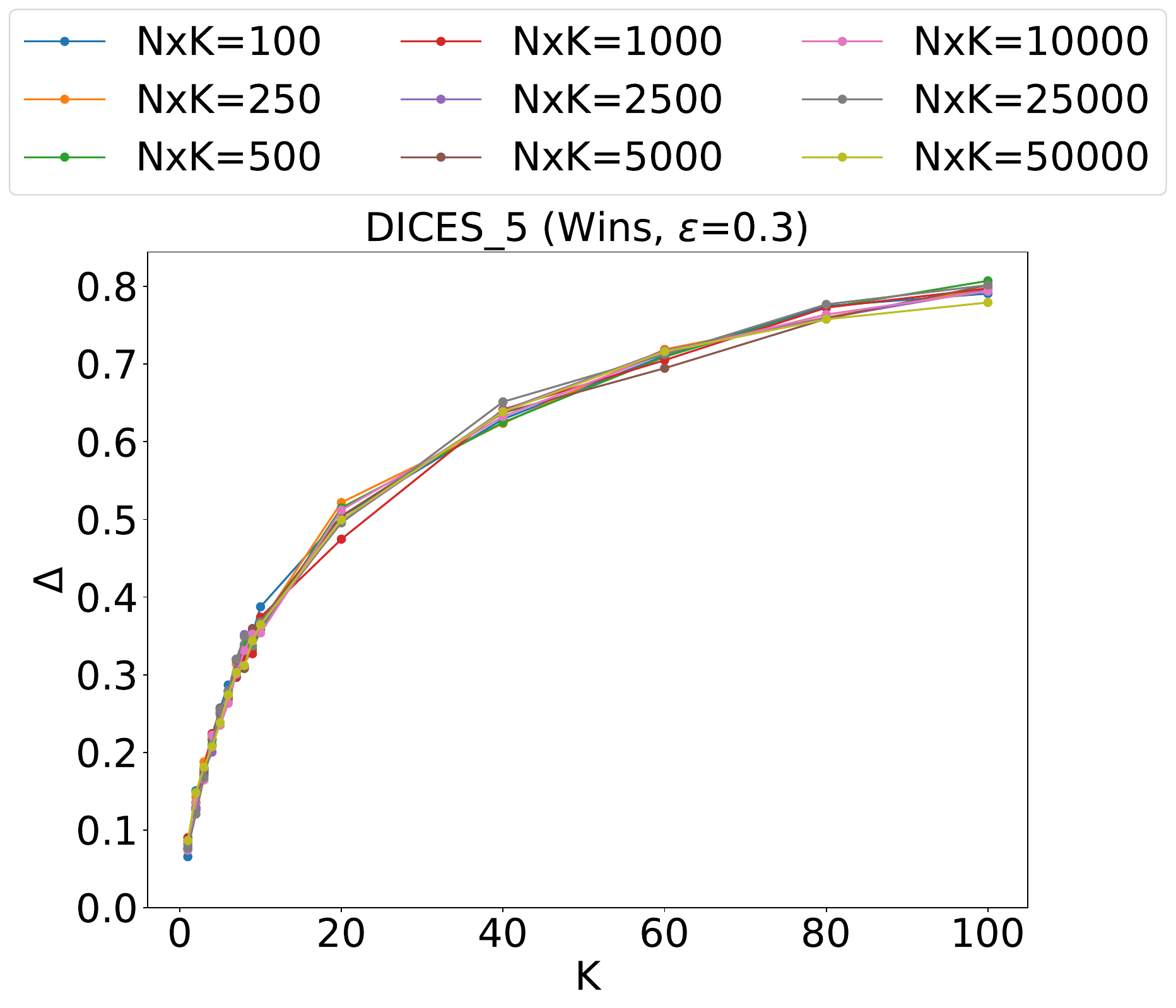}
    \caption{$\epsilon = 0.3$}
    \label{fig:dices_5_delta_wins_e03}
  \end{subfigure} \hfill
  \begin{subfigure}[b]{0.24\linewidth}
    \centering
    \includegraphics[width=\linewidth]{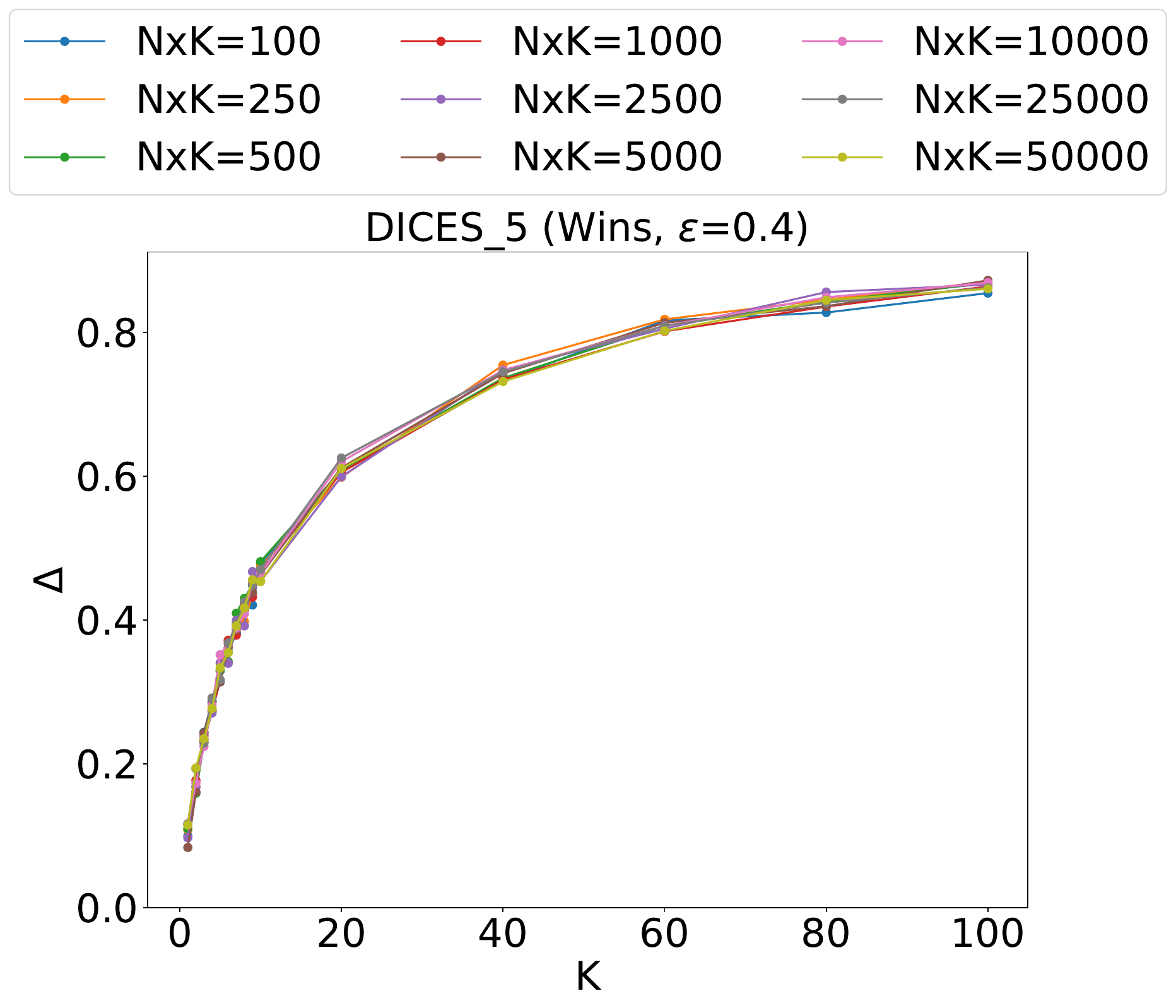}
    \caption{$\epsilon = 0.4$}
    \label{fig:dices_5_delta_wins_e04}
  \end{subfigure}
  \caption{S1: Effect sizes ($\Delta$) for DICES 5 rater sample with Wins as the metric}
  \label{fig:dices_5_delta_wins}
\end{figure*}

\subsection{S2}

\paragraph{DICES}

\begin{figure*}
  \centering
  \begin{subfigure}[b]{0.24\linewidth}
    \centering
    \includegraphics[width=\linewidth]{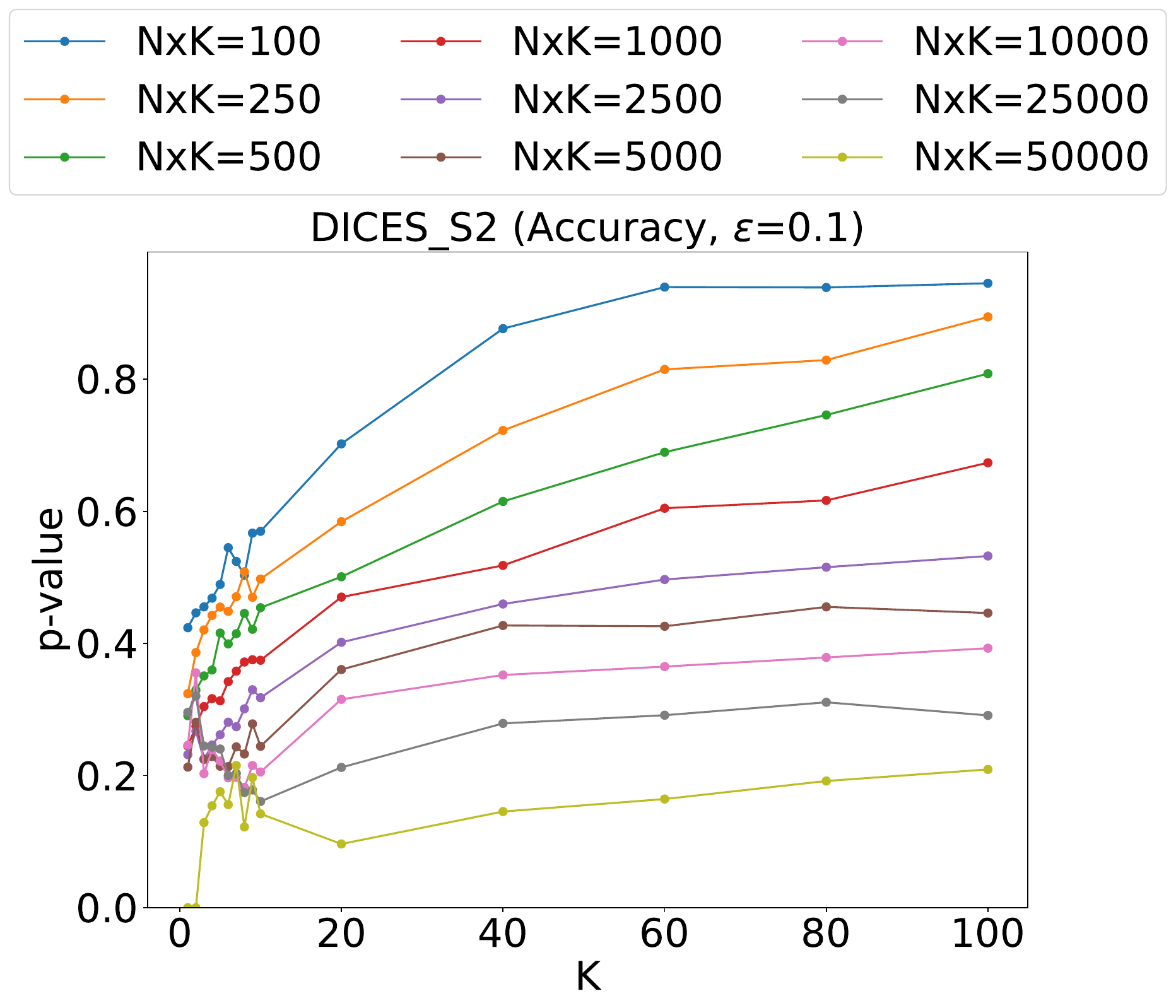}
    \caption{$\epsilon = 0.1$}
    \label{fig:dices_s2_acc_e01}
  \end{subfigure} \hfill
  \begin{subfigure}[b]{0.24\linewidth}
    \centering
    \includegraphics[width=\linewidth]{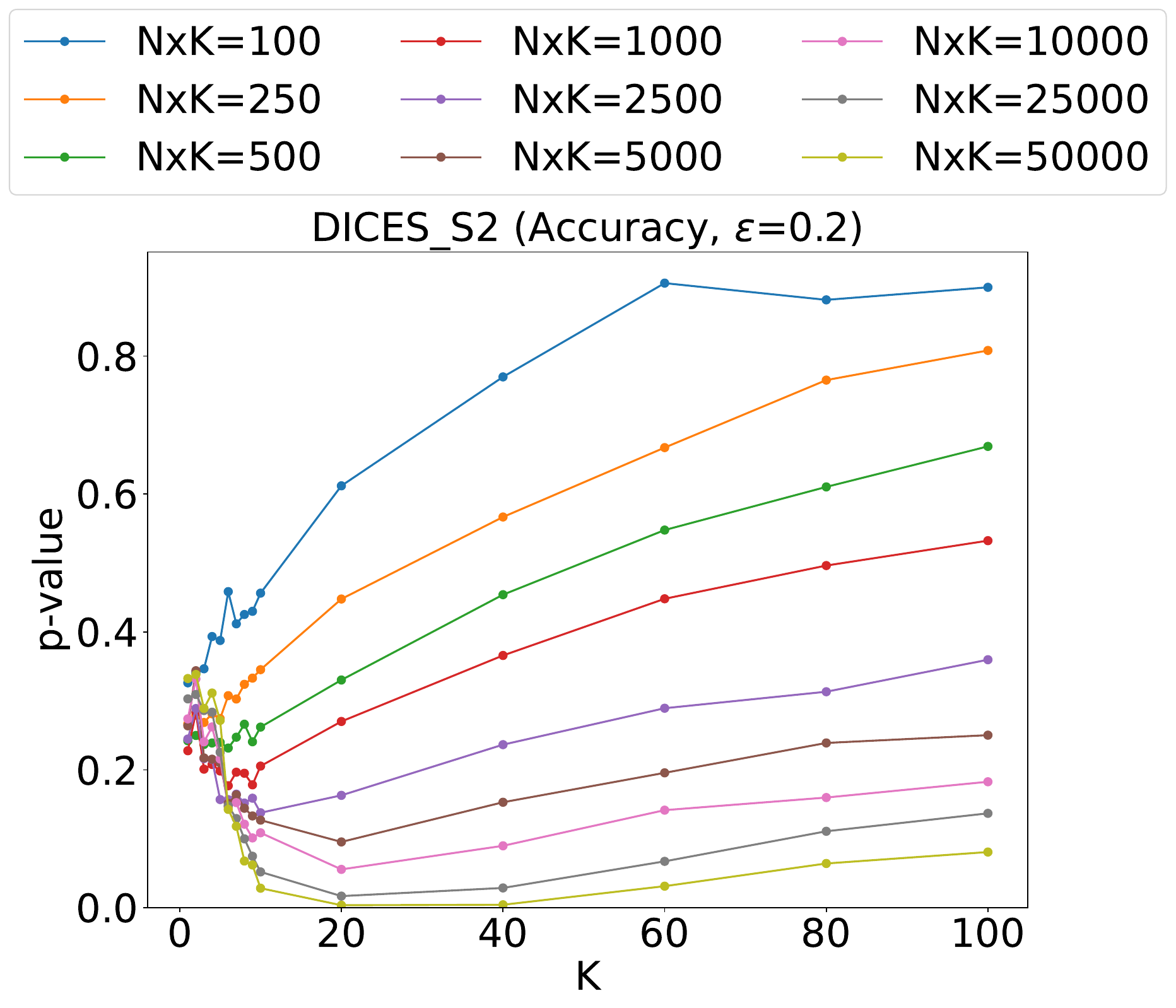}
    \caption{$\epsilon = 0.2$}
    \label{fig:dices_s2_acc_e02}
  \end{subfigure} \hfill
  \begin{subfigure}[b]{0.24\linewidth}
    \centering
    \includegraphics[width=\linewidth]{figures/pvals_plots/DICES_S2/DICES_S2_p_vals_Accuracy_K_100_e_0.3.pdf}
    \caption{$\epsilon = 0.3$}
    \label{fig:dices_s2_acc_e03}
  \end{subfigure} \hfill
  \begin{subfigure}[b]{0.24\linewidth}
    \centering
    \includegraphics[width=\linewidth]{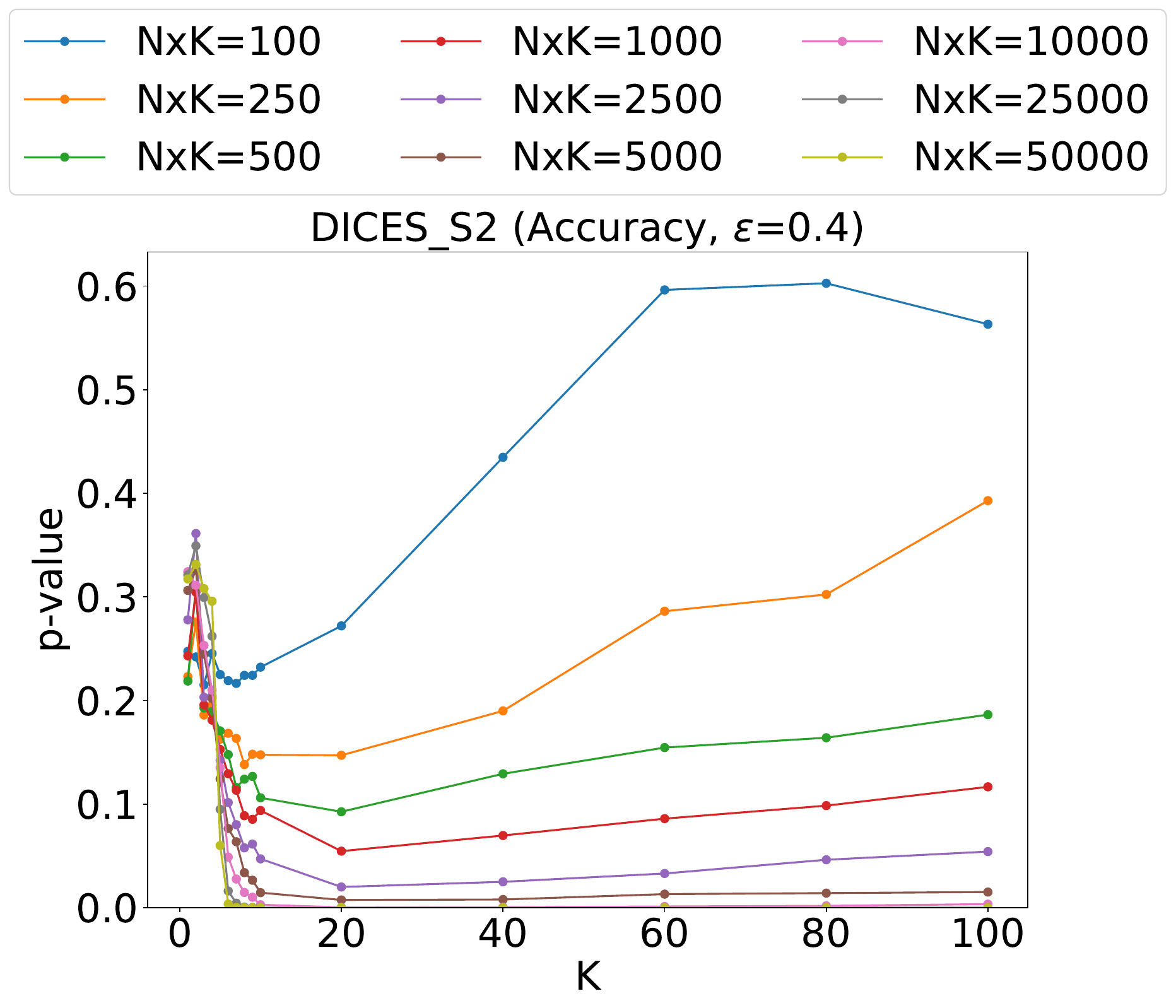}
    \caption{$\epsilon = 0.4$}
    \label{fig:dices_s2_acc_e04}
  \end{subfigure}
  \caption{S2: P-value plots for DICES dataset with Accuracy as the metric}
  \label{fig:dices_s2_accuracy}
\end{figure*}

\begin{figure*}
  \centering
  \begin{subfigure}[b]{0.24\linewidth}
    \centering
    \includegraphics[width=\linewidth]{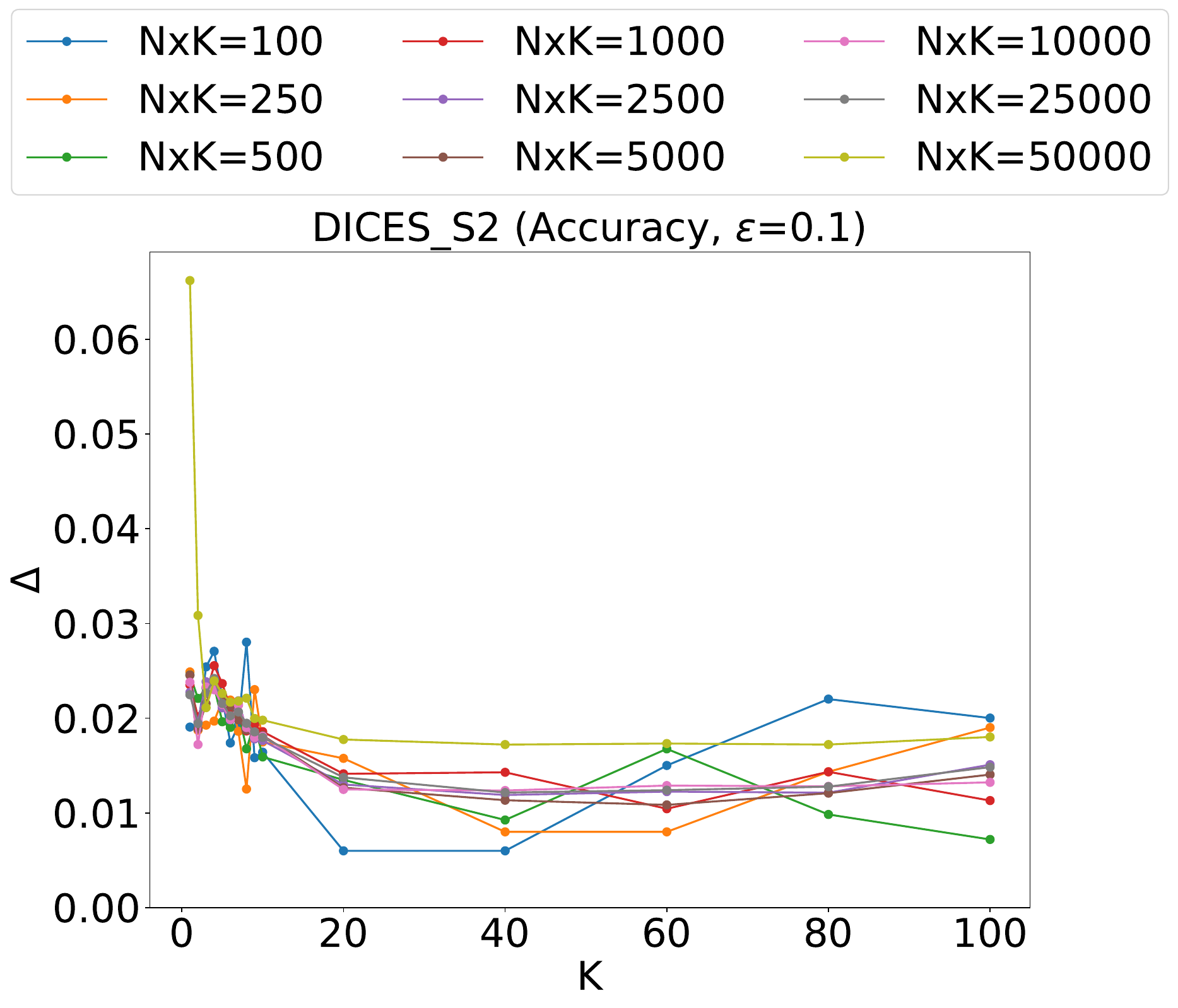}
    \caption{$\epsilon = 0.1$}
    \label{fig:dices_s2_delta_acc_e01}
  \end{subfigure} \hfill
  \begin{subfigure}[b]{0.24\linewidth}
    \centering
    \includegraphics[width=\linewidth]{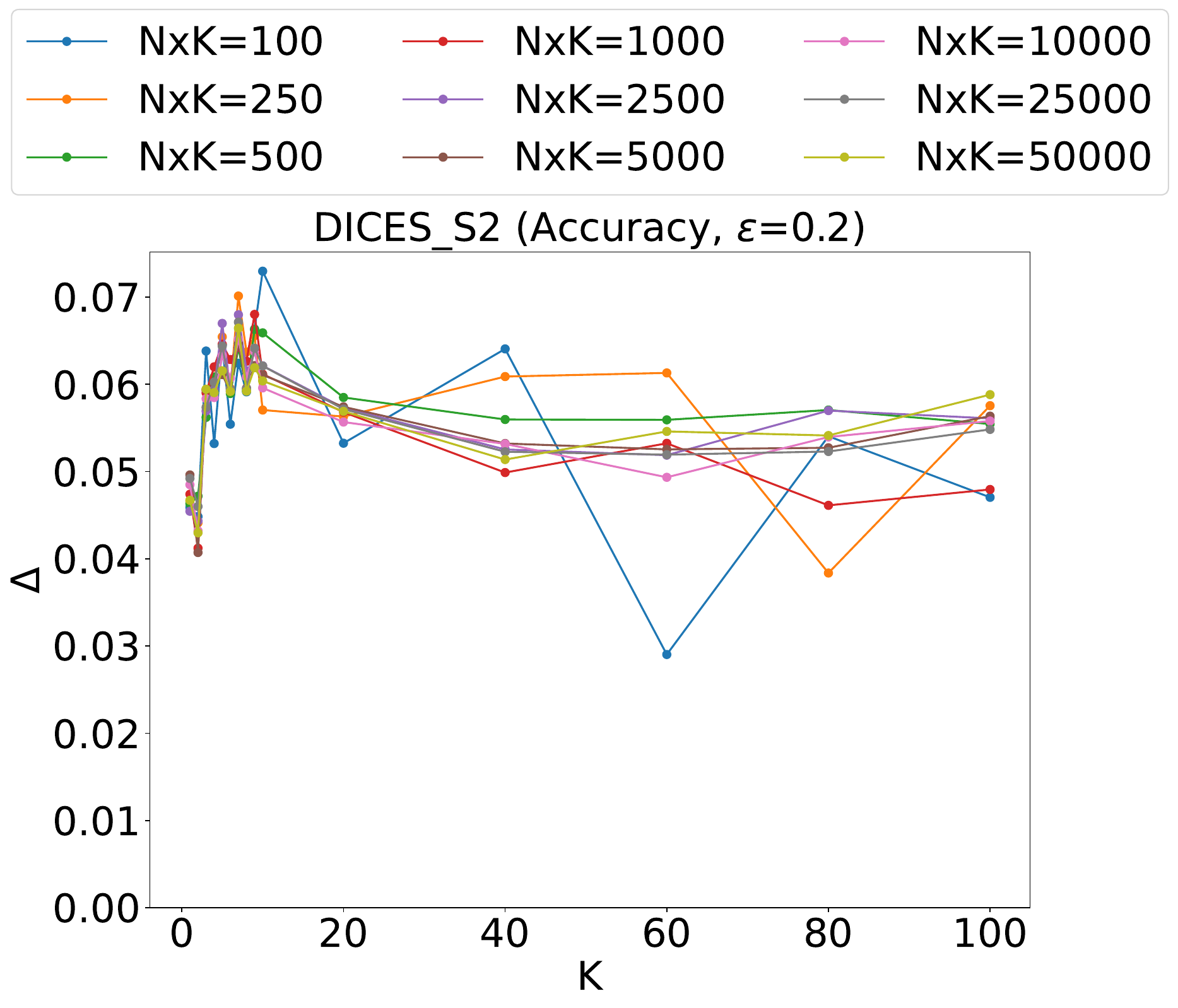}
    \caption{$\epsilon = 0.2$}
    \label{fig:dices_s2_delta_acc_e02}
  \end{subfigure} \hfill
  \begin{subfigure}[b]{0.24\linewidth}
    \centering
    \includegraphics[width=\linewidth]{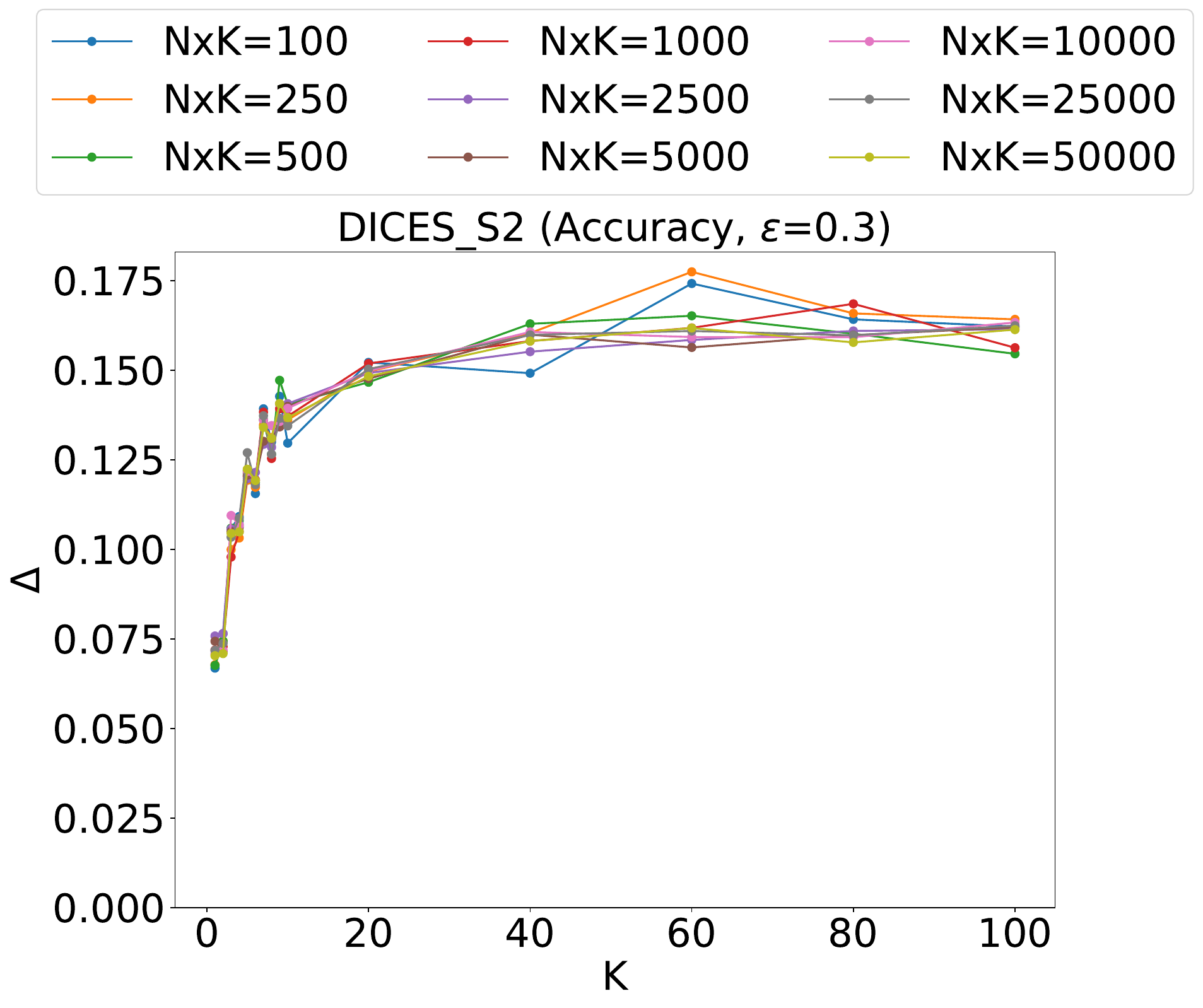}
    \caption{$\epsilon = 0.3$}
    \label{fig:dices_s2_delta_acc_e03}
  \end{subfigure} \hfill
  \begin{subfigure}[b]{0.24\linewidth}
    \centering
    \includegraphics[width=\linewidth]{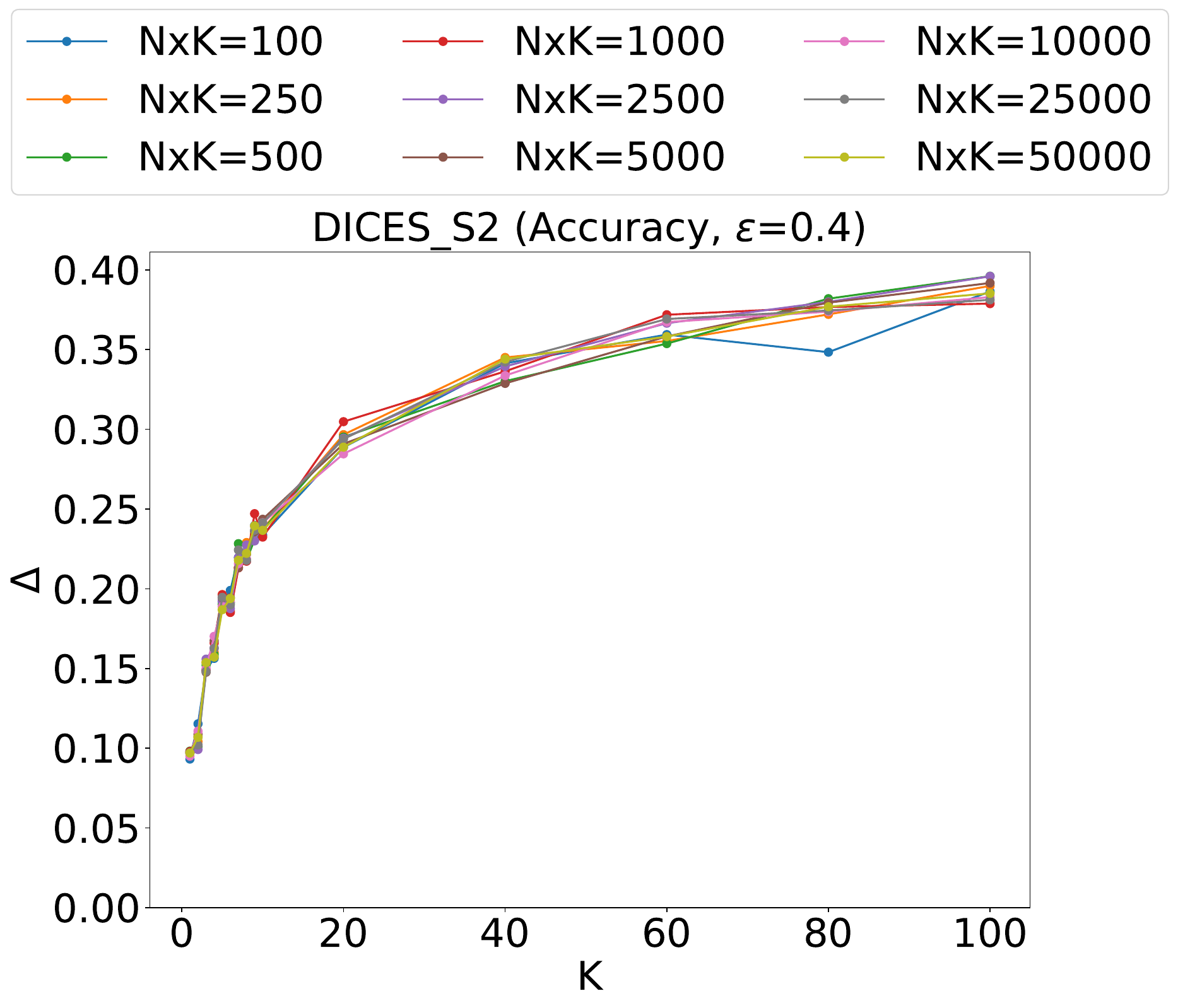}
    \caption{$\epsilon = 0.4$}
    \label{fig:dices_s2_delta_acc_e04}
  \end{subfigure}
  \caption{S2: Effect sizes ($\Delta$) for DICES dataset with Accuracy as the metric}
  \label{fig:dices_s2_delta_accuracy}
\end{figure*}

\begin{figure*}
  \centering
  \begin{subfigure}[b]{0.24\linewidth}
    \centering
    \includegraphics[width=\linewidth]{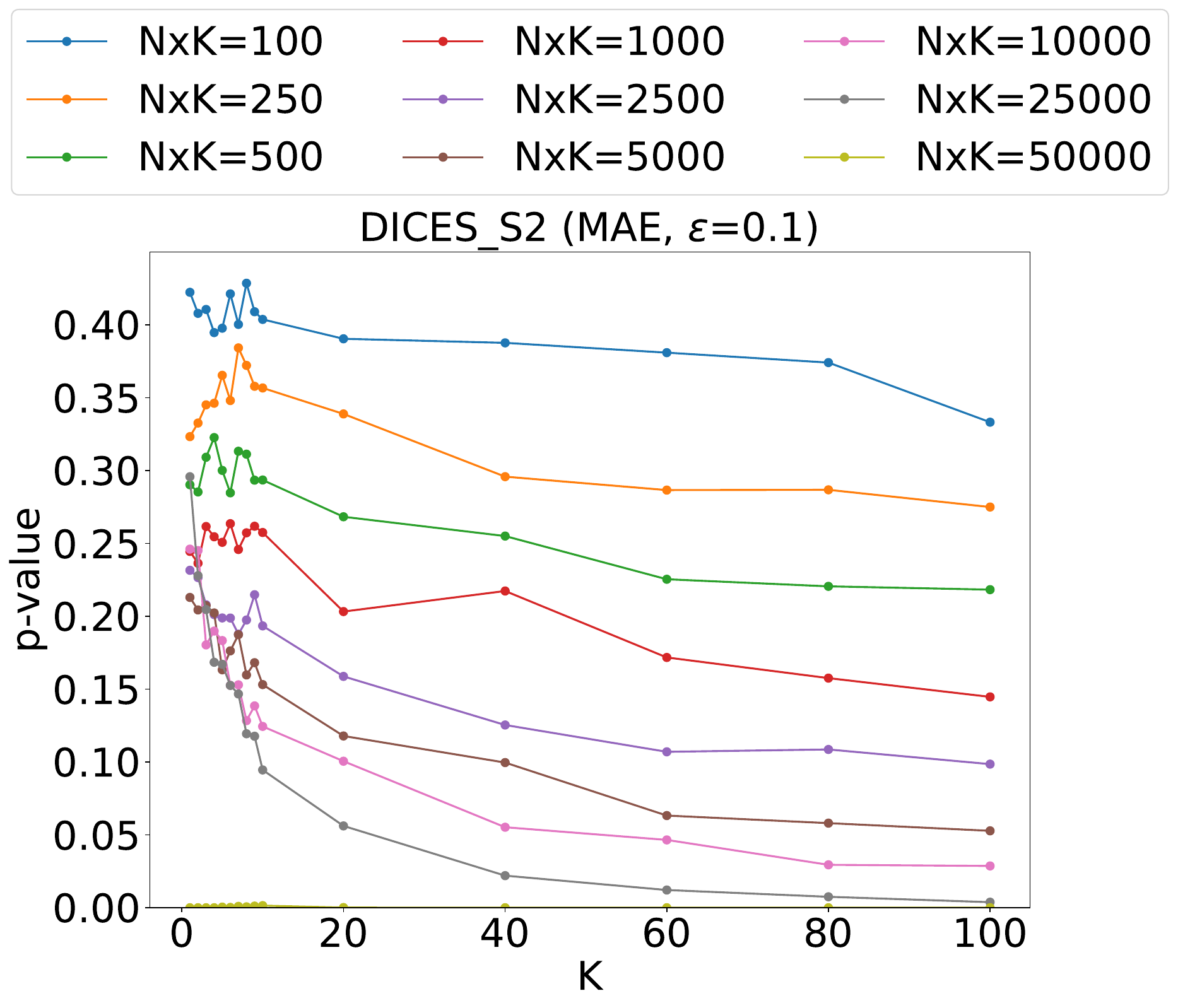}
    \caption{$\epsilon = 0.1$}
    \label{fig:dices_s2_MAE_e01}
  \end{subfigure} \hfill
  \begin{subfigure}[b]{0.24\linewidth}
    \centering
    \includegraphics[width=\linewidth]{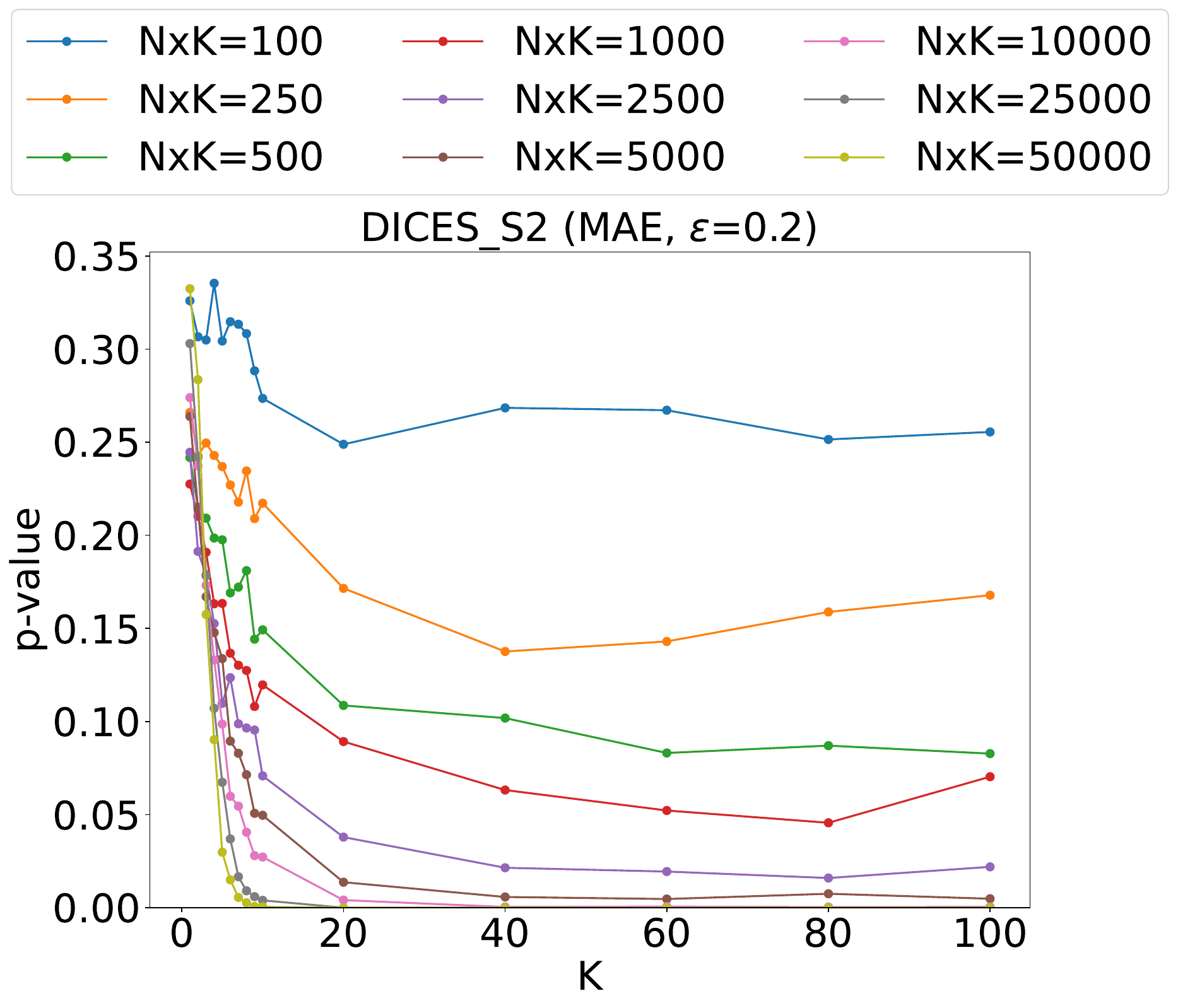}
    \caption{$\epsilon = 0.2$}
    \label{fig:dices_s2_MAE_e02}
  \end{subfigure} \hfill
  \begin{subfigure}[b]{0.24\linewidth}
    \centering
    \includegraphics[width=\linewidth]{figures/pvals_plots/DICES_S2/DICES_S2_p_vals_MAE_K_100_e_0.3.pdf}
    \caption{$\epsilon = 0.3$}
    \label{fig:dices_s2_MAE_e03}
  \end{subfigure} \hfill
  \begin{subfigure}[b]{0.24\linewidth}
    \centering
    \includegraphics[width=\linewidth]{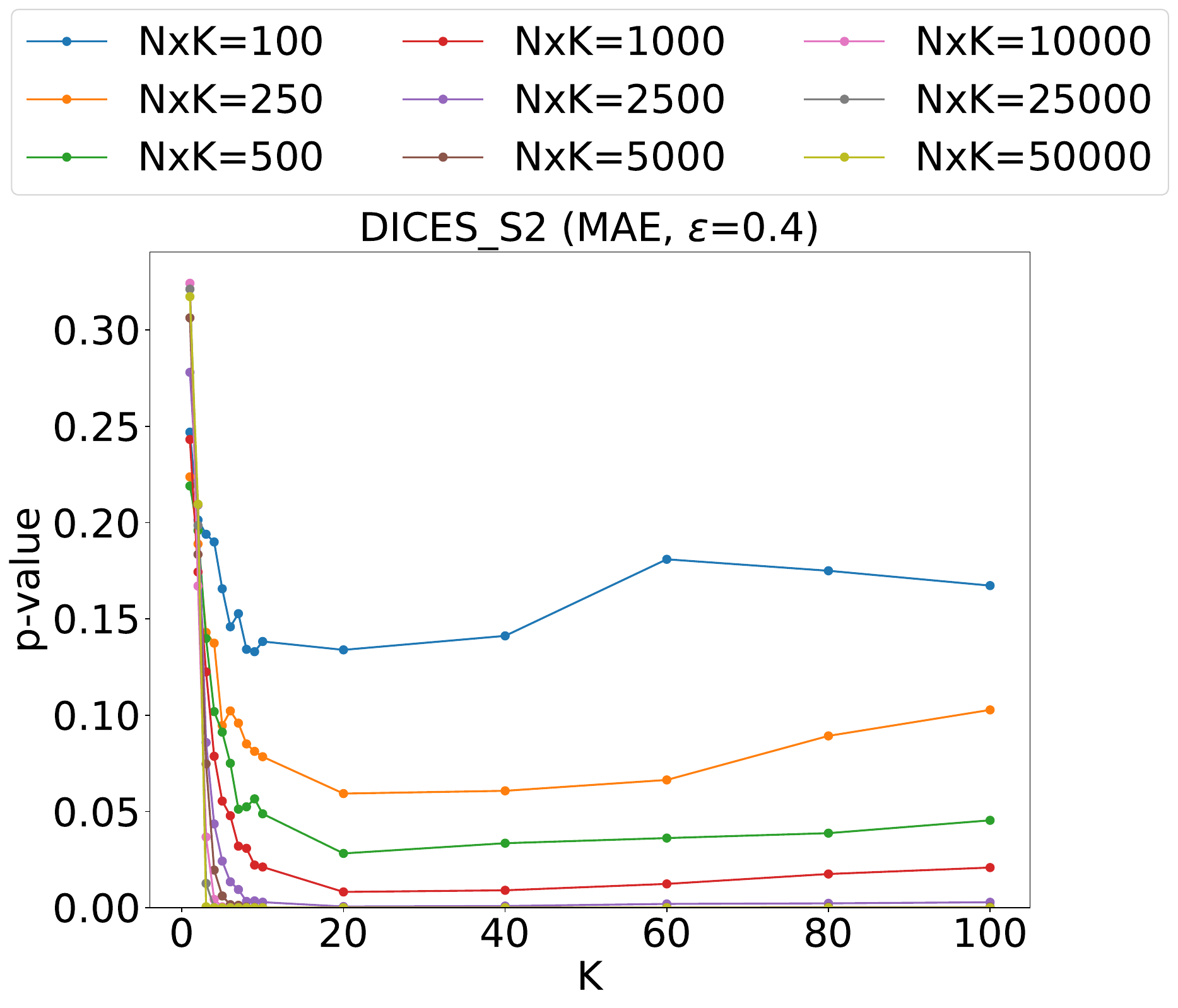}
    \caption{$\epsilon = 0.4$}
    \label{fig:dices_s2_MAE_e04}
  \end{subfigure}
  \caption{S2: P-value plots for DICES dataset with MAE as the metric}
  \label{fig:dices_s2_MAE}
\end{figure*}

\begin{figure*}
  \centering
  \begin{subfigure}[b]{0.24\linewidth}
    \centering
    \includegraphics[width=\linewidth]{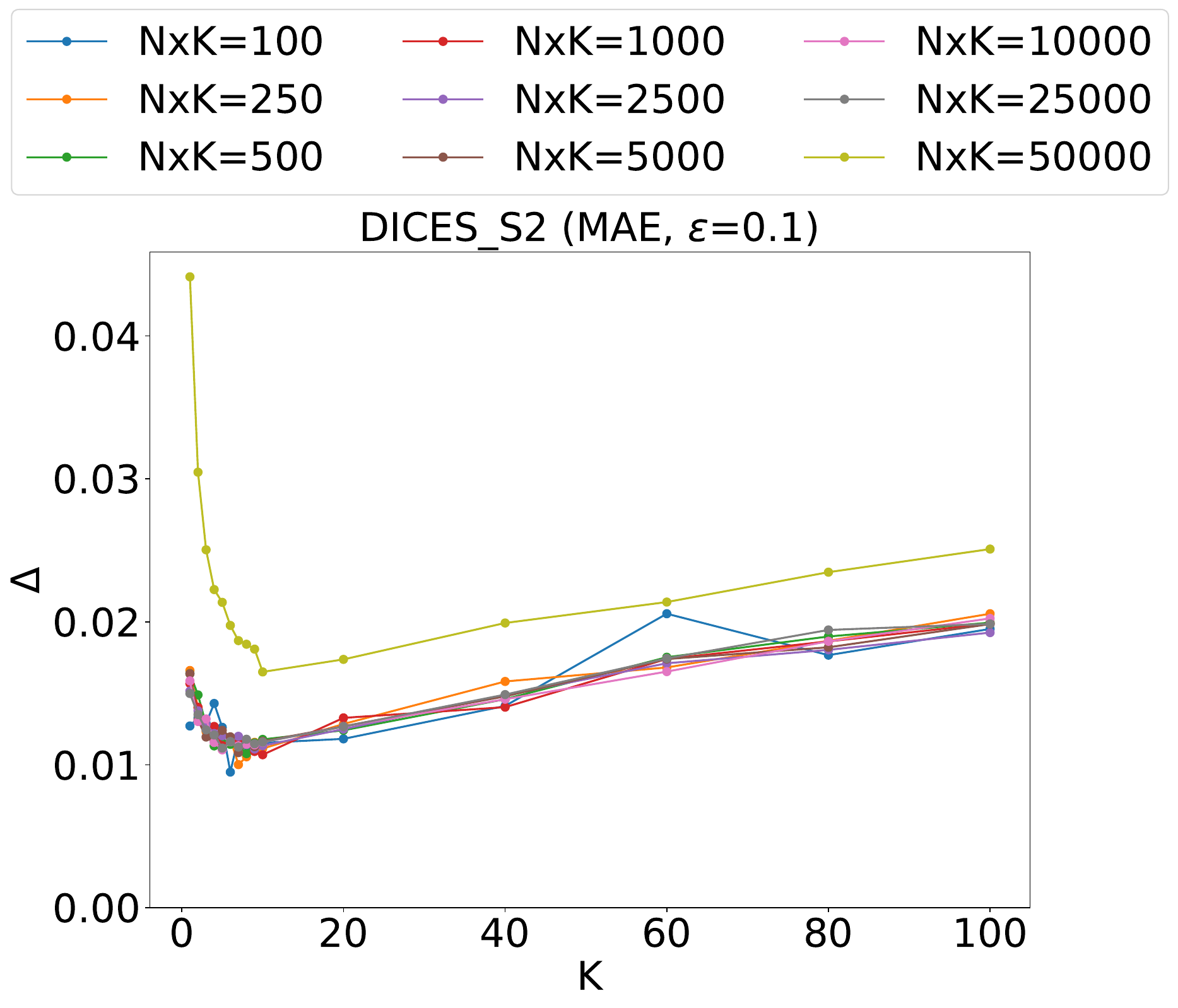}
    \caption{$\epsilon = 0.1$}
    \label{fig:dices_s2_delta_MAE_e01}
  \end{subfigure} \hfill
  \begin{subfigure}[b]{0.24\linewidth}
    \centering
    \includegraphics[width=\linewidth]{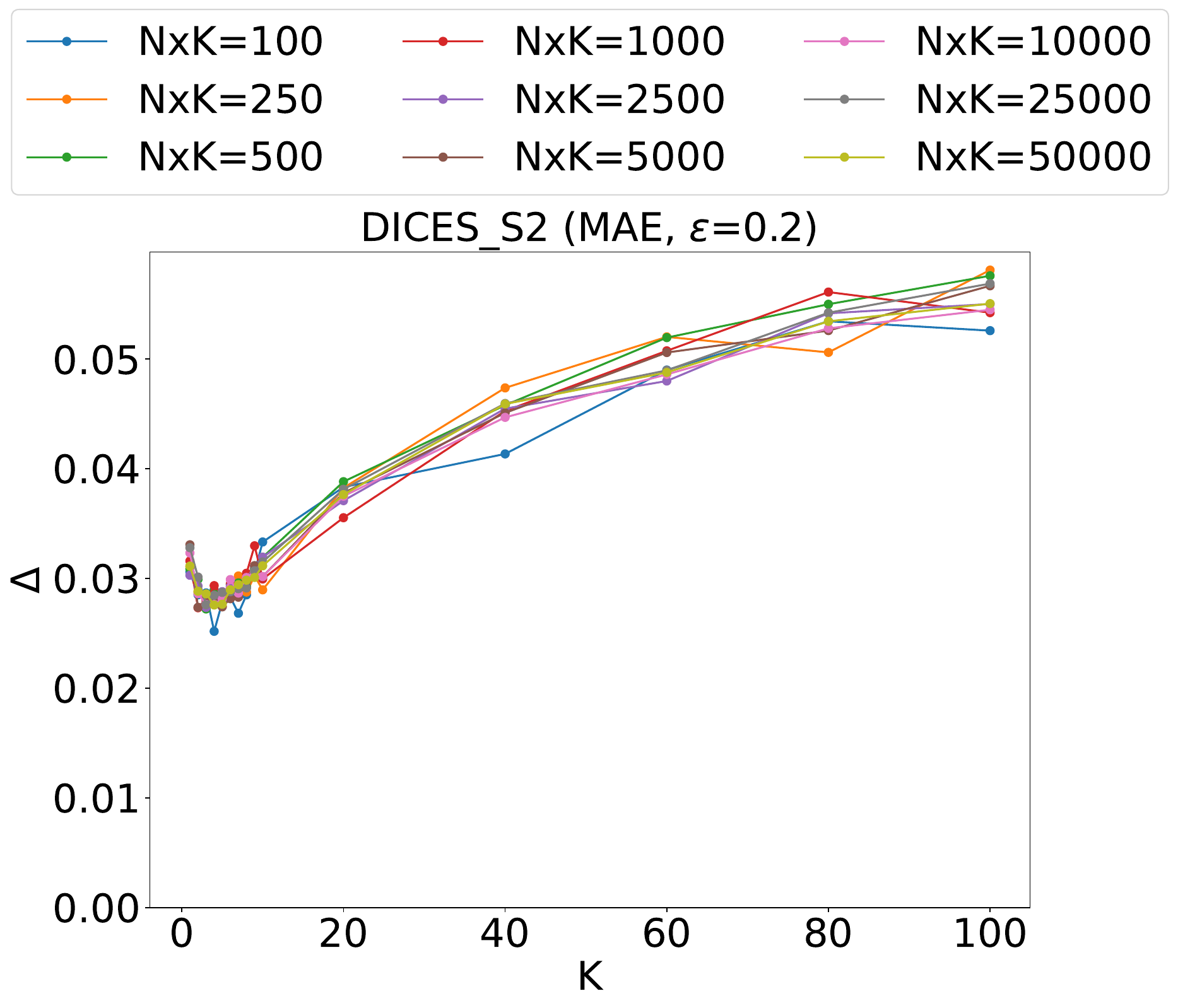}
    \caption{$\epsilon = 0.2$}
    \label{fig:dices_s2_delta_MAE_e02}
  \end{subfigure} \hfill
  \begin{subfigure}[b]{0.24\linewidth}
    \centering
    \includegraphics[width=\linewidth]{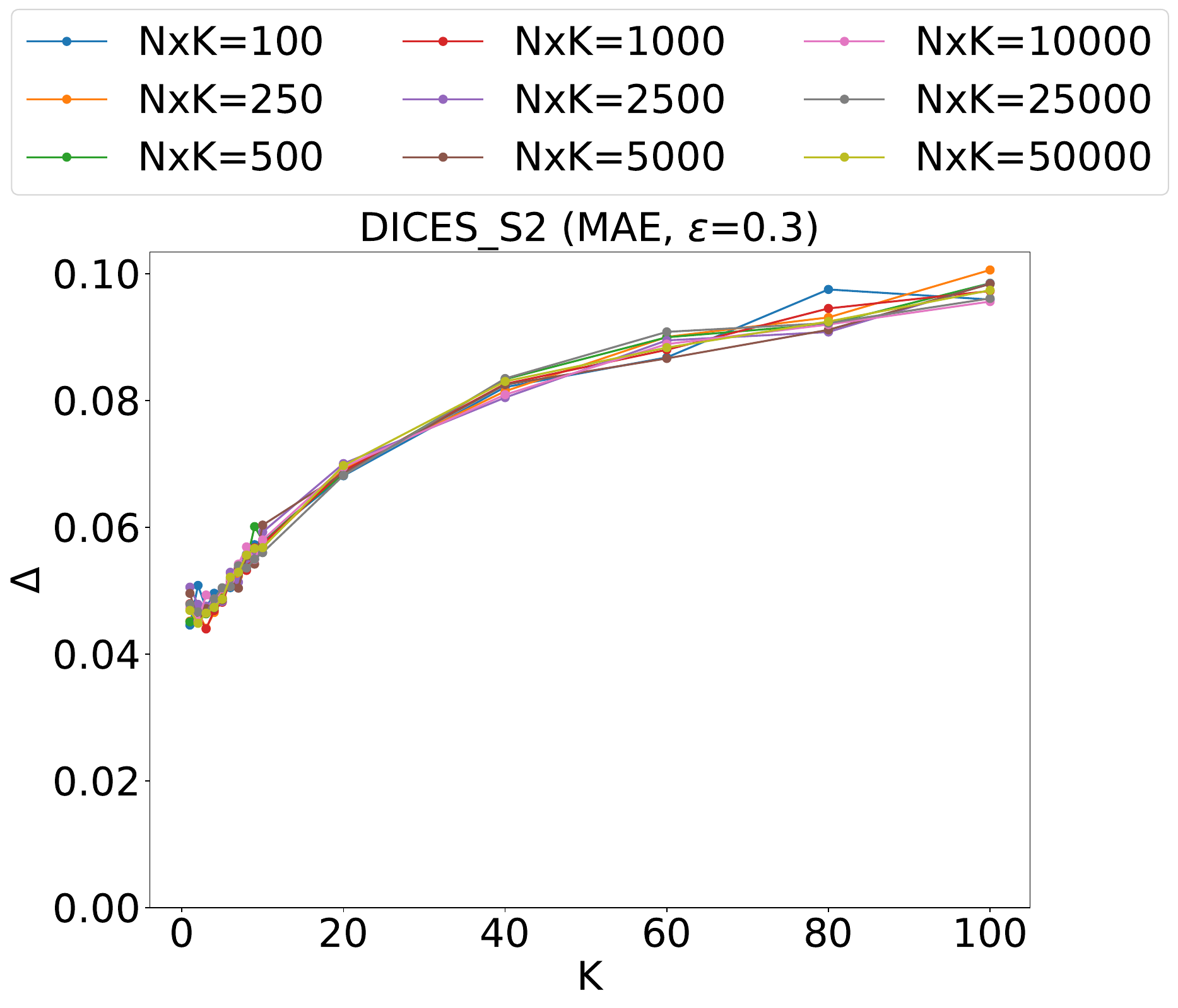}
    \caption{$\epsilon = 0.3$}
    \label{fig:dices_s2_delta_MAE_e03}
  \end{subfigure} \hfill
  \begin{subfigure}[b]{0.24\linewidth}
    \centering
    \includegraphics[width=\linewidth]{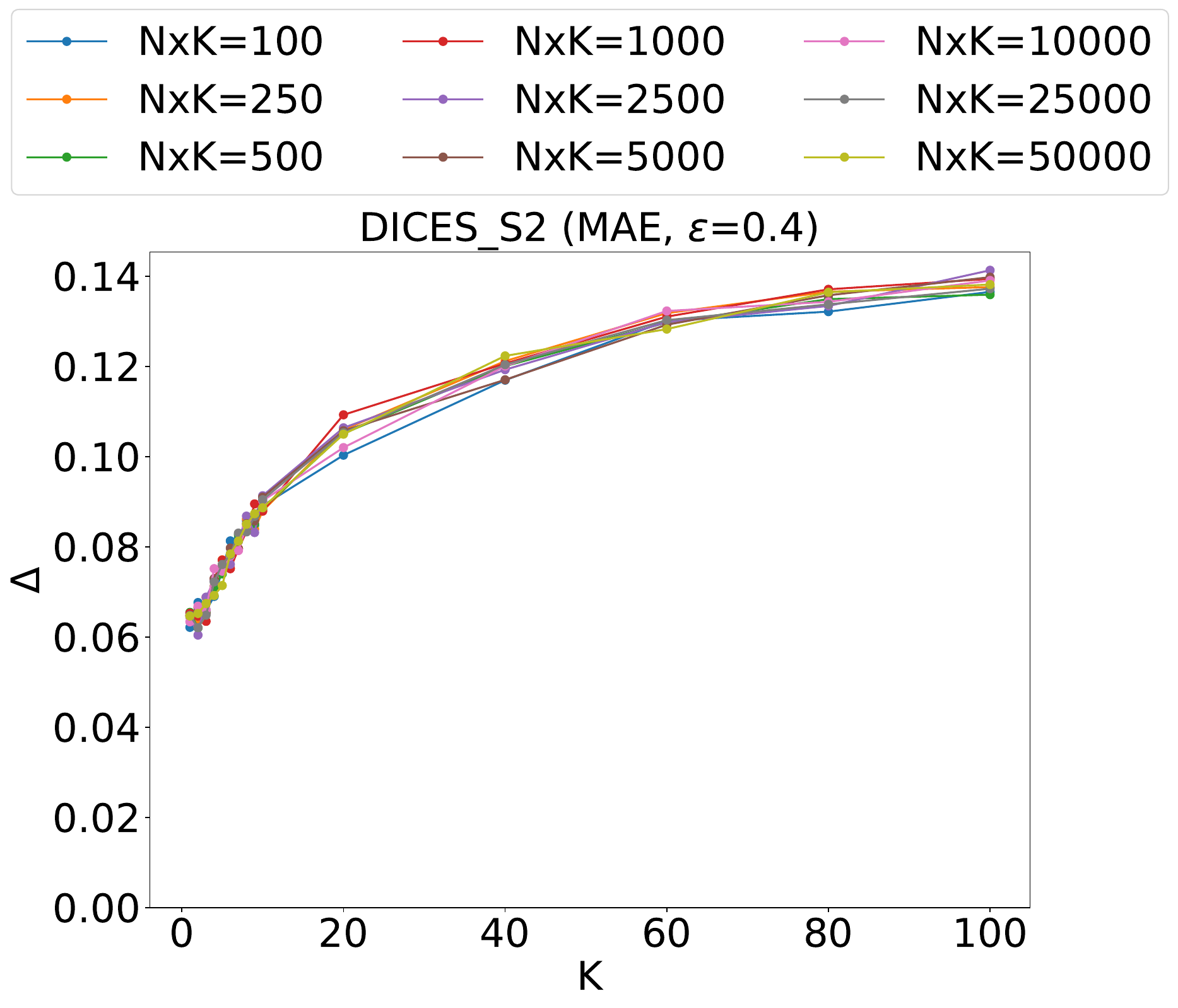}
    \caption{$\epsilon = 0.4$}
    \label{fig:dices_s2_delta_MAE_e04}
  \end{subfigure}
  \caption{S2: Effect sizes ($\Delta$) for DICES dataset with MAE as the metric}
  \label{fig:dices_s2_delta_MAE}
\end{figure*}

\begin{figure*}
  \centering
  \begin{subfigure}[b]{0.24\linewidth}
    \centering
    \includegraphics[width=\linewidth]{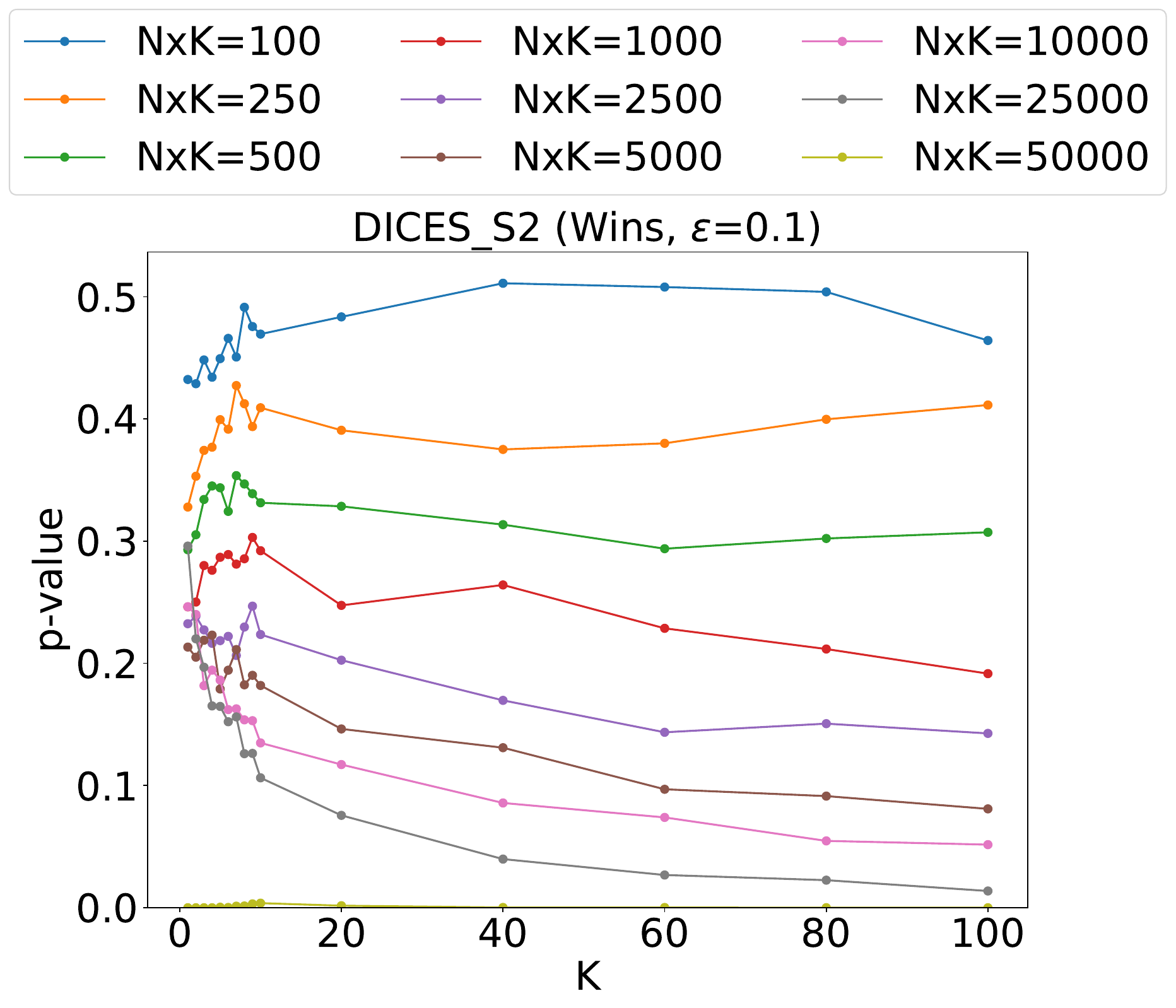}
    \caption{$\epsilon = 0.1$}
    \label{fig:dices_s2_wins_e01}
  \end{subfigure} \hfill
  \begin{subfigure}[b]{0.24\linewidth}
    \centering
    \includegraphics[width=\linewidth]{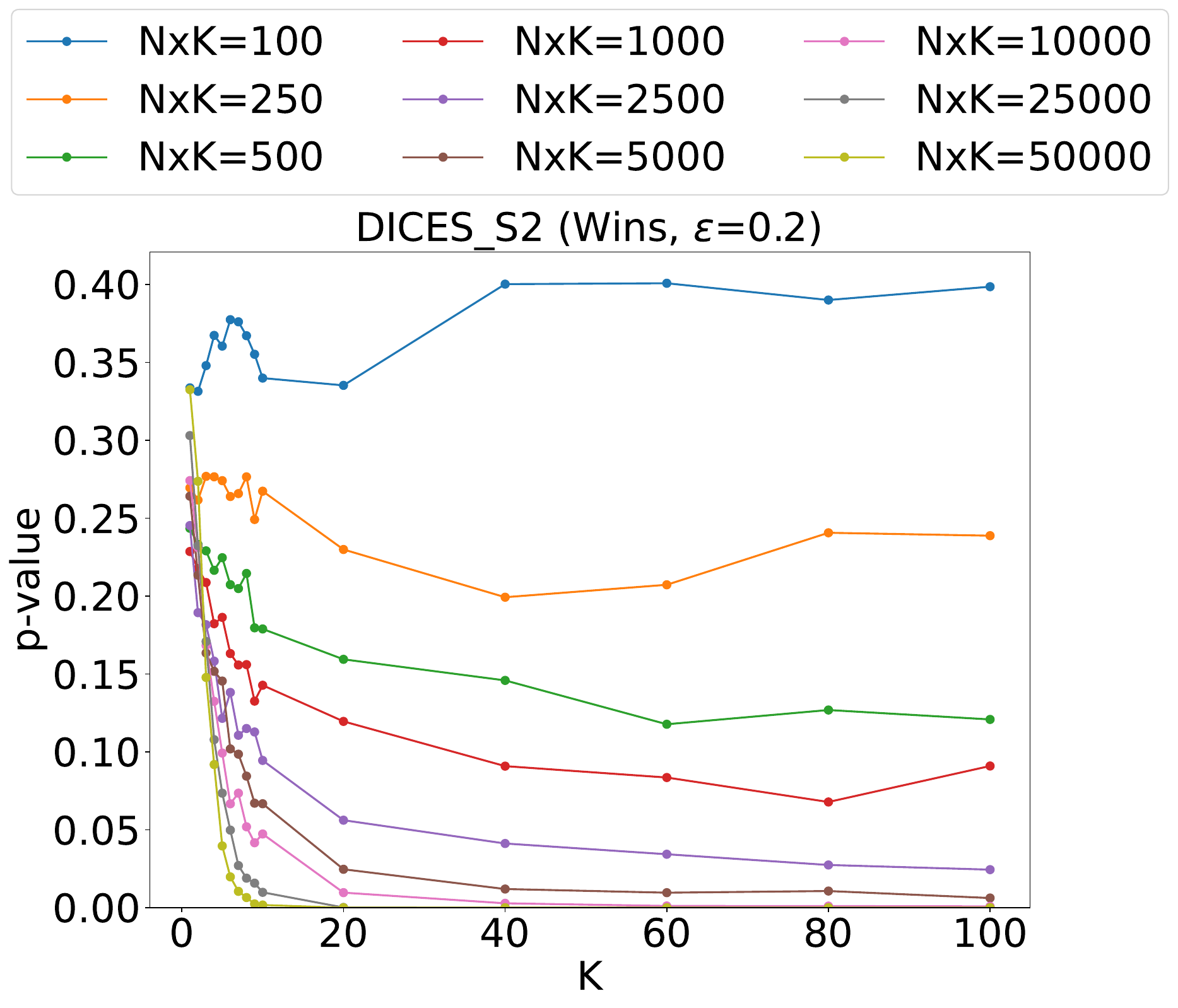}
    \caption{$\epsilon = 0.2$}
    \label{fig:dices_s2_wins_e02}
  \end{subfigure} \hfill
  \begin{subfigure}[b]{0.24\linewidth}
    \centering
    \includegraphics[width=\linewidth]{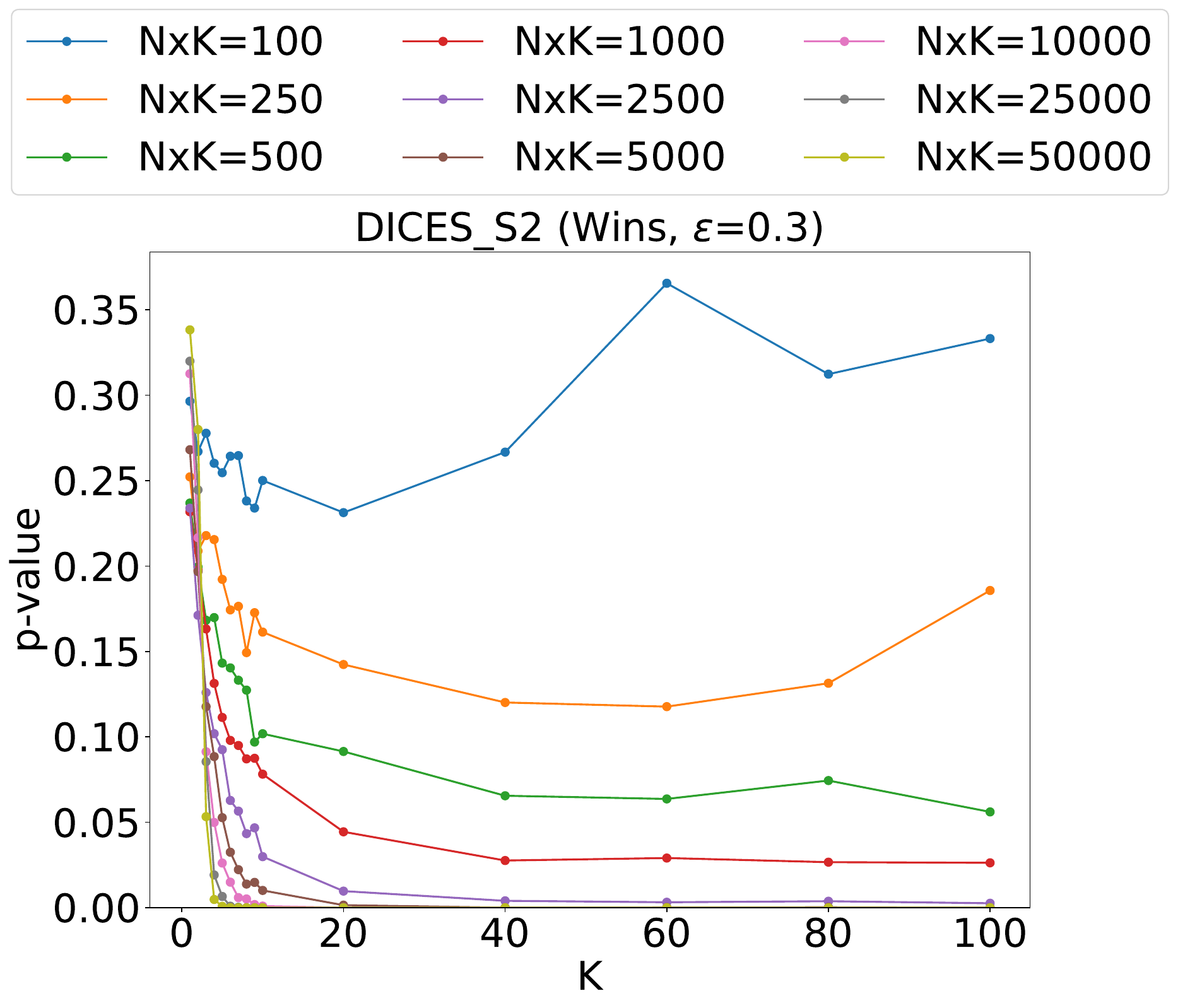}
    \caption{$\epsilon = 0.3$}
    \label{fig:dices_s2_wins_e03}
  \end{subfigure} \hfill
  \begin{subfigure}[b]{0.24\linewidth}
    \centering
    \includegraphics[width=\linewidth]{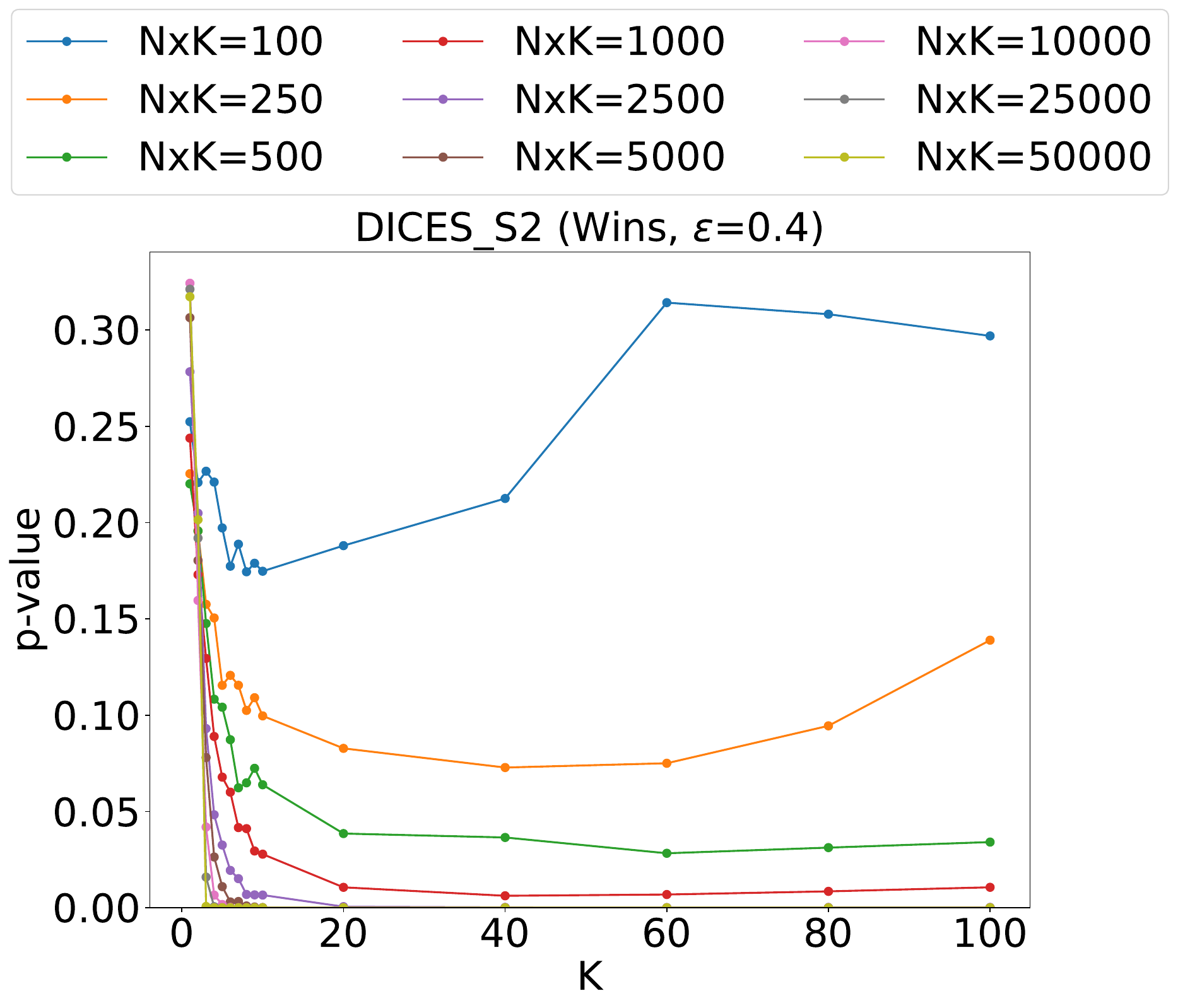}
    \caption{$\epsilon = 0.4$}
    \label{fig:dices_s2_wins_e04}
  \end{subfigure}
  \caption{S2: P-value plots for DICES dataset with Wins as the metric}
  \label{fig:dices_s2_wins}
\end{figure*}

\begin{figure*}
  \centering
  \begin{subfigure}[b]{0.24\linewidth}
    \centering
    \includegraphics[width=\linewidth]{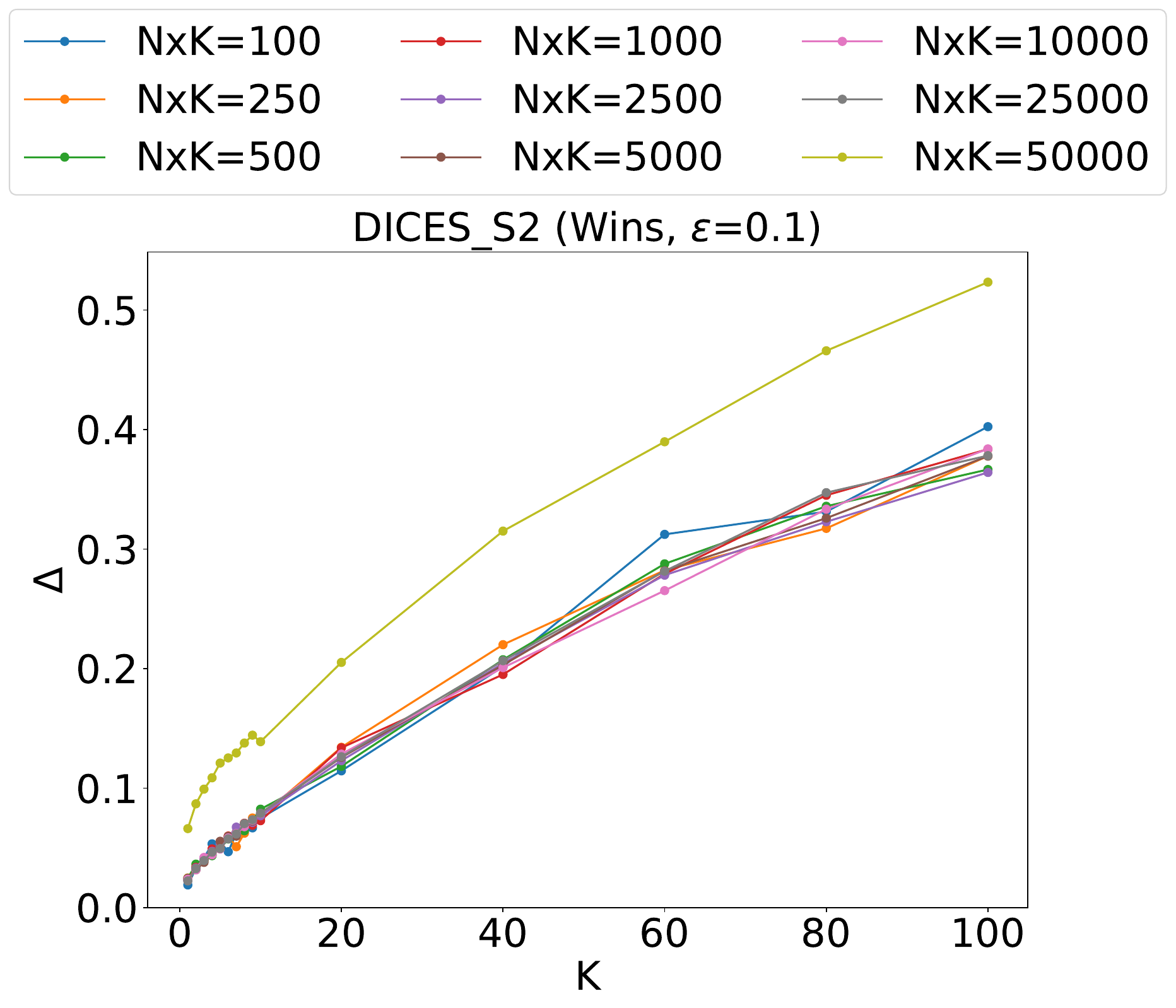}
    \caption{$\epsilon = 0.1$}
    \label{fig:dices_s2_delta_wins_e01}
  \end{subfigure} \hfill
  \begin{subfigure}[b]{0.24\linewidth}
    \centering
    \includegraphics[width=\linewidth]{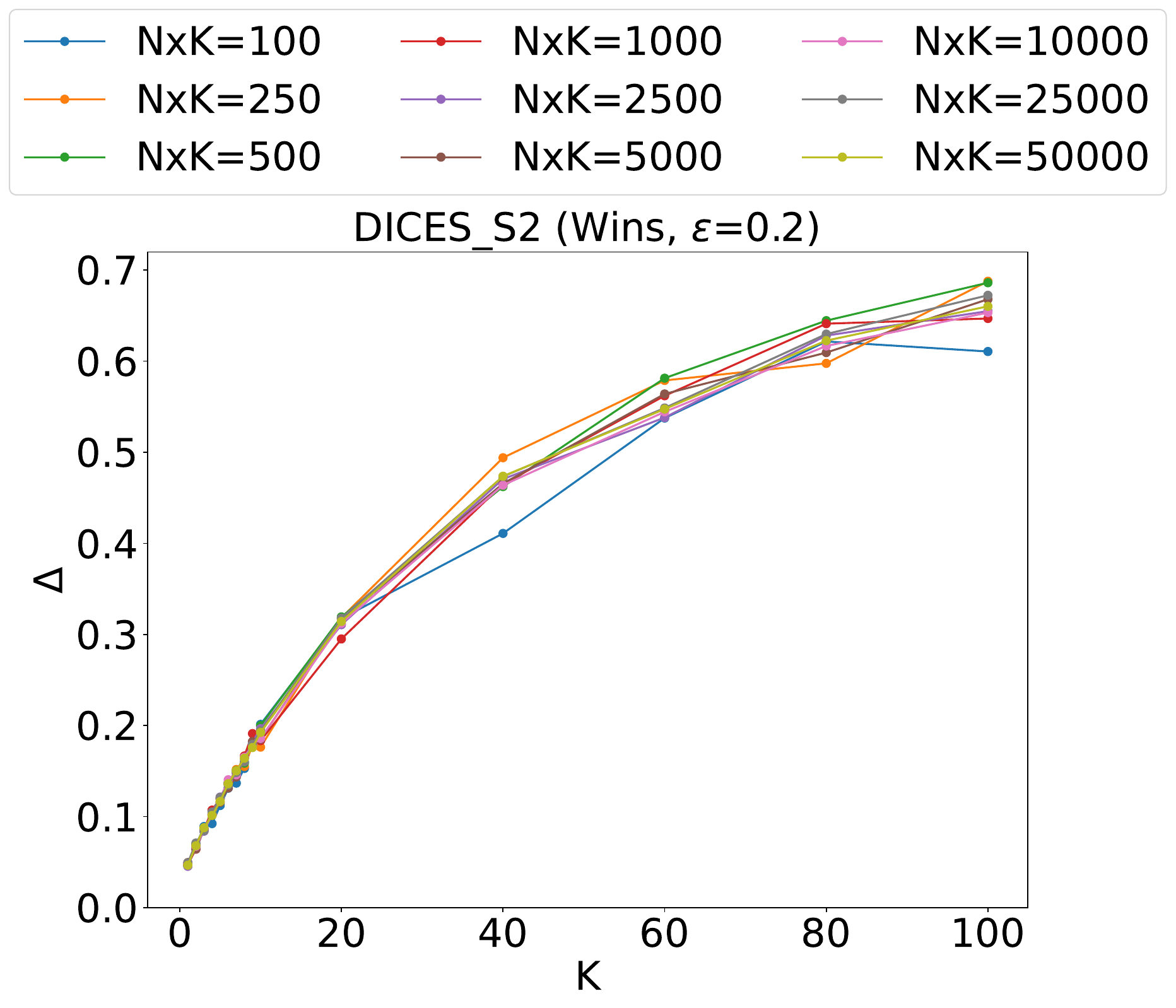}
    \caption{$\epsilon = 0.2$}
    \label{fig:dices_s2_delta_wins_e02}
  \end{subfigure} \hfill
  \begin{subfigure}[b]{0.24\linewidth}
    \centering
    \includegraphics[width=\linewidth]{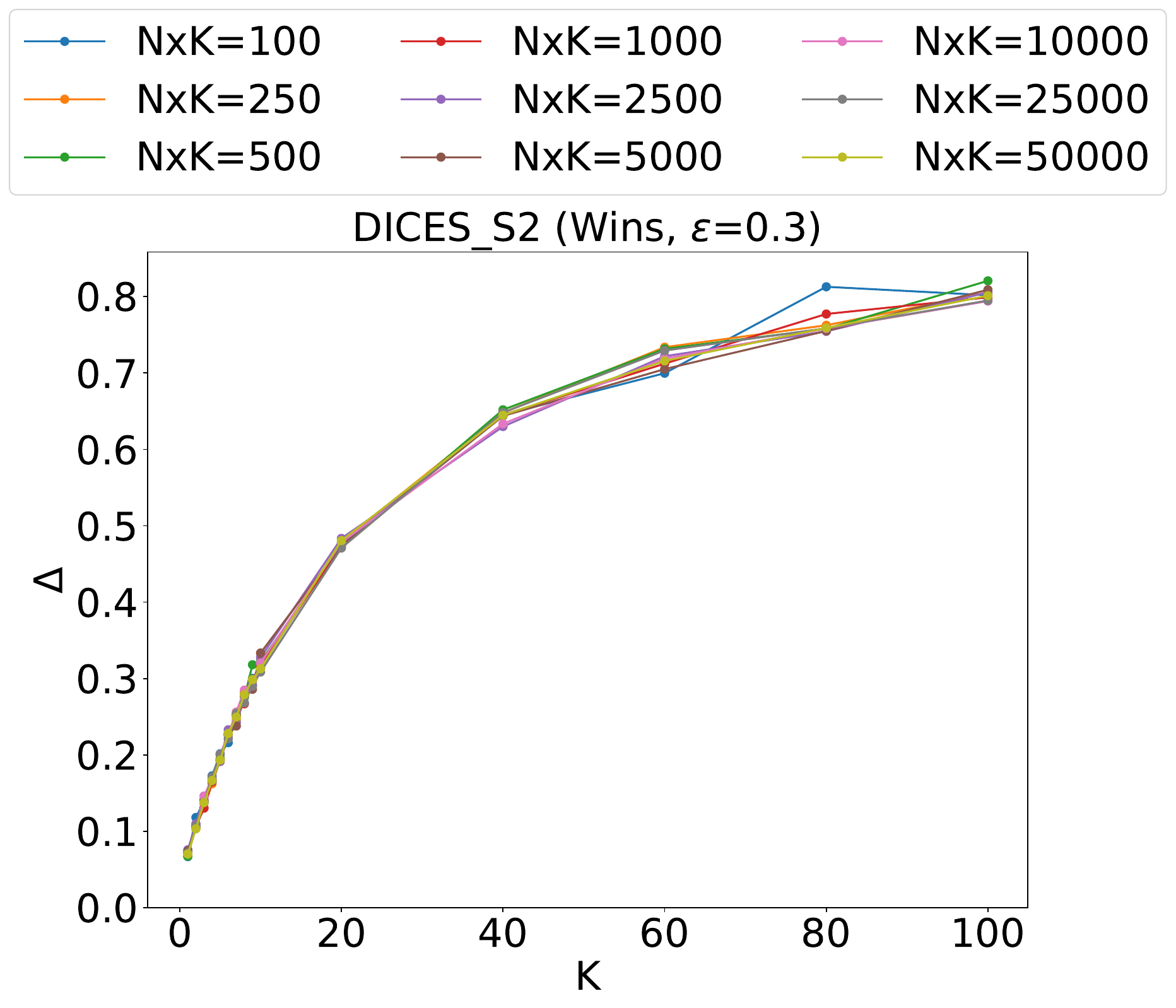}
    \caption{$\epsilon = 0.3$}
    \label{fig:dices_s2_delta_wins_e03}
  \end{subfigure} \hfill
  \begin{subfigure}[b]{0.24\linewidth}
    \centering
    \includegraphics[width=\linewidth]{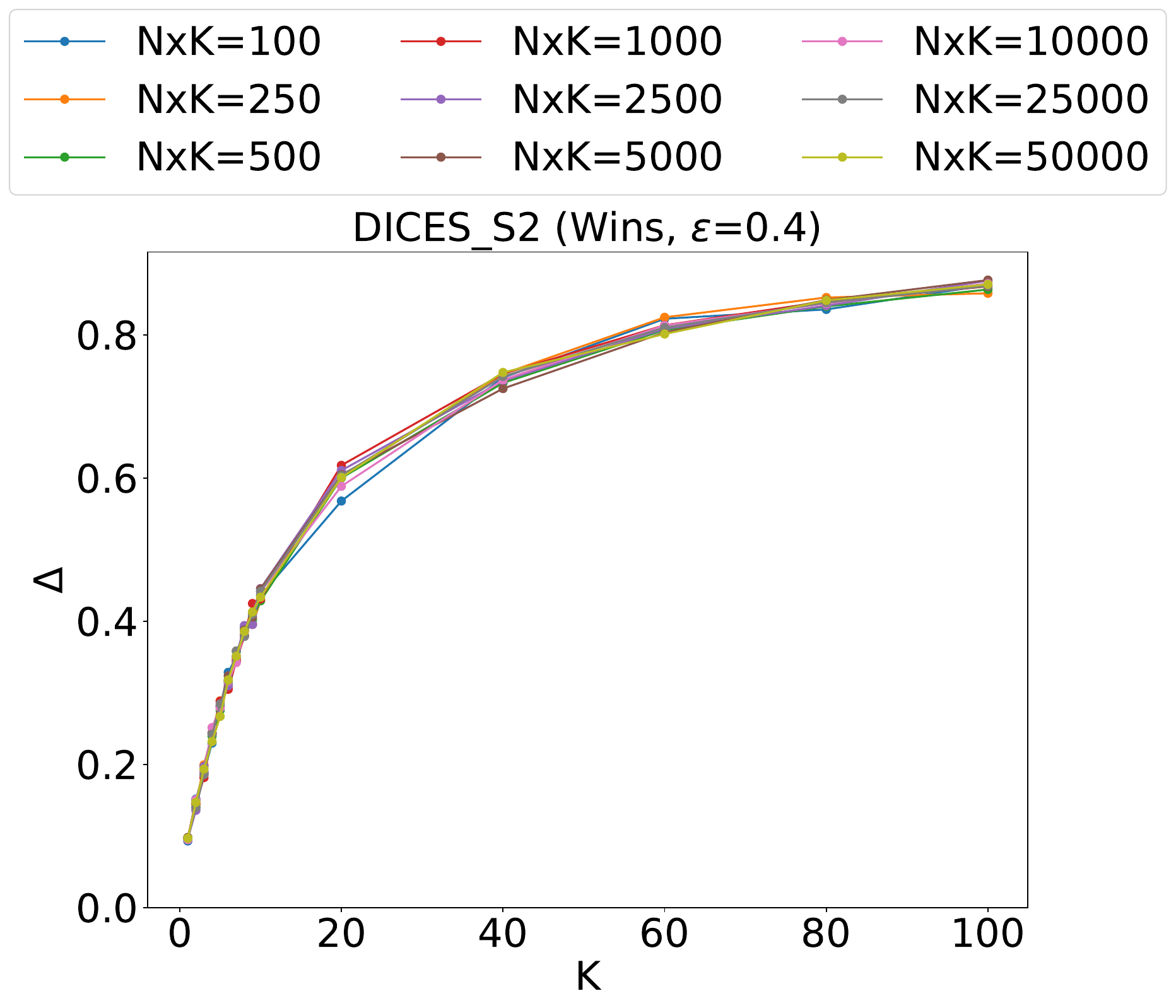}
    \caption{$\epsilon = 0.4$}
    \label{fig:dices_s2_delta_wins_e04}
  \end{subfigure}
  \caption{S2: Effect sizes ($\Delta$) for DICES dataset with Wins as the metric}
  \label{fig:dices_s2_delta_wins}
\end{figure*}

\paragraph{DICES - 5 Rater Sample}

\begin{figure*}
  \centering
  \begin{subfigure}[b]{0.24\linewidth}
    \centering
    \includegraphics[width=\linewidth]{figures/pvals_plots/DICES_5_S2/DICES_5_S2_p_vals_Accuracy_K_100_e_0.1.pdf}
    \caption{$\epsilon = 0.1$}
    \label{fig:dices_5_s2_acc_e01}
  \end{subfigure} \hfill
  \begin{subfigure}[b]{0.24\linewidth}
    \centering
    \includegraphics[width=\linewidth]{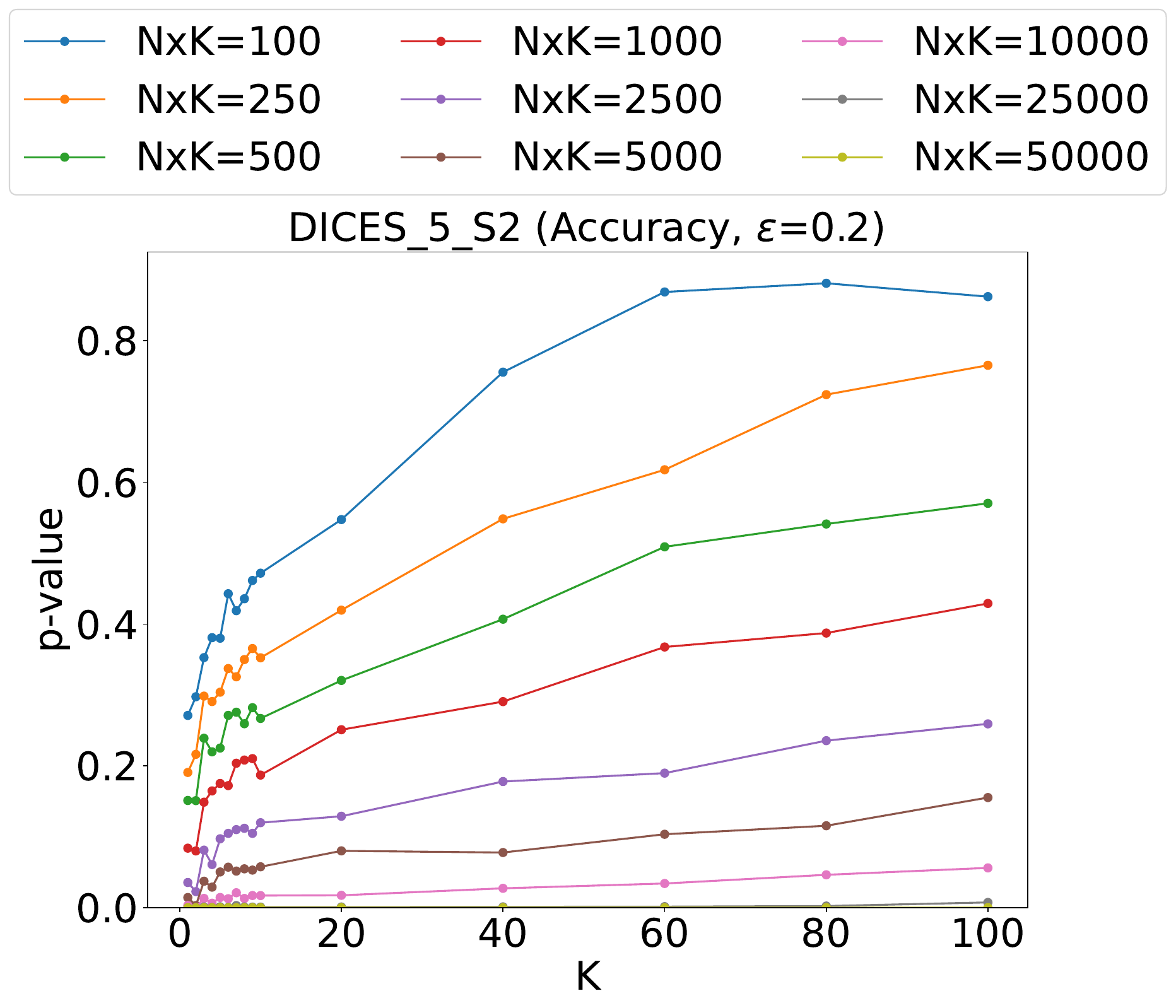}
    \caption{$\epsilon = 0.2$}
    \label{fig:dices_5_s2_acc_e02}
  \end{subfigure} \hfill
  \begin{subfigure}[b]{0.24\linewidth}
    \centering
    \includegraphics[width=\linewidth]{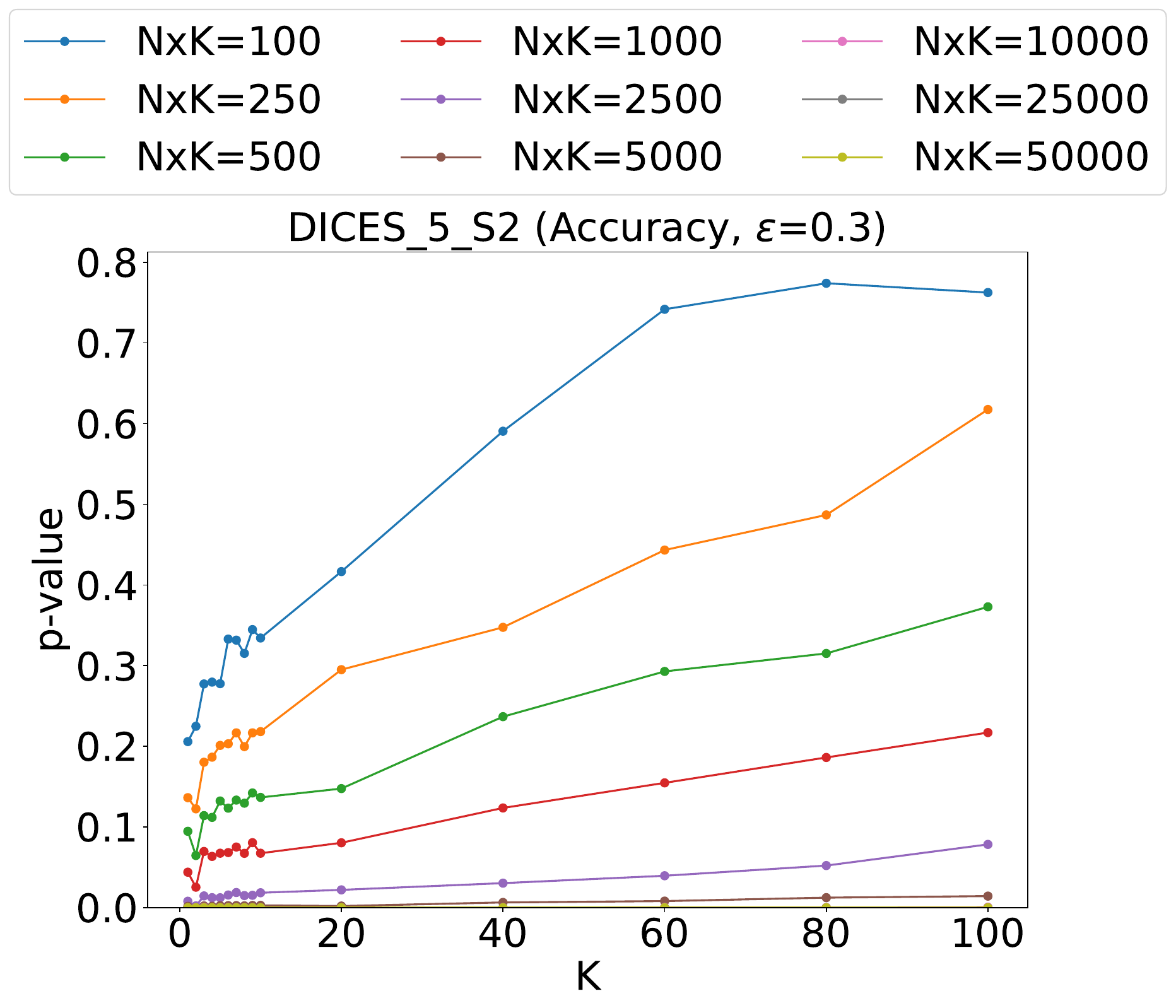}
    \caption{$\epsilon = 0.3$}
    \label{fig:dices_5_s2_acc_e03}
  \end{subfigure} \hfill
  \begin{subfigure}[b]{0.24\linewidth}
    \centering
    \includegraphics[width=\linewidth]{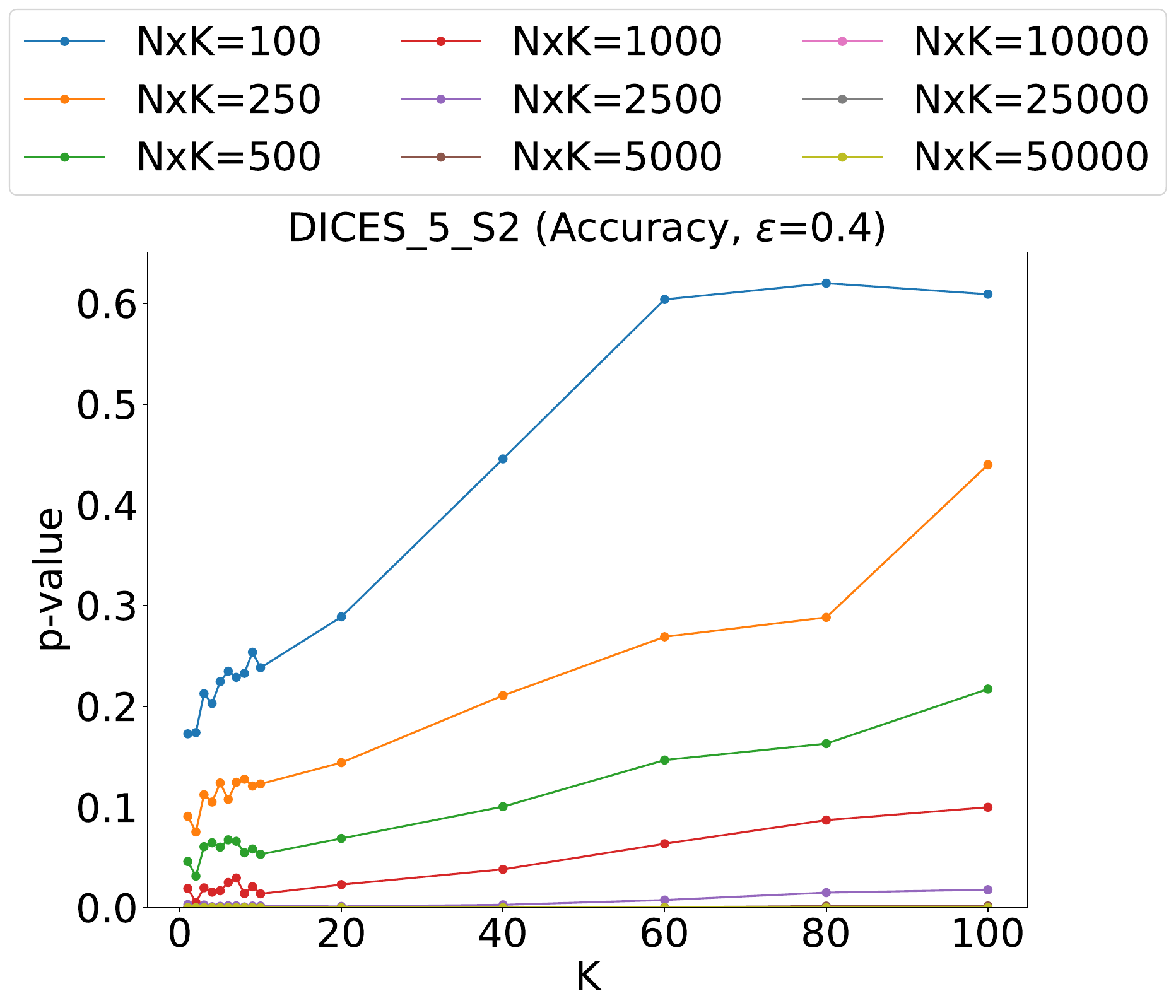}
    \caption{$\epsilon = 0.4$}
    \label{fig:dices_5_s2_acc_e04}
  \end{subfigure}
  \caption{S2: P-value plots for DICES 5 rater sample with Accuracy as the metric}
  \label{fig:dices_5_s2_accuracy}
\end{figure*}

\begin{figure*}
  \centering
  \begin{subfigure}[b]{0.24\linewidth}
    \centering
    \includegraphics[width=\linewidth]{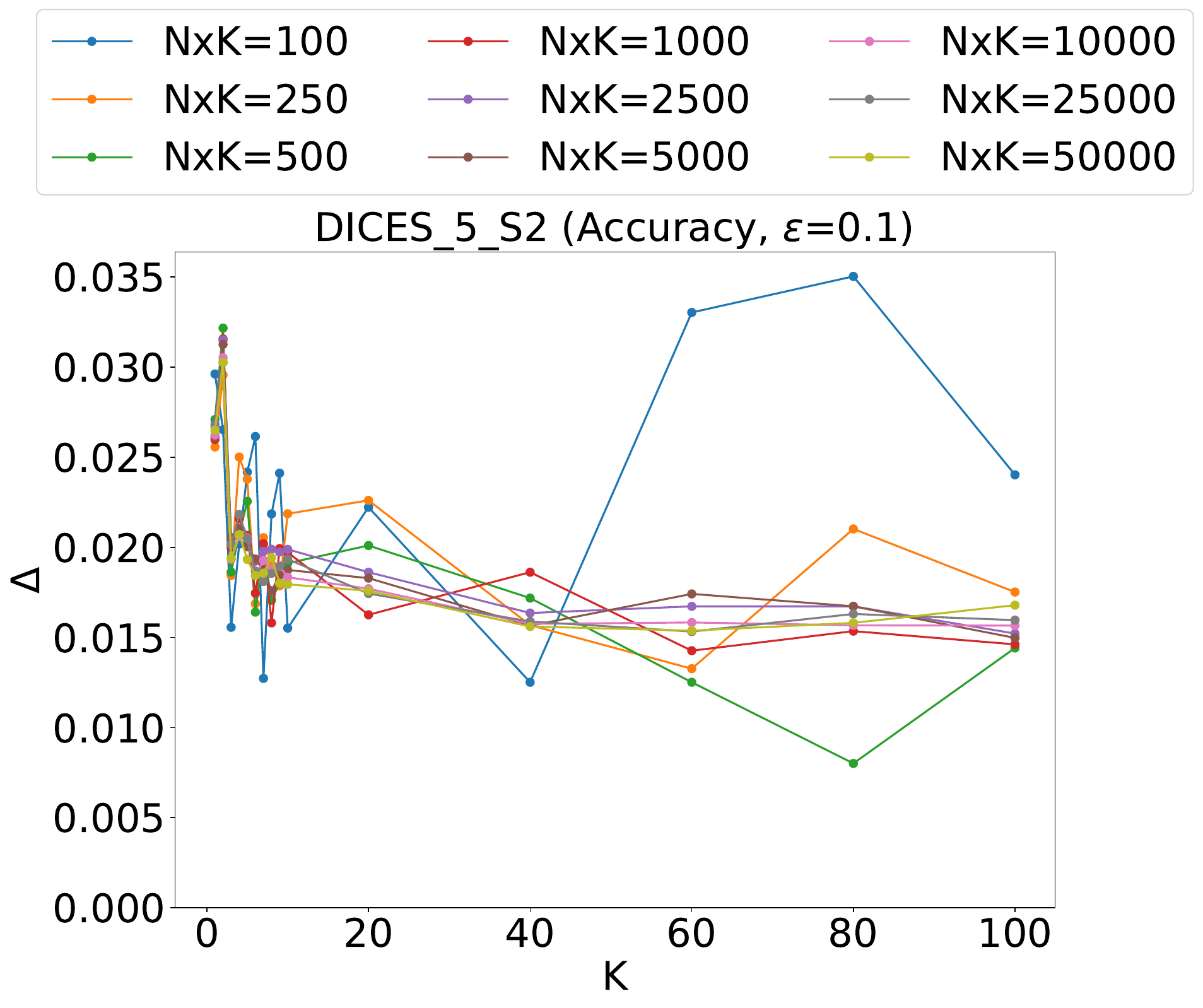}
    \caption{$\epsilon = 0.1$}
    \label{fig:dices_5_s2_delta_acc_e01}
  \end{subfigure} \hfill
  \begin{subfigure}[b]{0.24\linewidth}
    \centering
    \includegraphics[width=\linewidth]{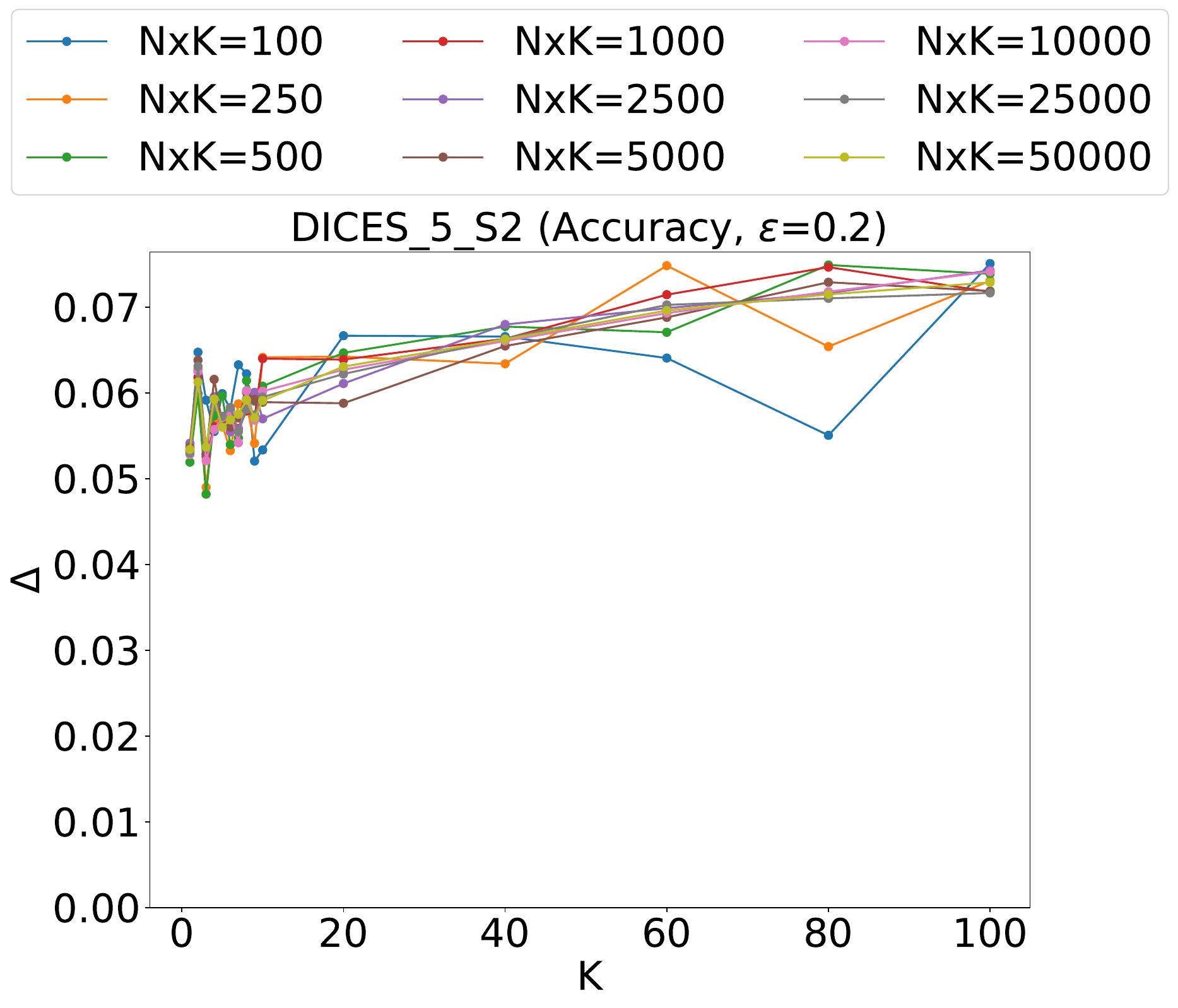}
    \caption{$\epsilon = 0.2$}
    \label{fig:dices_5_s2_delta_acc_e02}
  \end{subfigure} \hfill
  \begin{subfigure}[b]{0.24\linewidth}
    \centering
    \includegraphics[width=\linewidth]{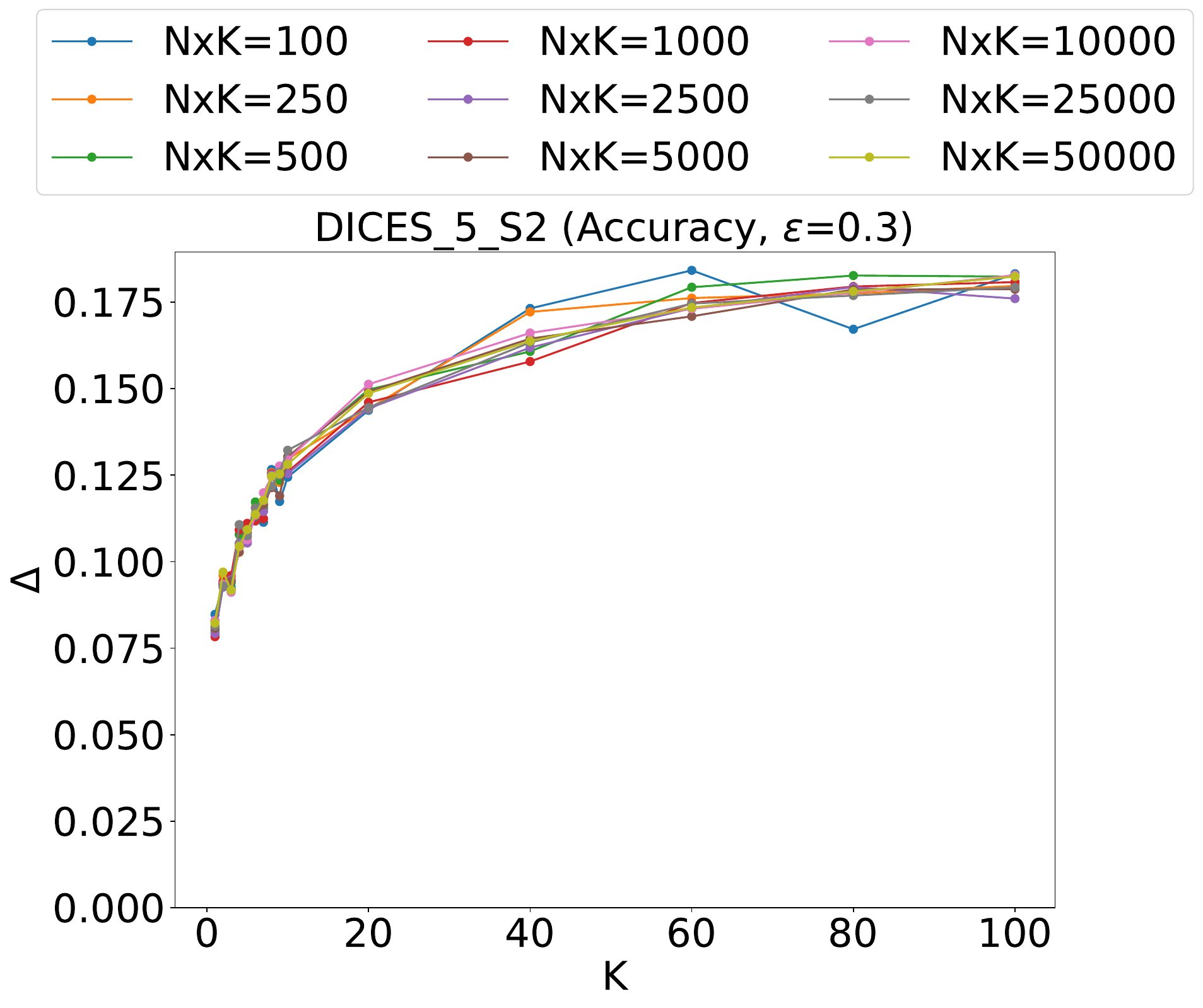}
    \caption{$\epsilon = 0.3$}
    \label{fig:dices_5_s2_delta_acc_e03}
  \end{subfigure} \hfill
  \begin{subfigure}[b]{0.24\linewidth}
    \centering
    \includegraphics[width=\linewidth]{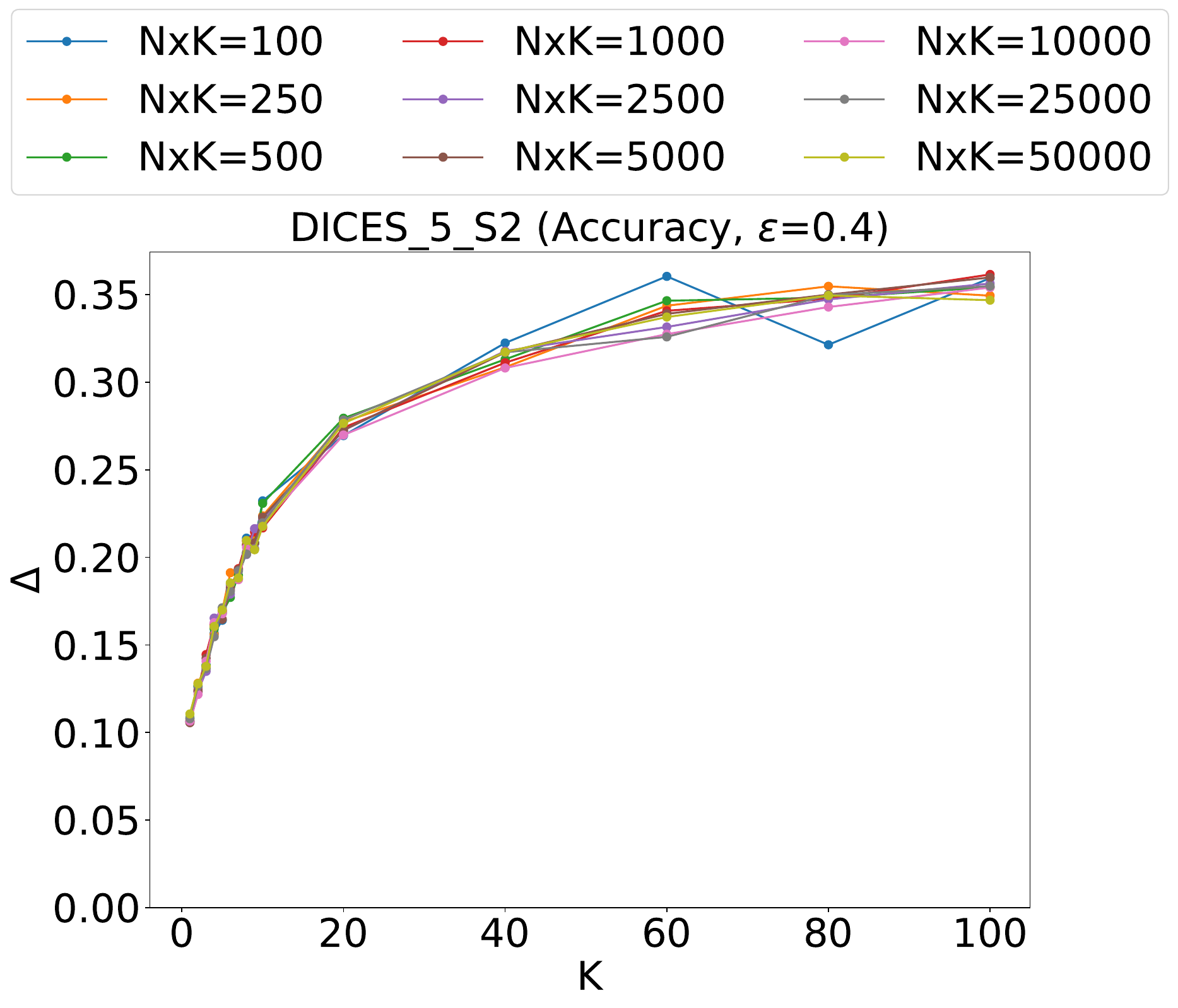}
    \caption{$\epsilon = 0.4$}
    \label{fig:dices_5_s2_delta_acc_e04}
  \end{subfigure}
  \caption{S2: Effect sizes ($\Delta$) for DICES 5 rater sample with Accuracy as the metric}
  \label{fig:dices_5_s2_delta_accuracy}
\end{figure*}

\begin{figure*}
  \centering
  \begin{subfigure}[b]{0.24\linewidth}
    \centering
    \includegraphics[width=\linewidth]{figures/pvals_plots/DICES_5_S2/DICES_5_S2_p_vals_MAE_K_100_e_0.1.pdf}
    \caption{$\epsilon = 0.1$}
    \label{fig:dices_5_s2_MAE_e01}
  \end{subfigure} \hfill
  \begin{subfigure}[b]{0.24\linewidth}
    \centering
    \includegraphics[width=\linewidth]{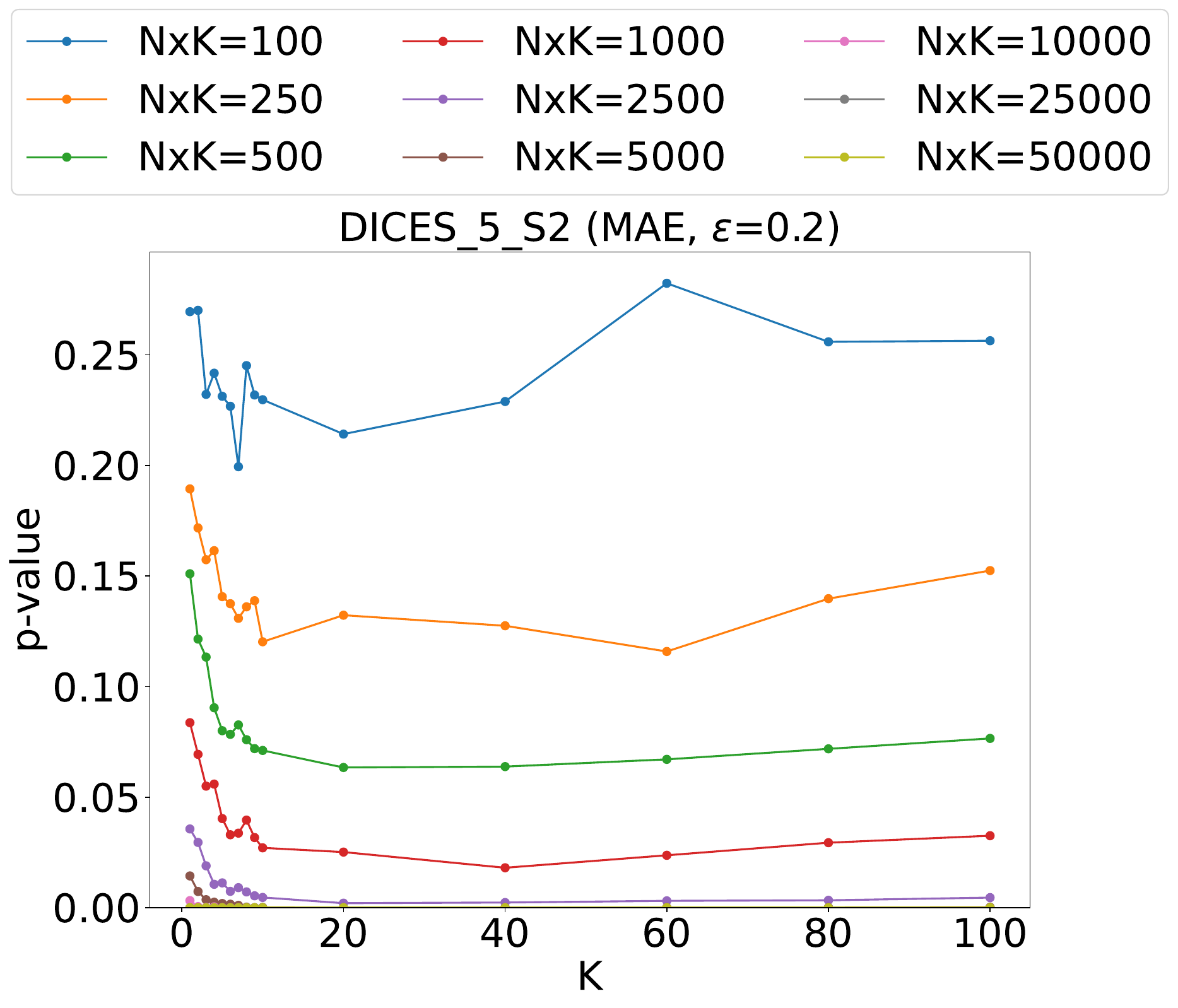}
    \caption{$\epsilon = 0.2$}
    \label{fig:dices_5_s2_MAE_e02}
  \end{subfigure} \hfill
  \begin{subfigure}[b]{0.24\linewidth}
    \centering
    \includegraphics[width=\linewidth]{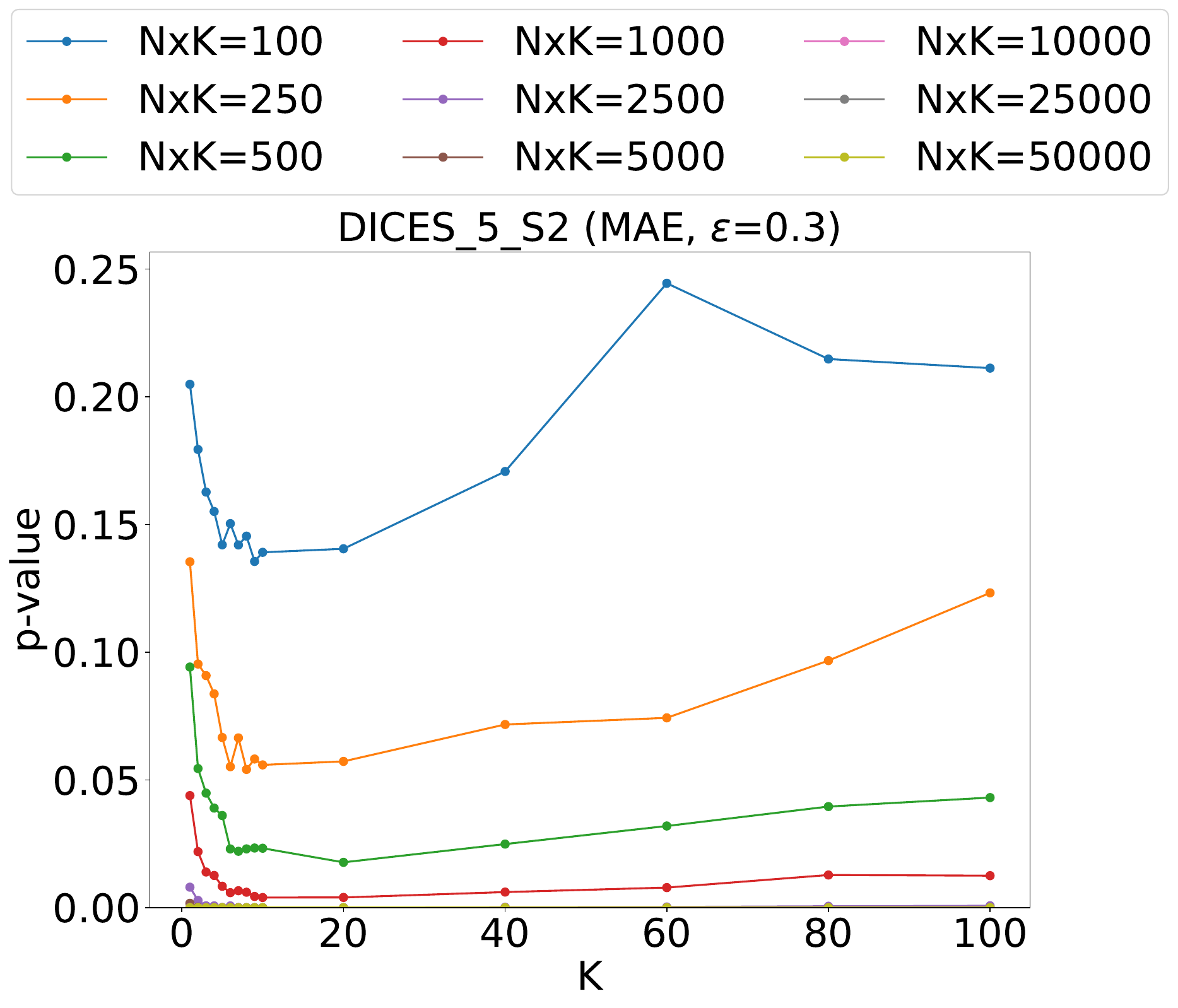}
    \caption{$\epsilon = 0.3$}
    \label{fig:dices_5_s2_MAE_e03}
  \end{subfigure} \hfill
  \begin{subfigure}[b]{0.24\linewidth}
    \centering
    \includegraphics[width=\linewidth]{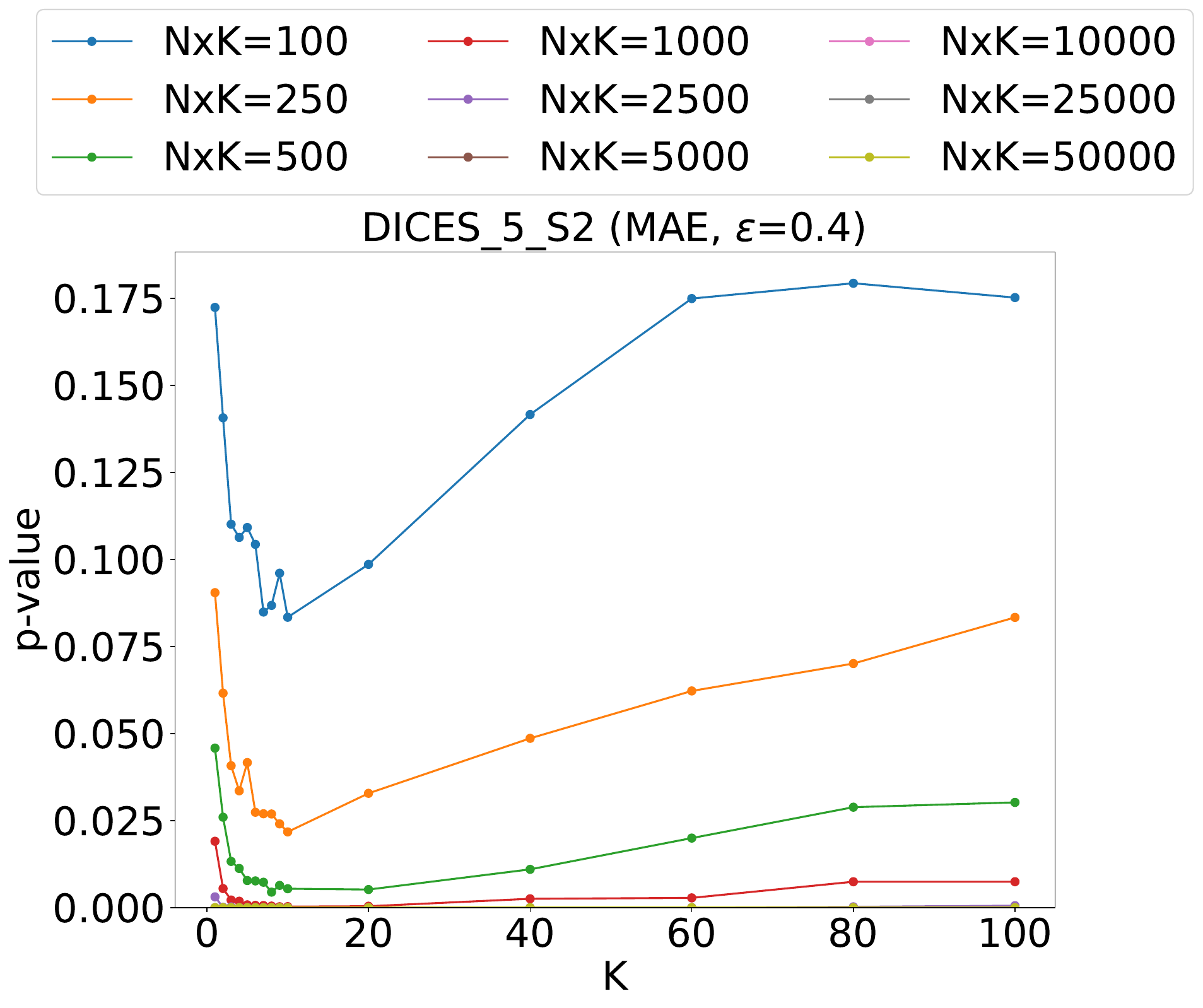}
    \caption{$\epsilon = 0.4$}
    \label{fig:dices_5_s2_MAE_e04}
  \end{subfigure}
  \caption{S2: P-value plots for DICES 5 rater sample with MAE as the metric}
  \label{fig:dices_5_s2_MAE}
\end{figure*}

\begin{figure*}
  \centering
  \begin{subfigure}[b]{0.24\linewidth}
    \centering
    \includegraphics[width=\linewidth]{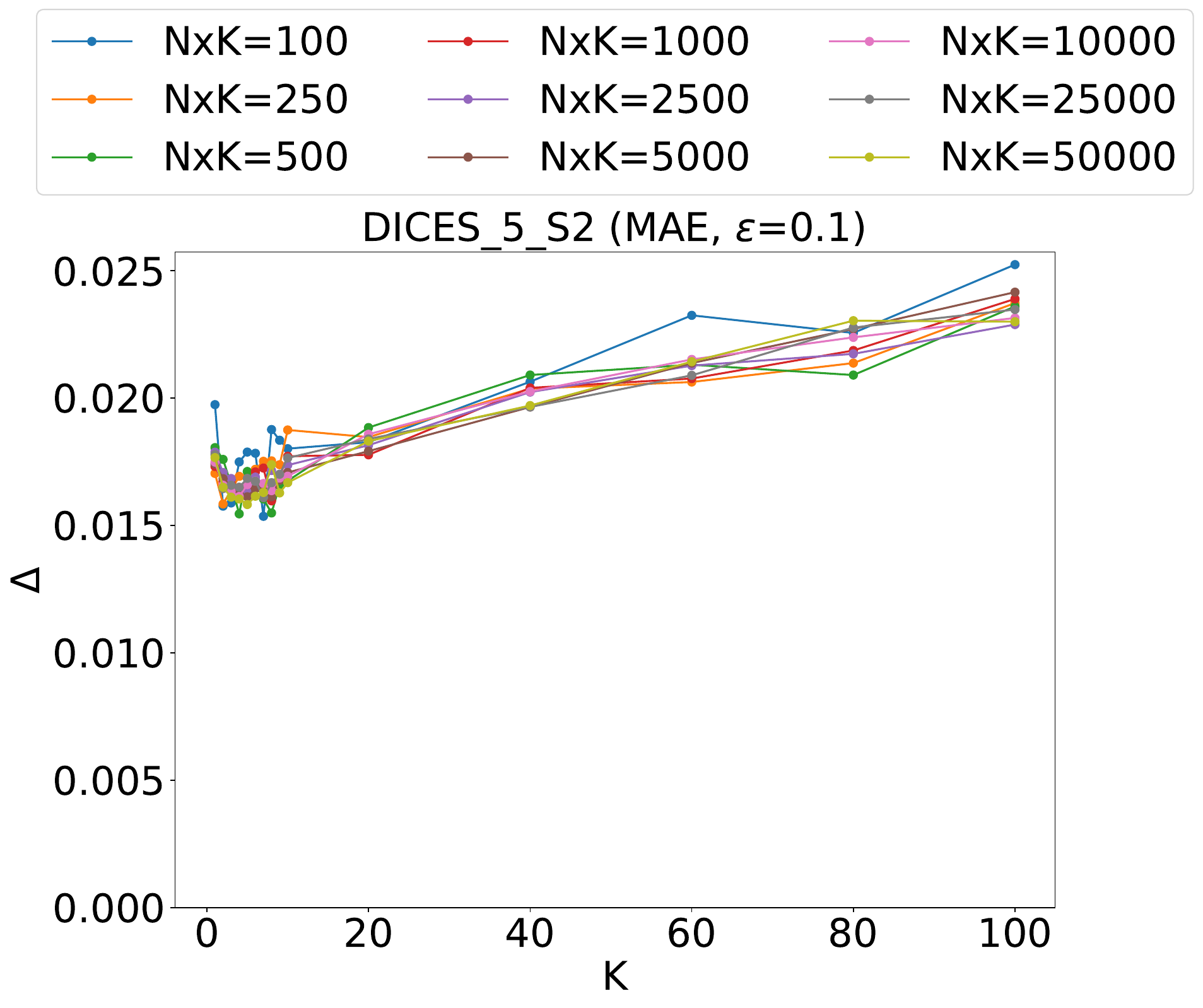}
    \caption{$\epsilon = 0.1$}
    \label{fig:dices_5_s2_delta_MAE_e01}
  \end{subfigure} \hfill
  \begin{subfigure}[b]{0.24\linewidth}
    \centering
    \includegraphics[width=\linewidth]{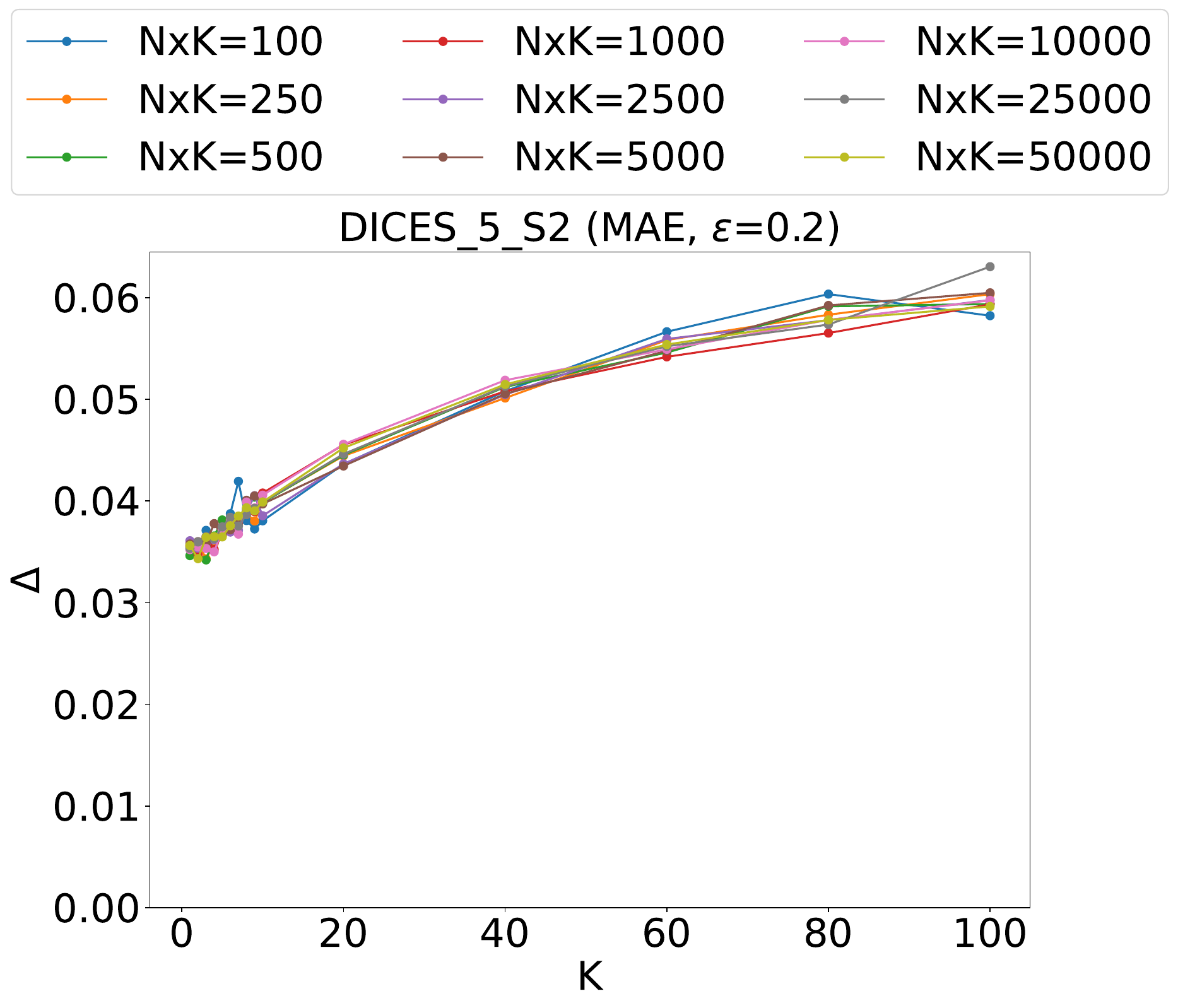}
    \caption{$\epsilon = 0.2$}
    \label{fig:dices_5_s2_delta_MAE_e02}
  \end{subfigure} \hfill
  \begin{subfigure}[b]{0.24\linewidth}
    \centering
    \includegraphics[width=\linewidth]{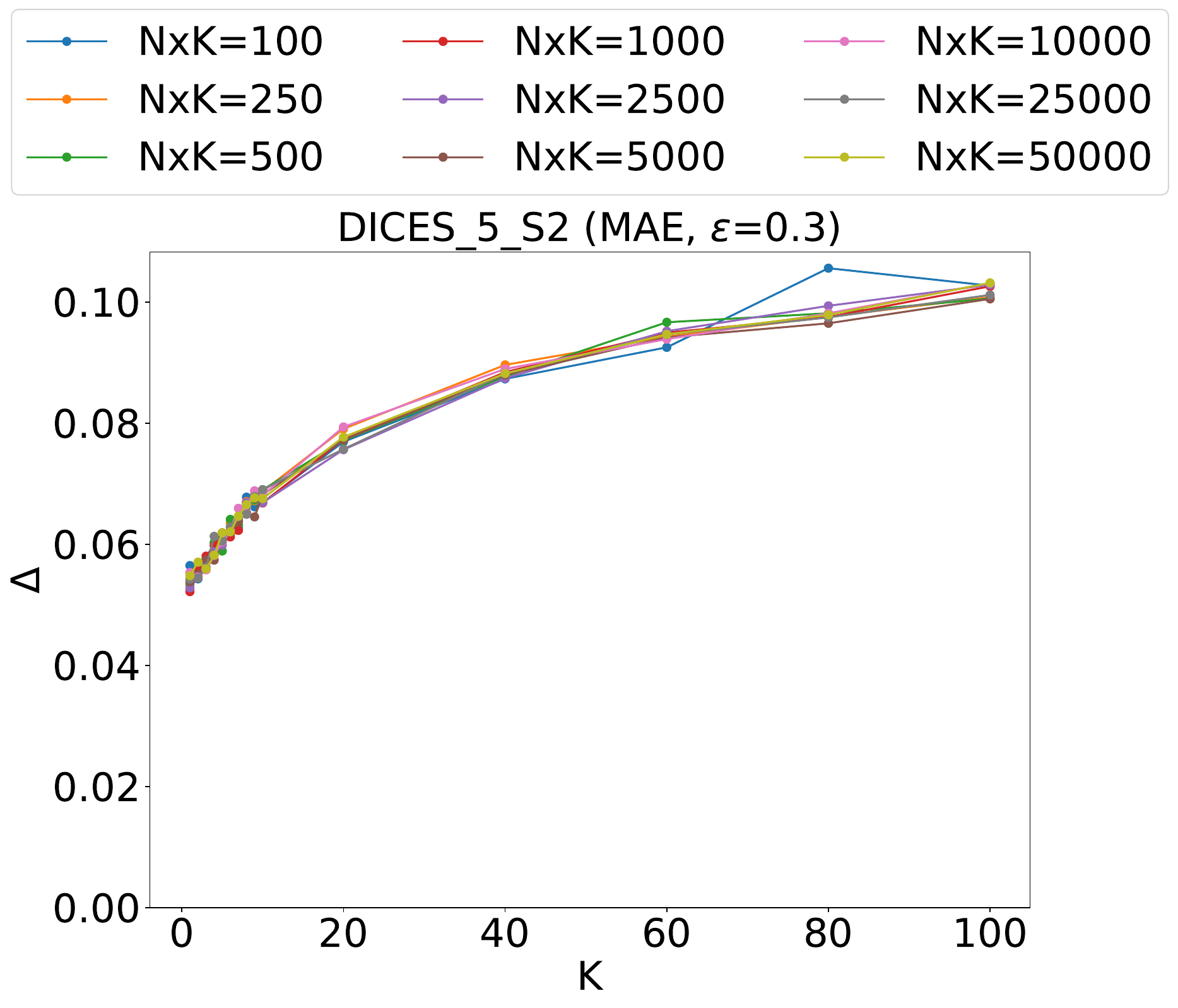}
    \caption{$\epsilon = 0.3$}
    \label{fig:dices_5_s2_delta_MAE_e03}
  \end{subfigure} \hfill
  \begin{subfigure}[b]{0.24\linewidth}
    \centering
    \includegraphics[width=\linewidth]{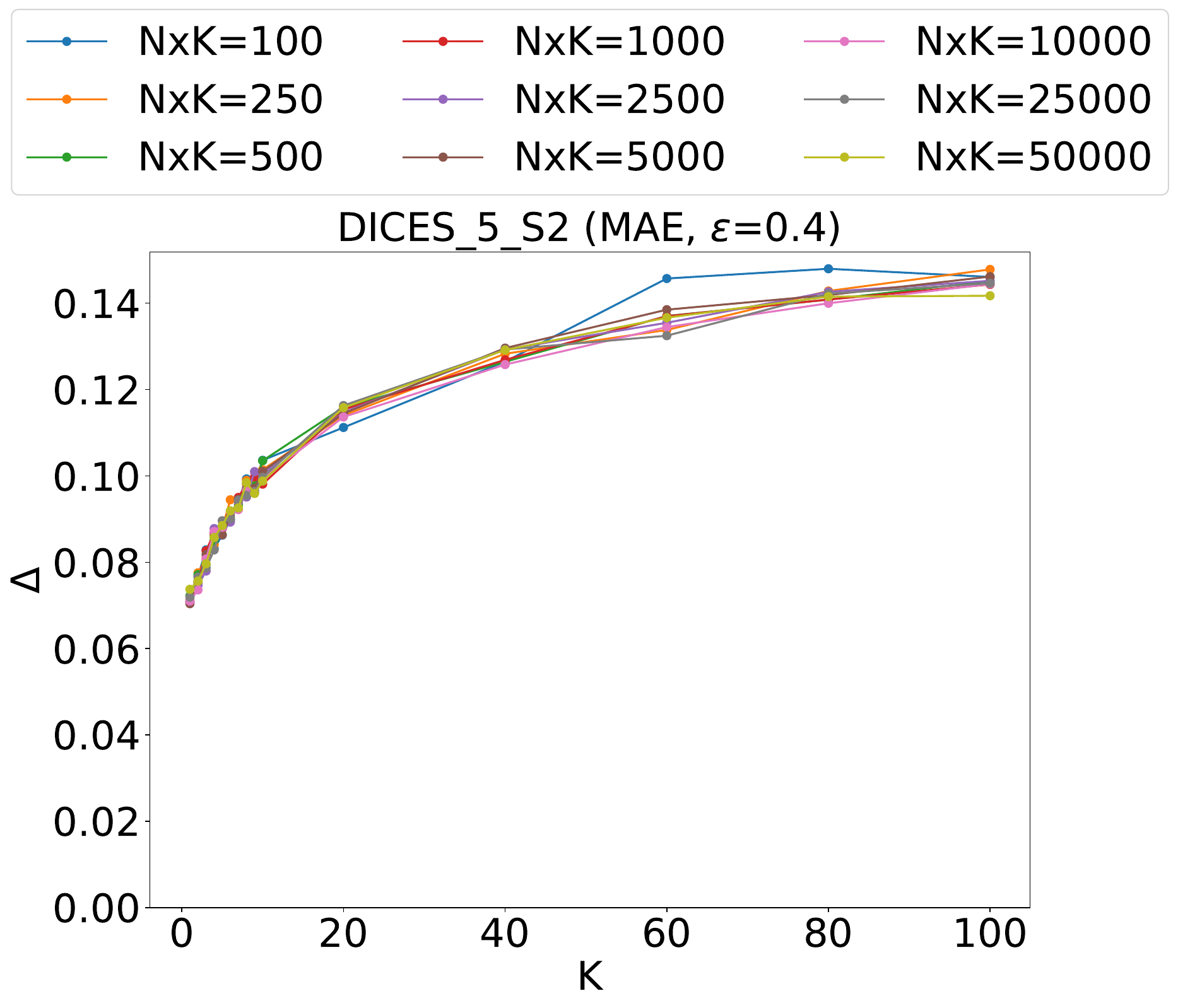}
    \caption{$\epsilon = 0.4$}
    \label{fig:dices_5_s2_delta_MAE_e04}
  \end{subfigure}
  \caption{S2: Effect sizes ($\Delta$) for DICES 5 rater sample with MAE as the metric}
  \label{fig:dices_5_s2_delta_MAE}
\end{figure*}

\begin{figure*}
  \centering
  \begin{subfigure}[b]{0.24\linewidth}
    \centering
    \includegraphics[width=\linewidth]{figures/pvals_plots/DICES_5_S2/DICES_5_S2_p_vals_Wins_K_100_e_0.1.pdf}
    \caption{$\epsilon = 0.1$}
    \label{fig:dices_5_s2_wins_e01}
  \end{subfigure} \hfill
  \begin{subfigure}[b]{0.24\linewidth}
    \centering
    \includegraphics[width=\linewidth]{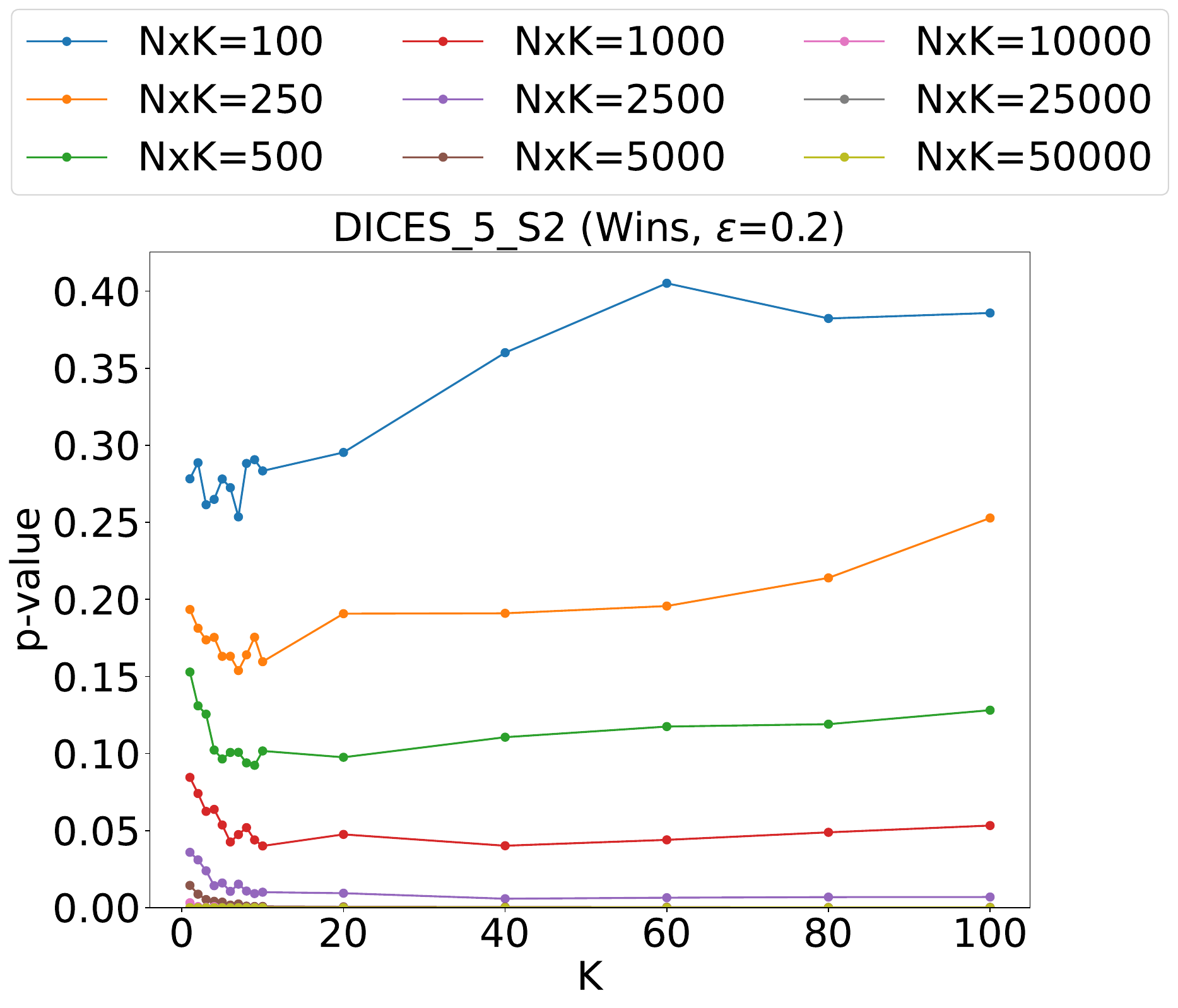}
    \caption{$\epsilon = 0.2$}
    \label{fig:dices_5_s2_wins_e02}
  \end{subfigure} \hfill
  \begin{subfigure}[b]{0.24\linewidth}
    \centering
    \includegraphics[width=\linewidth]{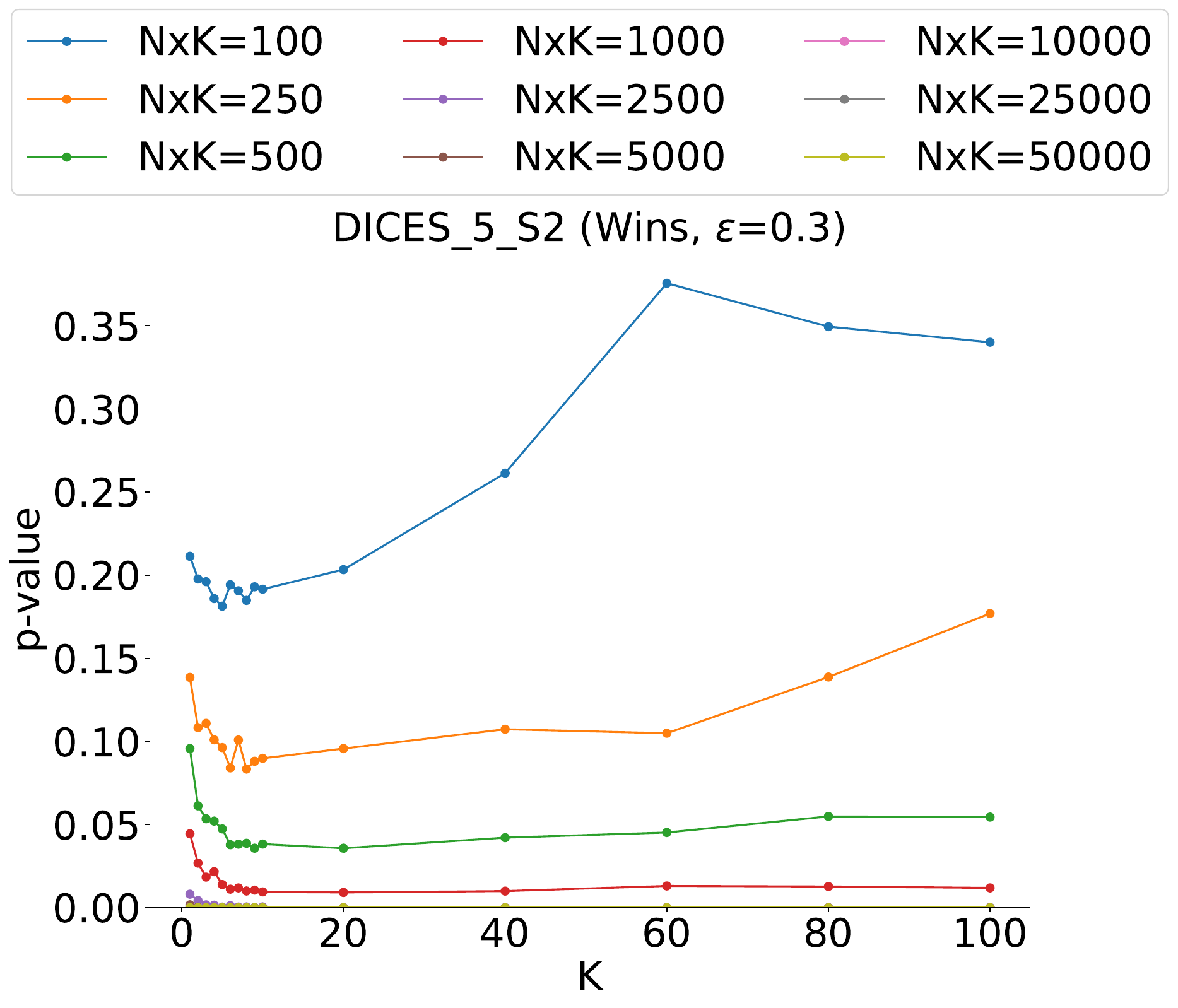}
    \caption{$\epsilon = 0.3$}
    \label{fig:dices_5_s2_wins_e03}
  \end{subfigure} \hfill
  \begin{subfigure}[b]{0.24\linewidth}
    \centering
    \includegraphics[width=\linewidth]{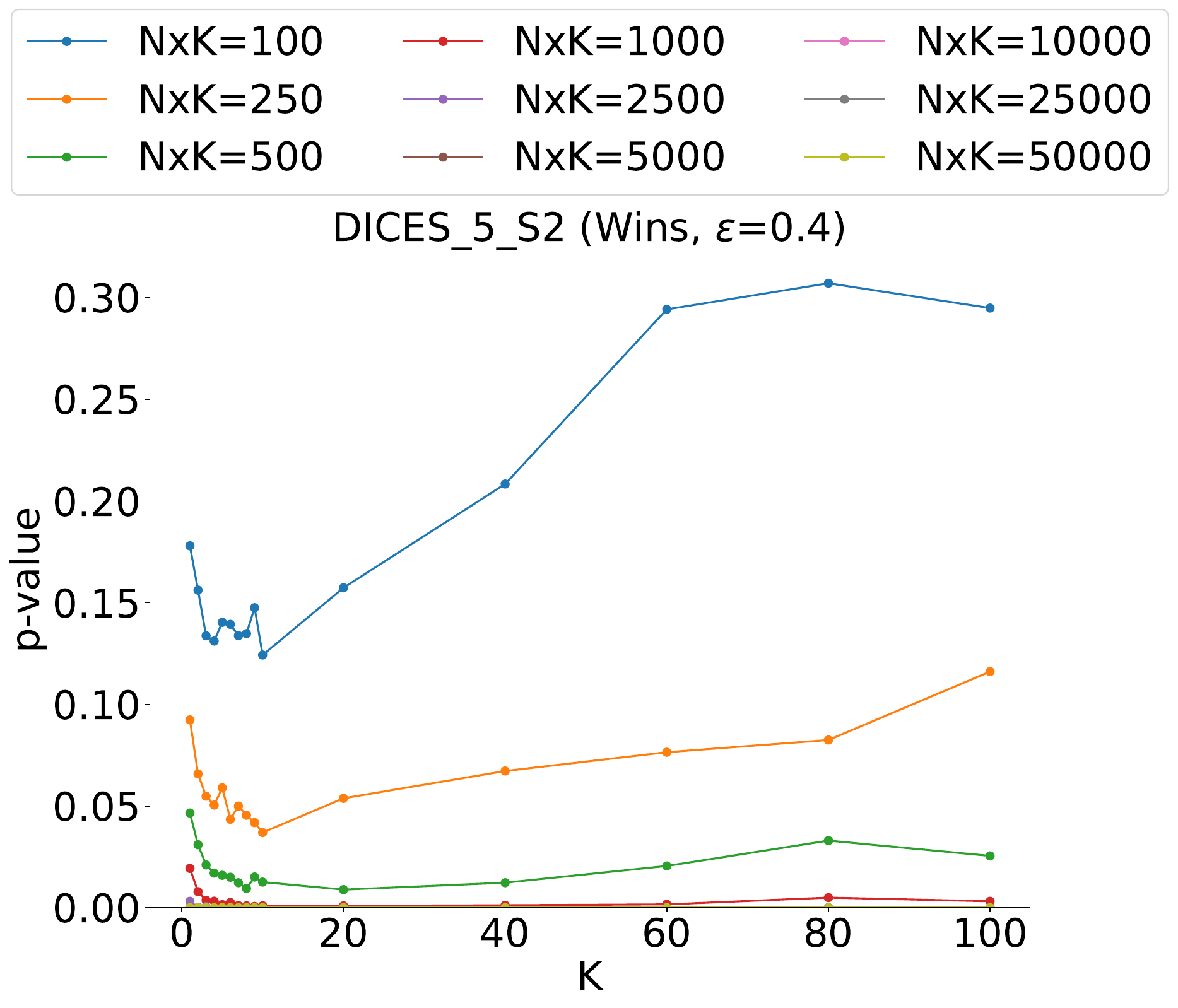}
    \caption{$\epsilon = 0.4$}
    \label{fig:dices_5_s2_wins_e04}
  \end{subfigure}
  \caption{S2: P-value plots for DICES 5 rater sample with Wins as the metric}
  \label{fig:dices_5_s2_wins}
\end{figure*}

\begin{figure*}
  \centering
  \begin{subfigure}[b]{0.24\linewidth}
    \centering
    \includegraphics[width=\linewidth]{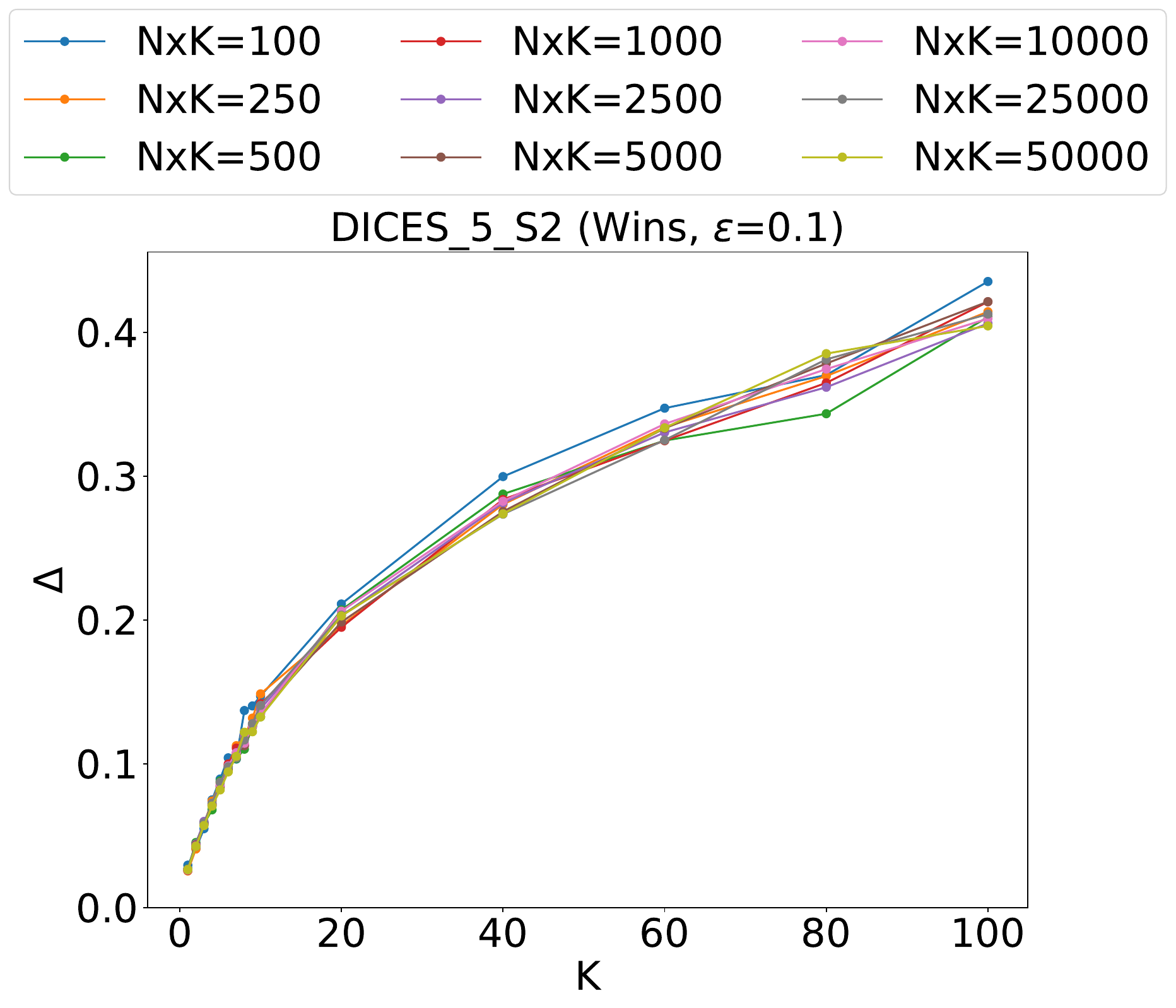}
    \caption{$\epsilon = 0.1$}
    \label{fig:dices_5_s2_delta_wins_e01}
  \end{subfigure} \hfill
  \begin{subfigure}[b]{0.24\linewidth}
    \centering
    \includegraphics[width=\linewidth]{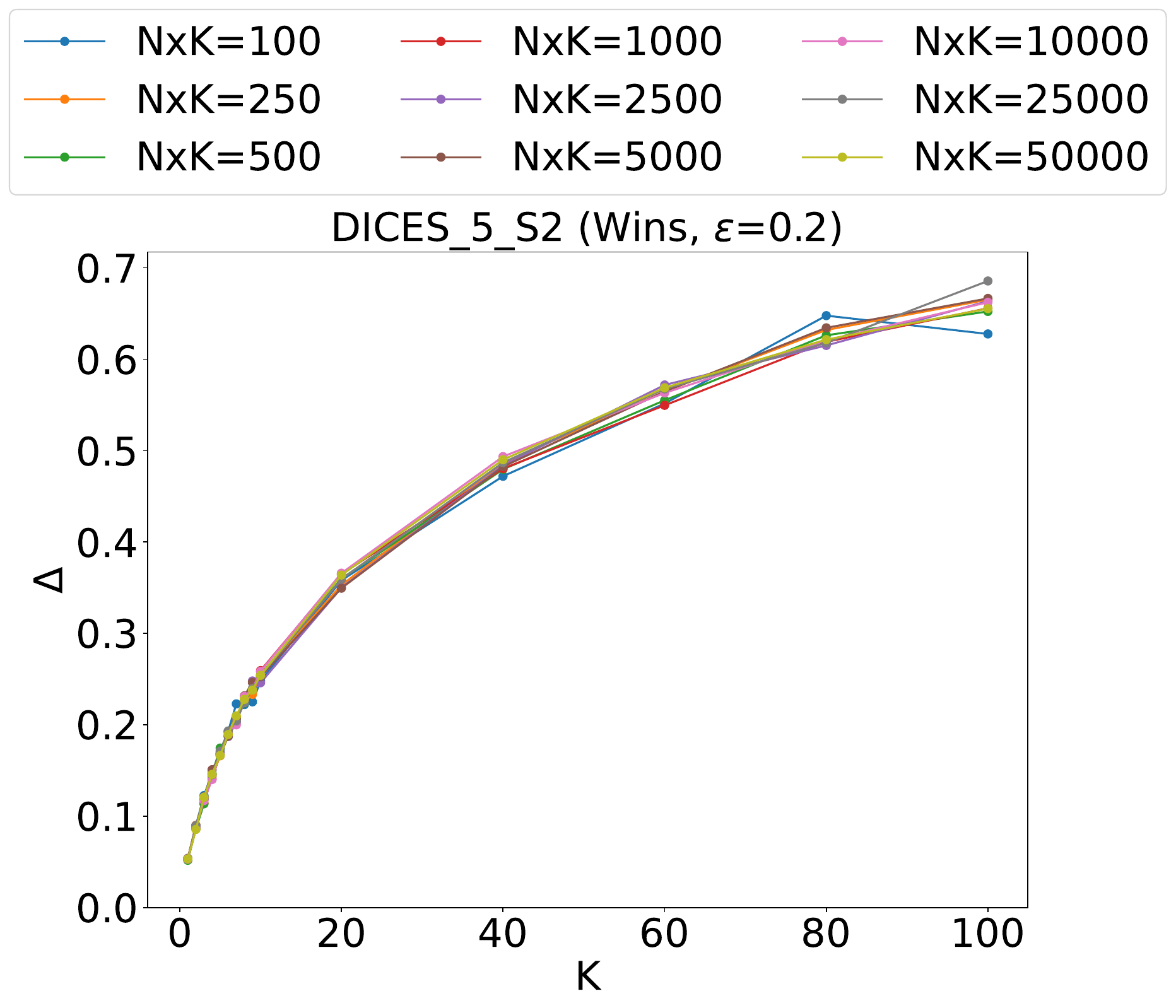}
    \caption{$\epsilon = 0.2$}
    \label{fig:dices_5_s2_delta_wins_e02}
  \end{subfigure} \hfill
  \begin{subfigure}[b]{0.24\linewidth}
    \centering
    \includegraphics[width=\linewidth]{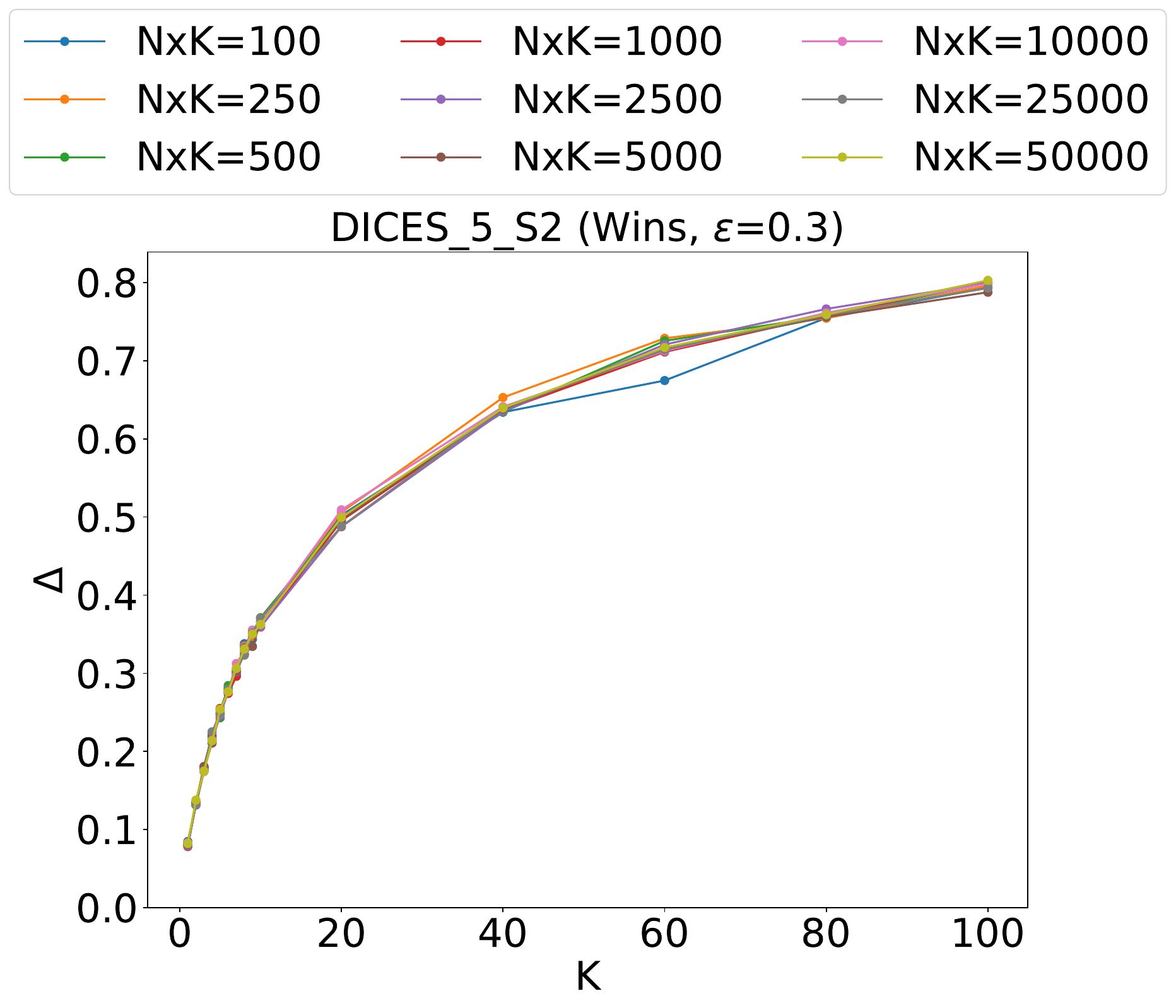}
    \caption{$\epsilon = 0.3$}
    \label{fig:dices_5_s2_delta_wins_e03}
  \end{subfigure} \hfill
  \begin{subfigure}[b]{0.24\linewidth}
    \centering
    \includegraphics[width=\linewidth]{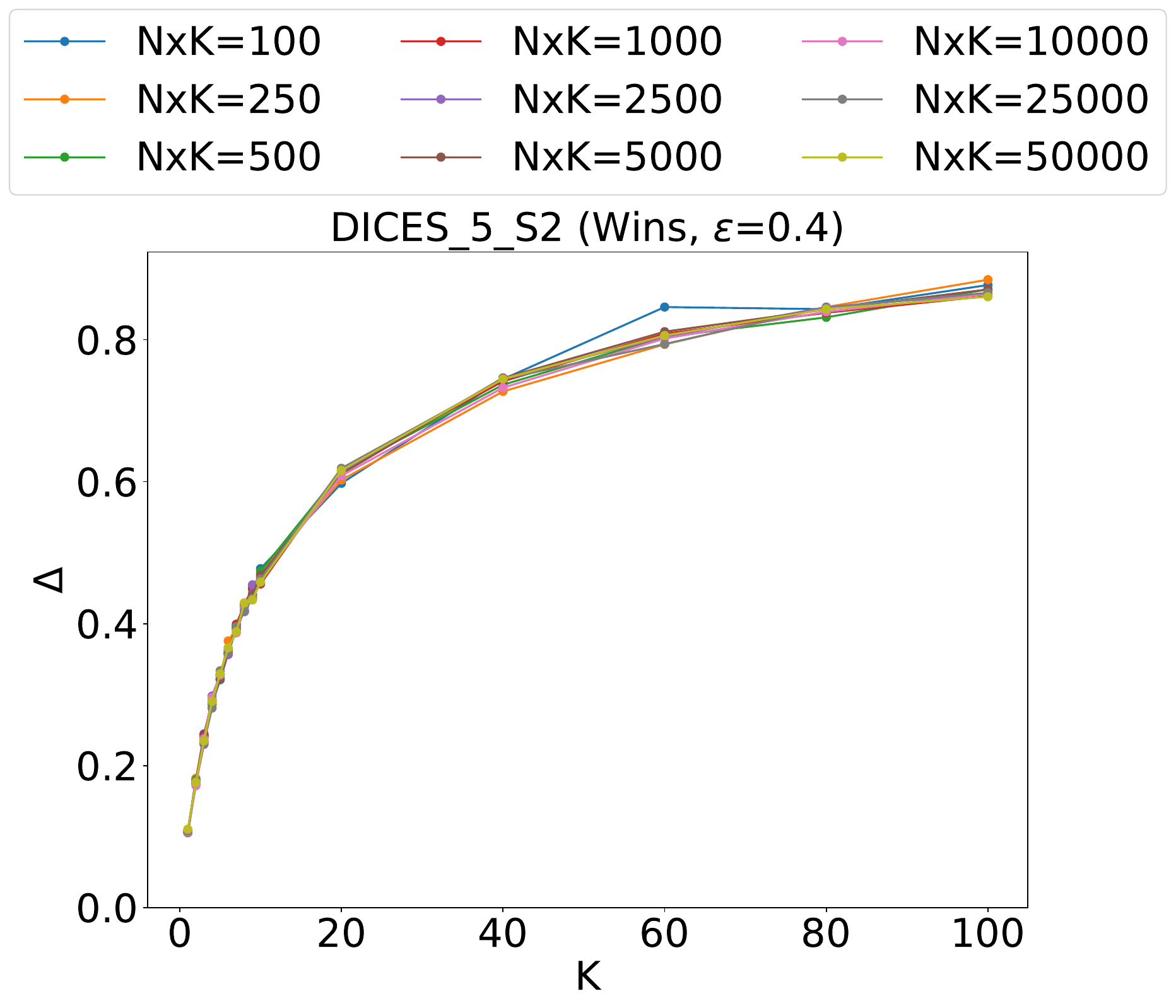}
    \caption{$\epsilon = 0.4$}
    \label{fig:dices_5_s2_delta_wins_e04}
  \end{subfigure}
  \caption{S2: Effect sizes ($\Delta$) for DICES 5 rater sample with Wins as the metric}
  \label{fig:dices_5_s2_delta_wins}
\end{figure*}

\paragraph{Toxicity}

\begin{figure*}
  \centering
  \begin{subfigure}[b]{0.24\linewidth}
    \centering
    \includegraphics[width=\linewidth]{figures/pvals_plots/Toxicity_S2/Toxicity_S2_p_vals_Accuracy_K_100_e_0.1.pdf}
    \caption{$\epsilon = 0.1$}
    \label{fig:toxicity_s2_acc_e01}
  \end{subfigure} \hfill
  \begin{subfigure}[b]{0.24\linewidth}
    \centering
    \includegraphics[width=\linewidth]{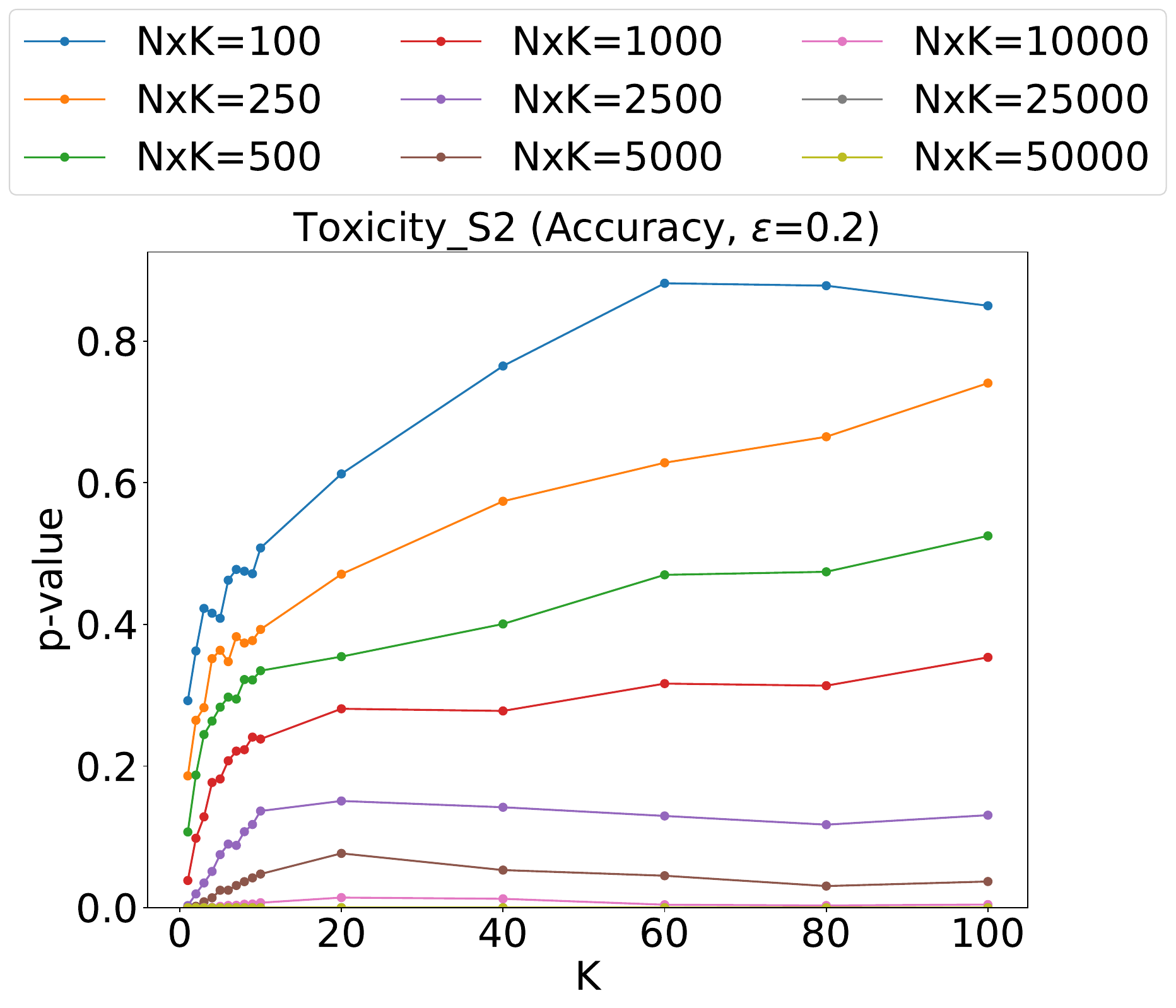}
    \caption{$\epsilon = 0.2$}
    \label{fig:toxicity_s2_acc_e02}
  \end{subfigure} \hfill
  \begin{subfigure}[b]{0.24\linewidth}
    \centering
    \includegraphics[width=\linewidth]{figures/pvals_plots/Toxicity_S2/Toxicity_S2_p_vals_Accuracy_K_100_e_0.3.pdf}
    \caption{$\epsilon = 0.3$}
    \label{fig:toxicity_s2_acc_e03}
  \end{subfigure} \hfill
  \begin{subfigure}[b]{0.24\linewidth}
    \centering
    \includegraphics[width=\linewidth]{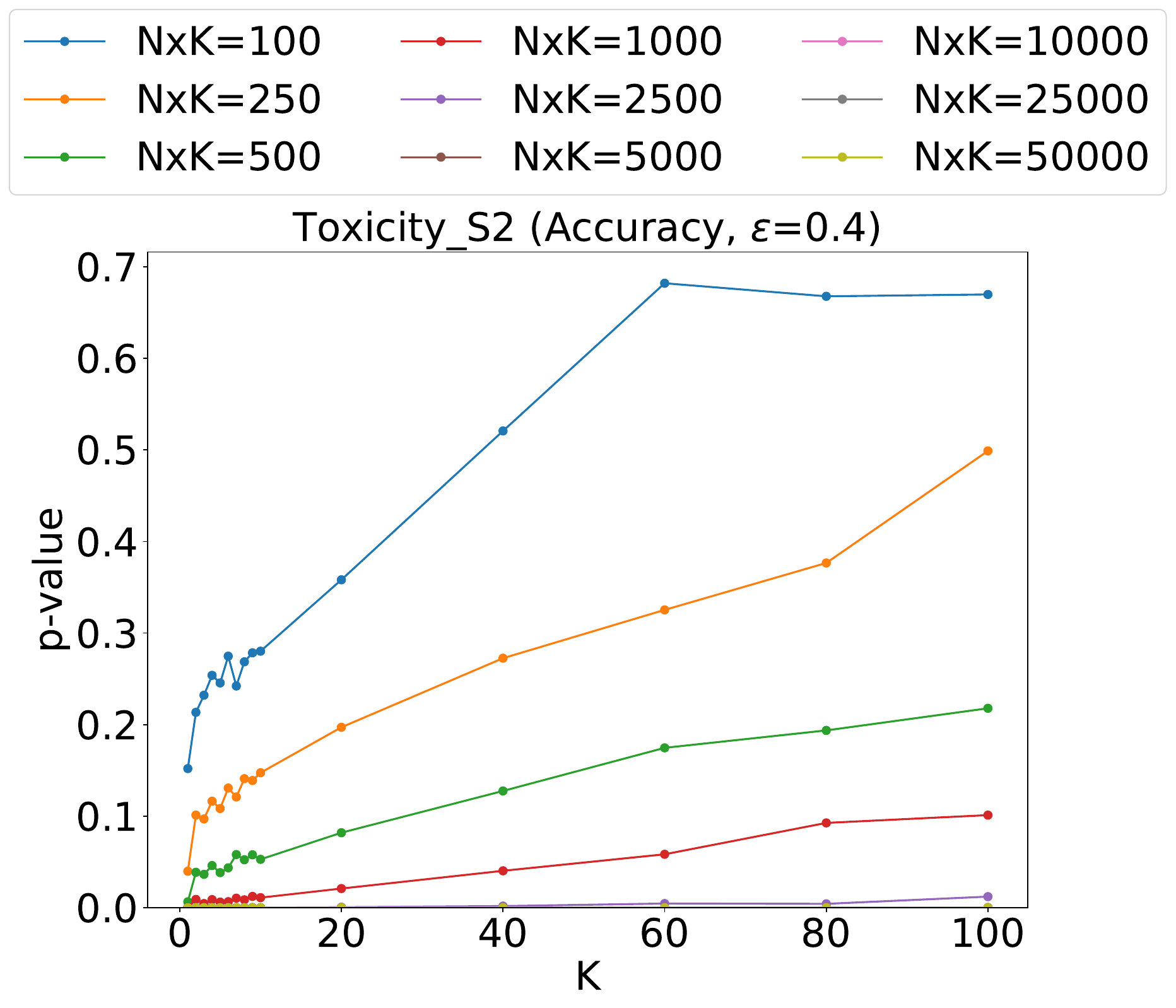}
    \caption{$\epsilon = 0.4$}
    \label{fig:toxicity_s2_acc_e04}
  \end{subfigure}
  \caption{S2: P-value plots for Toxicity dataset with Accuracy as the metric}
  \label{fig:toxicity_s2_accuracy}
\end{figure*}

\begin{figure*}
  \centering
  \begin{subfigure}[b]{0.24\linewidth}
    \centering
    \includegraphics[width=\linewidth]{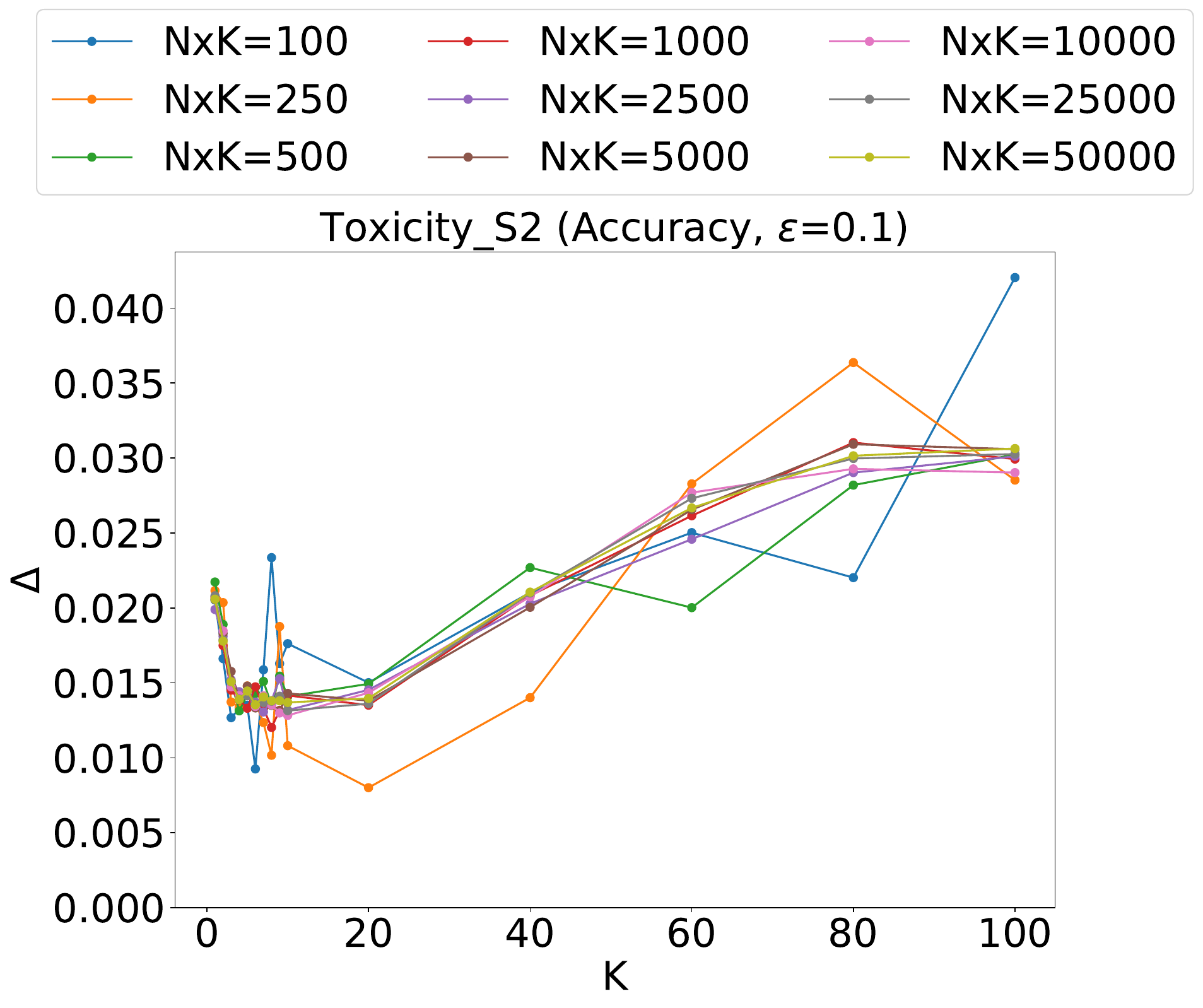}
    \caption{$\epsilon = 0.1$}
    \label{fig:toxicity_s2_delta_acc_e01}
  \end{subfigure} \hfill
  \begin{subfigure}[b]{0.24\linewidth}
    \centering
    \includegraphics[width=\linewidth]{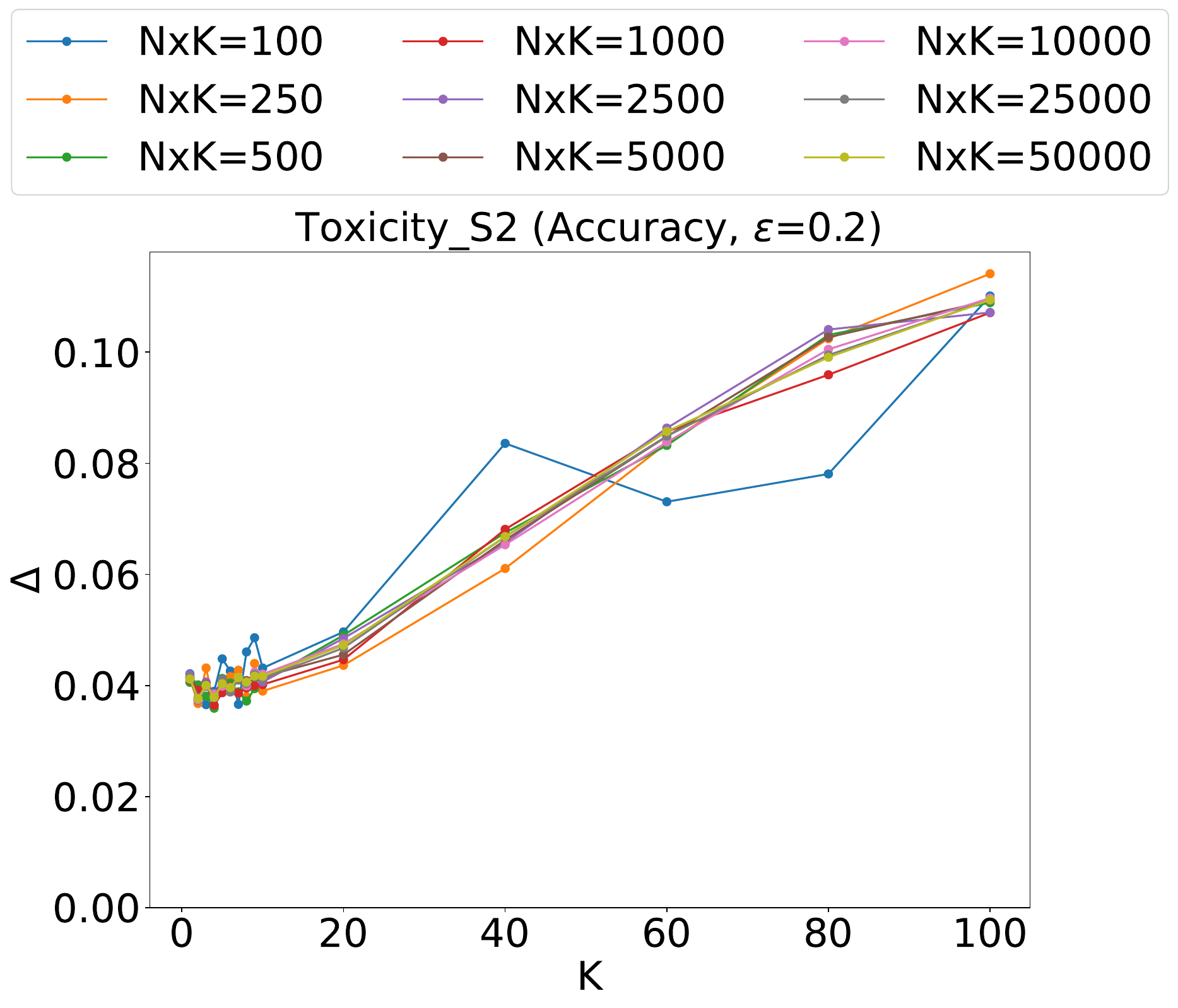}
    \caption{$\epsilon = 0.2$}
    \label{fig:toxicity_s2_delta_acc_e02}
  \end{subfigure} \hfill
  \begin{subfigure}[b]{0.24\linewidth}
    \centering
    \includegraphics[width=\linewidth]{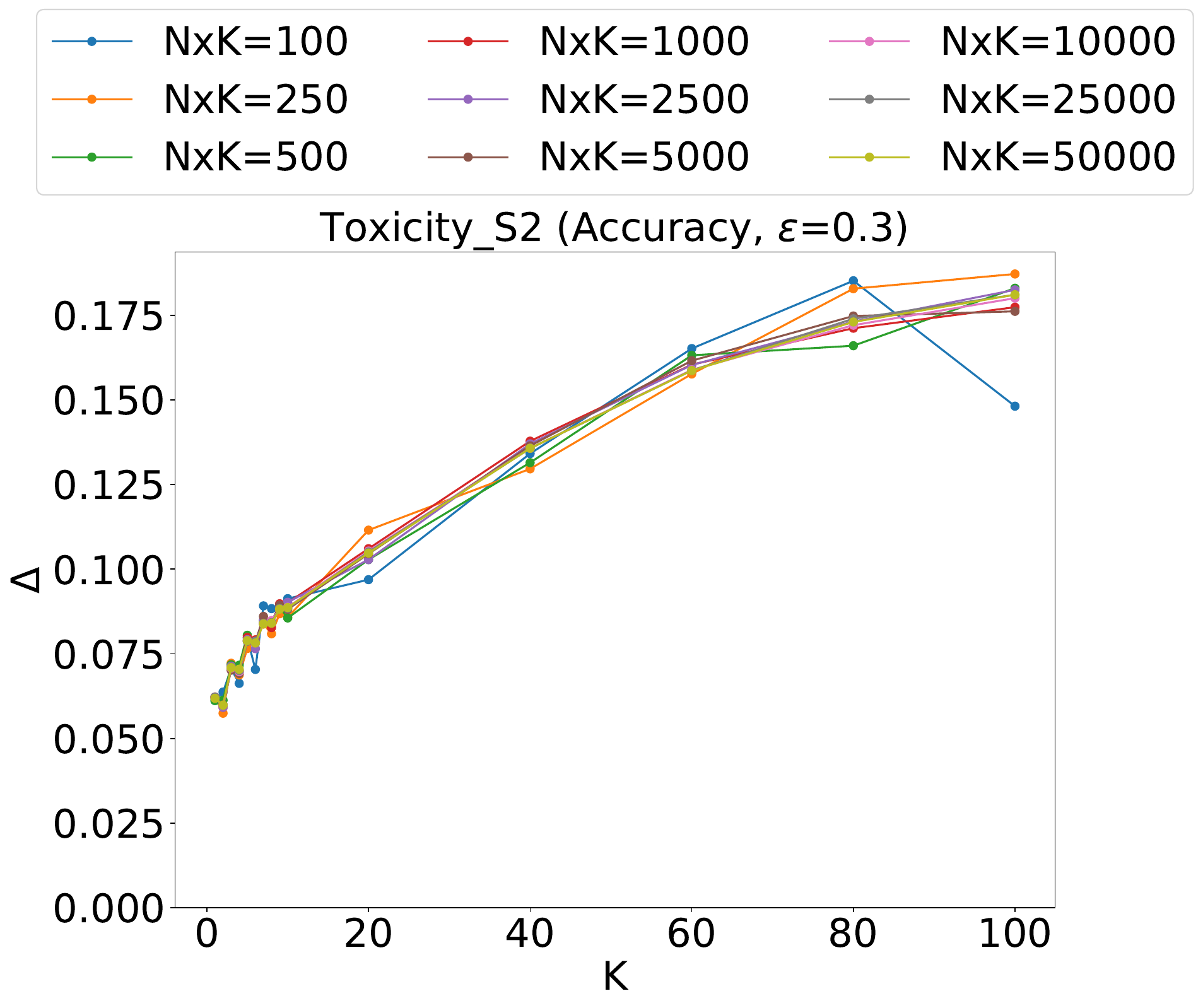}
    \caption{$\epsilon = 0.3$}
    \label{fig:toxicity_s2_delta_acc_e03}
  \end{subfigure} \hfill
  \begin{subfigure}[b]{0.24\linewidth}
    \centering
    \includegraphics[width=\linewidth]{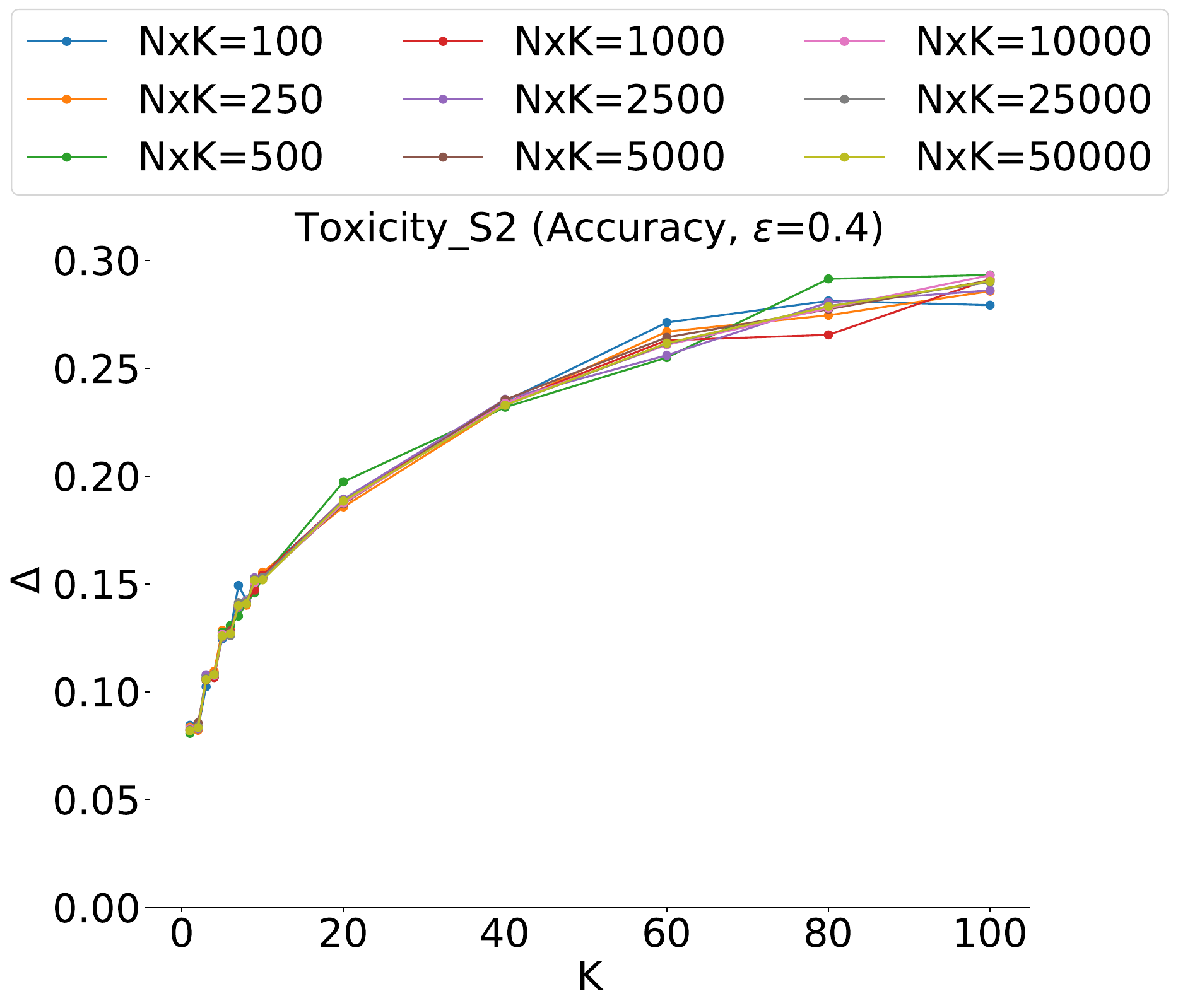}
    \caption{$\epsilon = 0.4$}
    \label{fig:toxicity_s2_delta_acc_e04}
  \end{subfigure}
  \caption{S2: Effect sizes ($\Delta$) for Toxicity dataset with Accuracy as the metric}
  \label{fig:toxicity_s2_delta_accuracy}
\end{figure*}

\begin{figure*}
  \centering
  \begin{subfigure}[b]{0.24\linewidth}
    \centering
    \includegraphics[width=\linewidth]{figures/pvals_plots/Toxicity_S2/Toxicity_S2_p_vals_MAE_K_100_e_0.1.pdf}
    \caption{$\epsilon = 0.1$}
    \label{fig:toxicity_s2_MAE_e01}
  \end{subfigure} \hfill
  \begin{subfigure}[b]{0.24\linewidth}
    \centering
    \includegraphics[width=\linewidth]{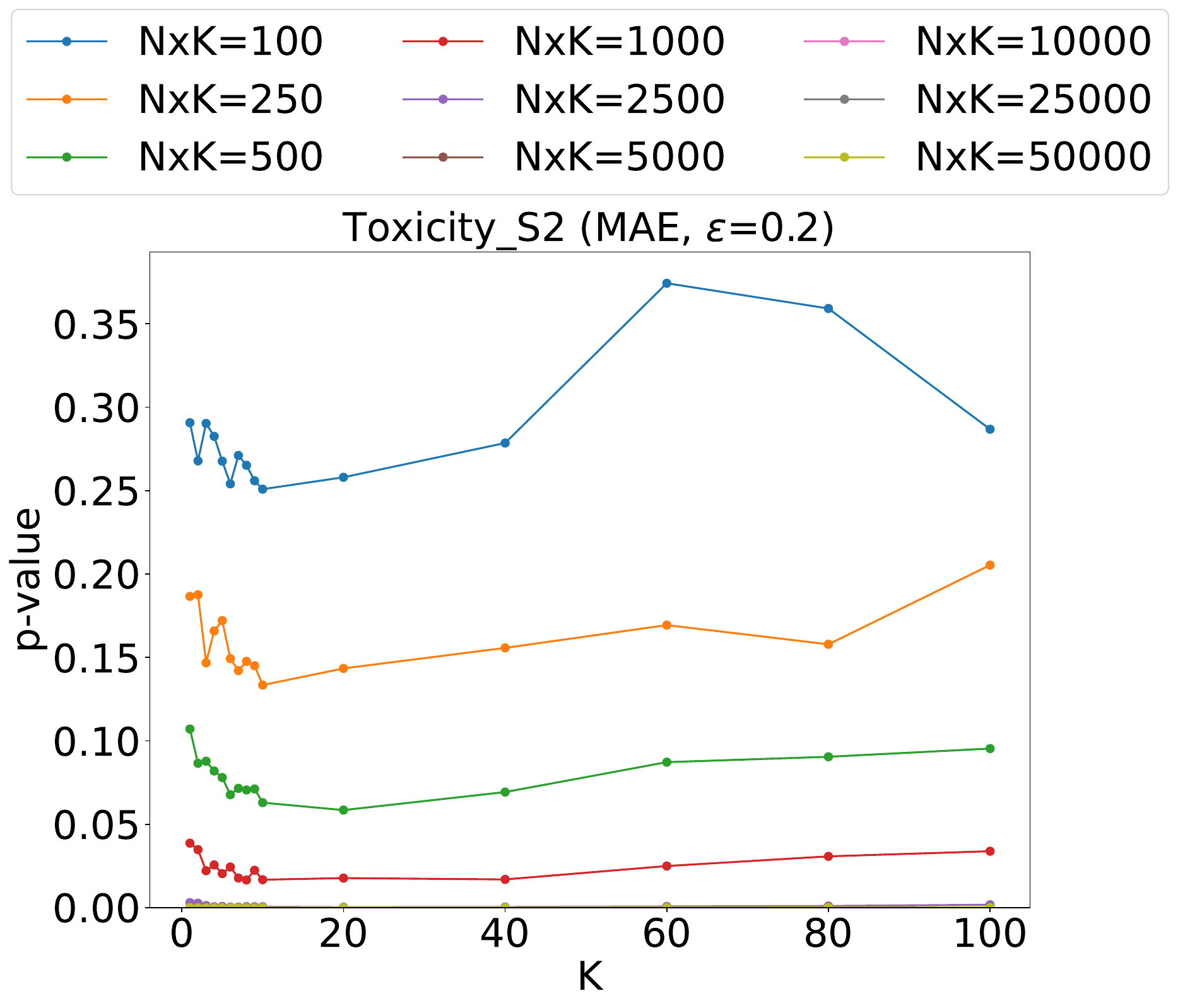}
    \caption{$\epsilon = 0.2$}
    \label{fig:toxicity_s2_MAE_e02}
  \end{subfigure} \hfill
  \begin{subfigure}[b]{0.24\linewidth}
    \centering
    \includegraphics[width=\linewidth]{figures/pvals_plots/Toxicity_S2/Toxicity_S2_p_vals_MAE_K_100_e_0.3.pdf}
    \caption{$\epsilon = 0.3$}
    \label{fig:toxicity_s2_MAE_e03}
  \end{subfigure} \hfill
  \begin{subfigure}[b]{0.24\linewidth}
    \centering
    \includegraphics[width=\linewidth]{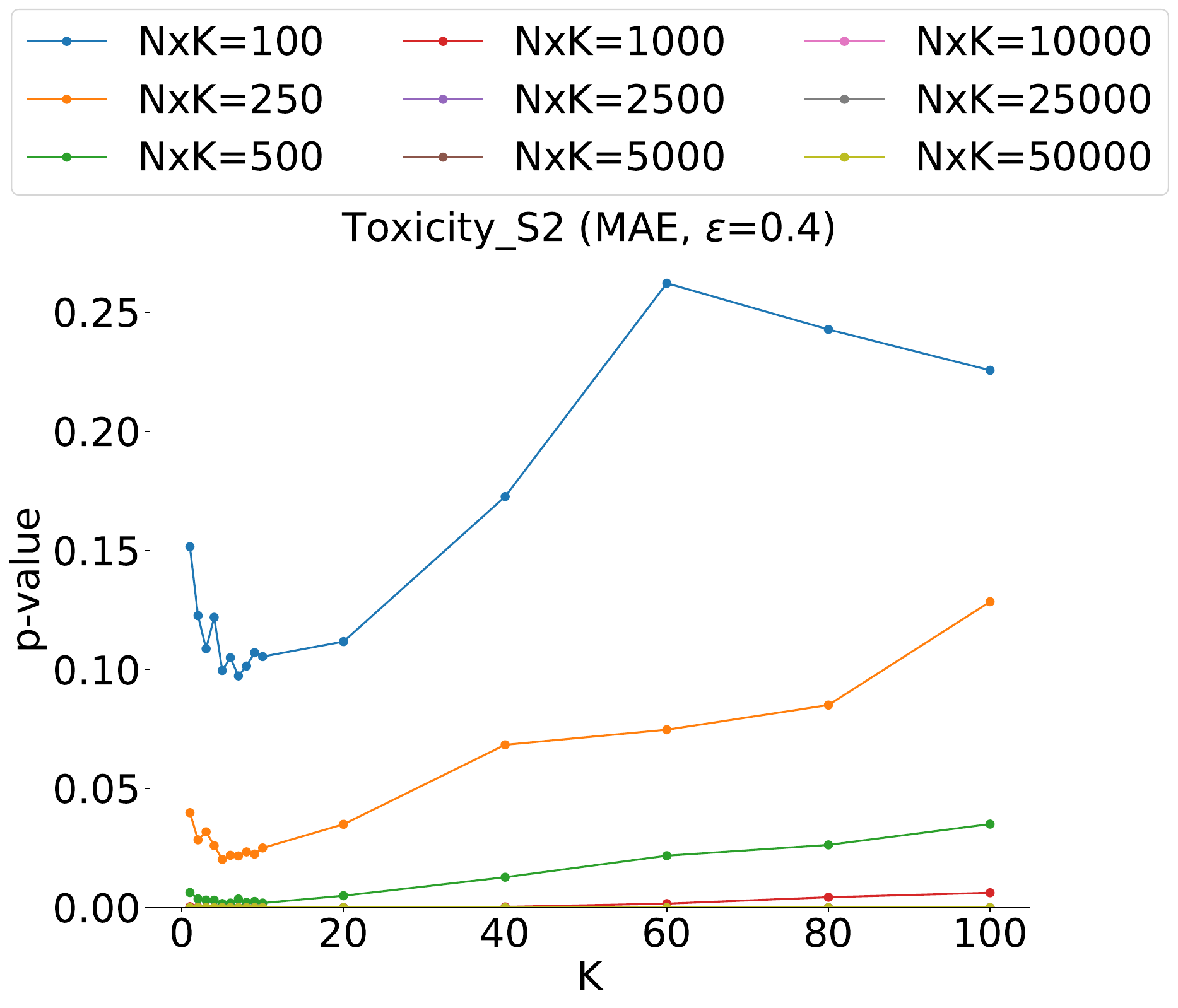}
    \caption{$\epsilon = 0.4$}
    \label{fig:toxicity_s2_MAE_e04}
  \end{subfigure}
  \caption{S2: P-value plots for Toxicity dataset with MAE as the metric}
  \label{fig:toxicity_s2_MAE}
\end{figure*}

\begin{figure*}
  \centering
  \begin{subfigure}[b]{0.24\linewidth}
    \centering
    \includegraphics[width=\linewidth]{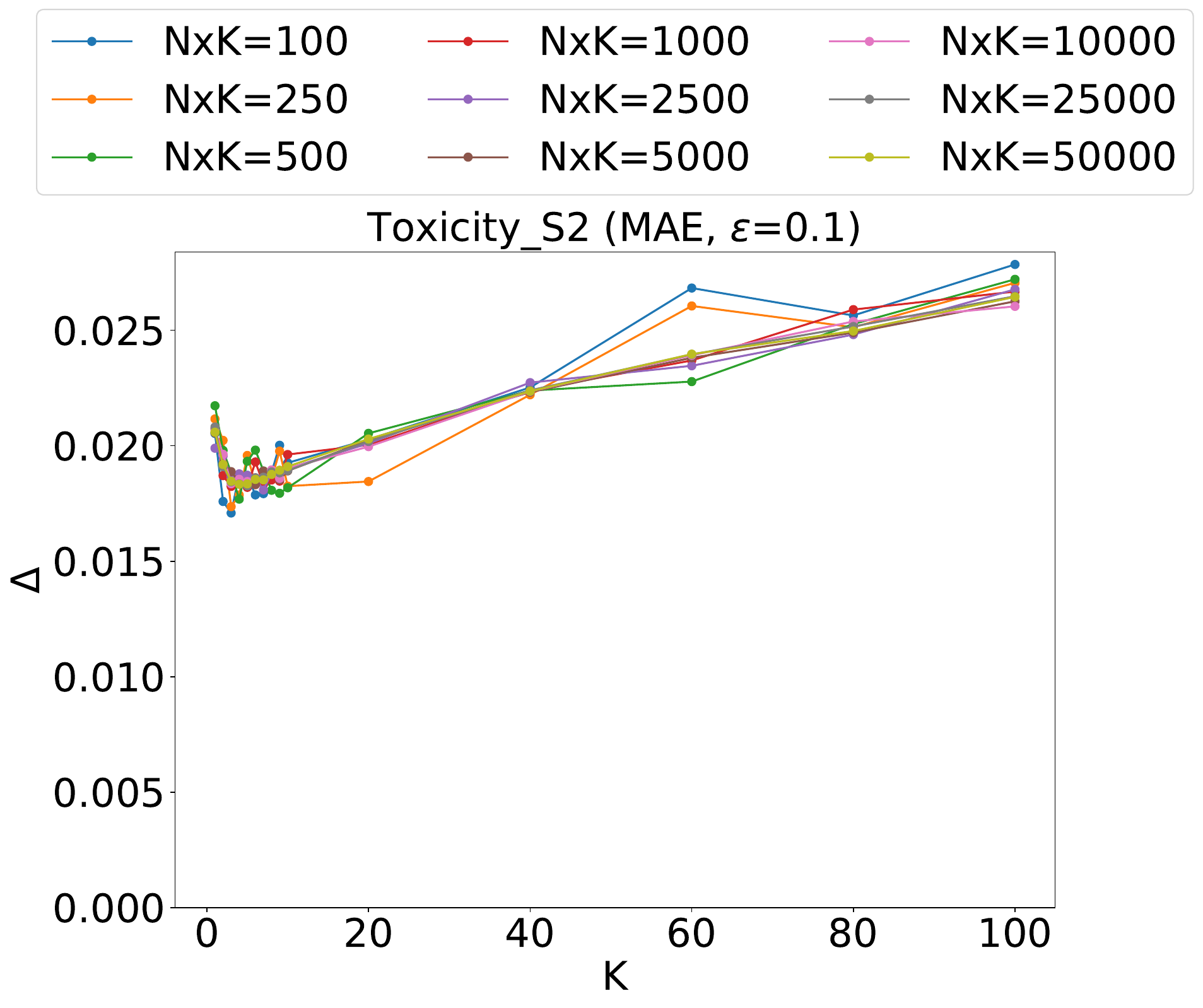}
    \caption{$\epsilon = 0.1$}
    \label{fig:toxicity_s2_delta_MAE_e01}
  \end{subfigure} \hfill
  \begin{subfigure}[b]{0.24\linewidth}
    \centering
    \includegraphics[width=\linewidth]{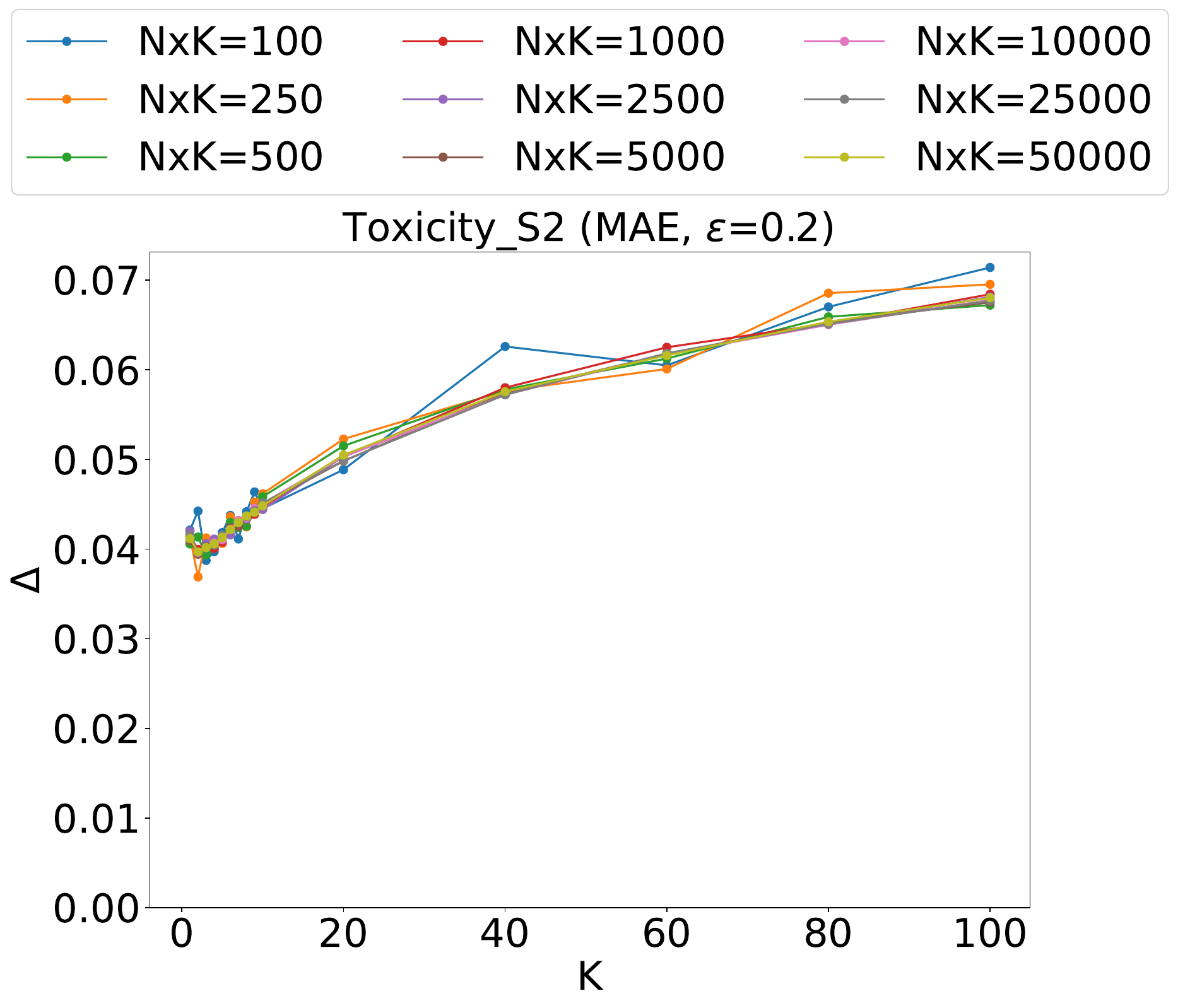}
    \caption{$\epsilon = 0.2$}
    \label{fig:toxicity_s2_delta_MAE_e02}
  \end{subfigure} \hfill
  \begin{subfigure}[b]{0.24\linewidth}
    \centering
    \includegraphics[width=\linewidth]{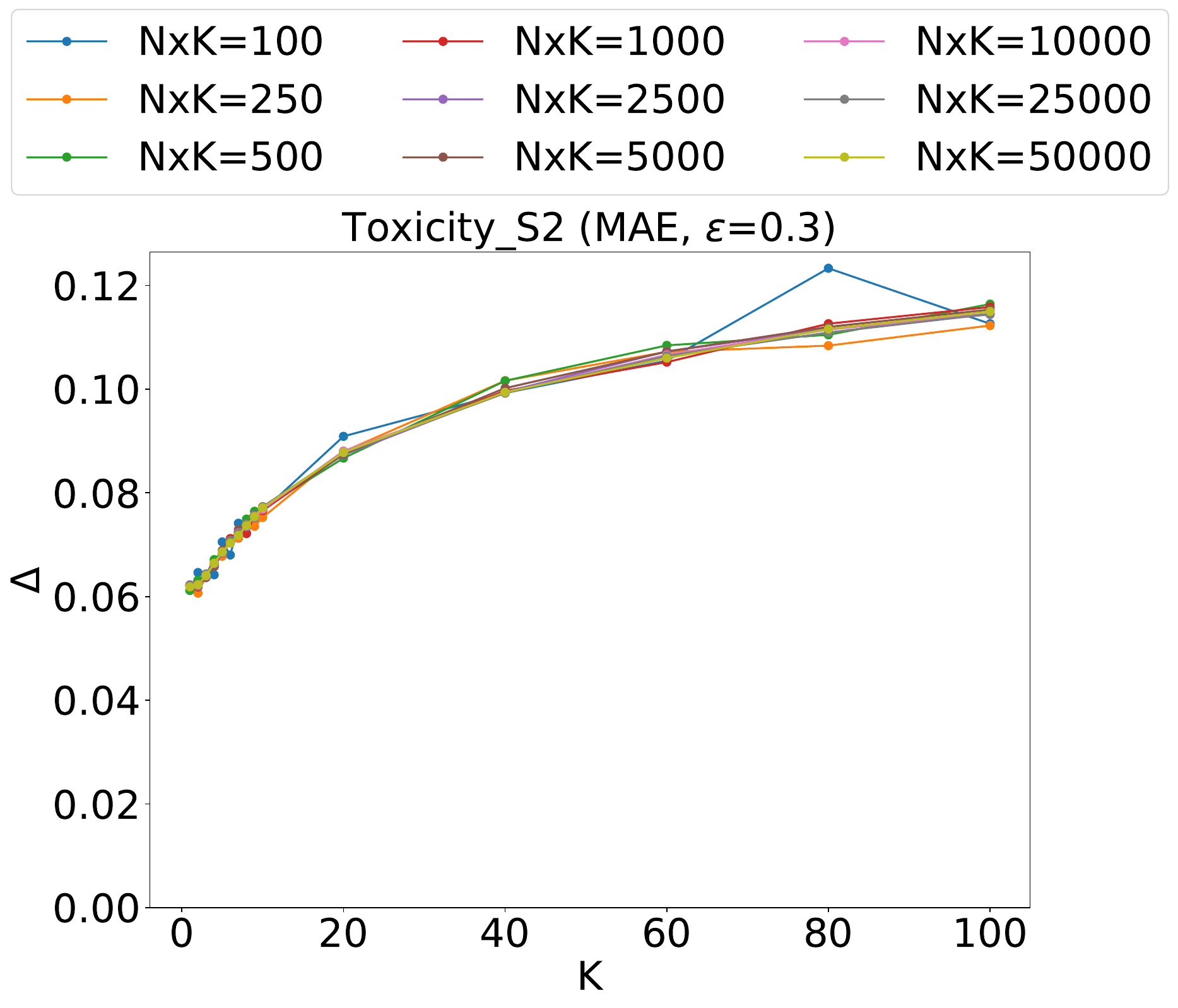}
    \caption{$\epsilon = 0.3$}
    \label{fig:toxicity_s2_delta_MAE_e03}
  \end{subfigure} \hfill
  \begin{subfigure}[b]{0.24\linewidth}
    \centering
    \includegraphics[width=\linewidth]{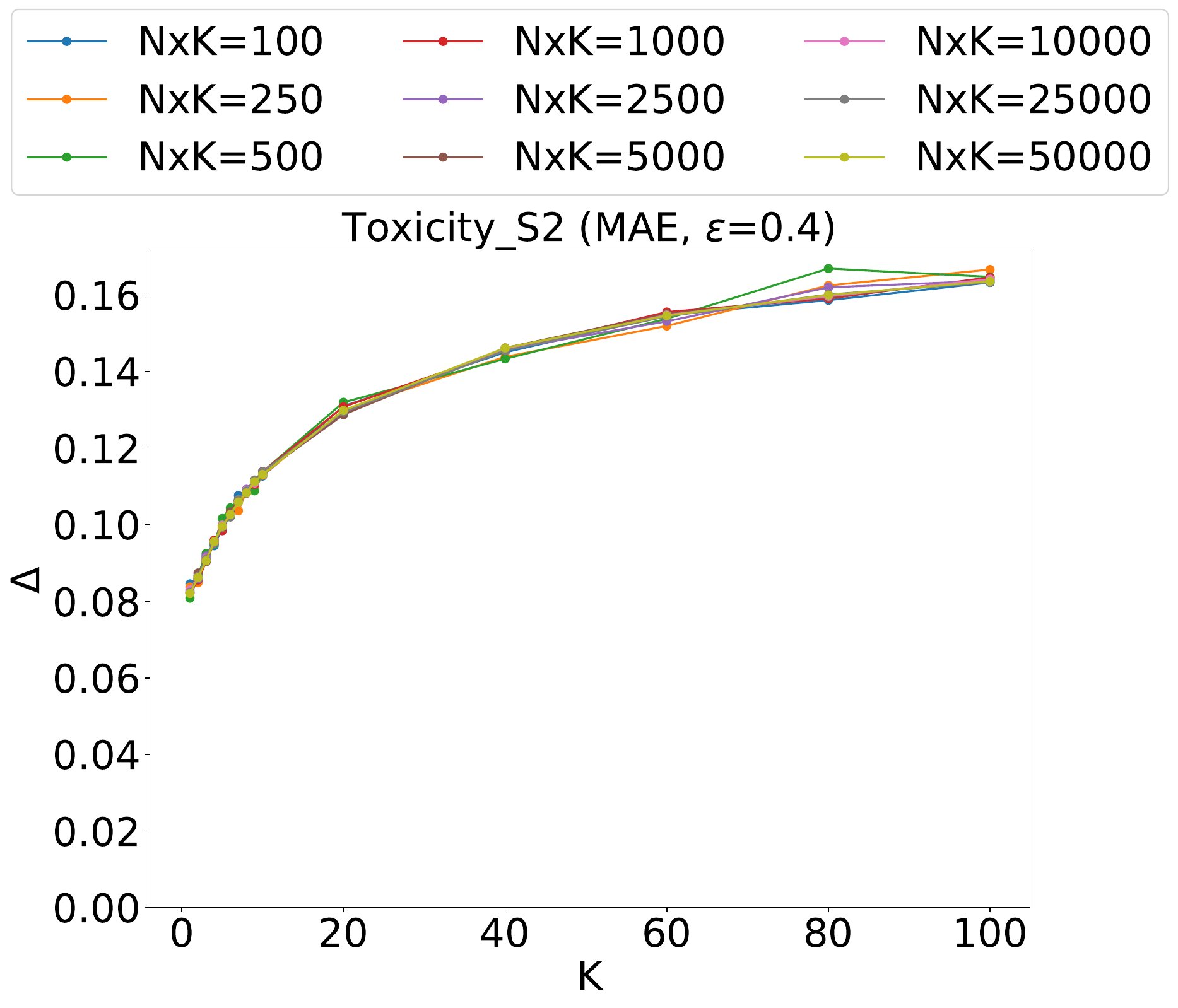}
    \caption{$\epsilon = 0.4$}
    \label{fig:toxicity_s2_delta_MAE_e04}
  \end{subfigure}
  \caption{S2: Effect sizes ($\Delta$) for Toxicity dataset with MAE as the metric}
  \label{fig:toxicity_s2_delta_MAE}
\end{figure*}

\begin{figure*}
  \centering
  \begin{subfigure}[b]{0.24\linewidth}
    \centering
    \includegraphics[width=\linewidth]{figures/pvals_plots/Toxicity_S2/Toxicity_S2_p_vals_Wins_K_100_e_0.1.pdf}
    \caption{$\epsilon = 0.1$}
    \label{fig:toxicity_s2_wins_e01}
  \end{subfigure} \hfill
  \begin{subfigure}[b]{0.24\linewidth}
    \centering
    \includegraphics[width=\linewidth]{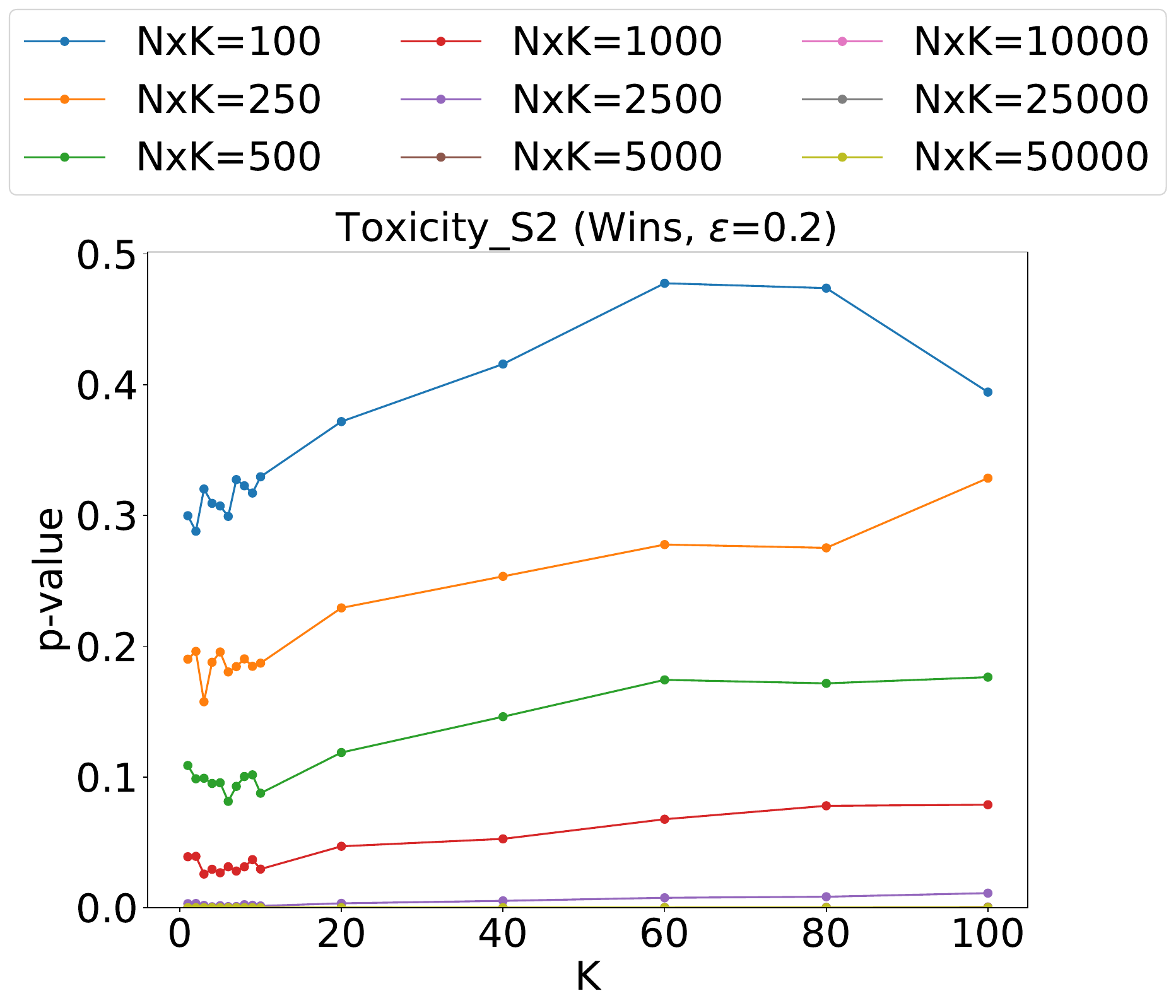}
    \caption{$\epsilon = 0.2$}
    \label{fig:toxicity_s2_wins_e02}
  \end{subfigure} \hfill
  \begin{subfigure}[b]{0.24\linewidth}
    \centering
    \includegraphics[width=\linewidth]{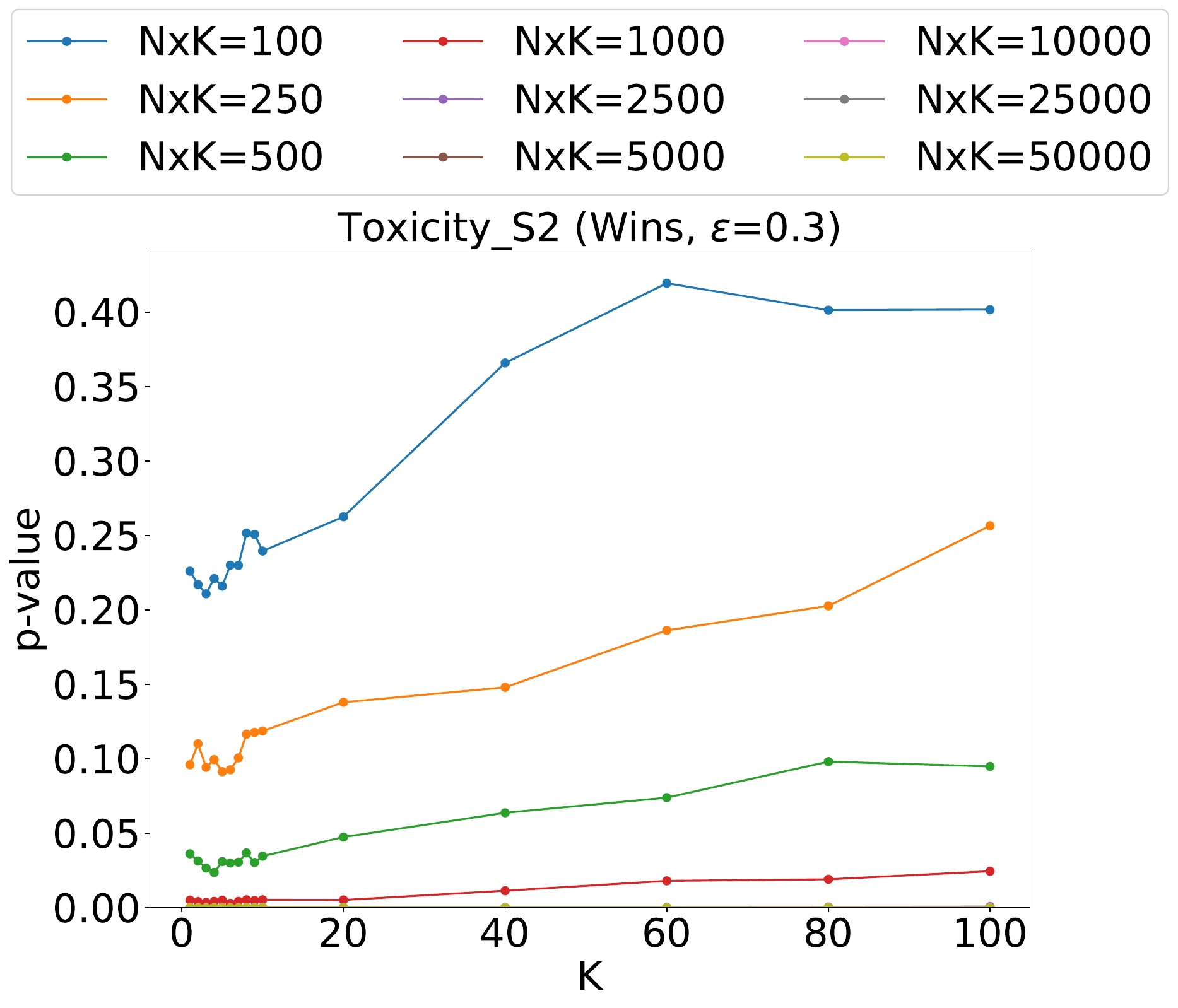}
    \caption{$\epsilon = 0.3$}
    \label{fig:toxicity_s2_wins_e03}
  \end{subfigure} \hfill
  \begin{subfigure}[b]{0.24\linewidth}
    \centering
    \includegraphics[width=\linewidth]{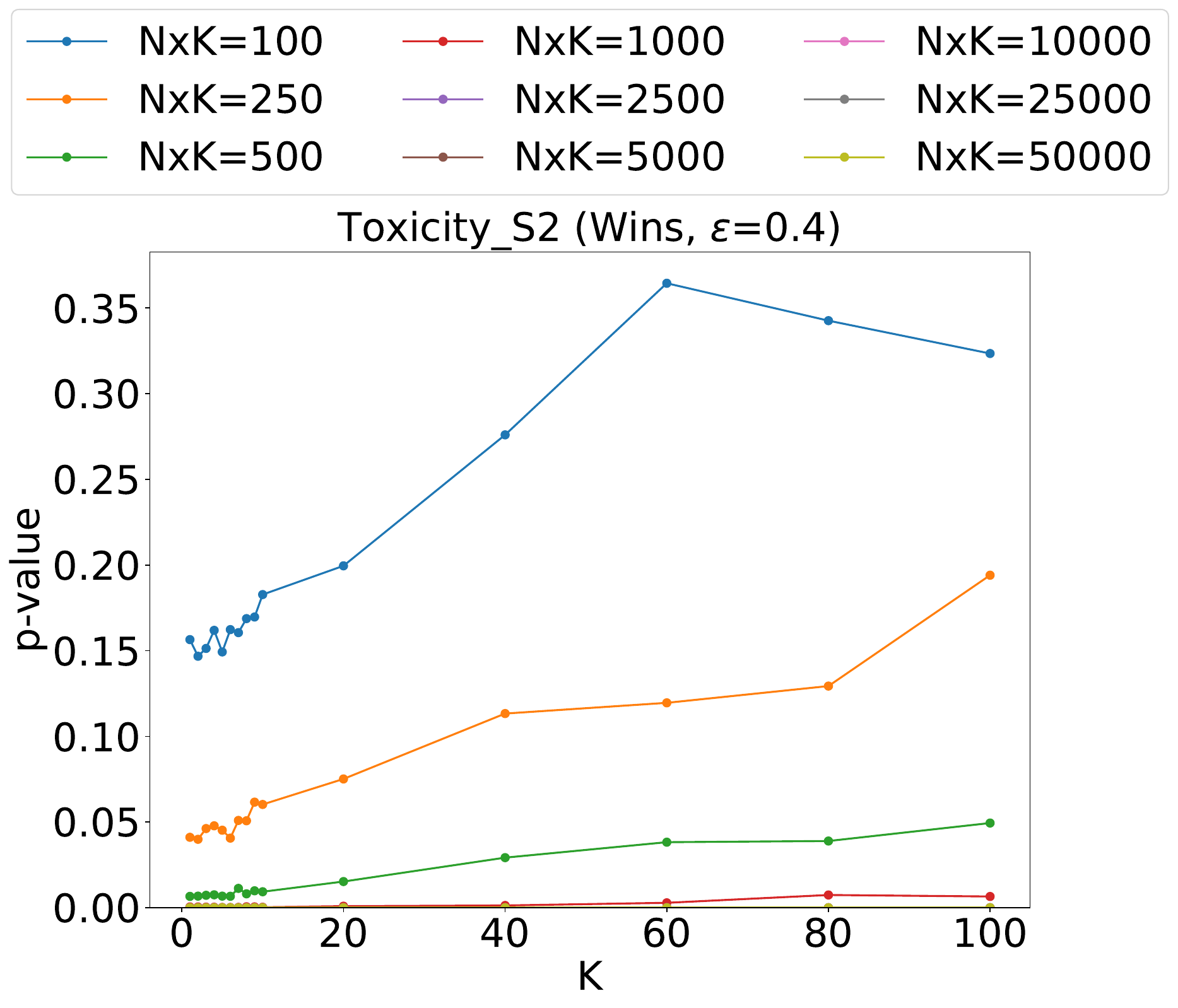}
    \caption{$\epsilon = 0.4$}
    \label{fig:toxicity_s2_wins_e04}
  \end{subfigure}
  \caption{S2: P-value plots for Toxicity dataset with Wins as the metric}
  \label{fig:toxicity_s2_wins}
\end{figure*}

\begin{figure*}
  \centering
  \begin{subfigure}[b]{0.24\linewidth}
    \centering
    \includegraphics[width=\linewidth]{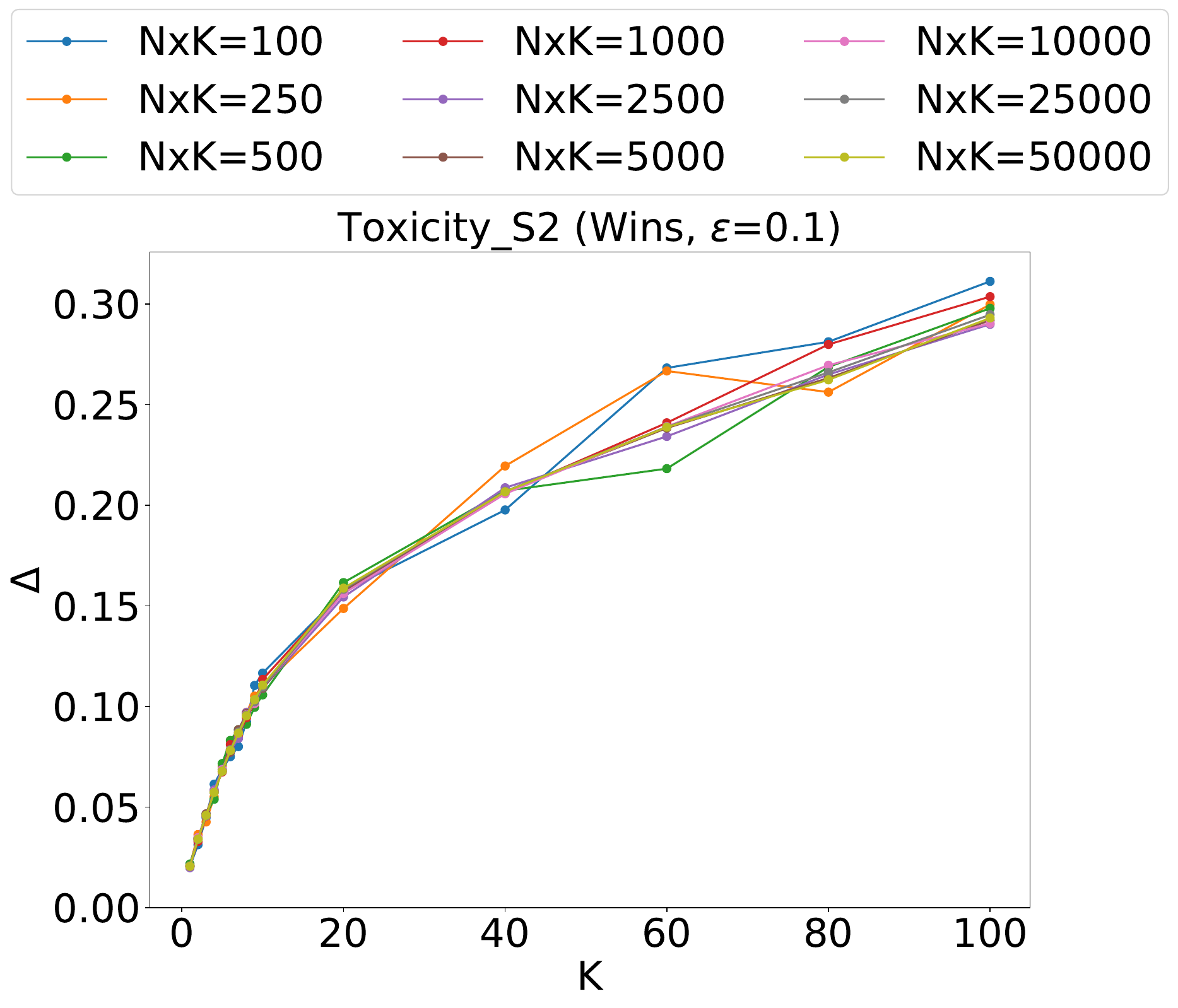}
    \caption{$\epsilon = 0.1$}
    \label{fig:toxicity_s2_delta_wins_e01}
  \end{subfigure} \hfill
  \begin{subfigure}[b]{0.24\linewidth}
    \centering
    \includegraphics[width=\linewidth]{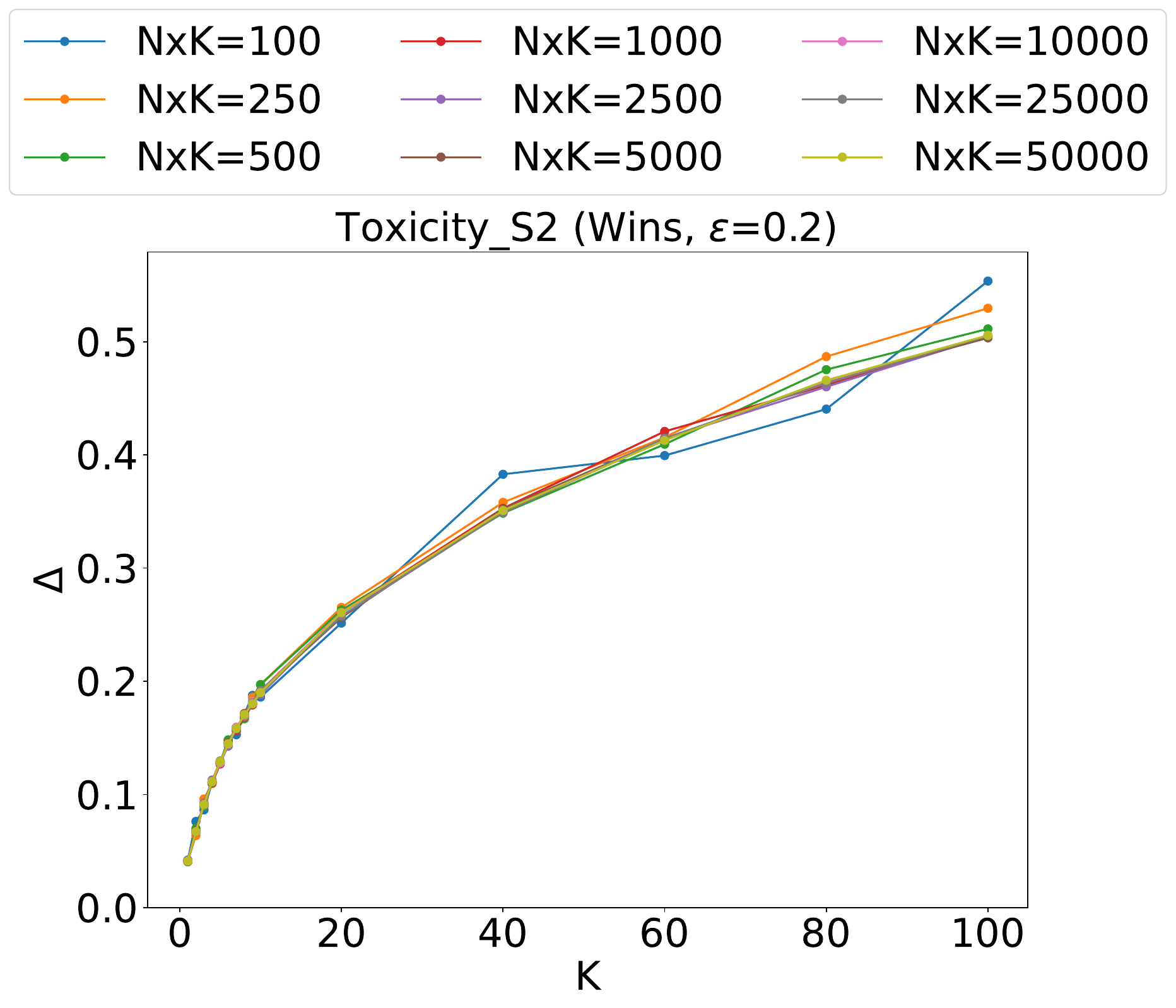}
    \caption{$\epsilon = 0.2$}
    \label{fig:toxicity_s2_delta_wins_e02}
  \end{subfigure} \hfill
  \begin{subfigure}[b]{0.24\linewidth}
    \centering
    \includegraphics[width=\linewidth]{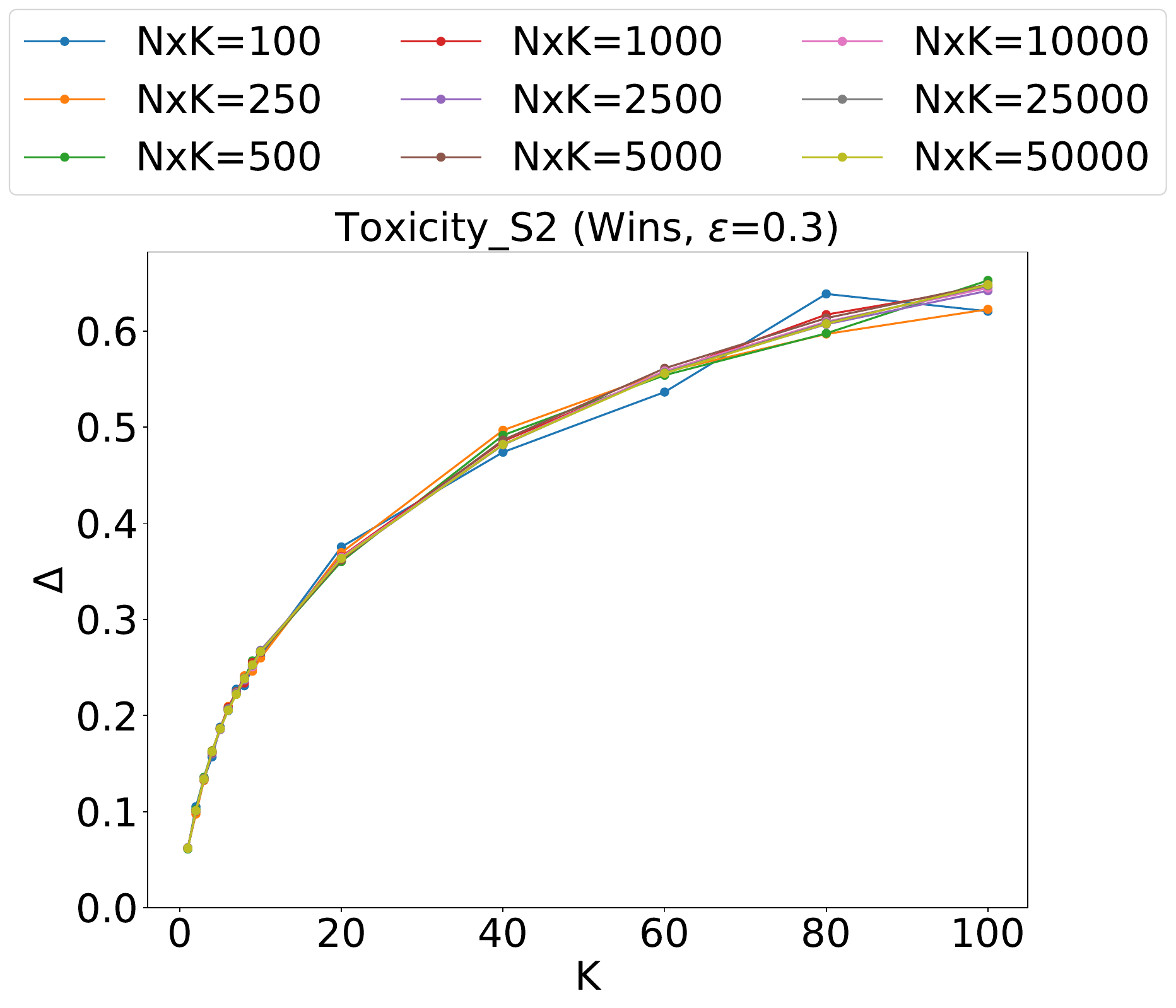}
    \caption{$\epsilon = 0.3$}
    \label{fig:toxicity_s2_delta_wins_e03}
  \end{subfigure} \hfill
  \begin{subfigure}[b]{0.24\linewidth}
    \centering
    \includegraphics[width=\linewidth]{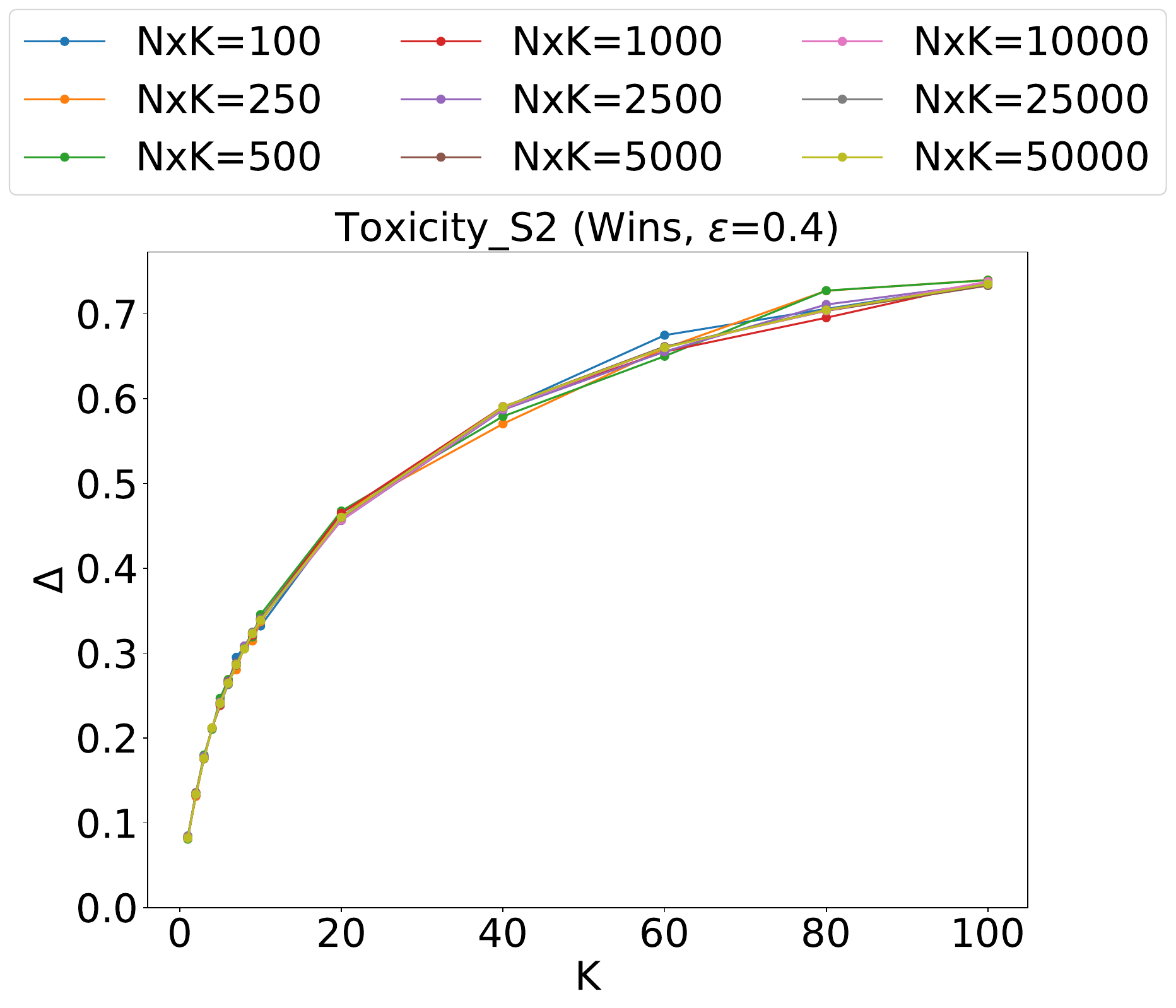}
    \caption{$\epsilon = 0.4$}
    \label{fig:toxicity_s2_delta_wins_e04}
  \end{subfigure}
  \caption{S2: Effect sizes ($\Delta$) for Toxicity dataset with Wins as the metric}
  \label{fig:toxicity_s2_delta_wins}
\end{figure*}

\paragraph{D3code}

\begin{figure*}
  \centering
  \begin{subfigure}[b]{0.24\linewidth}
    \centering
    \includegraphics[width=\linewidth]{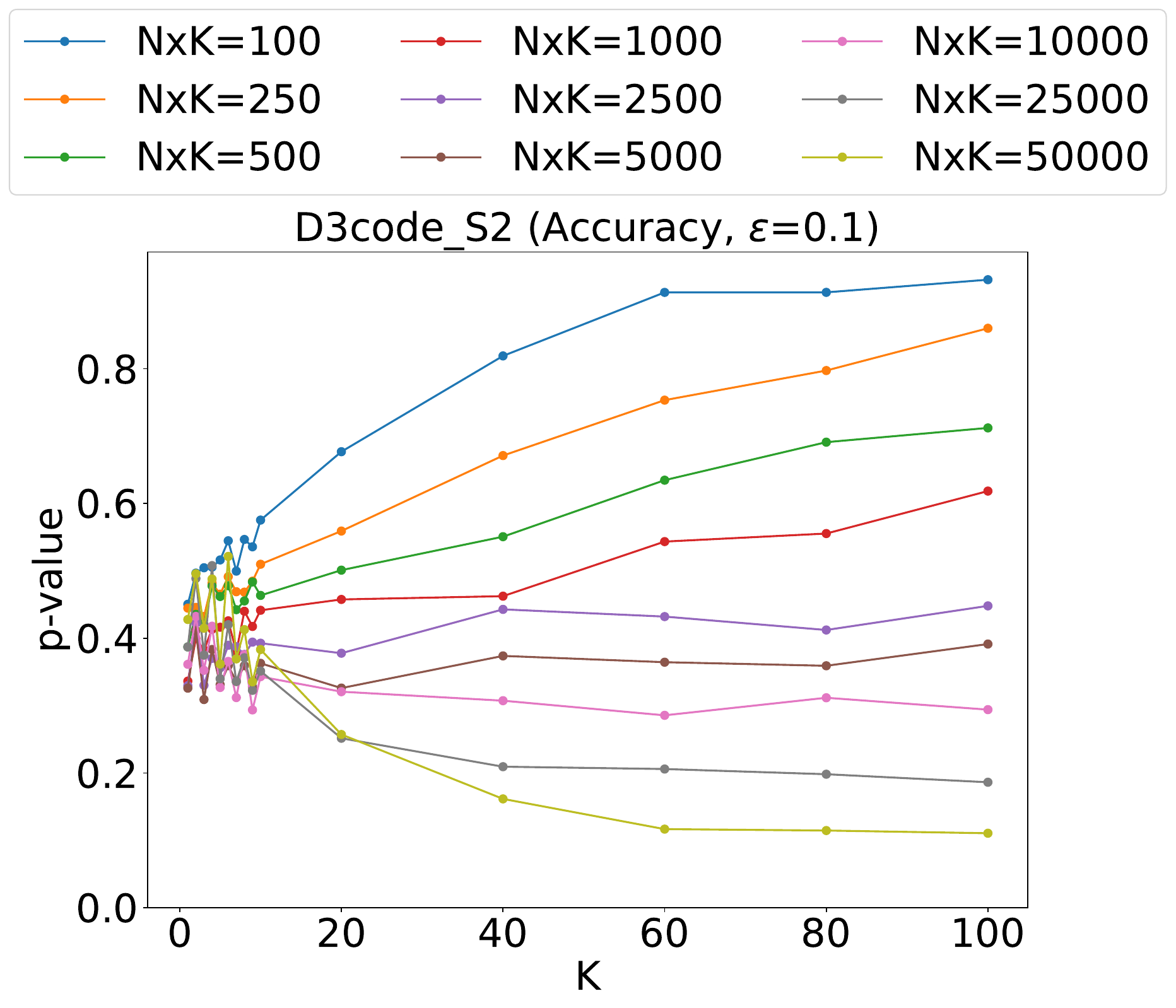}
    \caption{$\epsilon = 0.1$}
    \label{fig:d3code_s2_acc_e01}
  \end{subfigure} \hfill
  \begin{subfigure}[b]{0.24\linewidth}
    \centering
    \includegraphics[width=\linewidth]{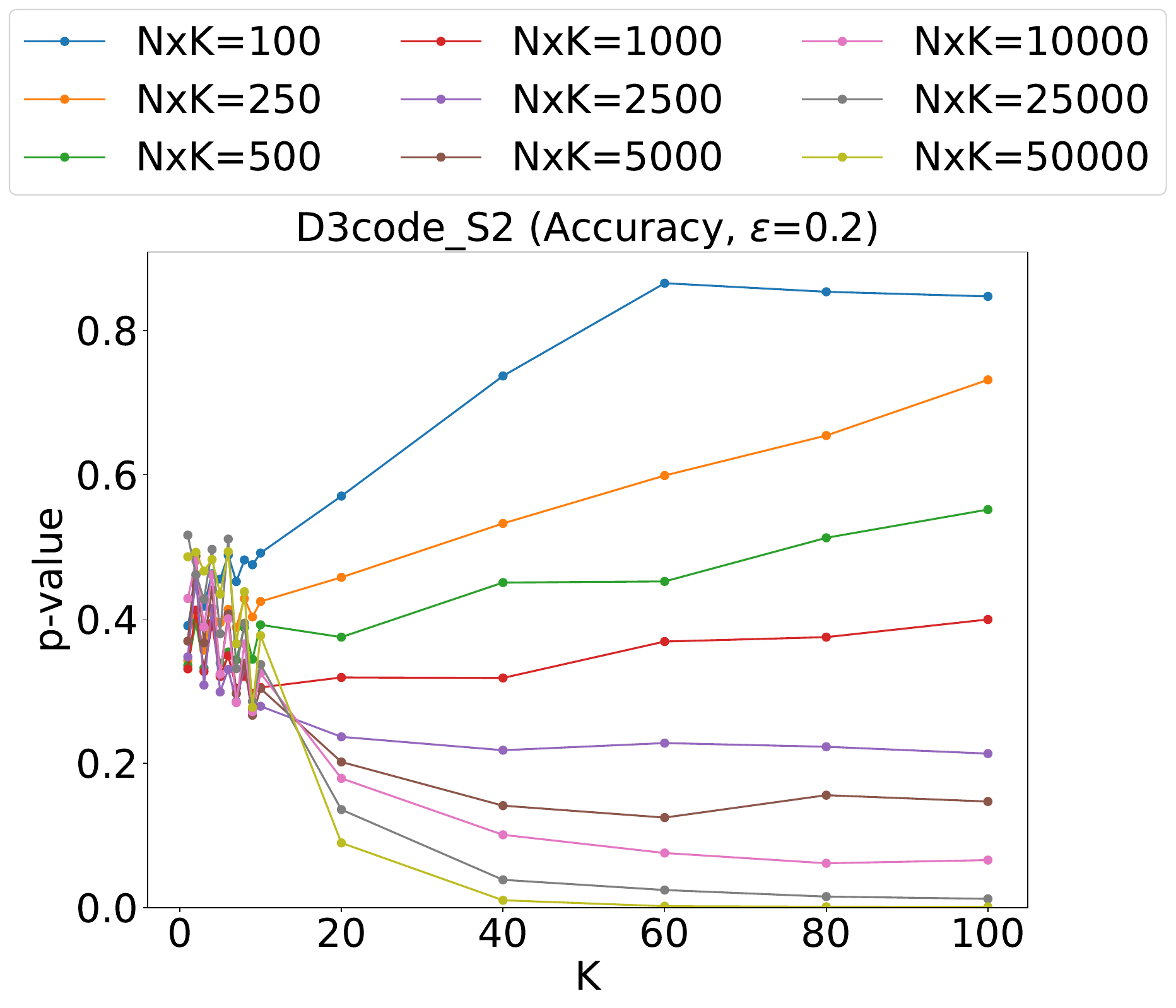}
    \caption{$\epsilon = 0.2$}
    \label{fig:d3code_s2_acc_e02}
  \end{subfigure} \hfill
  \begin{subfigure}[b]{0.24\linewidth}
    \centering
    \includegraphics[width=\linewidth]{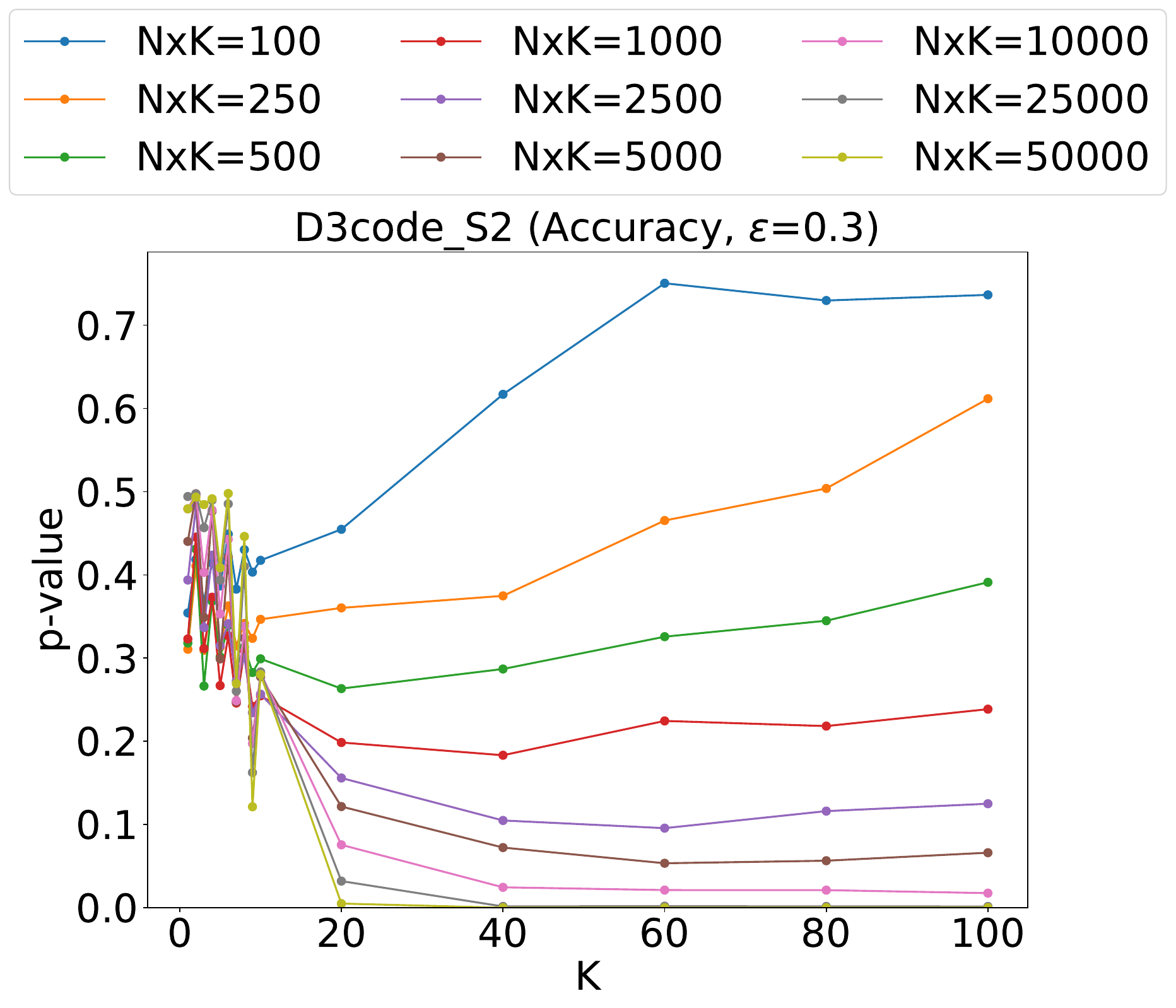}
    \caption{$\epsilon = 0.3$}
    \label{fig:d3code_s2_acc_e03}
  \end{subfigure} \hfill
  \begin{subfigure}[b]{0.24\linewidth}
    \centering
    \includegraphics[width=\linewidth]{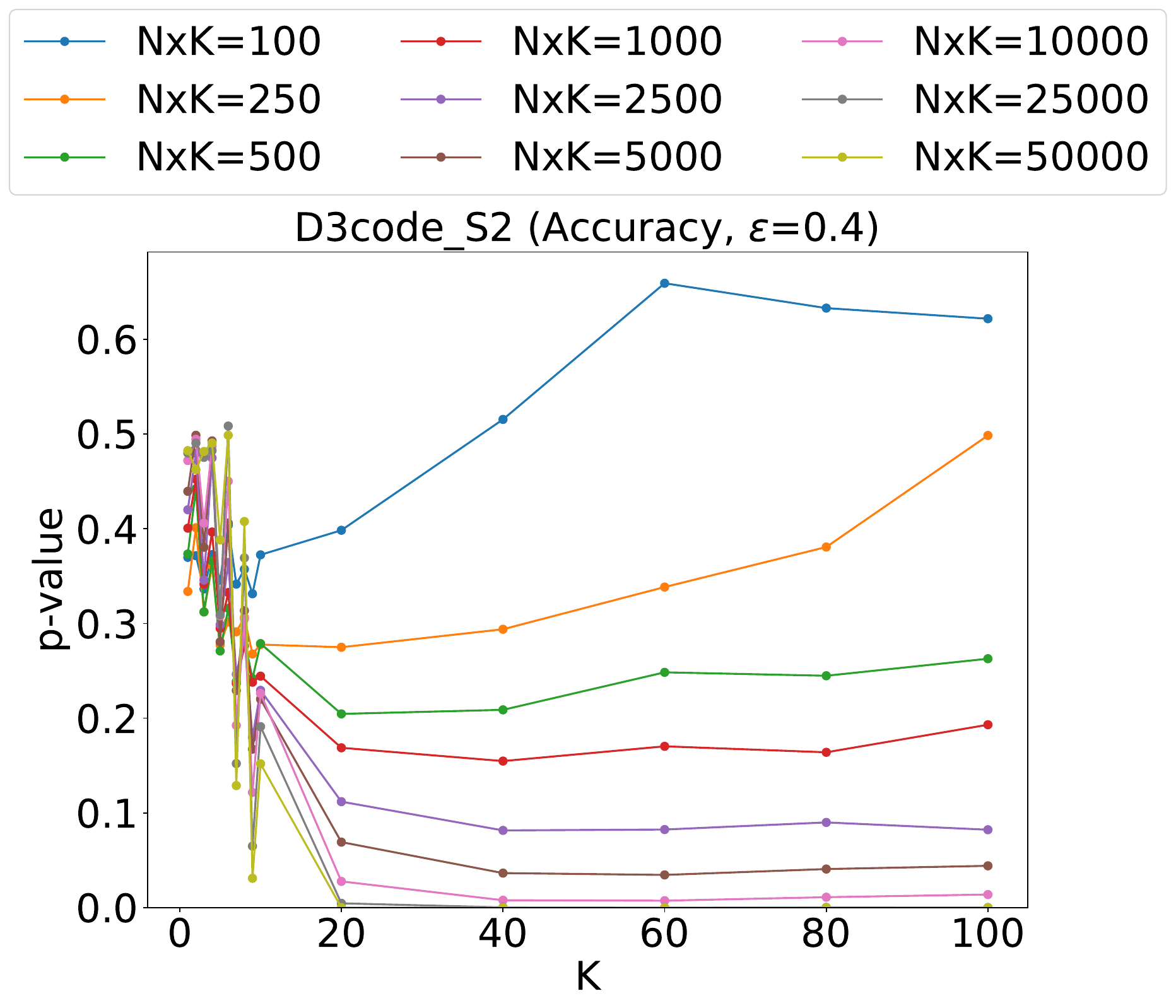}
    \caption{$\epsilon = 0.4$}
    \label{fig:d3code_s2_acc_e04}
  \end{subfigure}
  \caption{S2: P-value plots for D3code dataset with Accuracy as the metric}
  \label{fig:d3code_s2_accuracy}
\end{figure*}

\begin{figure*}
  \centering
  \begin{subfigure}[b]{0.24\linewidth}
    \centering
    \includegraphics[width=\linewidth]{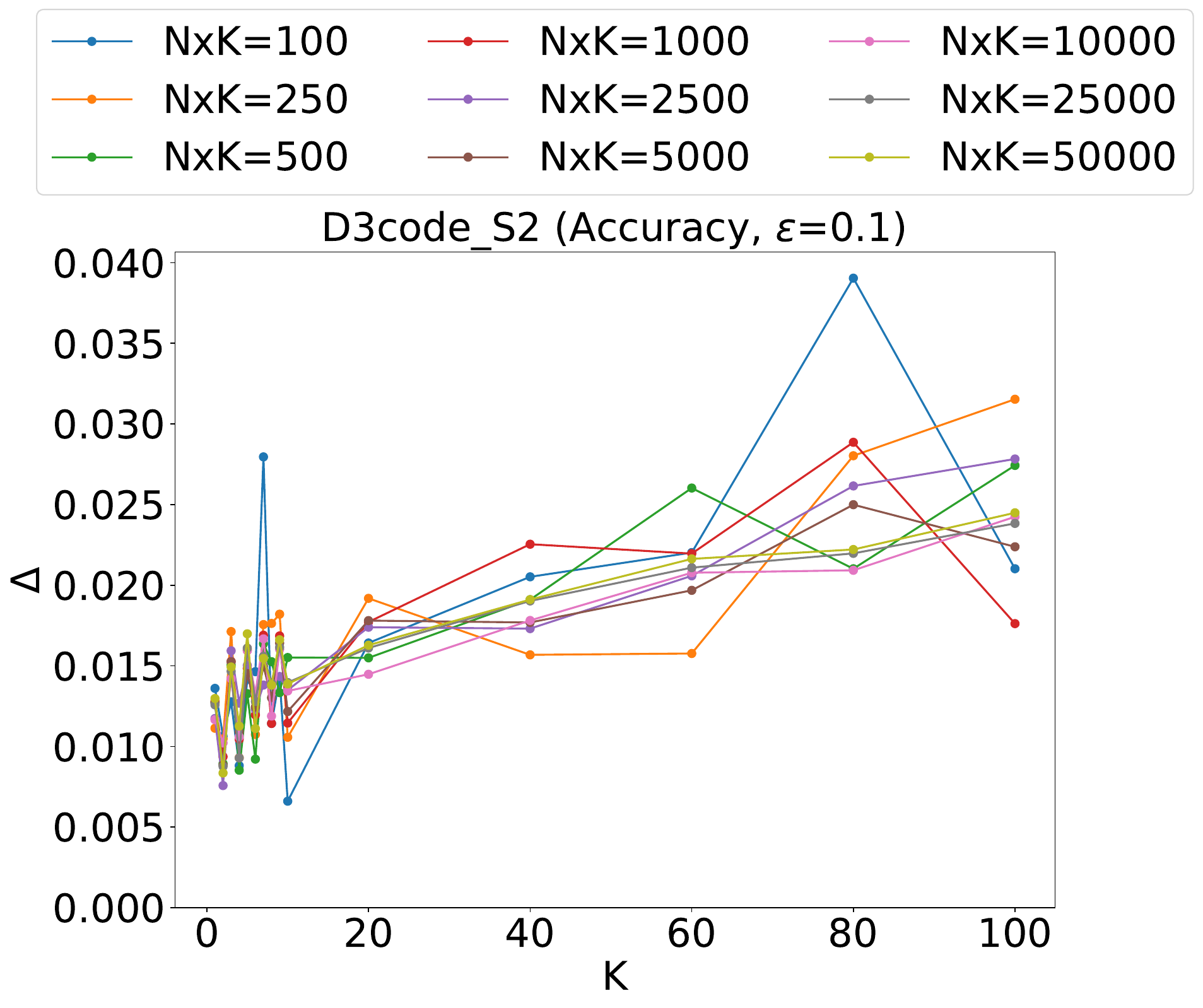}
    \caption{$\epsilon = 0.1$}
    \label{fig:d3code_s2_delta_acc_e01}
  \end{subfigure} \hfill
  \begin{subfigure}[b]{0.24\linewidth}
    \centering
    \includegraphics[width=\linewidth]{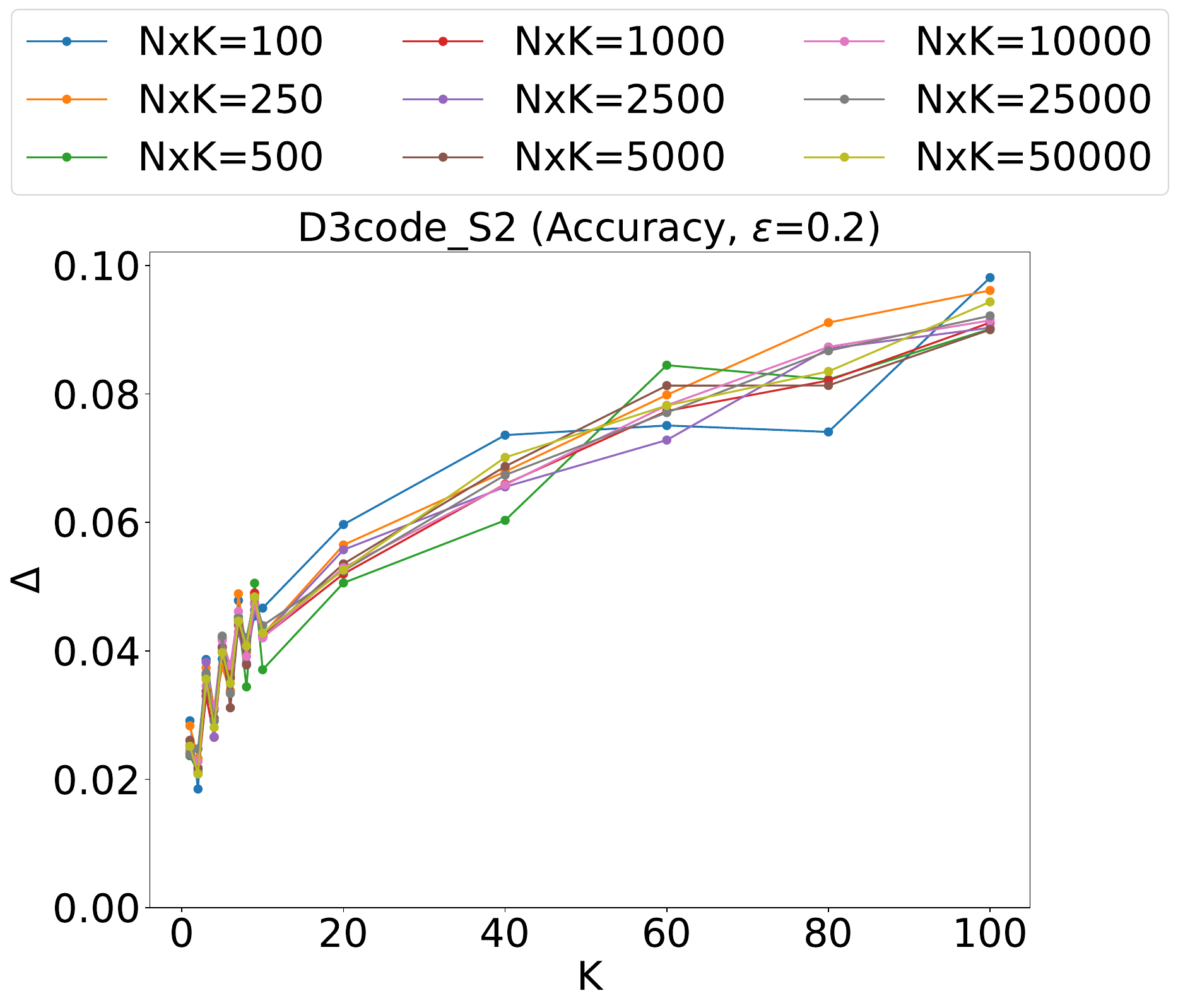}
    \caption{$\epsilon = 0.2$}
    \label{fig:d3code_s2_delta_acc_e02}
  \end{subfigure} \hfill
  \begin{subfigure}[b]{0.24\linewidth}
    \centering
    \includegraphics[width=\linewidth]{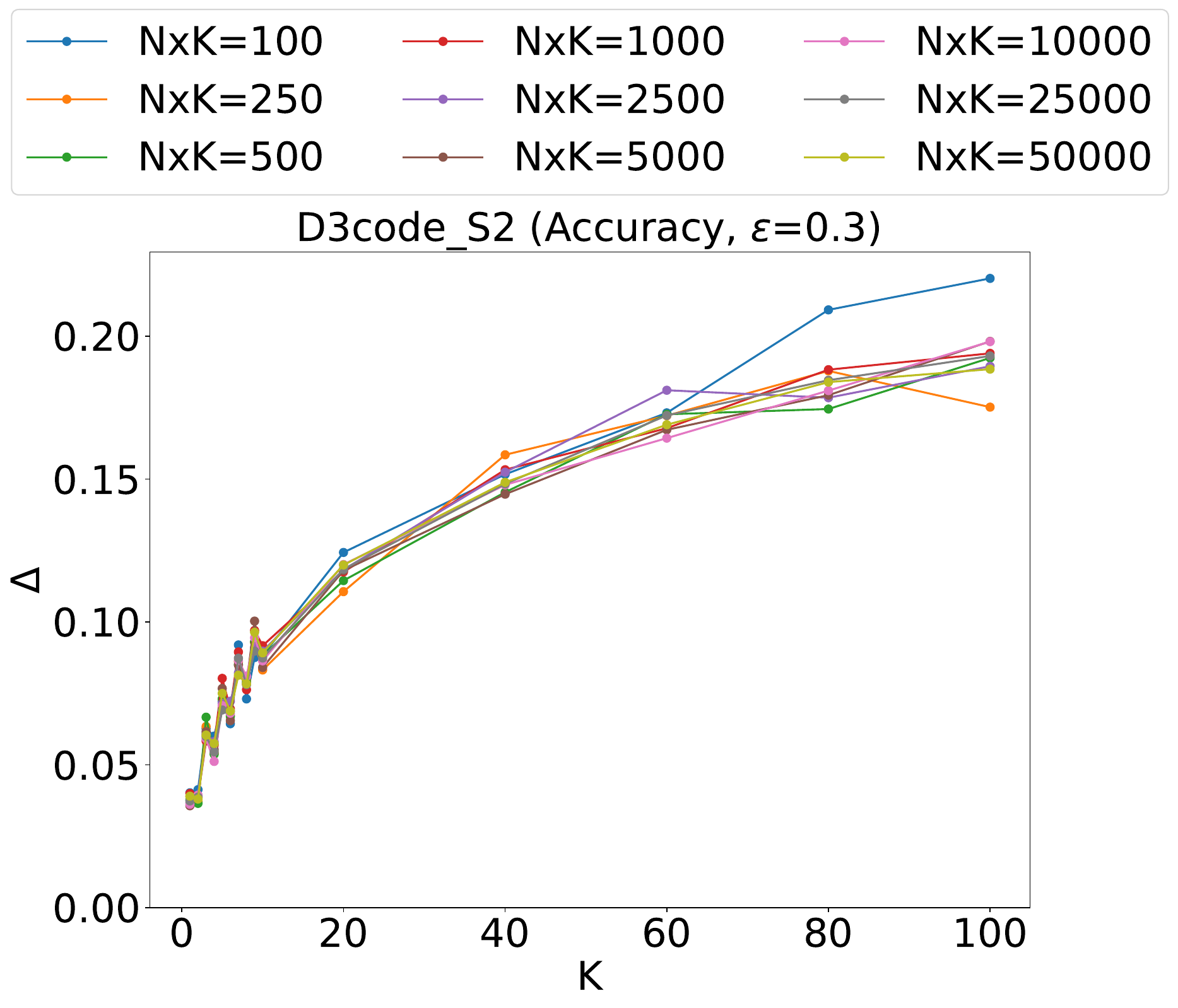}
    \caption{$\epsilon = 0.3$}
    \label{fig:d3code_s2_delta_acc_e03}
  \end{subfigure} \hfill
  \begin{subfigure}[b]{0.24\linewidth}
    \centering
    \includegraphics[width=\linewidth]{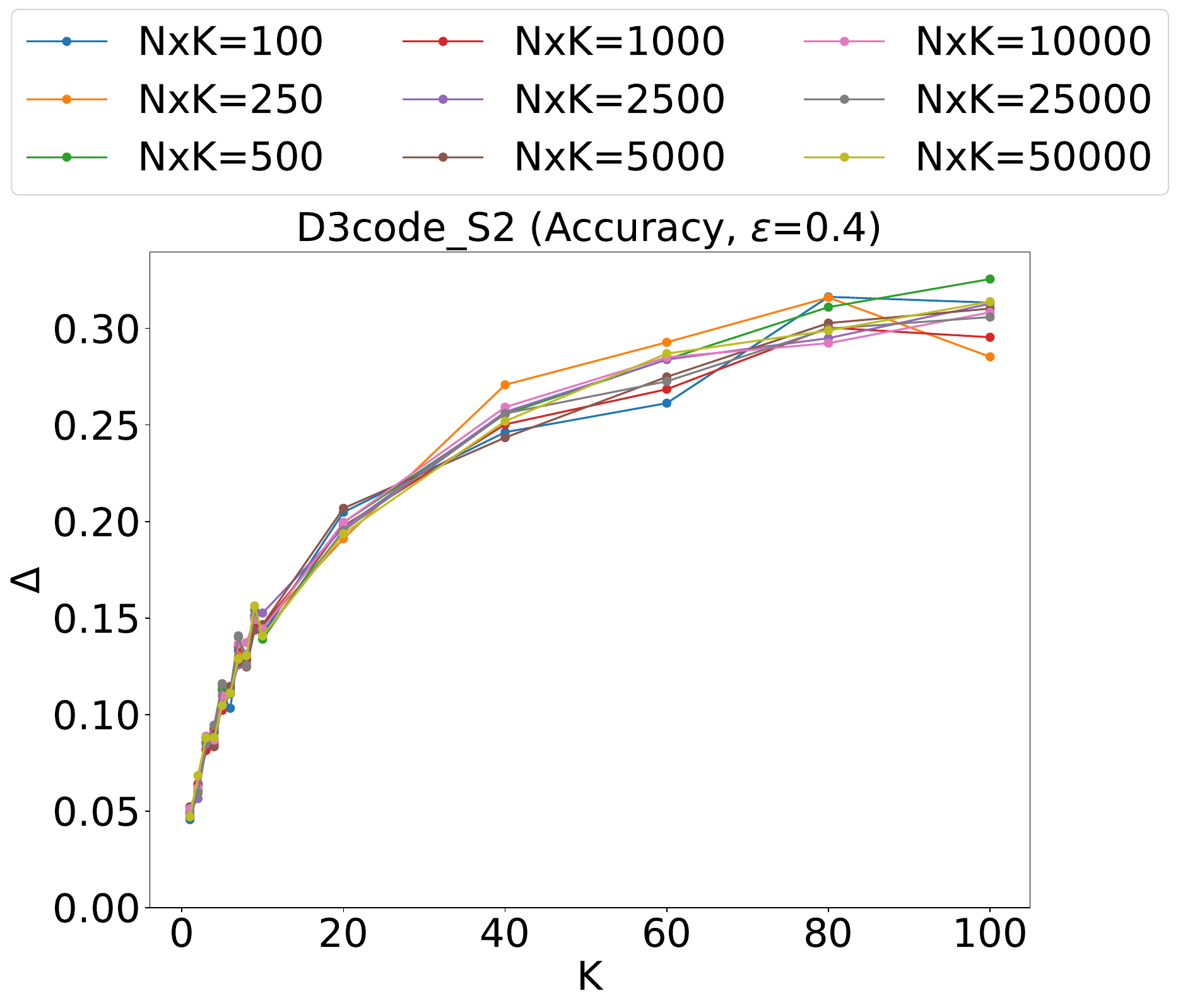}
    \caption{$\epsilon = 0.4$}
    \label{fig:d3code_s2_delta_acc_e04}
  \end{subfigure}
  \caption{S2: Effect sizes ($\Delta$) for D3code dataset with Accuracy as the metric}
  \label{fig:d3code_s2_delta_accuracy}
\end{figure*}

\begin{figure*}
  \centering
  \begin{subfigure}[b]{0.24\linewidth}
    \centering
    \includegraphics[width=\linewidth]{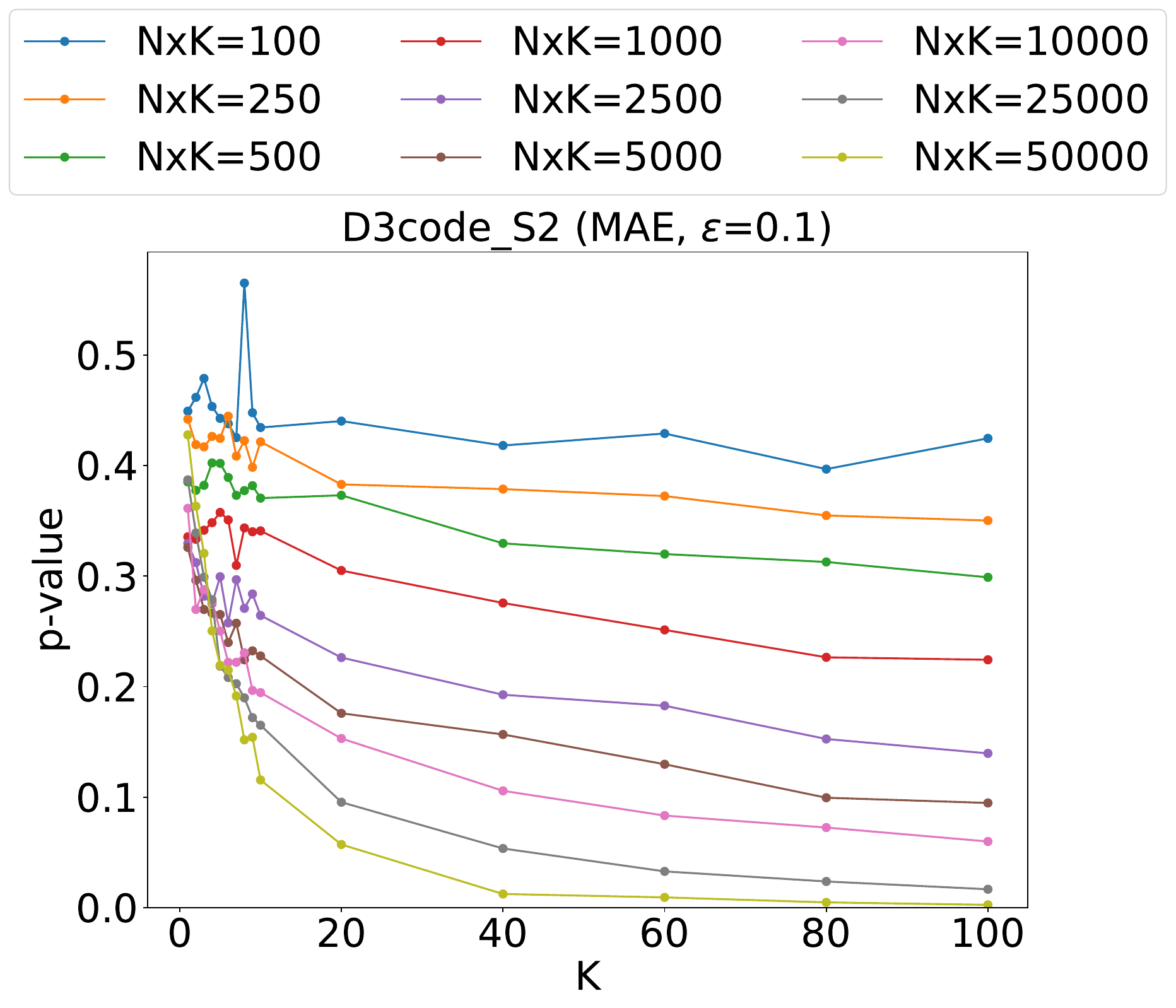}
    \caption{$\epsilon = 0.1$}
    \label{fig:d3code_s2_MAE_e01}
  \end{subfigure} \hfill
  \begin{subfigure}[b]{0.24\linewidth}
    \centering
    \includegraphics[width=\linewidth]{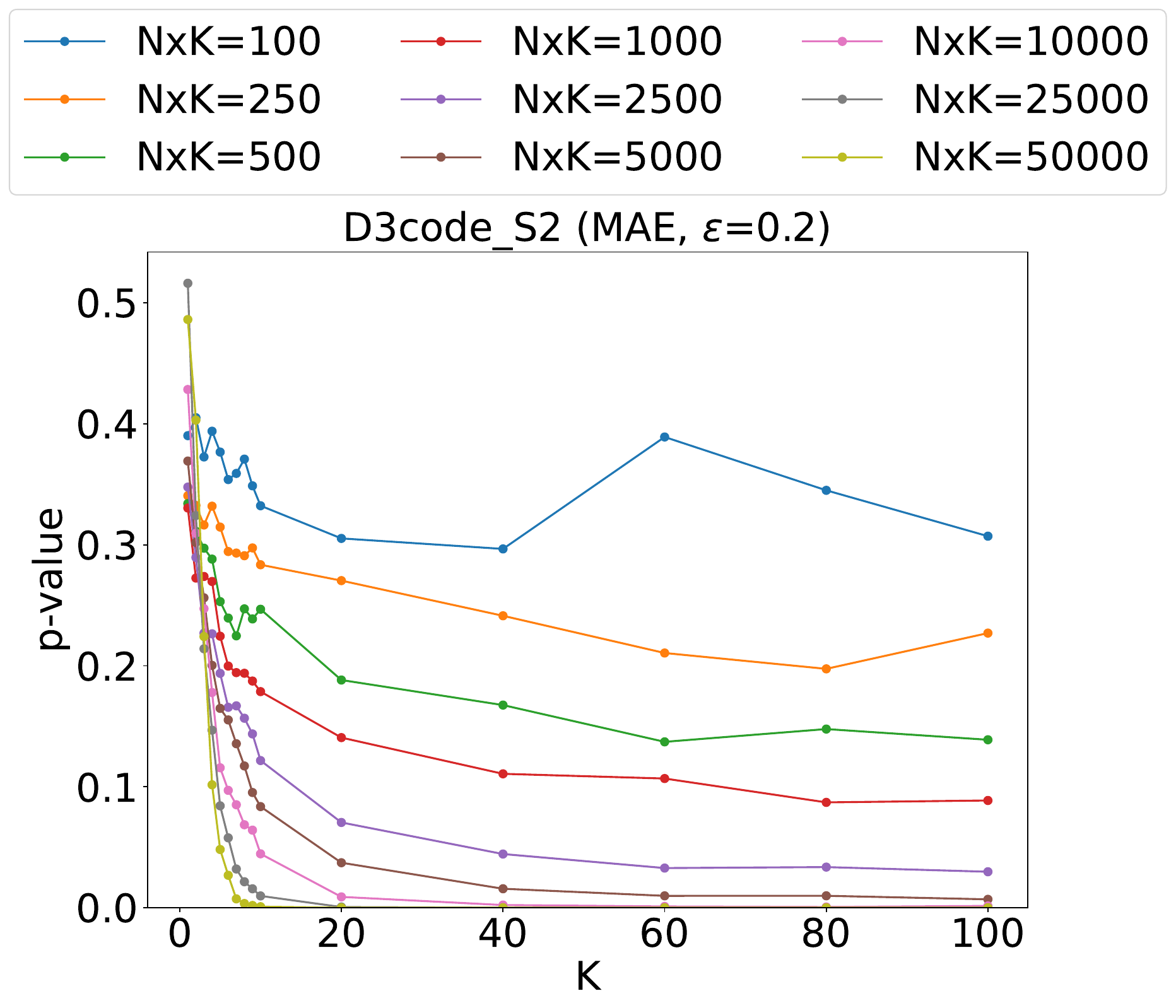}
    \caption{$\epsilon = 0.2$}
    \label{fig:d3code_s2_MAE_e02}
  \end{subfigure} \hfill
  \begin{subfigure}[b]{0.24\linewidth}
    \centering
    \includegraphics[width=\linewidth]{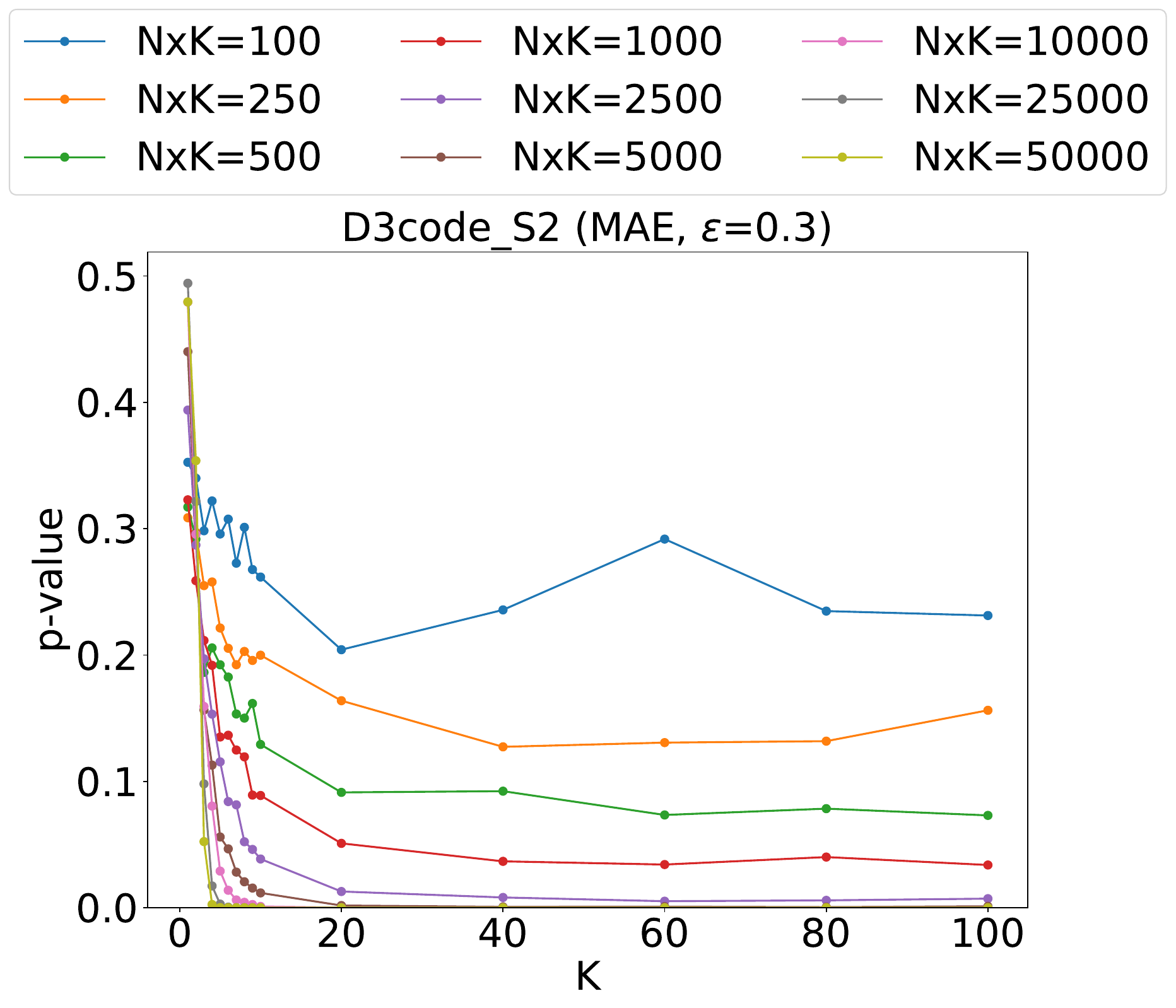}
    \caption{$\epsilon = 0.3$}
    \label{fig:d3code_s2_MAE_e03}
  \end{subfigure} \hfill
  \begin{subfigure}[b]{0.24\linewidth}
    \centering
    \includegraphics[width=\linewidth]{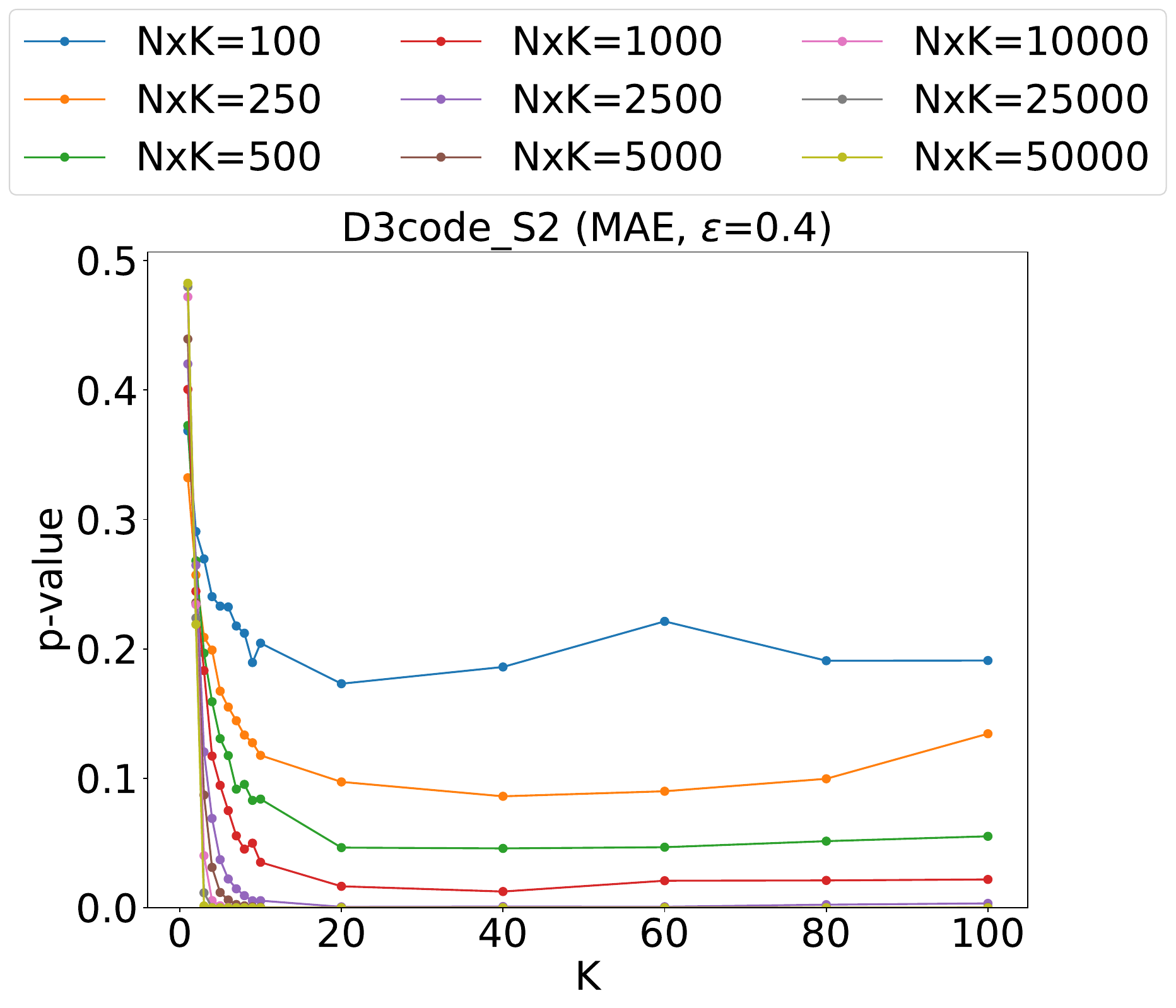}
    \caption{$\epsilon = 0.4$}
    \label{fig:d3code_s2_MAE_e04}
  \end{subfigure}
  \caption{S2: P-value plots for D3code dataset with MAE as the metric}
  \label{fig:d3code_s2_MAE}
\end{figure*}

\begin{figure*}
  \centering
  \begin{subfigure}[b]{0.24\linewidth}
    \centering
    \includegraphics[width=\linewidth]{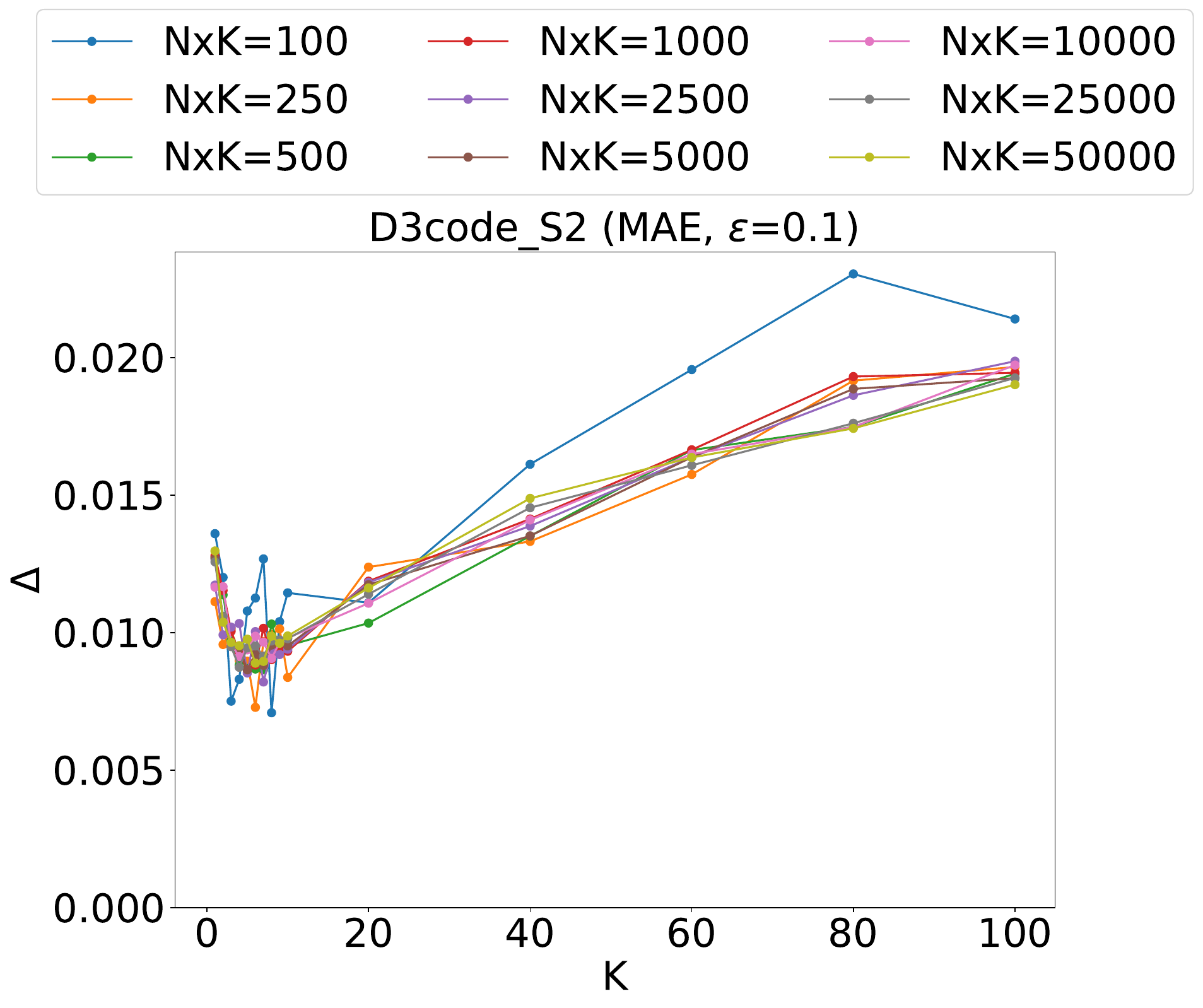}
    \caption{$\epsilon = 0.1$}
    \label{fig:d3code_s2_delta_MAE_e01}
  \end{subfigure} \hfill
  \begin{subfigure}[b]{0.24\linewidth}
    \centering
    \includegraphics[width=\linewidth]{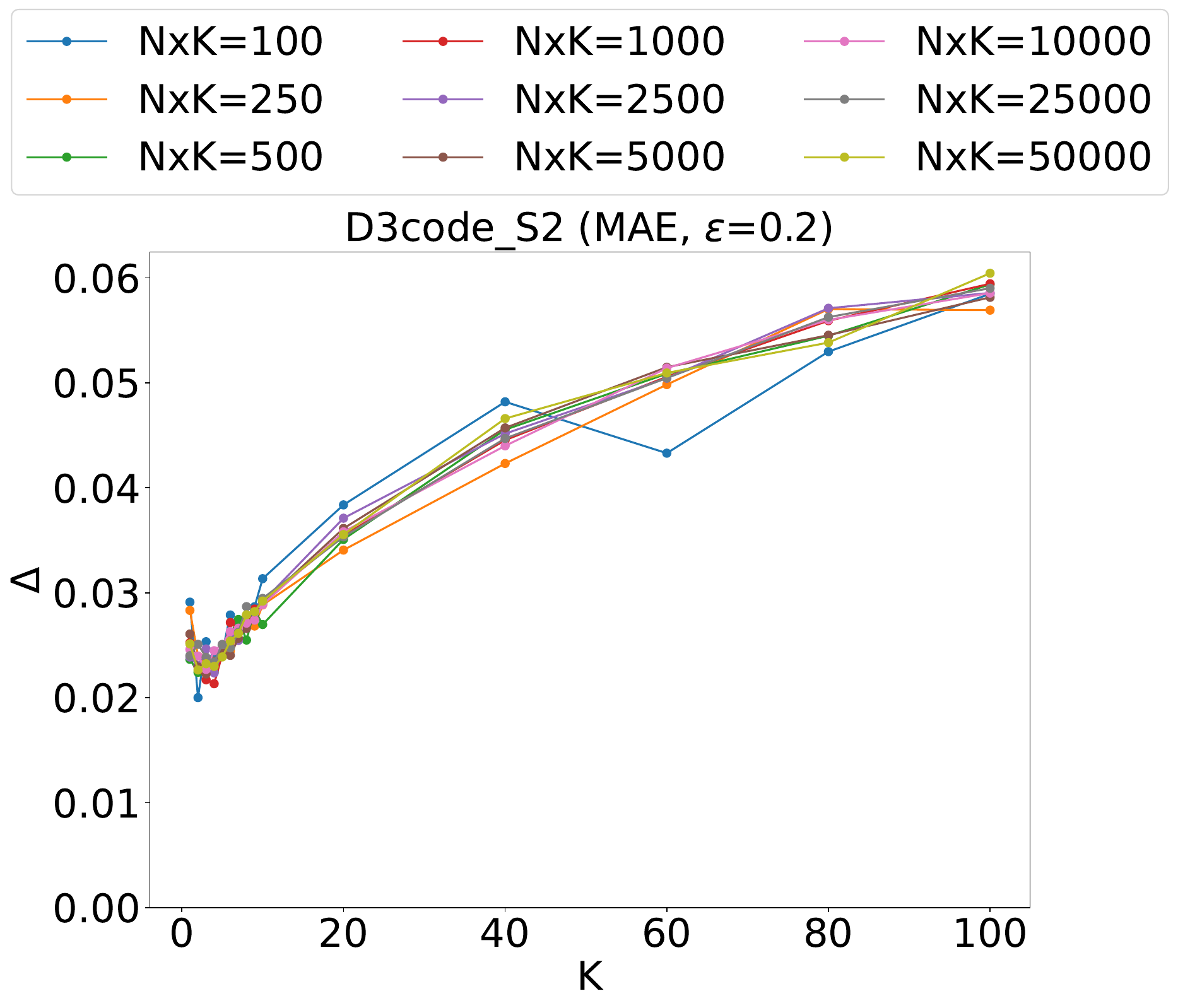}
    \caption{$\epsilon = 0.2$}
    \label{fig:d3code_s2_delta_MAE_e02}
  \end{subfigure} \hfill
  \begin{subfigure}[b]{0.24\linewidth}
    \centering
    \includegraphics[width=\linewidth]{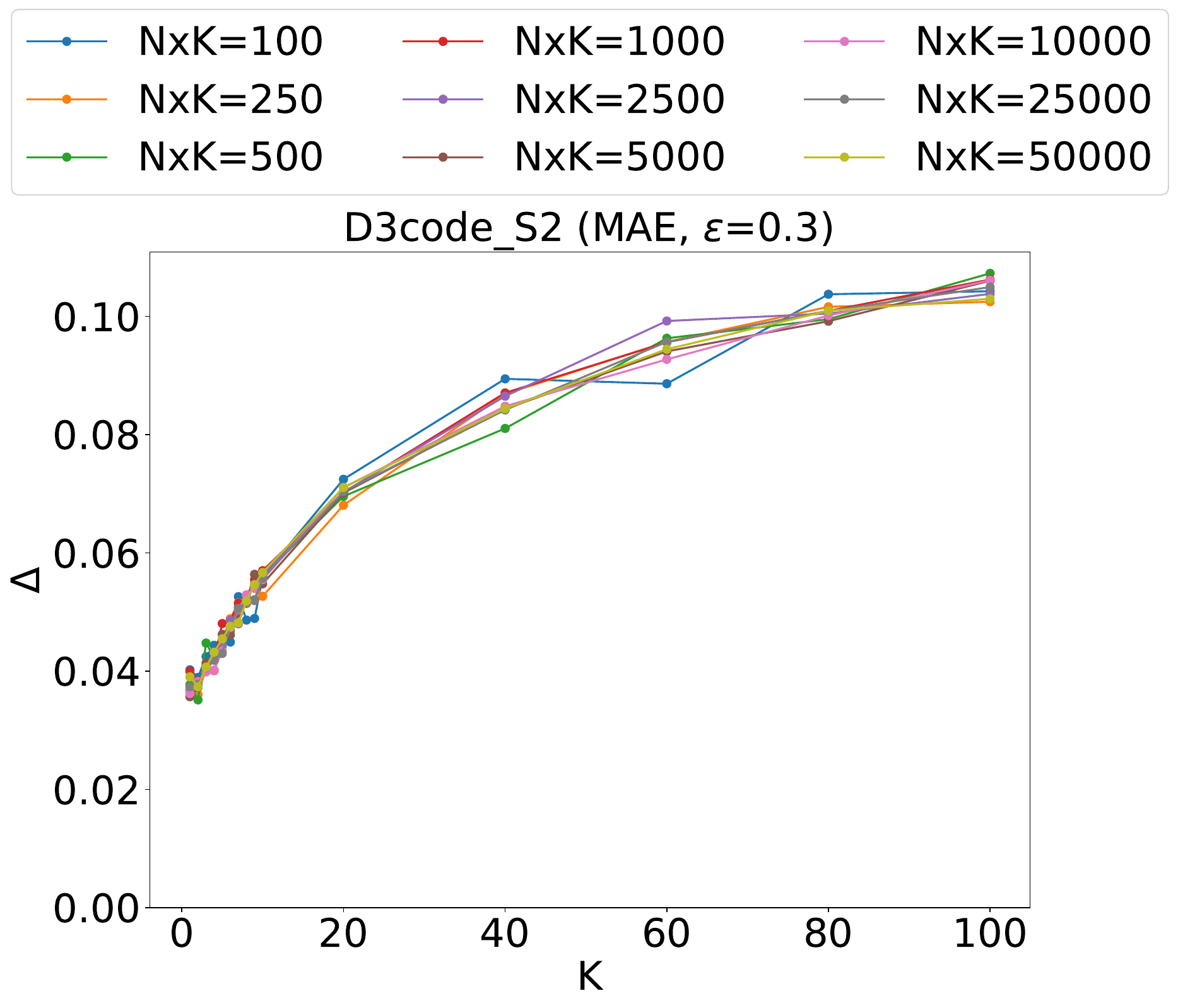}
    \caption{$\epsilon = 0.3$}
    \label{fig:d3code_s2_delta_MAE_e03}
  \end{subfigure} \hfill
  \begin{subfigure}[b]{0.24\linewidth}
    \centering
    \includegraphics[width=\linewidth]{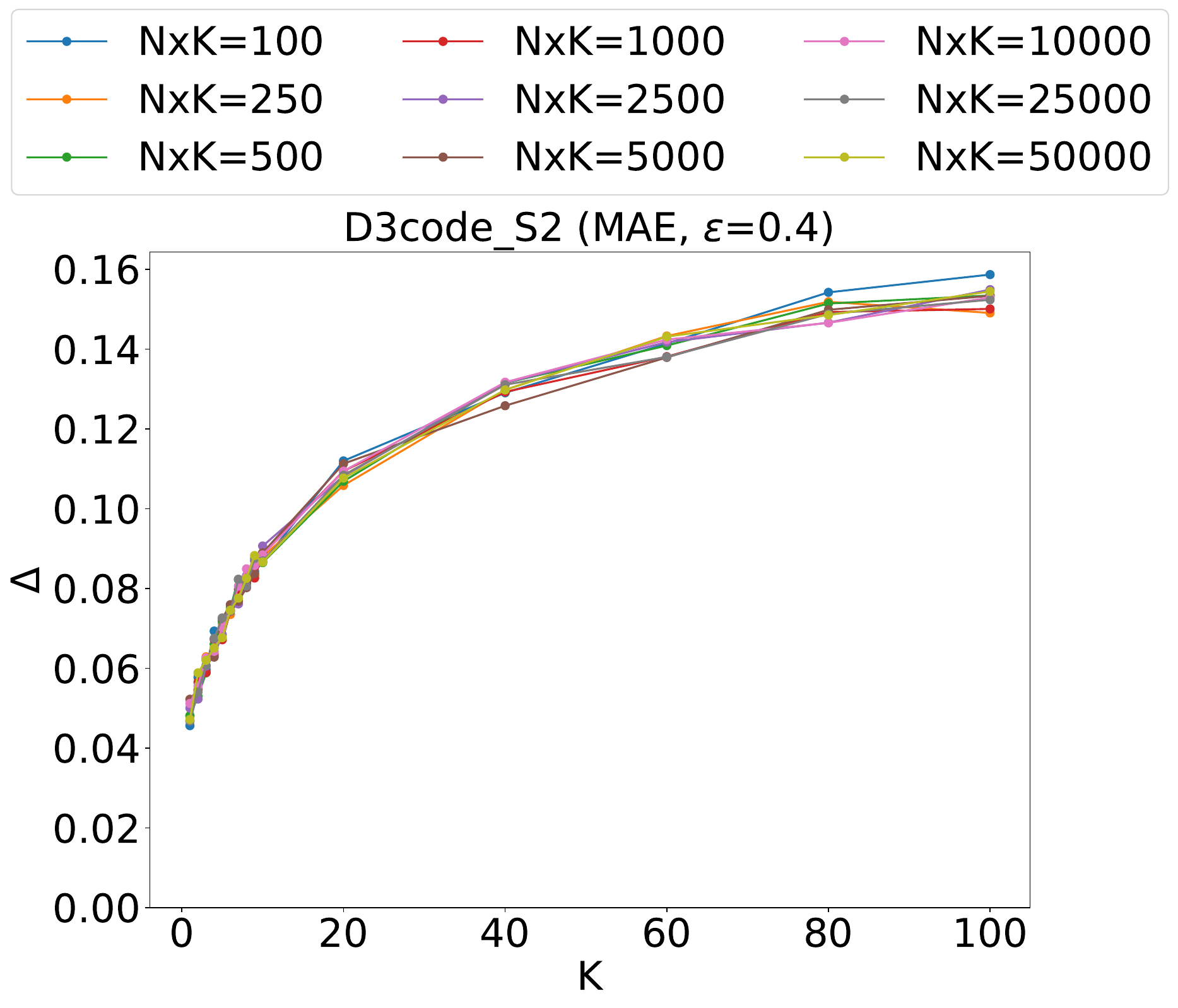}
    \caption{$\epsilon = 0.4$}
    \label{fig:d3code_s2_delta_MAE_e04}
  \end{subfigure}
  \caption{S2: Effect sizes ($\Delta$) for D3code dataset with MAE as the metric}
  \label{fig:d3code_s2_delta_MAE}
\end{figure*}

\begin{figure*}
  \centering
  \begin{subfigure}[b]{0.24\linewidth}
    \centering
    \includegraphics[width=\linewidth]{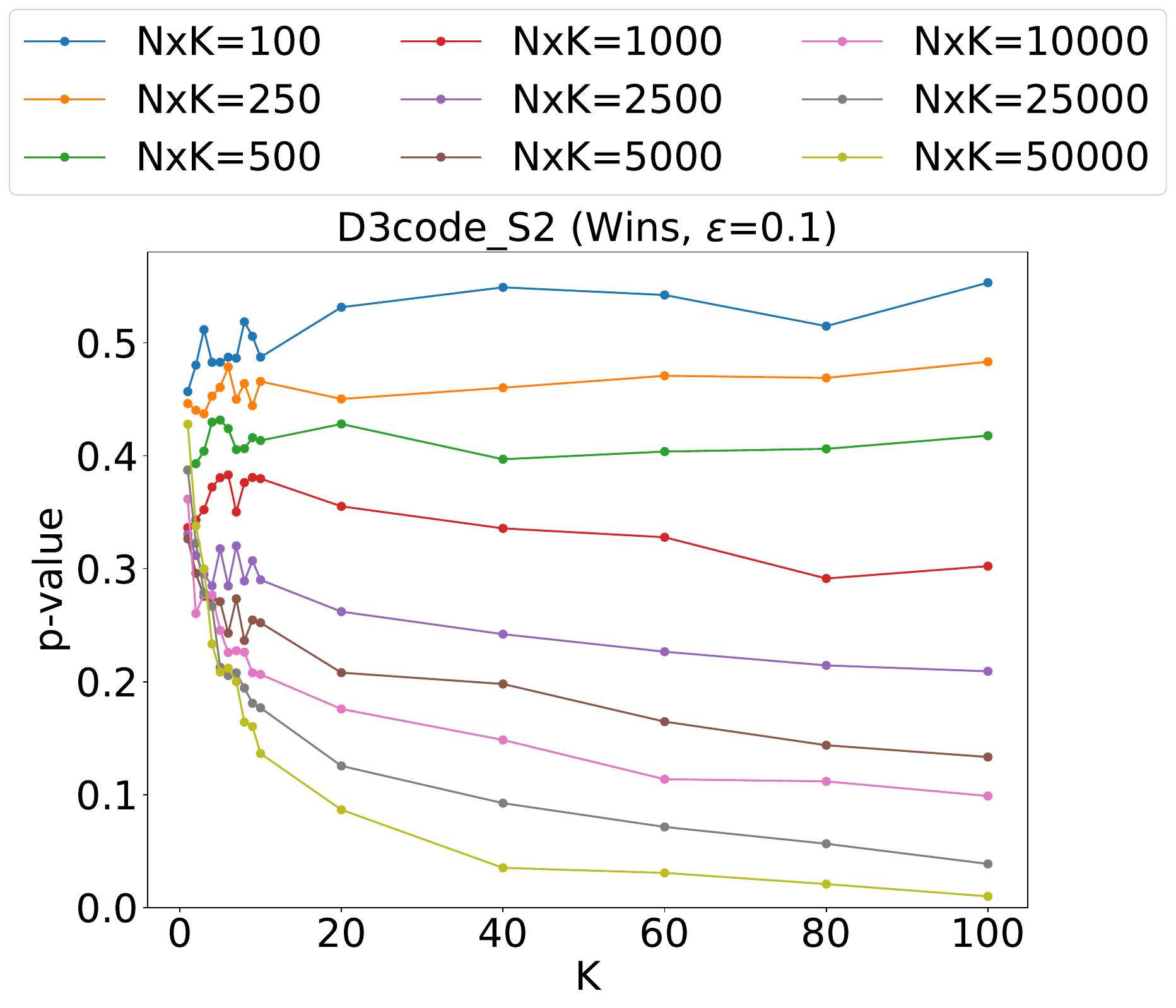}
    \caption{$\epsilon = 0.1$}
    \label{fig:d3code_s2_wins_e01}
  \end{subfigure} \hfill
  \begin{subfigure}[b]{0.24\linewidth}
    \centering
    \includegraphics[width=\linewidth]{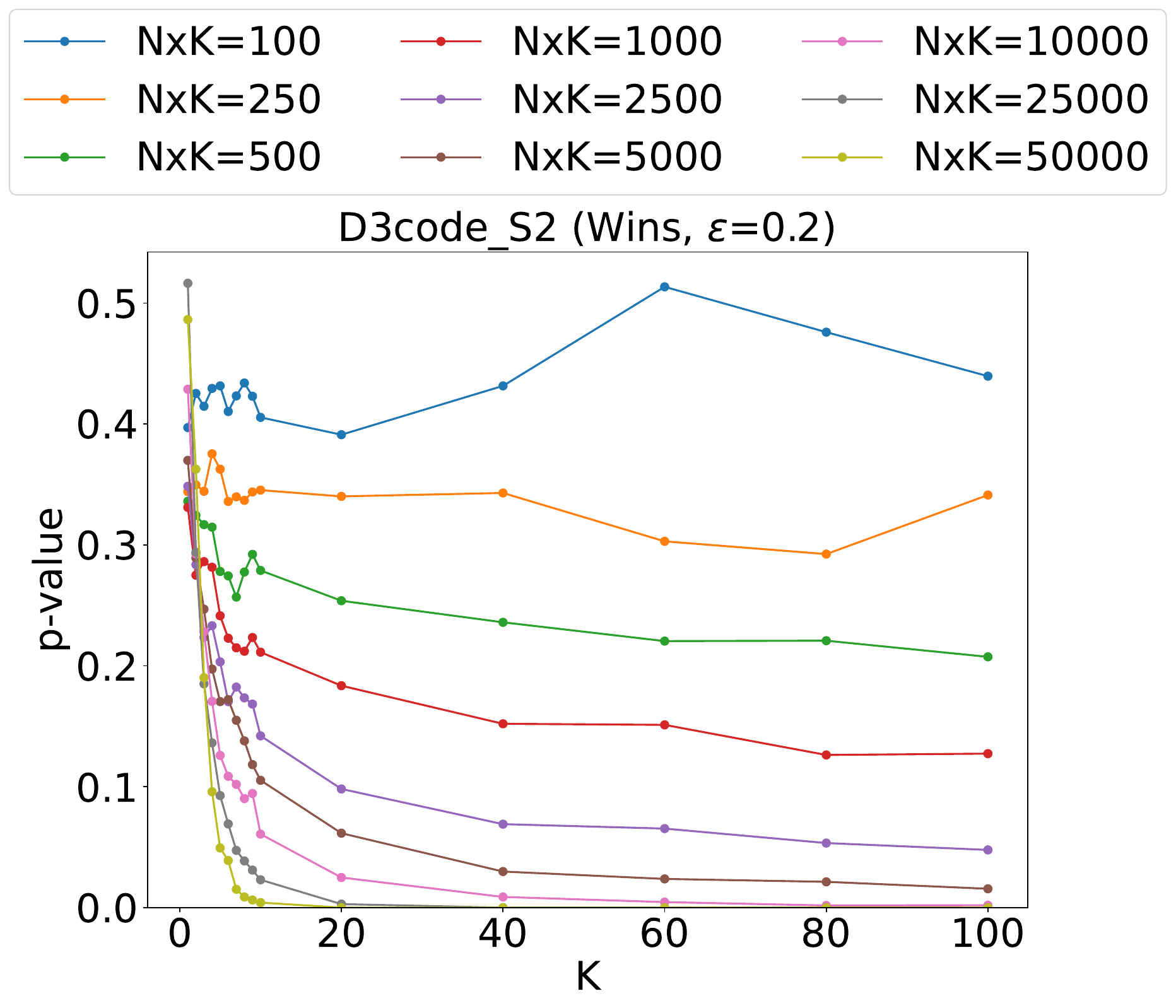}
    \caption{$\epsilon = 0.2$}
    \label{fig:d3code_s2_wins_e02}
  \end{subfigure} \hfill
  \begin{subfigure}[b]{0.24\linewidth}
    \centering
    \includegraphics[width=\linewidth]{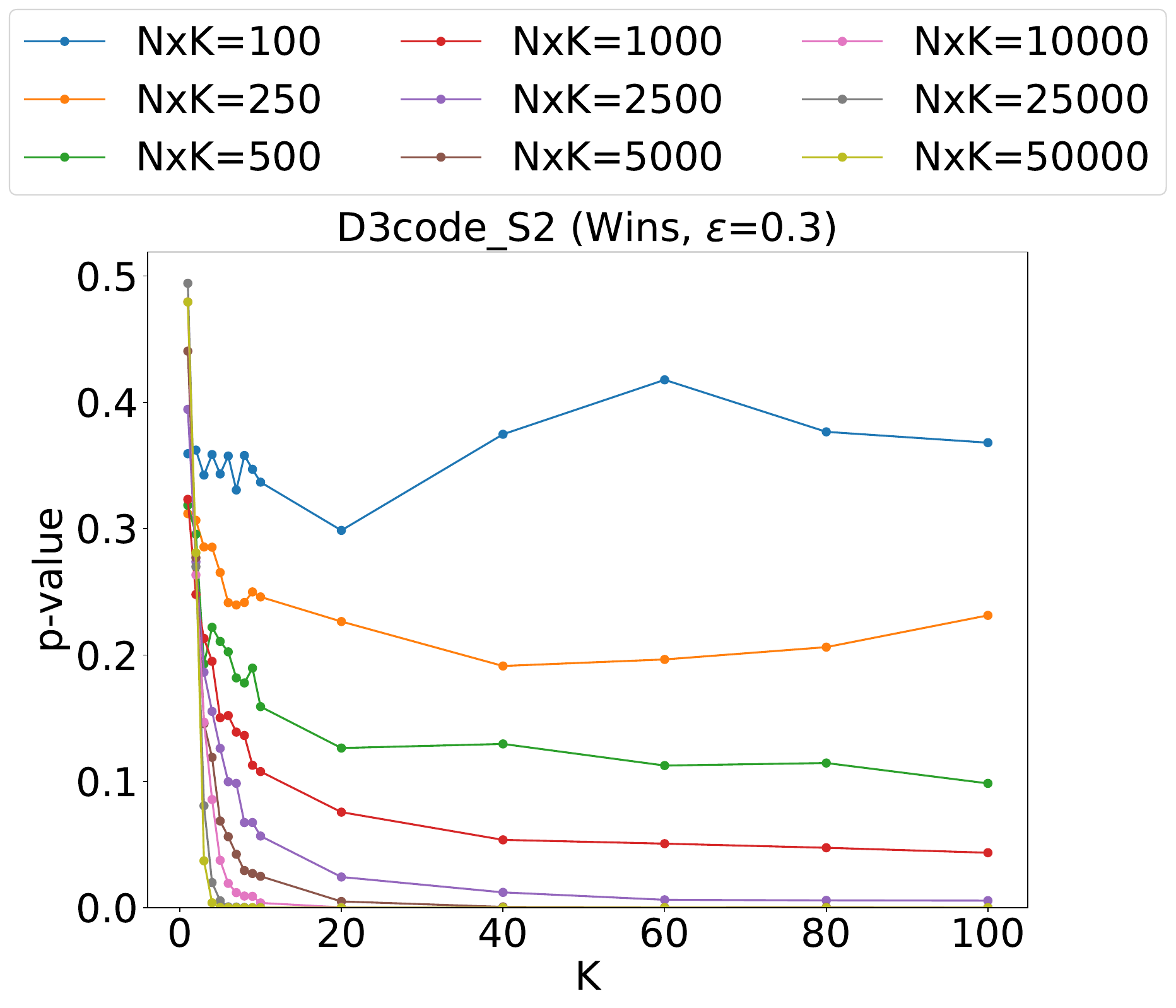}
    \caption{$\epsilon = 0.3$}
    \label{fig:d3code_s2_wins_e03}
  \end{subfigure} \hfill
  \begin{subfigure}[b]{0.24\linewidth}
    \centering
    \includegraphics[width=\linewidth]{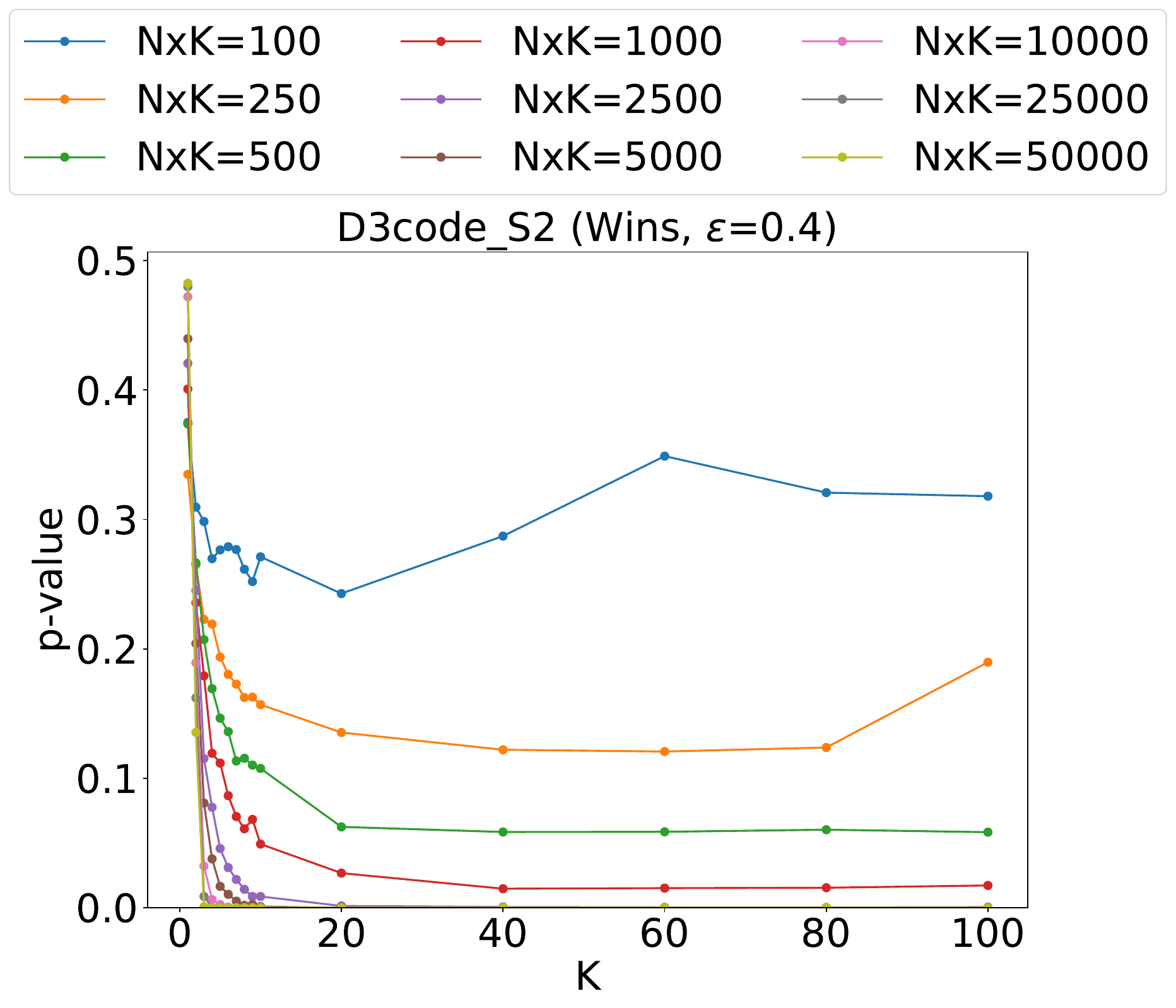}
    \caption{$\epsilon = 0.4$}
    \label{fig:d3code_s2_wins_e04}
  \end{subfigure}
  \caption{S2: P-value plots for D3code dataset with Wins as the metric}
  \label{fig:d3code_s2_wins}
\end{figure*}

\begin{figure*}
  \centering
  \begin{subfigure}[b]{0.24\linewidth}
    \centering
    \includegraphics[width=\linewidth]{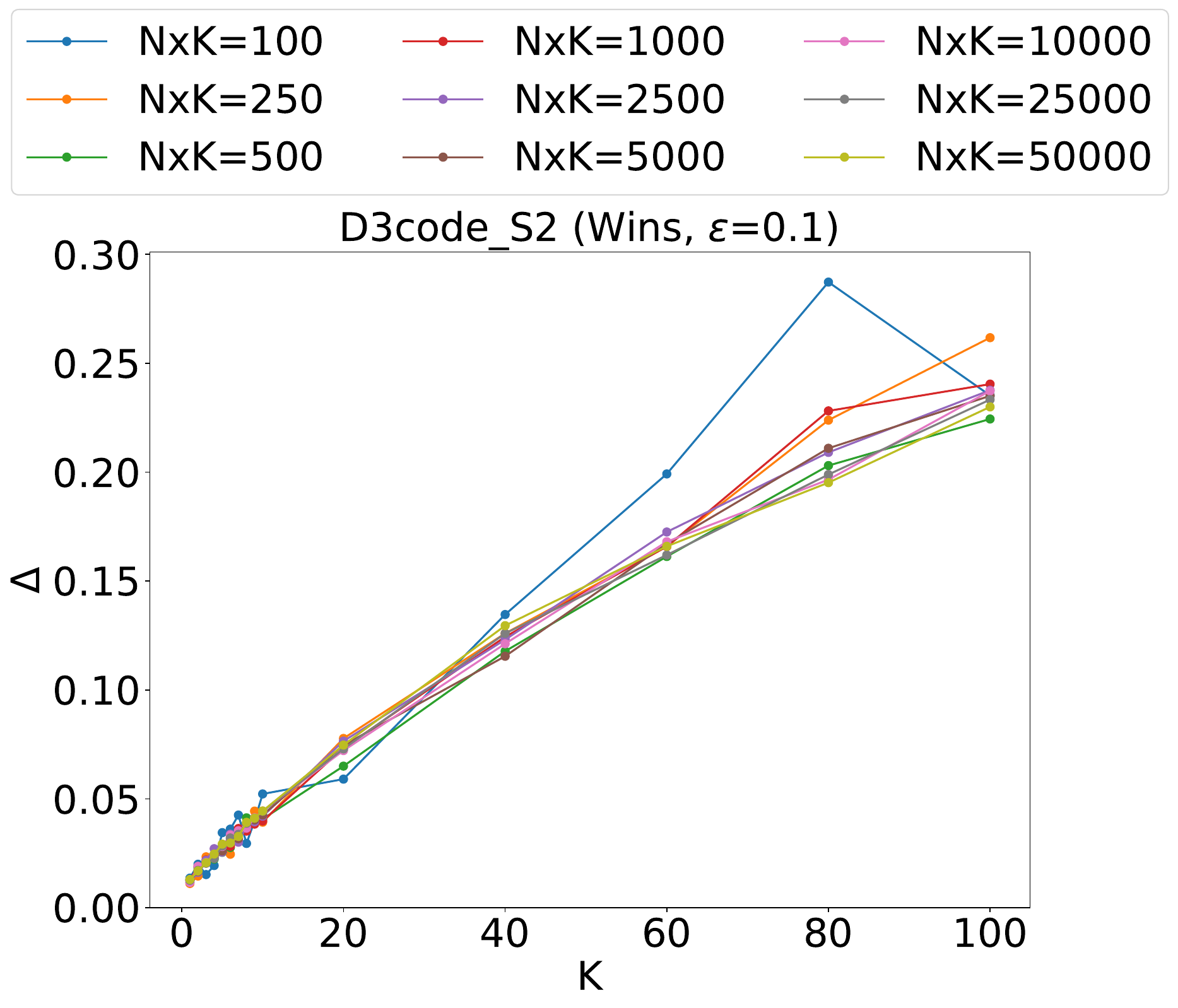}
    \caption{$\epsilon = 0.1$}
    \label{fig:d3code_s2_delta_wins_e01}
  \end{subfigure} \hfill
  \begin{subfigure}[b]{0.24\linewidth}
    \centering
    \includegraphics[width=\linewidth]{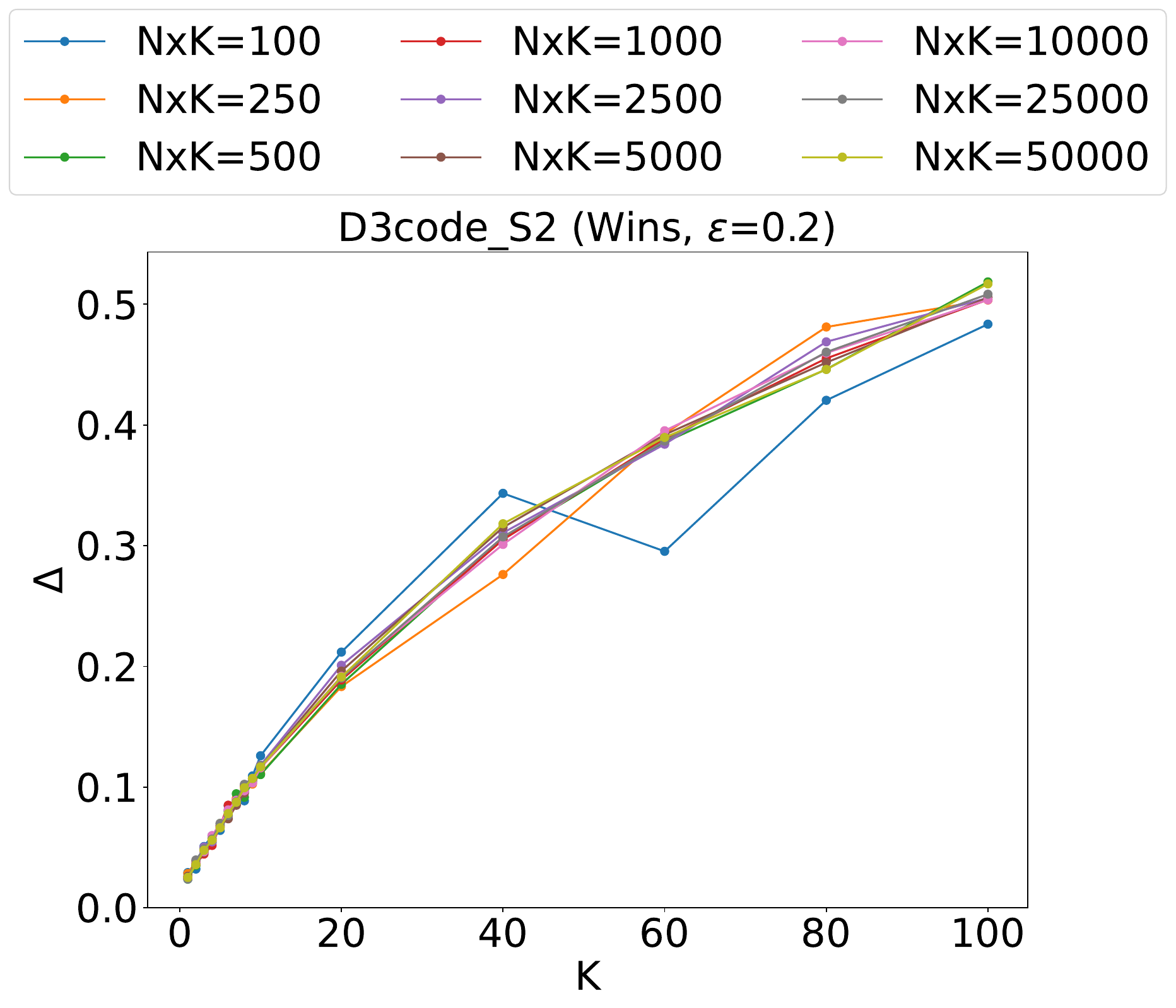}
    \caption{$\epsilon = 0.2$}
    \label{fig:d3code_s2_delta_wins_e02}
  \end{subfigure} \hfill
  \begin{subfigure}[b]{0.24\linewidth}
    \centering
    \includegraphics[width=\linewidth]{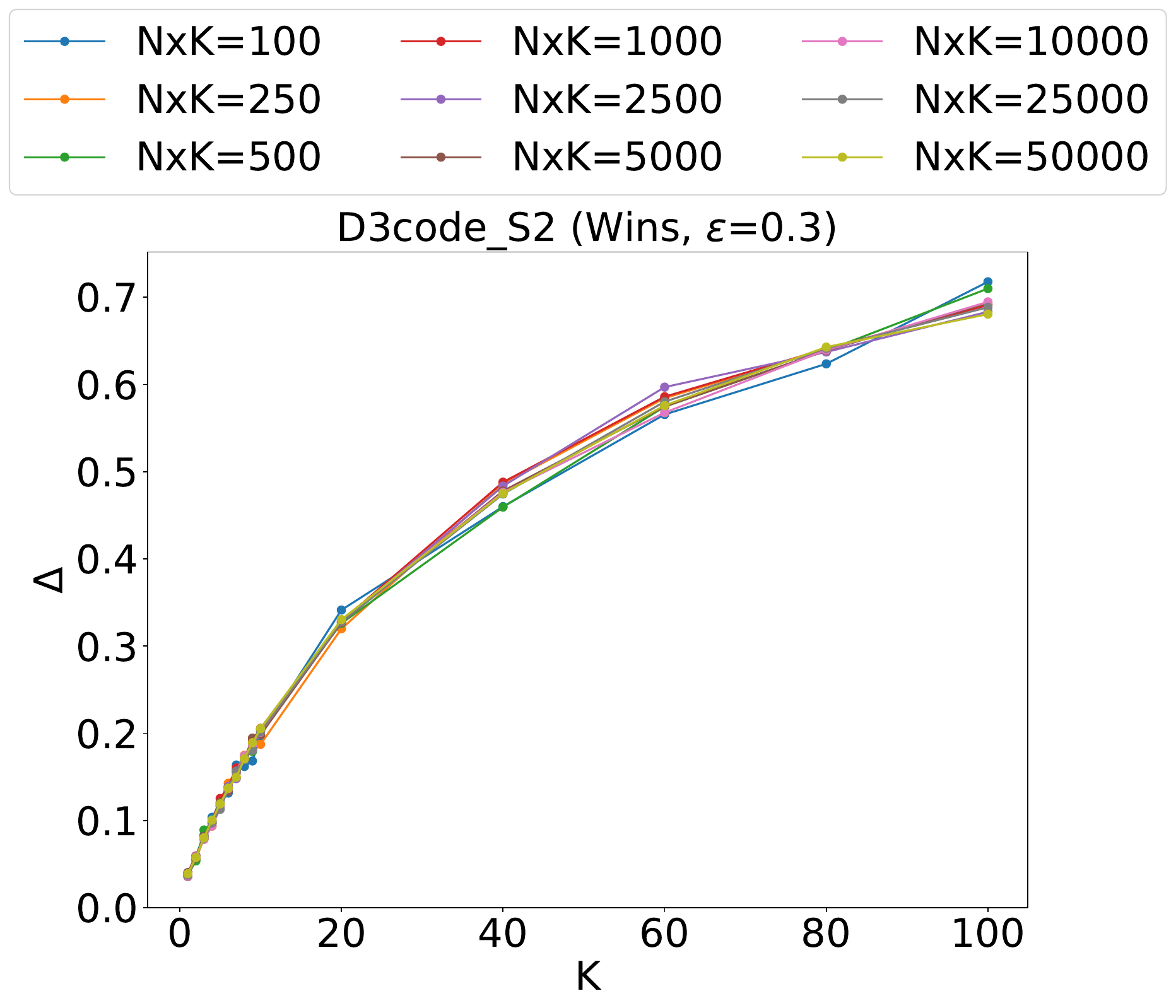}
    \caption{$\epsilon = 0.3$}
    \label{fig:d3code_s2_delta_wins_e03}
  \end{subfigure} \hfill
  \begin{subfigure}[b]{0.24\linewidth}
    \centering
    \includegraphics[width=\linewidth]{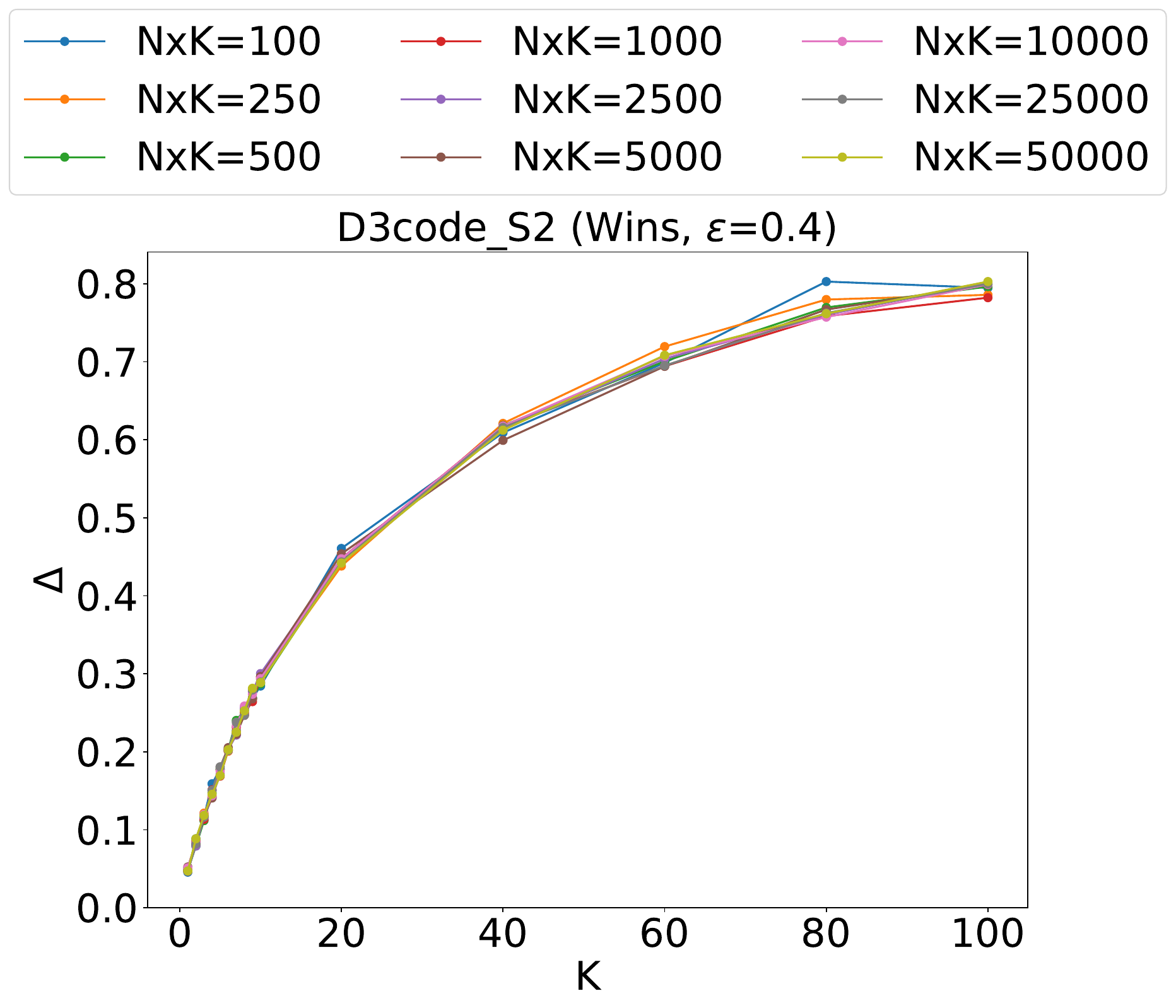}
    \caption{$\epsilon = 0.4$}
    \label{fig:d3code_s2_delta_wins_e04}
  \end{subfigure}
  \caption{S2: Effect sizes ($\Delta$) for D3code dataset with Wins as the metric}
  \label{fig:d3code_s2_delta_wins}
\end{figure*}

\subsection{S3}

\paragraph{Toxicity}

\begin{figure*}
  \centering
  \begin{subfigure}[b]{0.24\linewidth}
    \centering
    \includegraphics[width=\linewidth]{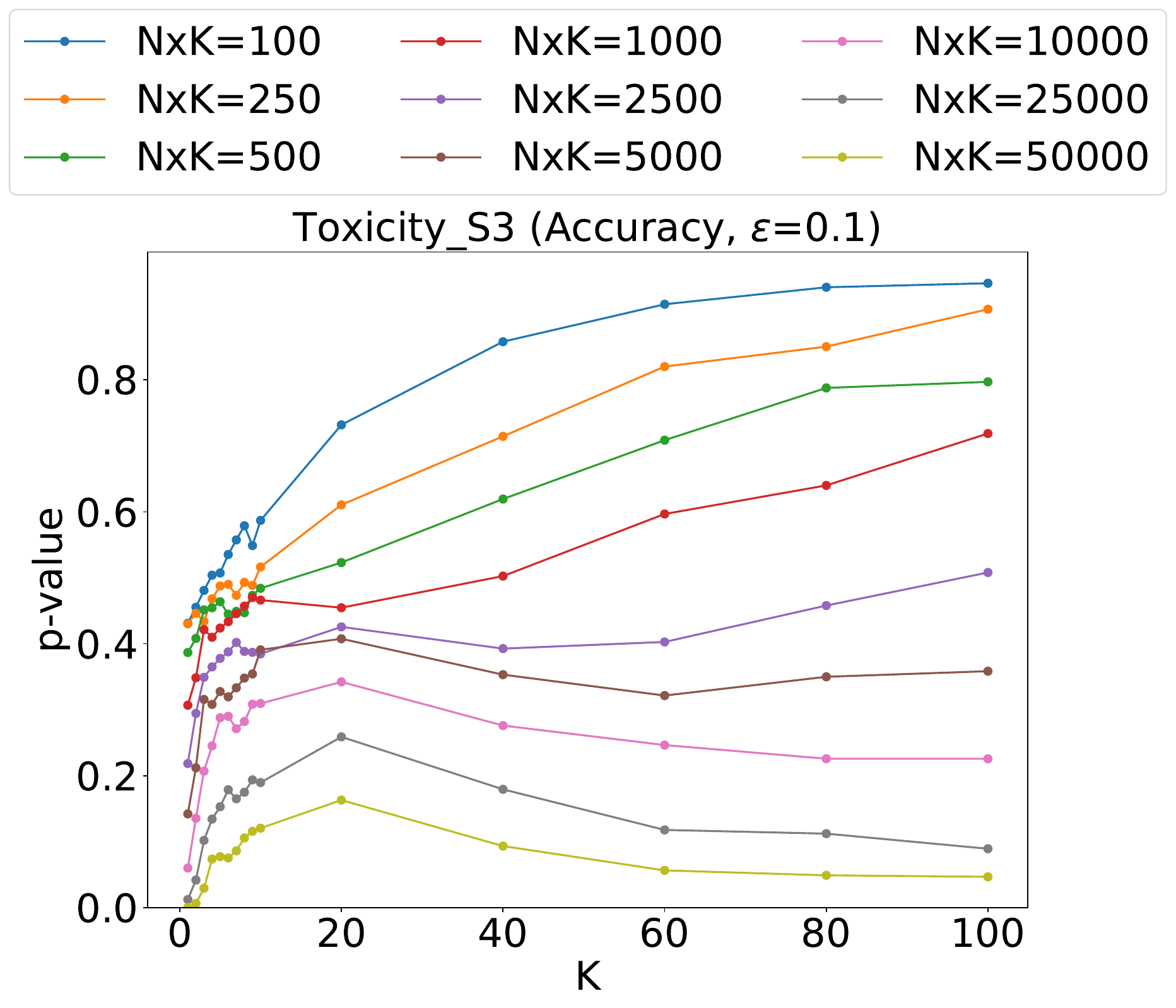}
    \caption{$\epsilon = 0.1$}
    \label{fig:toxicity_s3_acc_e01}
  \end{subfigure} \hfill
  \begin{subfigure}[b]{0.24\linewidth}
    \centering
    \includegraphics[width=\linewidth]{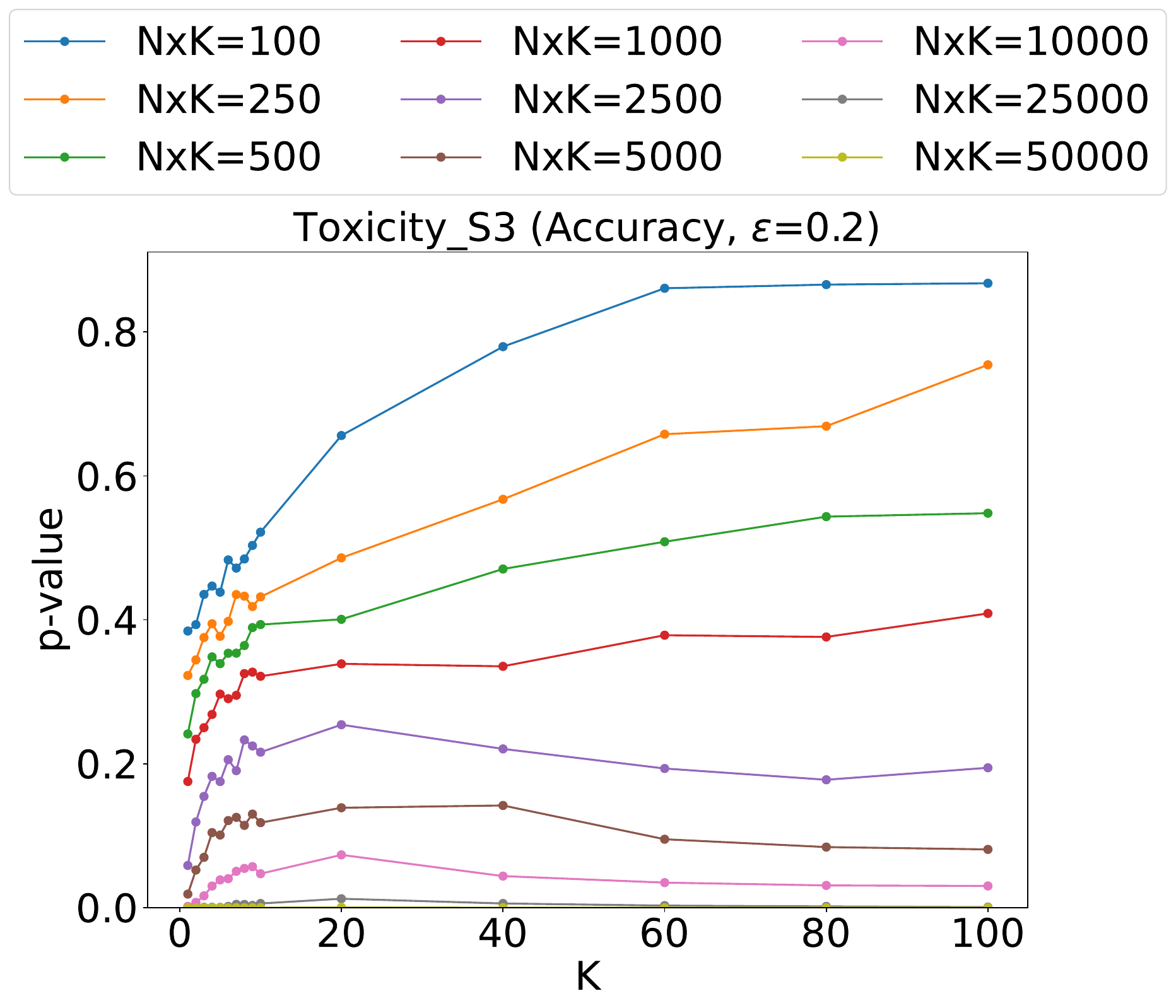}
    \caption{$\epsilon = 0.2$}
    \label{fig:toxicity_s3_acc_e02}
  \end{subfigure} \hfill
  \begin{subfigure}[b]{0.24\linewidth}
    \centering
    \includegraphics[width=\linewidth]{figures/pvals_plots/Toxicity_S3/Toxicity_S3_p_vals_Accuracy_K_100_e_0.3.pdf}
    \caption{$\epsilon = 0.3$}
    \label{fig:toxicity_s3_acc_e03}
  \end{subfigure} \hfill
  \begin{subfigure}[b]{0.24\linewidth}
    \centering
    \includegraphics[width=\linewidth]{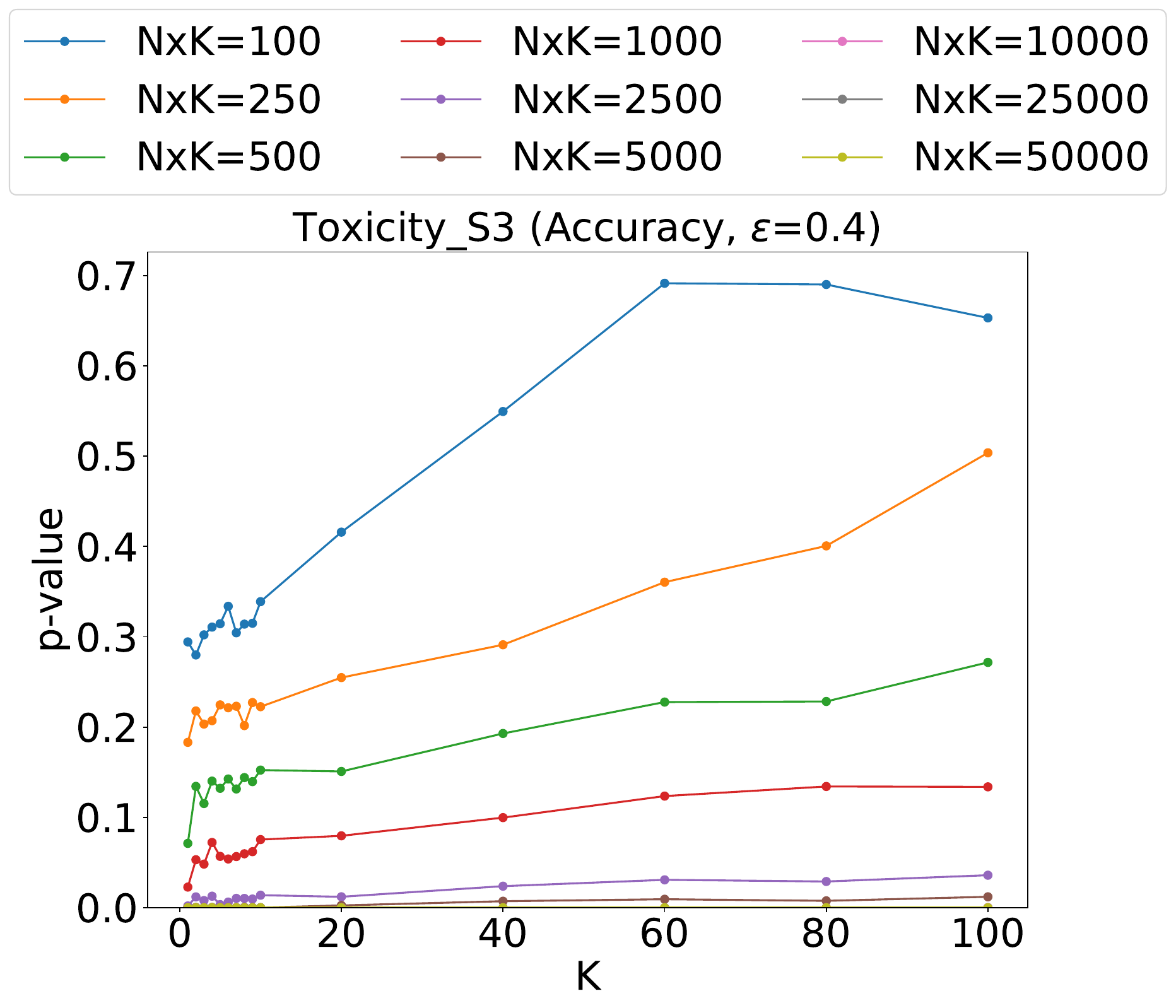}
    \caption{$\epsilon = 0.4$}
    \label{fig:toxicity_s3_acc_e04}
  \end{subfigure}
  \caption{S3: P-value plots for Toxicity dataset with Accuracy as the metric}
  \label{fig:toxicity_s3_accuracy}
\end{figure*}

\begin{figure*}
  \centering
  \begin{subfigure}[b]{0.24\linewidth}
    \centering
    \includegraphics[width=\linewidth]{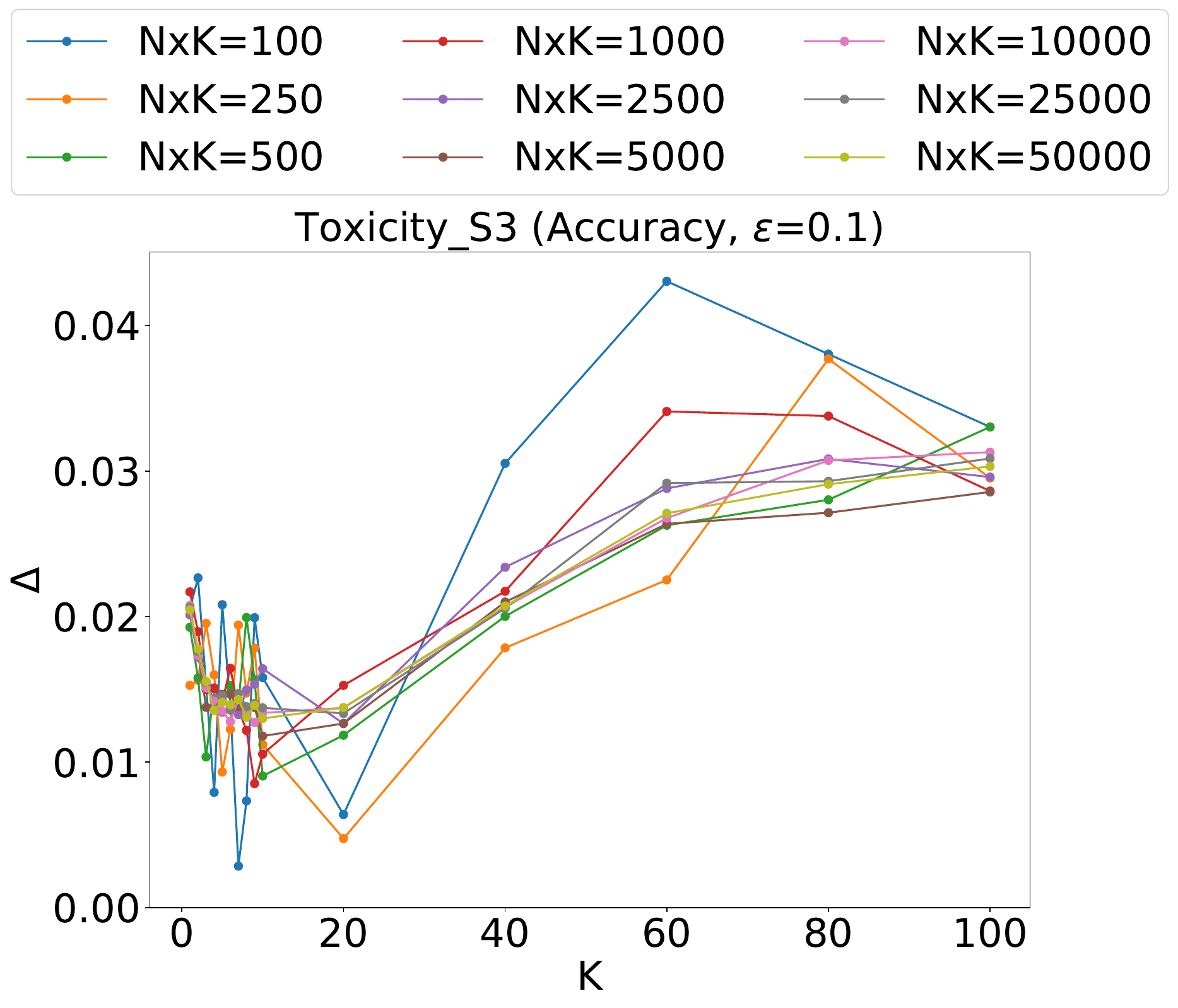}
    \caption{$\epsilon = 0.1$}
    \label{fig:toxicity_s3_delta_acc_e01}
  \end{subfigure} \hfill
  \begin{subfigure}[b]{0.24\linewidth}
    \centering
    \includegraphics[width=\linewidth]{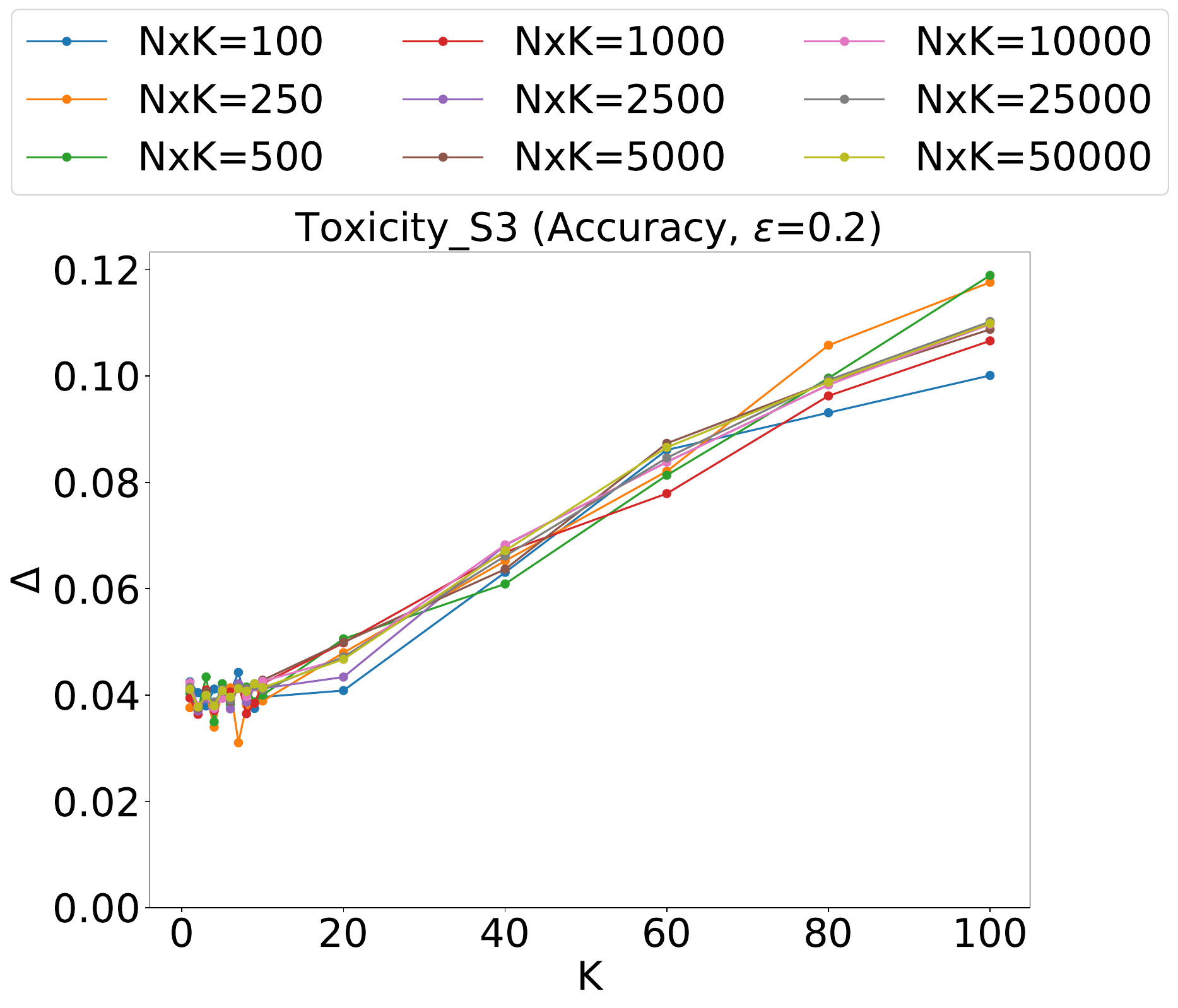}
    \caption{$\epsilon = 0.2$}
    \label{fig:toxicity_s3_delta_acc_e02}
  \end{subfigure} \hfill
  \begin{subfigure}[b]{0.24\linewidth}
    \centering
    \includegraphics[width=\linewidth]{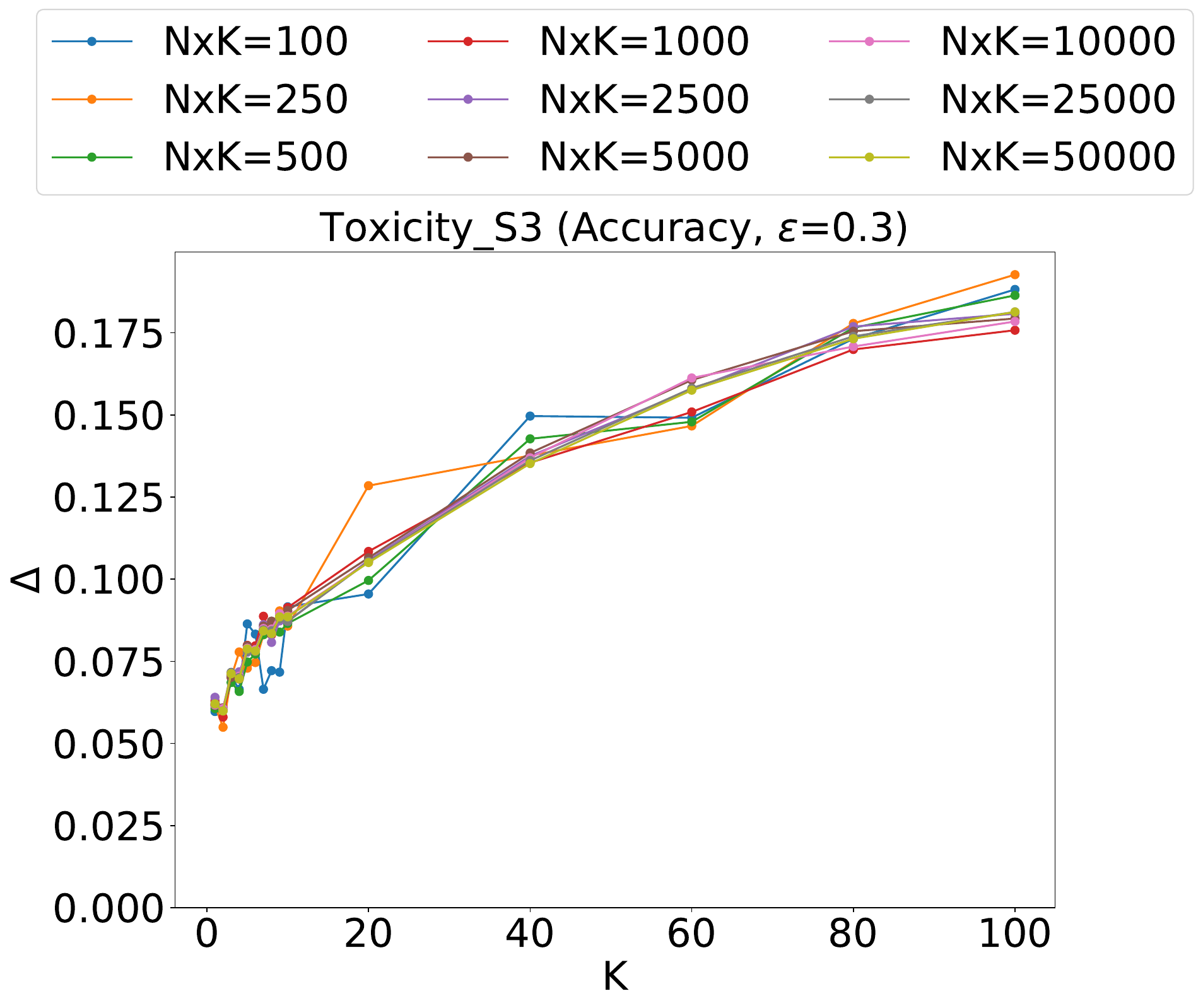}
    \caption{$\epsilon = 0.3$}
    \label{fig:toxicity_s3_delta_acc_e03}
  \end{subfigure} \hfill
  \begin{subfigure}[b]{0.24\linewidth}
    \centering
    \includegraphics[width=\linewidth]{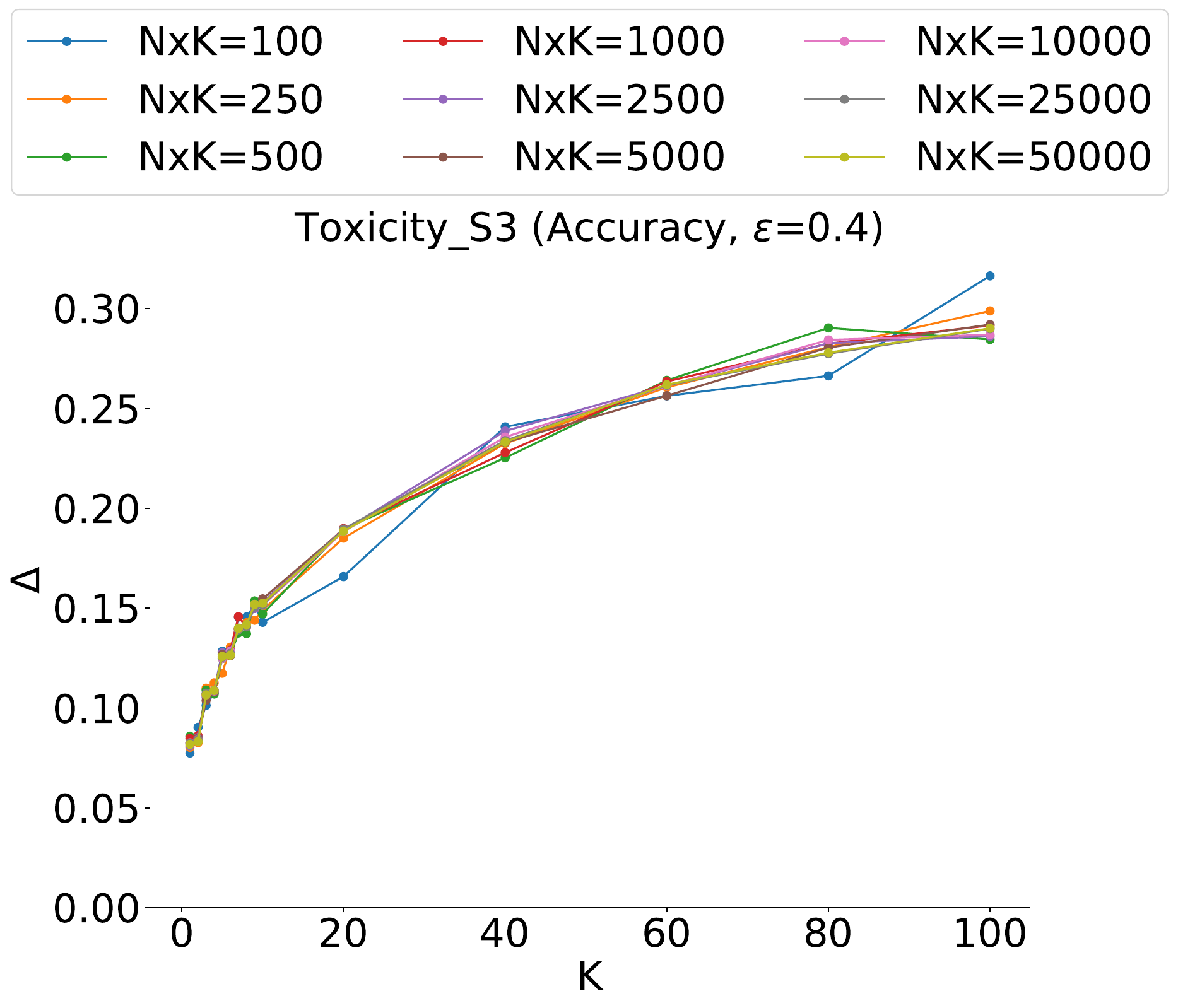}
    \caption{$\epsilon = 0.4$}
    \label{fig:toxicity_s3_delta_acc_e04}
  \end{subfigure}
  \caption{S3: Effect sizes ($\Delta$) for Toxicity dataset with Accuracy as the metric}
  \label{fig:toxicity_s3_delta_accuracy}
\end{figure*}

\begin{figure*}
  \centering
  \begin{subfigure}[b]{0.24\linewidth}
    \centering
    \includegraphics[width=\linewidth]{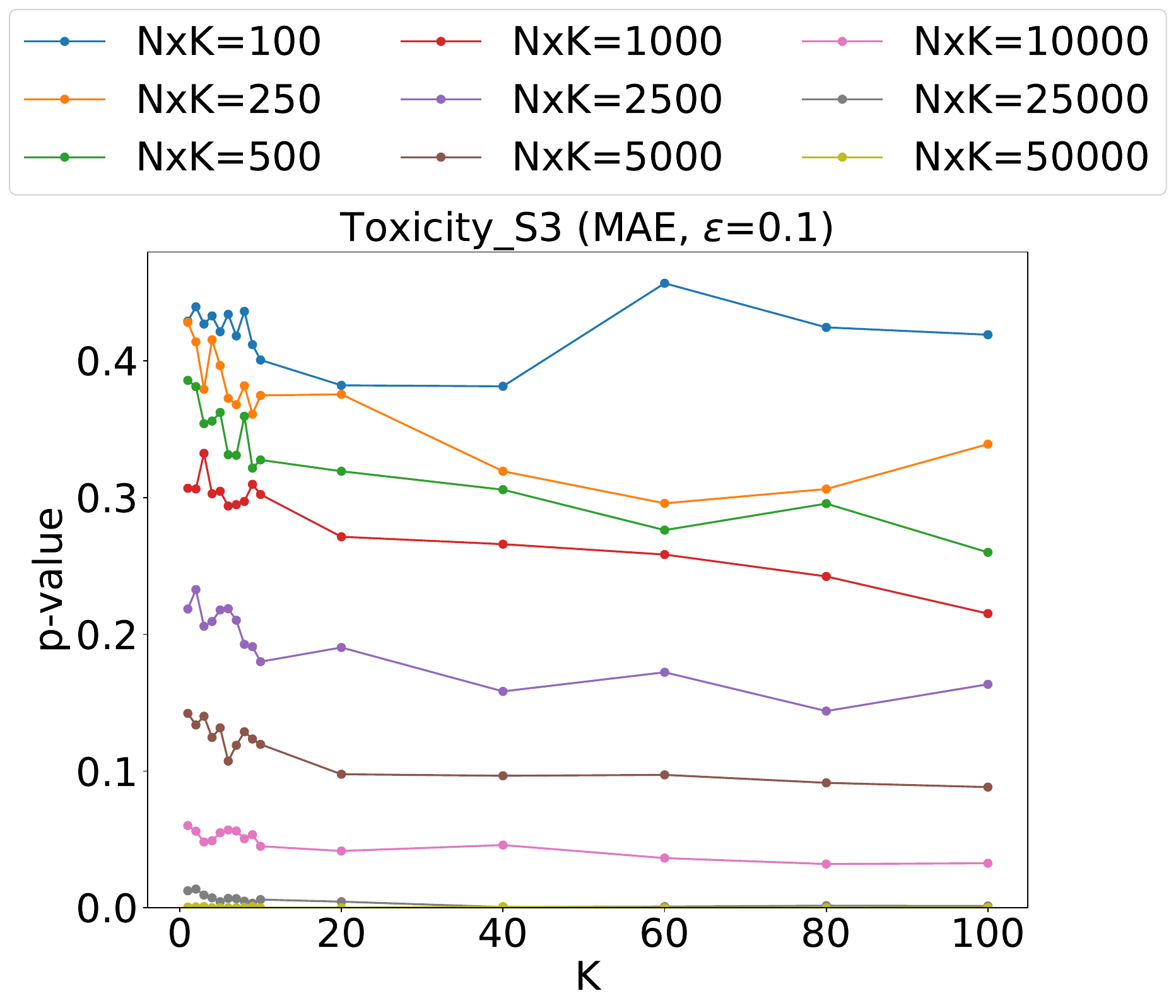}
    \caption{$\epsilon = 0.1$}
    \label{fig:toxicity_s3_MAE_e01}
  \end{subfigure} \hfill
  \begin{subfigure}[b]{0.24\linewidth}
    \centering
    \includegraphics[width=\linewidth]{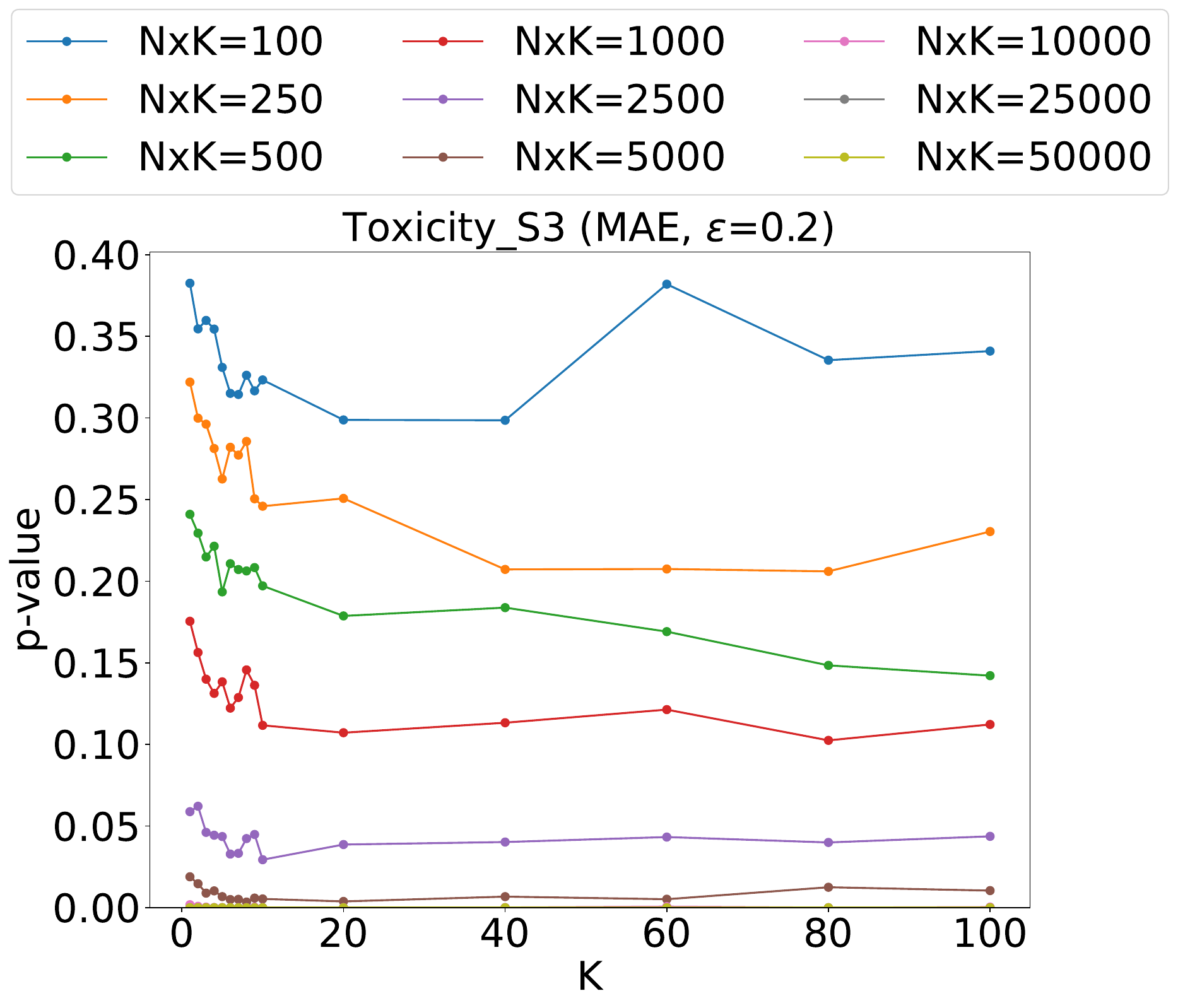}
    \caption{$\epsilon = 0.2$}
    \label{fig:toxicity_s3_MAE_e02}
  \end{subfigure} \hfill
  \begin{subfigure}[b]{0.24\linewidth}
    \centering
    \includegraphics[width=\linewidth]{figures/pvals_plots/Toxicity_S3/Toxicity_S3_p_vals_MAE_K_100_e_0.3.pdf}
    \caption{$\epsilon = 0.3$}
    \label{fig:toxicity_s3_MAE_e03}
  \end{subfigure} \hfill
  \begin{subfigure}[b]{0.24\linewidth}
    \centering
    \includegraphics[width=\linewidth]{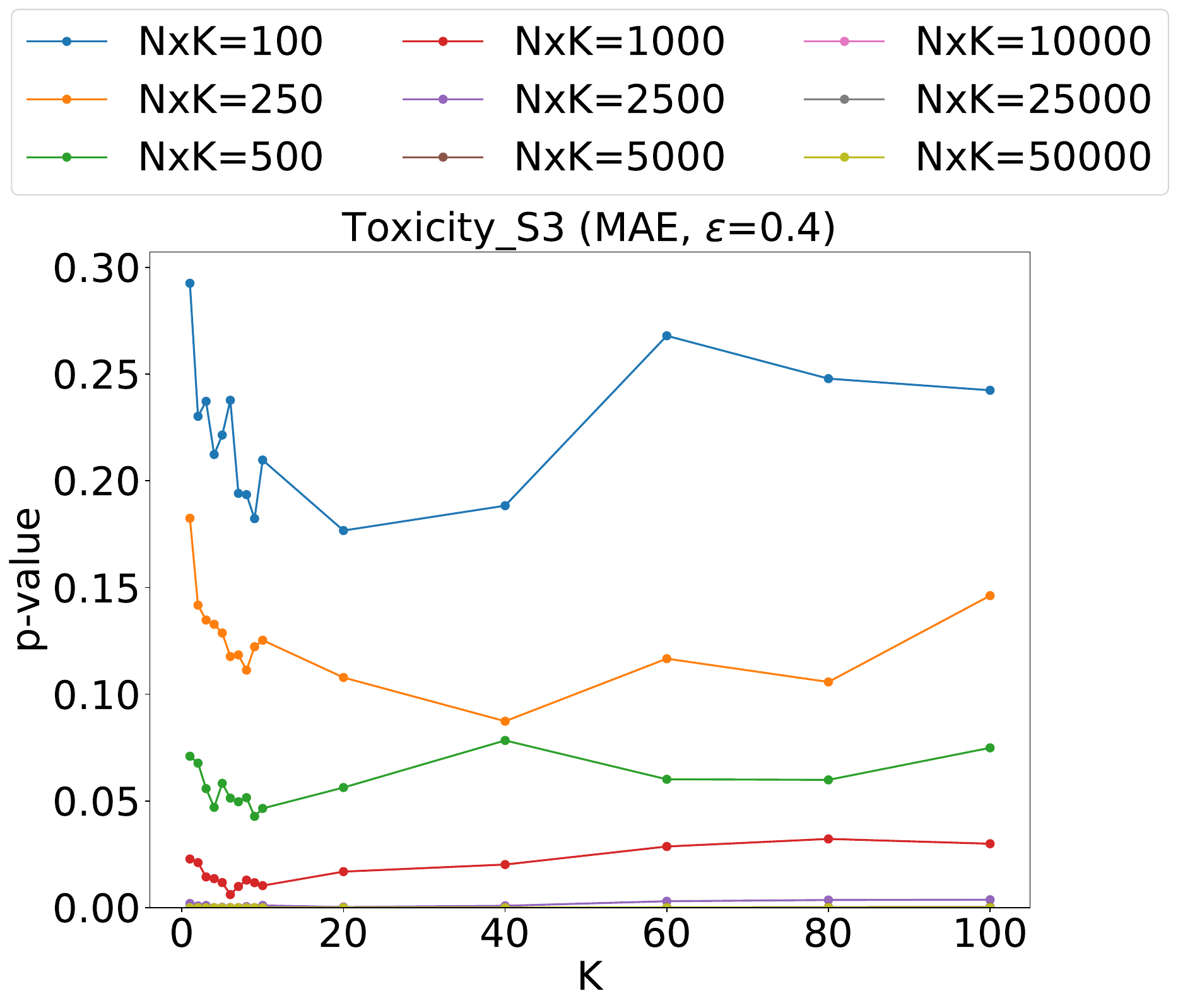}
    \caption{$\epsilon = 0.4$}
    \label{fig:toxicity_s3_MAE_e04}
  \end{subfigure}
  \caption{S3: P-value plots for Toxicity dataset with MAE as the metric}
  \label{fig:toxicity_s3_MAE}
\end{figure*}

\begin{figure*}
  \centering
  \begin{subfigure}[b]{0.24\linewidth}
    \centering
    \includegraphics[width=\linewidth]{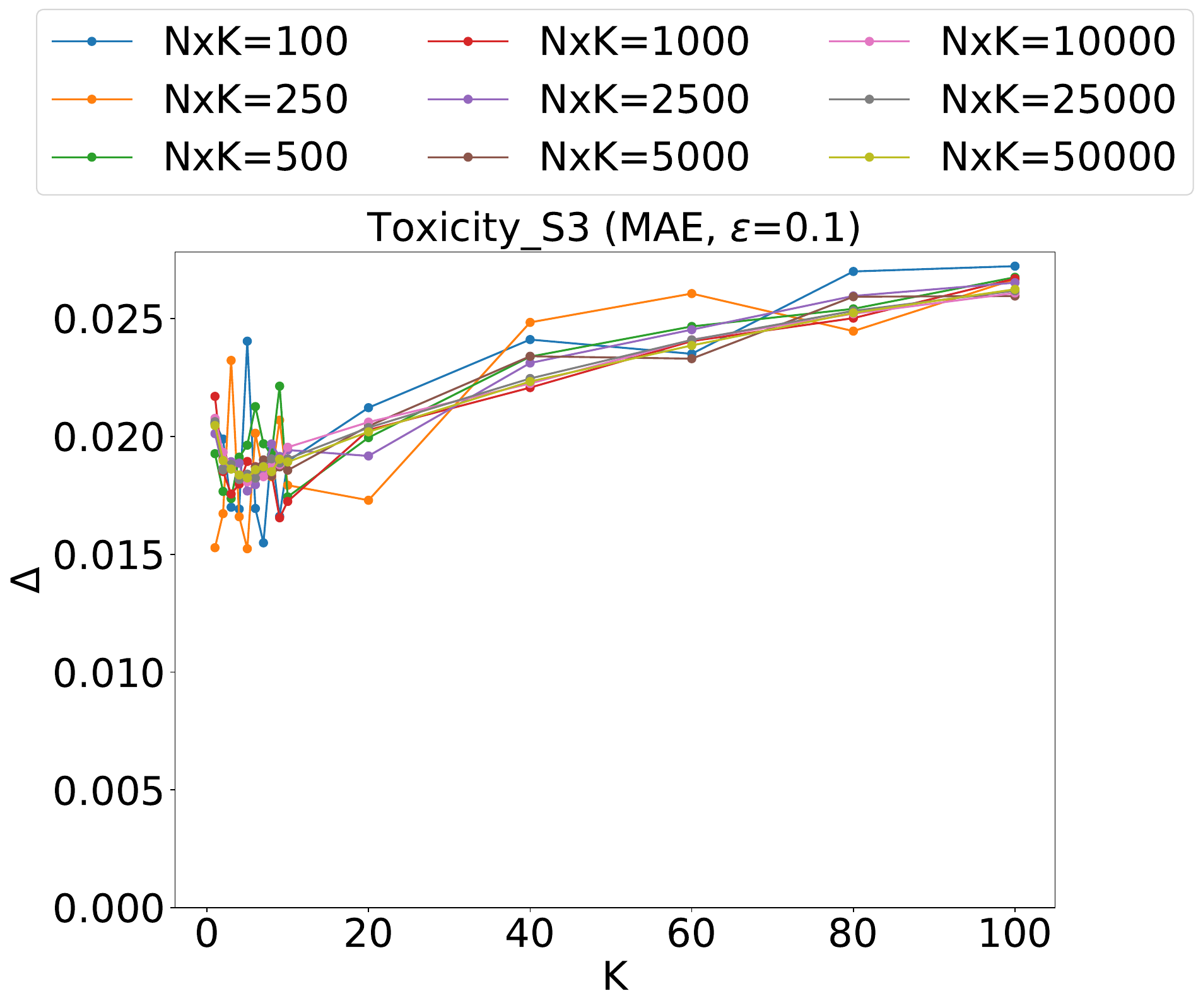}
    \caption{$\epsilon = 0.1$}
    \label{fig:toxicity_s3_delta_MAE_e01}
  \end{subfigure} \hfill
  \begin{subfigure}[b]{0.24\linewidth}
    \centering
    \includegraphics[width=\linewidth]{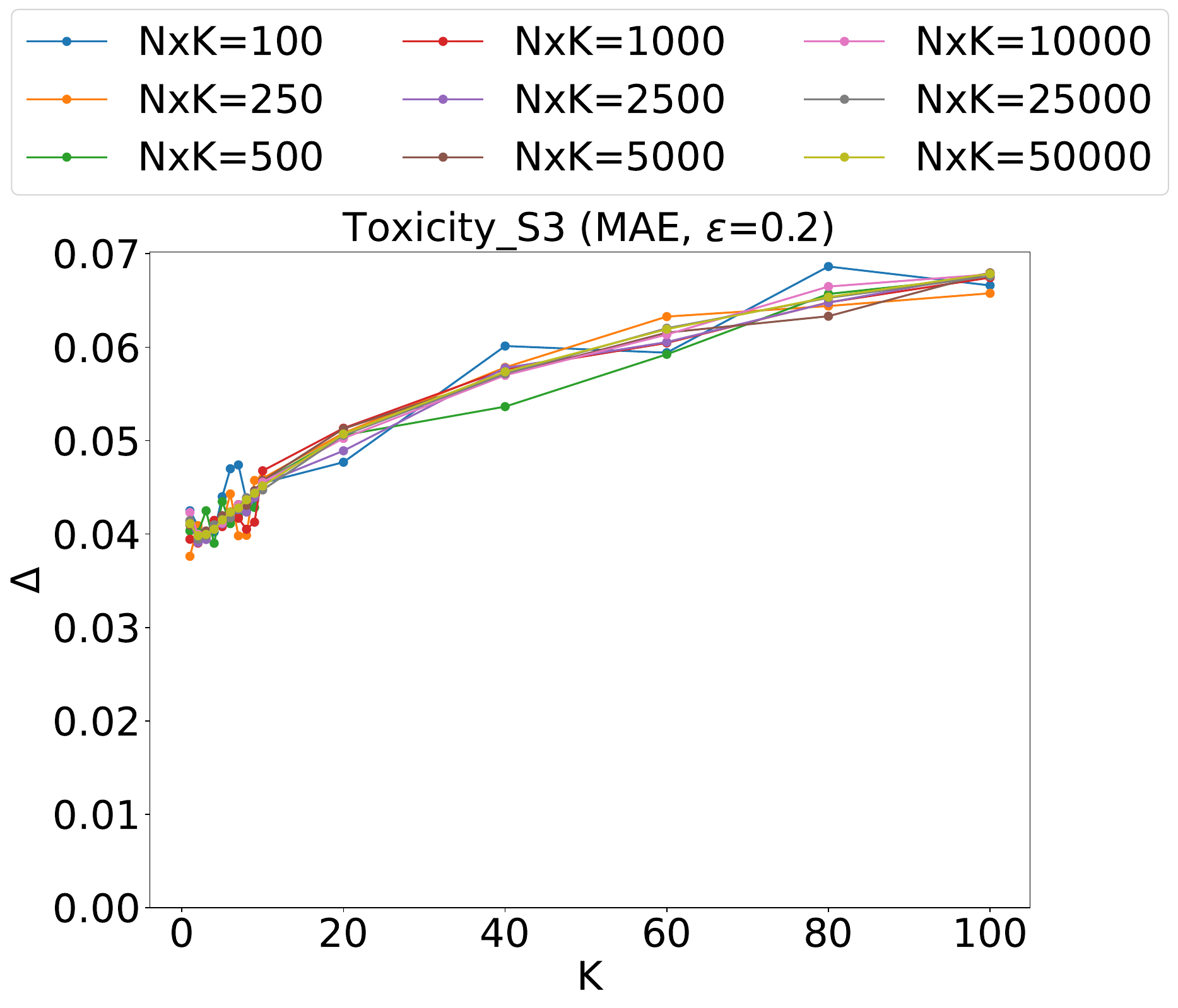}
    \caption{$\epsilon = 0.2$}
    \label{fig:toxicity_s3_delta_MAE_e02}
  \end{subfigure} \hfill
  \begin{subfigure}[b]{0.24\linewidth}
    \centering
    \includegraphics[width=\linewidth]{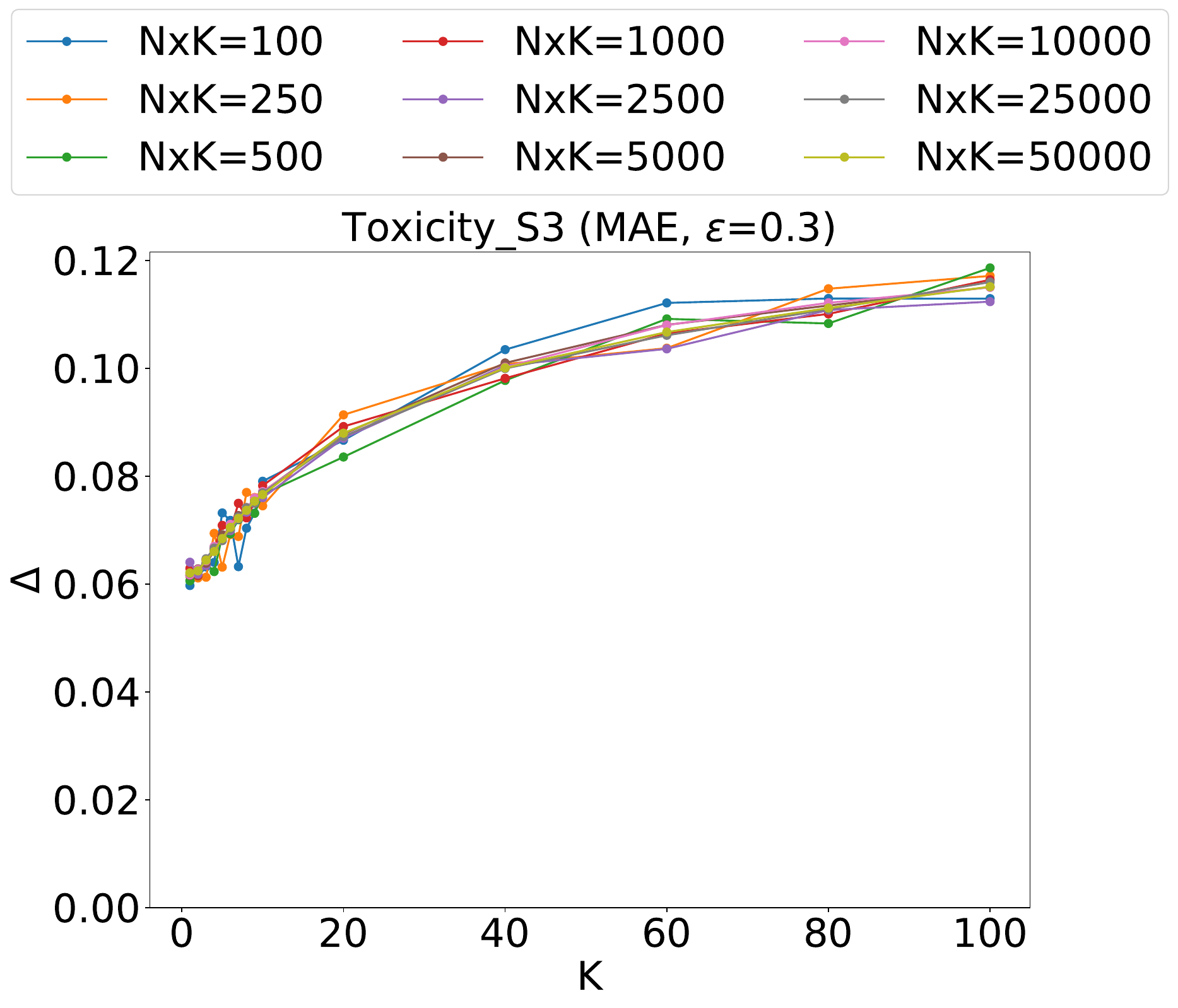}
    \caption{$\epsilon = 0.3$}
    \label{fig:toxicity_s3_delta_MAE_e03}
  \end{subfigure} \hfill
  \begin{subfigure}[b]{0.24\linewidth}
    \centering
    \includegraphics[width=\linewidth]{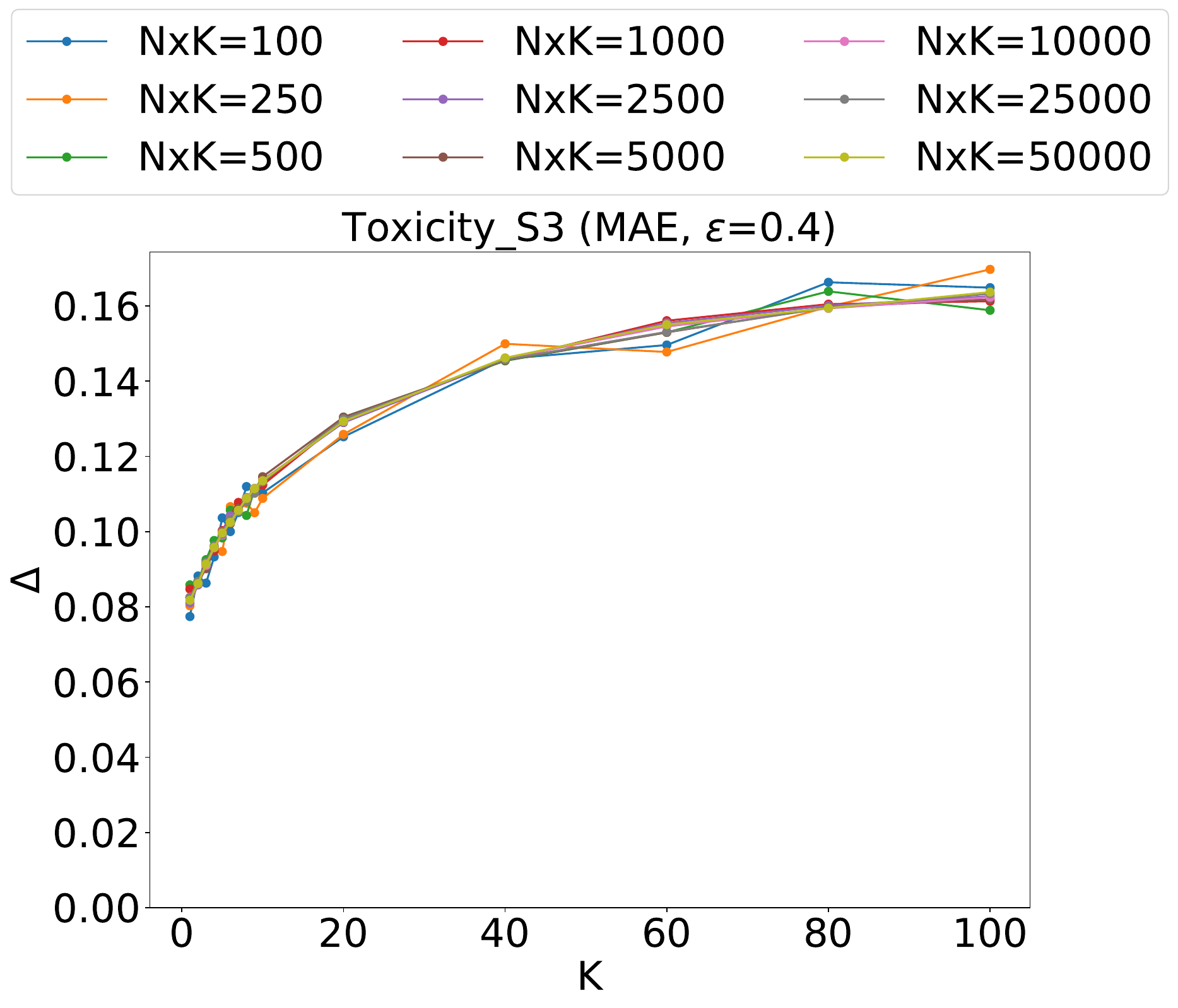}
    \caption{$\epsilon = 0.4$}
    \label{fig:toxicity_s3_delta_MAE_e04}
  \end{subfigure}
  \caption{S3: Effect sizes ($\Delta$) for Toxicity dataset with MAE as the metric}
  \label{fig:toxicity_s3_delta_MAE}
\end{figure*}

\begin{figure*}
  \centering
  \begin{subfigure}[b]{0.24\linewidth}
    \centering
    \includegraphics[width=\linewidth]{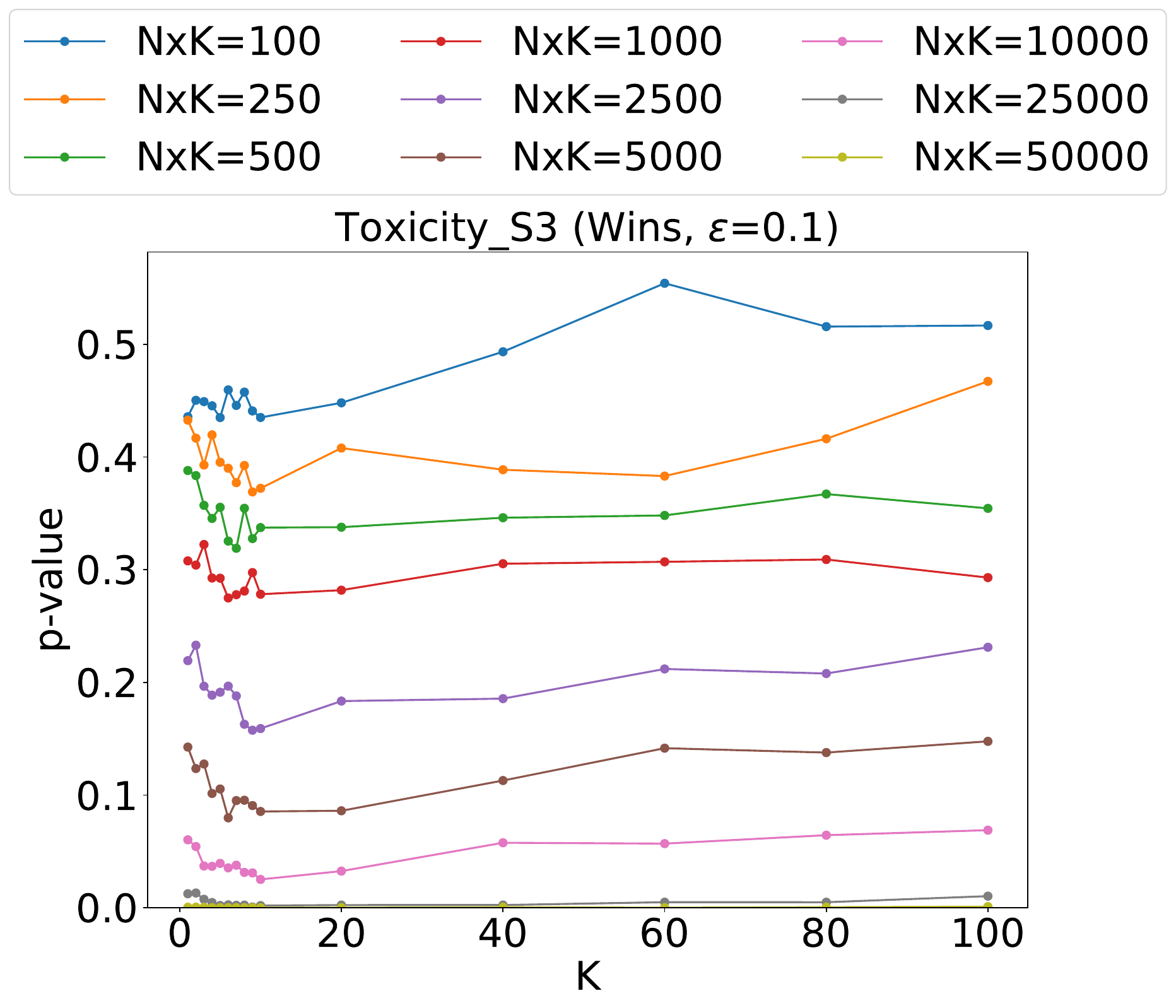}
    \caption{$\epsilon = 0.1$}
    \label{fig:toxicity_s3_wins_e01}
  \end{subfigure} \hfill
  \begin{subfigure}[b]{0.24\linewidth}
    \centering
    \includegraphics[width=\linewidth]{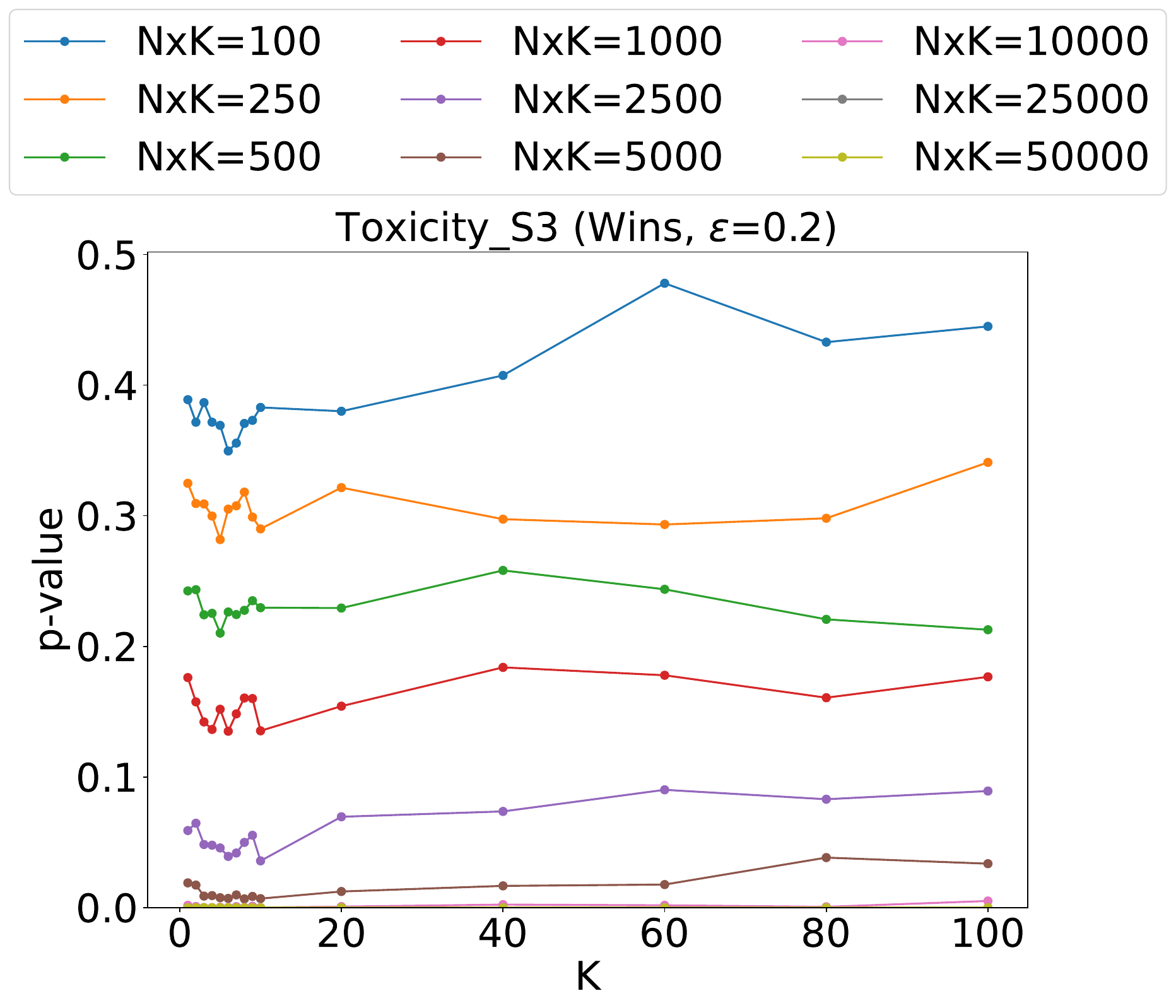}
    \caption{$\epsilon = 0.2$}
    \label{fig:toxicity_s3_wins_e02}
  \end{subfigure} \hfill
  \begin{subfigure}[b]{0.24\linewidth}
    \centering
    \includegraphics[width=\linewidth]{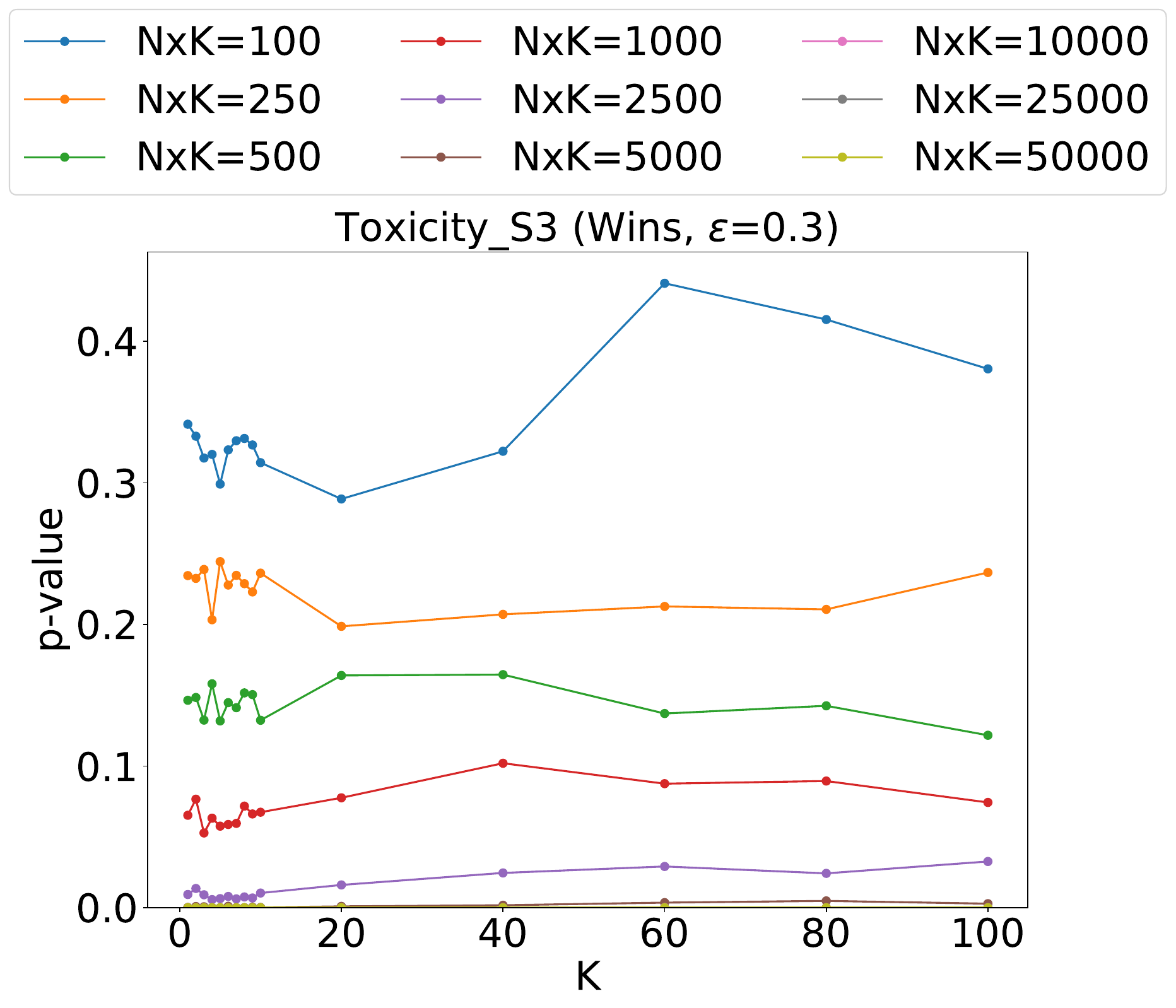}
    \caption{$\epsilon = 0.3$}
    \label{fig:toxicity_s3_wins_e03}
  \end{subfigure} \hfill
  \begin{subfigure}[b]{0.24\linewidth}
    \centering
    \includegraphics[width=\linewidth]{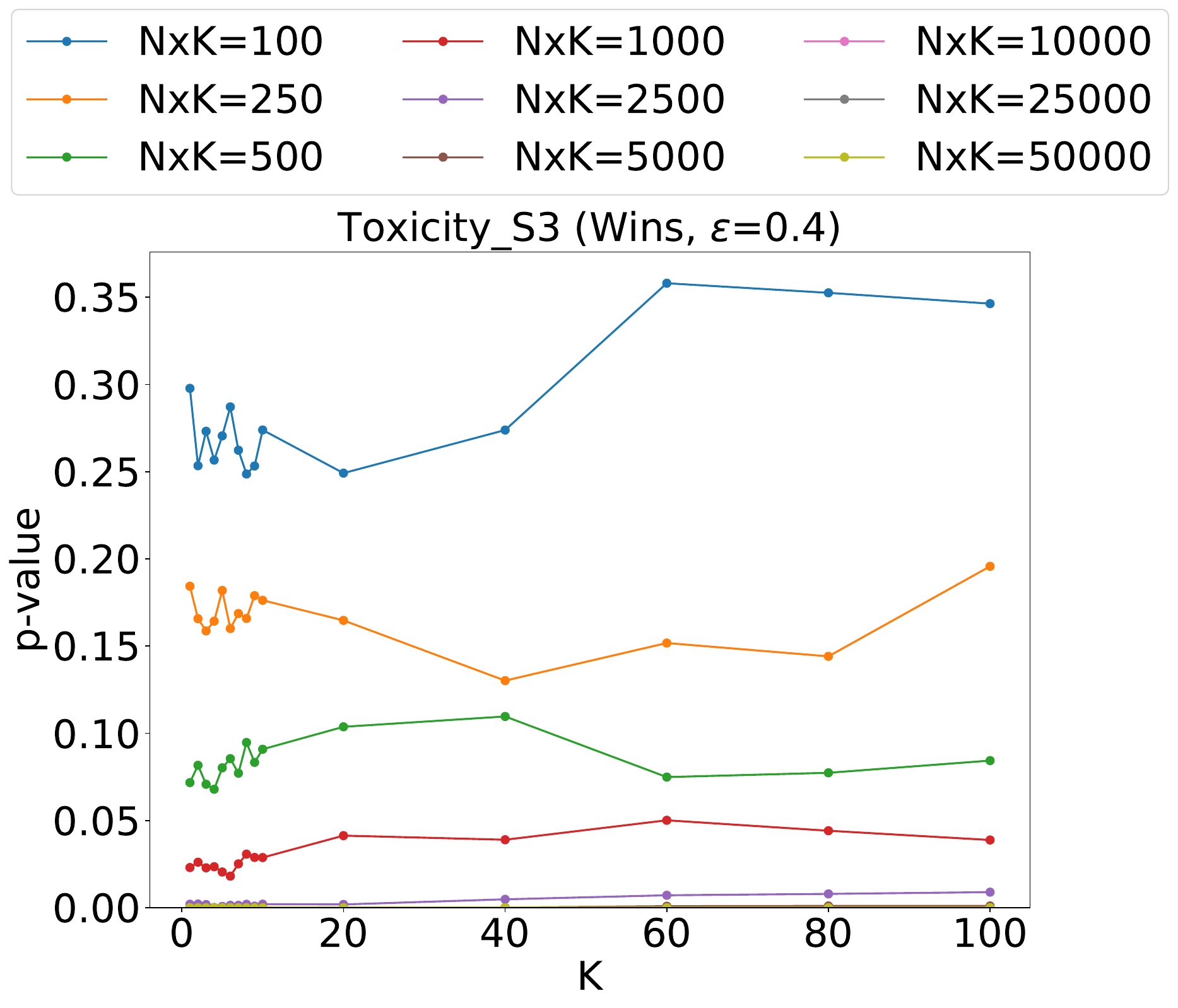}
    \caption{$\epsilon = 0.4$}
    \label{fig:toxicity_s3_wins_e04}
  \end{subfigure}
  \caption{S3: P-value plots for Toxicity dataset with Wins as the metric}
  \label{fig:toxicity_s3_wins}
\end{figure*}

\begin{figure*}
  \centering
  \begin{subfigure}[b]{0.24\linewidth}
    \centering
    \includegraphics[width=\linewidth]{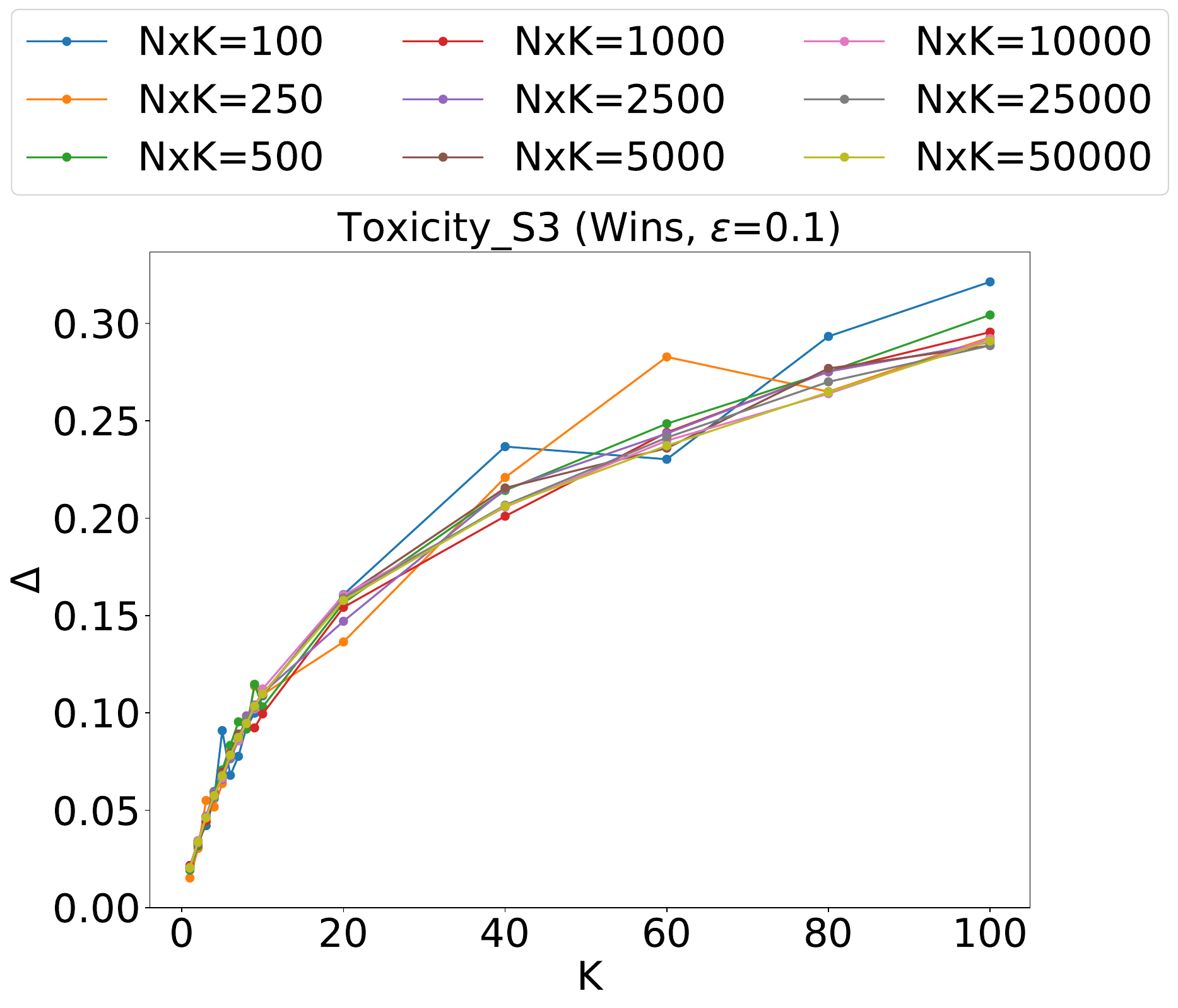}
    \caption{$\epsilon = 0.1$}
    \label{fig:toxicity_s3_delta_wins_e01}
  \end{subfigure} \hfill
  \begin{subfigure}[b]{0.24\linewidth}
    \centering
    \includegraphics[width=\linewidth]{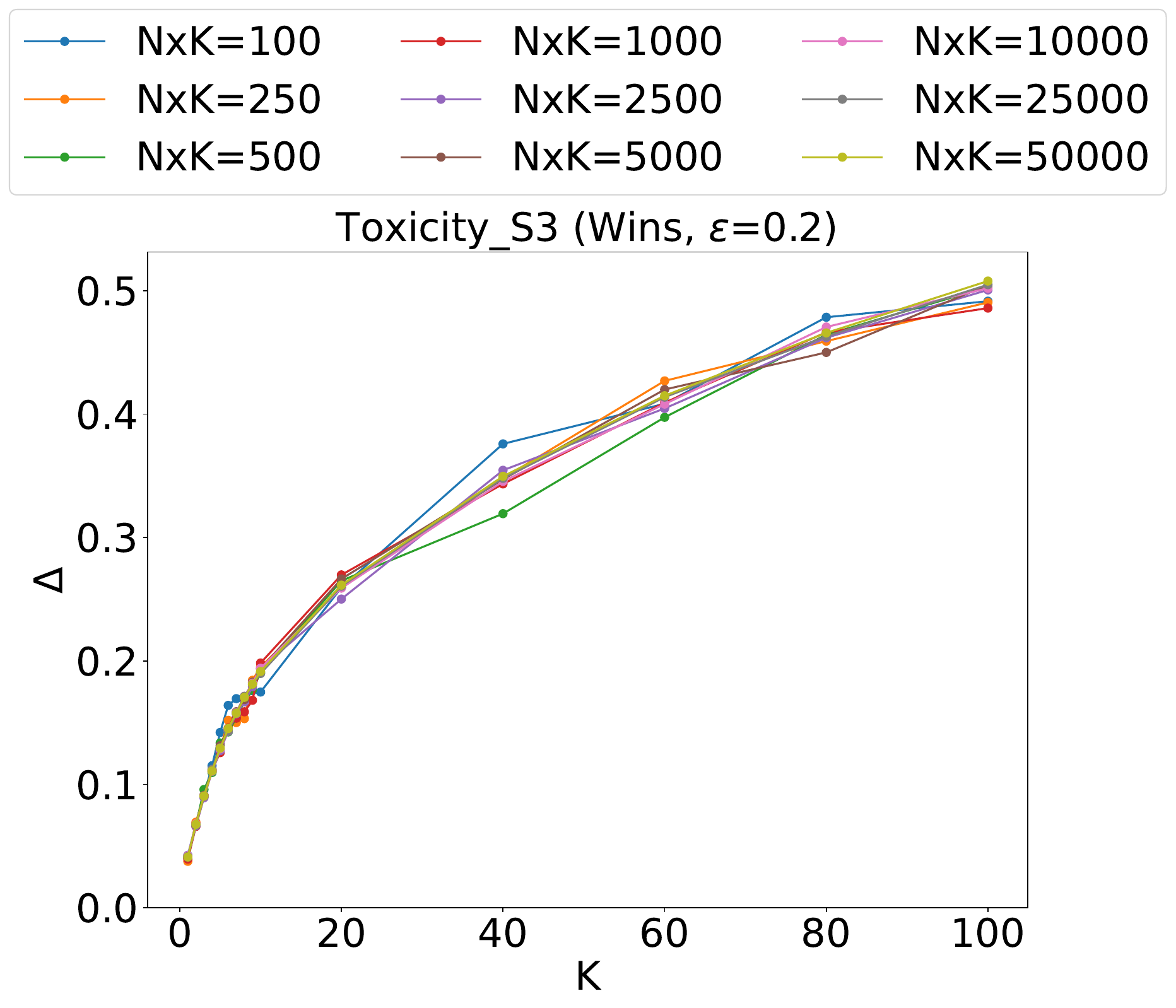}
    \caption{$\epsilon = 0.2$}
    \label{fig:toxicity_s3_delta_wins_e02}
  \end{subfigure} \hfill
  \begin{subfigure}[b]{0.24\linewidth}
    \centering
    \includegraphics[width=\linewidth]{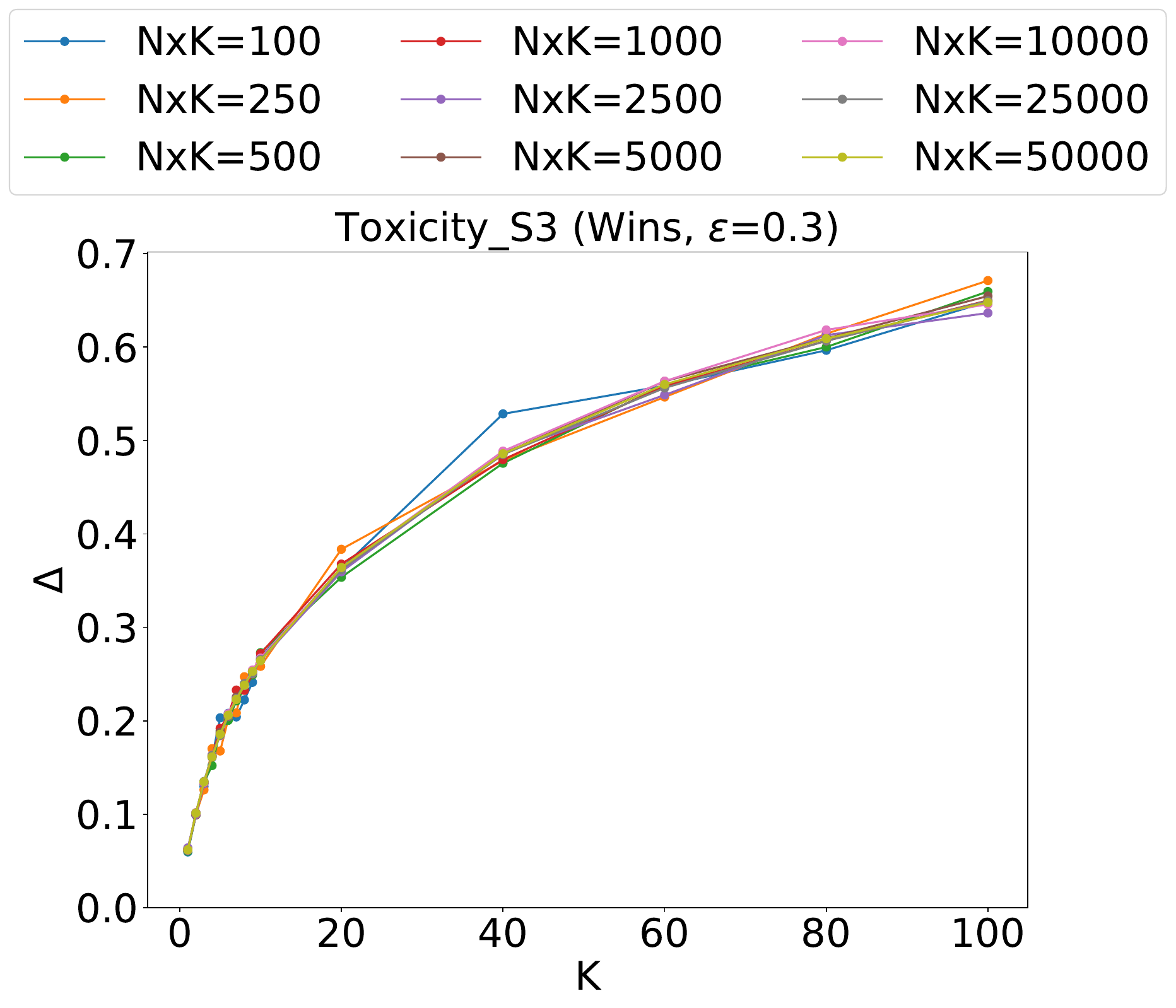}
    \caption{$\epsilon = 0.3$}
    \label{fig:toxicity_s3_delta_wins_e03}
  \end{subfigure} \hfill
  \begin{subfigure}[b]{0.24\linewidth}
    \centering
    \includegraphics[width=\linewidth]{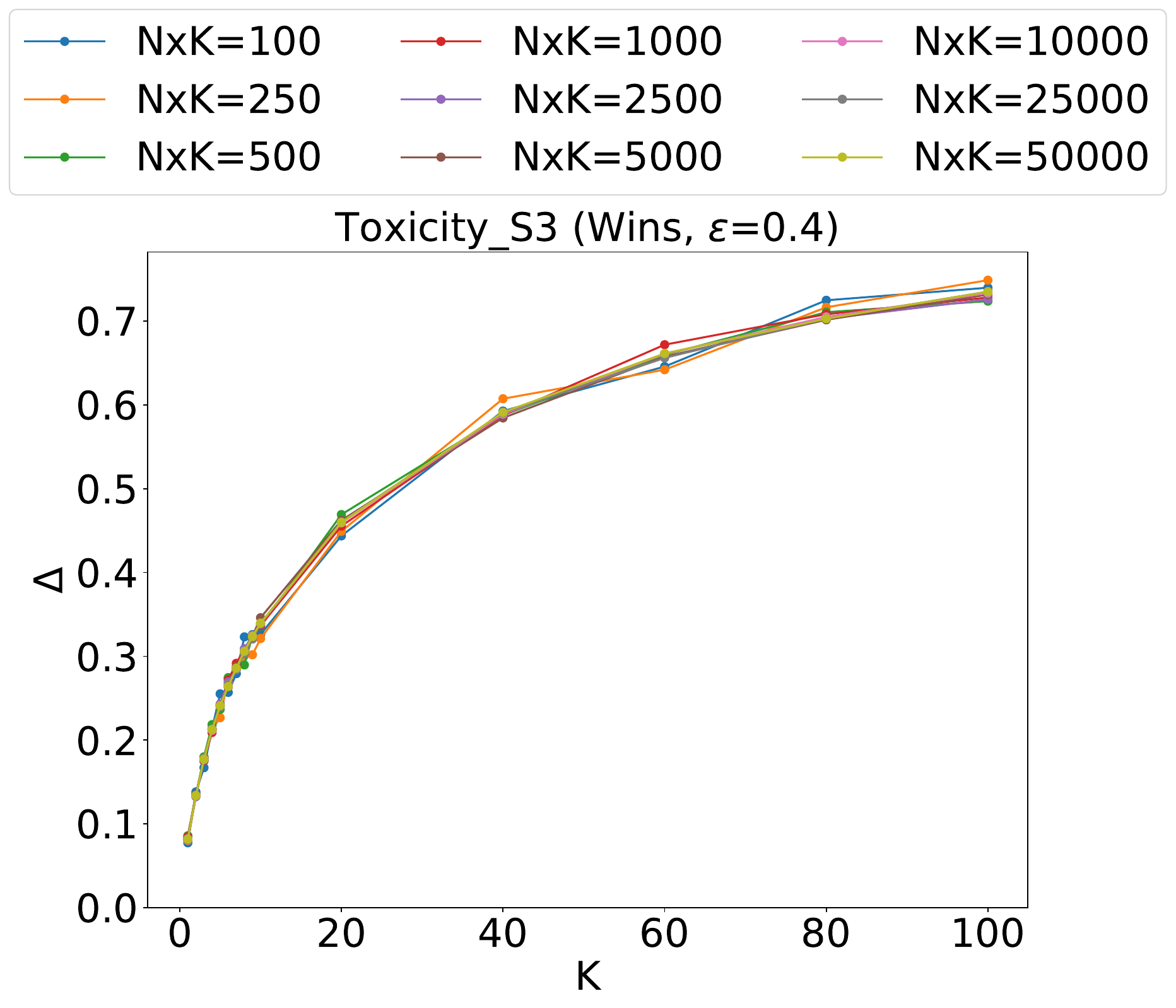}
    \caption{$\epsilon = 0.4$}
    \label{fig:toxicity_s3_delta_wins_e04}
  \end{subfigure}
  \caption{S3: Effect sizes ($\Delta$) for Toxicity dataset with Wins as the metric}
  \label{fig:toxicity_s3_delta_wins}
\end{figure*}

\end{document}